\documentclass[twoside]{article}
\usepackage[preprint]{aistats2027}
\usepackage[round]{natbib}
\renewcommand{\bibname}{References}
\renewcommand{\bibsection}{\subsubsection*{\bibname}}

\usepackage{amsmath,amsfonts,bm}

\def\eqref#1{equation~\ref{#1}}

\def\1{\bm{1}}

\DeclareMathAlphabet{\mathsfit}{\encodingdefault}{\sfdefault}{m}{sl}
\SetMathAlphabet{\mathsfit}{bold}{\encodingdefault}{\sfdefault}{bx}{n}

\usepackage{hyperref}
\usepackage{url}
\usepackage{booktabs}
\usepackage{multirow}
\usepackage{graphicx}
\usepackage{amsmath,amssymb}
\usepackage{xcolor}
\usepackage{colortbl}
\usepackage{tikz}
\usepackage{pifont}

\newcommand{\gapchip}[2]{\tikz[baseline=(c.base)]\node(c)[fill=#1,text=black,rounded corners=4pt,inner xsep=2pt,inner ysep=0.3pt,minimum width=4.6em]{#2};}
\newcommand{\gapdot}[2]{\tikz[baseline=(c.base)]\node(c)[fill=#1,text=black,rounded corners=3pt,inner xsep=1.5pt,inner ysep=0.3pt,minimum width=2.6em]{#2};}
\newcommand{\fitdot}[2]{\tikz[baseline=(c.base)]\node(c)[fill=#1,text=black,rounded corners=3pt,inner xsep=1.5pt,inner ysep=0.3pt,minimum width=2.1em]{#2};} 
\usepackage{placeins}\usepackage{stfloats} 
\usepackage{pifont}


\newcommand{\nRuns}{2098}
\newcommand{\nModels}{10}
\newcommand{\nLargeModels}{2}

\newcommand{\gapToneMeanQwenSmallUOneGsm}{$-0.2\pm0.4$}

\newcommand{\gapToneMeanQwenSmallUTwoGsm}{$+0.0\pm0.3$}

\newcommand{\gapToneMeanQwenSmallUFourGsm}{$+0.7\pm0.7$}

\newcommand{\gapToneMeanQwenSmallUSixteenGsm}{$+3.7\pm2.6$}

\newcommand{\gapToneMeanQwenSmallUSixtyFourGsm}{$-5.3\pm1.7$}

\newcommand{\gapGreedyGreedyQwenSmallUOneGsm}{$+1.6\pm1.4$}

\newcommand{\gapGreedyGreedyQwenSmallUTwoGsm}{$+0.9\pm0.7$}

\newcommand{\gapGreedyGreedyQwenSmallUFourGsm}{$+1.2\pm1.4$}

\newcommand{\gapToneMeanQwenSmallUSixteenMathFive}{$+1.9\pm1.7$}

\newcommand{\gainToneMeanQwenUTwoGsm}{$+6.9$}

\newcommand{\gainGrpoMeanQwenUTwoGsm}{$+2.1$}

\newcommand{\gapBFourToneMeanQwenUEightGsm}{$-7.5\pm4.0$}

\newcommand{\gainToneMeanQwenUOneMathFive}{$+5.8$}

\newcommand{\gainGrpoMeanQwenUOneMathFive}{$+4.8$}

\newcommand{\gainToneMeanQwenUTwoMathFive}{$+7.1$}

\newcommand{\gainGrpoMeanQwenUTwoMathFive}{$+2.2$}
\newcommand{\gapToneMeanQwenUTwoMathFive}{$+4.8\pm1.4$}

\newcommand{\gainToneMeanQwenUFourMathFive}{$+10.3$}

\newcommand{\gainGrpoMeanQwenUFourMathFive}{$+6.9$}

\newcommand{\gainToneMeanQwenUFourMathFiveMathTrain}{$+9.5$}

\newcommand{\gainGrpoMeanQwenUFourMathFiveMathTrain}{$+7.5$}

\newcommand{\gapToneMeanQwenUEightMathFive}{$-1.0\pm2.2$}

\newcommand{\gainGrpoMeanQwenUTwoFiftySixMathFive}{$+16.9$}

\newcommand{\gainGreedyGreedyQwenUOneMathFive}{$+1.5$}
\newcommand{\gainGrpoGreedyQwenUOneMathFive}{$+1.0$}

\newcommand{\gainGreedyGreedyQwenUTwoMathFive}{$+1.5$}
\newcommand{\gainGrpoGreedyQwenUTwoMathFive}{$+0.9$}

\newcommand{\gainGreedyGreedyQwenUFourMathFive}{$+2.4$}
\newcommand{\gainGrpoGreedyQwenUFourMathFive}{$+0.8$}

\newcommand{\gapGreedyGreedyQwenUFourMathFive}{$+3.2\pm0.4$}

\newcommand{\gapGreedyGreedyQwenUEightMathFive}{$-1.6\pm2.3$}

\newcommand{\gapToneMeanInstSmallUOneGsm}{$+0.2\pm0.2$}

\newcommand{\gapToneMeanInstSmallUTwoGsm}{$-0.4\pm0.1$}

\newcommand{\gapToneMeanInstSmallUFourGsm}{$-0.7\pm0.2$}

\newcommand{\gainToneMeanInstSmallUOneMathFive}{$-1.0$}

\newcommand{\gainGrpoMeanInstSmallUOneMathFive}{$+0.2$}

\newcommand{\gainToneMeanInstUOneMathFive}{$+0.1$}

\newcommand{\gainGrpoMeanInstUOneMathFive}{$+0.7$}

\newcommand{\gainToneMeanInstUFourMathFive}{$-1.0$}

\newcommand{\gainGrpoMeanInstUFourMathFive}{$+0.2$}

\newcommand{\gapGreedyGreedyInstUFourMathFive}{$-1.6\pm0.4$}

\newcommand{\gapToneMeanMathSmallUOneGsm}{$+0.7\pm0.7$}

\newcommand{\gapToneMeanMathSmallUTwoGsm}{$+1.1\pm0.7$}

\newcommand{\gapToneMeanMathSmallUFourGsm}{$-0.1\pm0.7$}

\newcommand{\gapToneMeanMathSmallUSixtyFourGsm}{$-5.2\pm3.2$}

\newcommand{\gapGreedyGreedyMathSmallUOneGsm}{$+7.4\pm4.1$}

\newcommand{\gainGreedyGreedyMathSmallUTwoGsm}{$+27.3$}
\newcommand{\gainGrpoGreedyMathSmallUTwoGsm}{$+3.0$}

\newcommand{\gapGreedyGreedyMathSmallUTwoGsm}{$+27.3\pm9.0$}

\newcommand{\gapGreedyGreedyMathSmallUFourGsm}{$+7.2\pm2.2$}

\newcommand{\gapGreedyGreedyMathSmallUEightGsm}{$+17.3\pm11.4$}

\newcommand{\gapGreedyGreedyMathSmallUSixteenGsm}{$+6.8\pm3.0$}

\newcommand{\gapGreedyGreedyMathSmallUSixtyFourGsm}{$+1.0\pm1.4$}

\newcommand{\gainGreedyGreedyMathSmallUTwoMathFive}{$+11.4$}
\newcommand{\gainGrpoGreedyMathSmallUTwoMathFive}{$-2.2$}

\newcommand{\gapGreedyGreedyMathSmallUTwoMathFive}{$+15.1\pm7.4$}

\newcommand{\gainToneMeanMathUFourGsm}{$+16.5$}

\newcommand{\gainGrpoMeanMathUFourGsm}{$+11.2$}
\newcommand{\gapToneMeanMathUFourGsm}{$+5.2\pm2.1$}

\newcommand{\gapToneMeanMathUEightGsm}{$+7.9\pm1.7$}

\newcommand{\gainGreedyGreedyMathUTwoGsm}{$+12.3$}
\newcommand{\gainGrpoGreedyMathUTwoGsm}{$+4.4$}

\newcommand{\gainGreedyGreedyMathUFourGsm}{$+11.9$}
\newcommand{\gainGrpoGreedyMathUFourGsm}{$+7.9$}

\newcommand{\gapGreedyGreedyMathUEightGsm}{$+3.1\pm3.2$}

\newcommand{\gainToneMeanMathUFourMathFive}{$+12.0$}

\newcommand{\gainGrpoMeanMathUFourMathFive}{$+8.5$}
\newcommand{\gapToneMeanMathUFourMathFive}{$+3.6\pm1.7$}

\newcommand{\gapToneMeanMathUEightMathFive}{$+1.9\pm0.2$}

\newcommand{\gainGrpoMeanMathUTwoFiftySixMathFive}{$+21.7$}

\newcommand{\gainGreedyGreedyMathUFourMathFive}{$+5.7$}
\newcommand{\gainGrpoGreedyMathUFourMathFive}{$+4.4$}

\newcommand{\gapGreedyGreedyMathUEightMathFive}{$+1.8\pm0.8$}

\newcommand{\gapToneMeanLlamaUOneGsm}{$+0.4\pm0.2$}

\newcommand{\gapGreedyGreedyLlamaUOneGsm}{$+1.1\pm0.5$}

\newcommand{\gapGreedyGreedyLlamaUTwoGsm}{$-0.1\pm0.5$}

\newcommand{\gapGreedyGreedyLlamaUFourGsm}{$+1.2\pm0.8$}

\newcommand{\gapGreedyGreedyLlamaUSixteenGsm}{$+2.2\pm0.4$}

\newcommand{\gapToneMeanMistralUFourGsm}{$+0.6\pm0.4$}

\newcommand{\gapToneMeanMistralUSixteenGsm}{$-2.3\pm1.1$}

\newcommand{\gapGreedyGreedyMistralUOneGsm}{$+0.7\pm0.5$}

\newcommand{\gapGreedyGreedyMistralUTwoGsm}{$+0.8\pm1.1$}

\newcommand{\gapGreedyGreedyMistralUFourGsm}{$+1.6\pm0.5$}

\newcommand{\gapGreedyGreedyMistralUSixteenGsm}{$+1.0\pm1.1$}

\newcommand{\gapToneMeanDeepseekBaseUFourGsm}{$+0.4\pm0.1$}

\newcommand{\gapToneMeanDeepseekUOneGsm}{$+0.2\pm0.3$}

\newcommand{\gapToneMeanDeepseekUTwoGsm}{$+0.3\pm0.3$}

\newcommand{\gapToneMeanDeepseekUFourGsm}{$+0.7\pm0.1$}

\newcommand{\gapToneMeanDeepseekUSixteenGsm}{$-3.1\pm1.6$}

\newcommand{\gapGreedyGreedyDeepseekUOneGsm}{$+5.6\pm3.0$}

\newcommand{\gapGreedyGreedyDeepseekUTwoGsm}{$+1.9\pm5.2$}

\newcommand{\gapGreedyGreedyDeepseekUFourGsm}{$+5.5\pm3.4$}

\newcommand{\gapGreedyGreedyDeepseekUFourMathFive}{$+2.5\pm0.4$}

\newcommand{\budgetEndGreedyMathUFourGsm}{$+9.7$}

\newcommand{\reachGreedyMathSmallUTwoGsm}{256}
\newcommand{\budgetEndGreedyMathSmallUTwoGsm}{$+27.3$}

\newcommand{\budgetEndGreedyMathSmallUTwoMathFive}{$+15.1$}

\newcommand{\reachGreedyMathSmallUOneGsmMathTrain}{256}

\newcommand{\budgetEndGreedyMathSmallUOneMathFiveMathTrain}{$+4.9$}

\newcommand{\budgetEndGreedyMathSmallUTwoGsmMathTrain}{$+23.2$}

\newcommand{\budgetEndGreedyMathSmallUTwoMathFiveMathTrain}{$+14.1$}

\newcommand{\budgetEndGreedyMathSmallUFourGsmMathTrain}{$+13.2$}

\newcommand{\reachPoolGreedyGsm}{2k}
\newcommand{\poolEndGreedyGsm}{$+2.6$}
\newcommand{\poolEndGrpoGsm}{$+1.1$}
\newcommand{\reachPoolGreedyMathFive}{2k}
\newcommand{\poolEndGreedyMathFive}{$+1.6$}
\newcommand{\poolEndGrpoMathFive}{$+1.0$}
\newcommand{\grpoCompletions}{101k}
\newcommand{\boCompletions}{123k}
\newcommand{\grpoPasses}{4}
\newcommand{\boComputeShare}{30}

\newcommand{\batchFourToneMeanUTwoGsm}{$-0.8$}

\newcommand{\batchFourToneMeanUTwoMathFive}{$-0.3$}

\newcommand{\batchFourGreedyGreedyUTwoGsm}{$+0.5$}

\newcommand{\batchFourGreedyGreedyUTwoMathFive}{$-0.5$}

\newcommand{\trainGapGreedyUOne}{$+0.3\pm0.6$}
\newcommand{\trainGapGreedyUTwo}{$+1.3\pm0.3$}
\newcommand{\trainGapGreedyUFour}{$+1.3\pm0.9$}

\newcommand{\trainGapGreedyUTwoSmall}{$+13.8\pm7.4$}

\newcommand{\objectiveUFourGsmBase}{$+3.2\pm1.3$}
\newcommand{\objectiveUFourMathFiveBase}{$+1.8\pm0.8$}

\newcommand{\objectiveUFourGsmMathVariants}{$+4.2\pm2.7$}
\newcommand{\objectiveUFourMathFiveMathVariants}{$+2.8\pm1.2$}

\newcommand{\objectiveUFourGsmSmall}{$+6.3\pm3.1$}
\newcommand{\objectiveUFourMathFiveSmall}{$+3.3\pm1.8$}

\newcommand{\objectiveQwenUFourGsm}{$-0.4$}
\newcommand{\objectiveQwenUFourMathFive}{$-0.8$}

\newcommand{\objectiveMathSmallUTwoGsm}{$+29.4$}

\newcommand{\poolGreedyGreedyUOneGsm}{$+2.6\pm1.1$}
\newcommand{\poolGreedyGreedyUOneMathFive}{$+0.5\pm0.5$}
\newcommand{\poolGreedyGreedyUTwoGsm}{$+1.0\pm1.0$}
\newcommand{\poolGreedyGreedyUTwoMathFive}{$+0.6\pm0.3$}
\newcommand{\poolGreedyGreedyUFourGsm}{$+2.9\pm1.9$}
\newcommand{\poolGreedyGreedyUFourMathFive}{$+1.7\pm0.6$}

\newcommand{\poolToneMeanUOneGsmBase}{$+0.6\pm0.3$}
\newcommand{\poolToneMeanUOneMathFiveBase}{$+0.3\pm0.3$}
\newcommand{\poolToneMeanUTwoGsmBase}{$+1.4\pm0.5$}
\newcommand{\poolToneMeanUTwoMathFiveBase}{$+1.3\pm0.5$}
\newcommand{\poolToneMeanUFourGsmBase}{$+0.9\pm0.5$}
\newcommand{\poolToneMeanUFourMathFiveBase}{$+1.0\pm0.4$}

\newcommand{\poolGreedyGreedyUOneGsmBase}{$+0.9\pm0.3$}

\newcommand{\poolGreedyGreedyUTwoMathFiveBase}{$+0.8\pm0.2$}
\newcommand{\poolGreedyGreedyUFourGsmBase}{$+0.5\pm0.4$}
\newcommand{\poolGreedyGreedyUFourMathFiveBase}{$+1.3\pm0.4$}

\newcommand{\poolToneMeanUTwoGsmMathVariants}{$+2.6\pm1.4$}
\newcommand{\poolToneMeanUTwoMathFiveMathVariants}{$+1.4\pm1.0$}
\newcommand{\poolToneMeanUFourGsmMathVariants}{$+3.0\pm1.2$}
\newcommand{\poolToneMeanUFourMathFiveMathVariants}{$+2.0\pm0.9$}
\newcommand{\poolToneMeanUEightGsmMathVariants}{$+4.6\pm1.7$}
\newcommand{\poolToneMeanUEightMathFiveMathVariants}{$+1.4\pm0.4$}

\newcommand{\poolGreedyGreedyUOneGsmMathVariants}{$+5.8\pm2.8$}

\newcommand{\poolGreedyGreedyUTwoGsmMathVariants}{$+2.8\pm3.1$}

\newcommand{\poolGreedyGreedyUFourGsmMathVariants}{$+7.6\pm5.4$}

\newcommand{\poolToneMeanUOneGsmSmall}{$+0.2\pm0.4$}
\newcommand{\poolToneMeanUOneMathFiveSmall}{$-0.1\pm0.4$}
\newcommand{\poolToneMeanUTwoGsmSmall}{$+0.6\pm0.4$}
\newcommand{\poolToneMeanUTwoMathFiveSmall}{$+0.1\pm0.4$}
\newcommand{\poolToneMeanUFourGsmSmall}{$+0.3\pm0.5$}
\newcommand{\poolToneMeanUFourMathFiveSmall}{$-0.7\pm0.4$}

\newcommand{\poolGreedyGreedyUOneGsmSmall}{$+4.5\pm2.3$}

\newcommand{\poolGreedyGreedyUTwoGsmSmall}{$+14.1\pm7.1$}

\newcommand{\poolGreedyGreedyUFourGsmSmall}{$+4.2\pm1.8$}

\newcommand{\poolToneMeanUOneGsmShiftLarge}{$+1.0\pm0.7$}
\newcommand{\poolToneMeanUOneMathFiveShiftLarge}{$+1.0\pm0.6$}
\newcommand{\poolToneMeanUTwoGsmShiftLarge}{$+4.9\pm1.4$}
\newcommand{\poolToneMeanUTwoMathFiveShiftLarge}{$+3.7\pm1.2$}
\newcommand{\poolToneMeanUFourGsmShiftLarge}{$+4.1\pm1.4$}
\newcommand{\poolToneMeanUFourMathFiveShiftLarge}{$+3.5\pm1.1$}

\newcommand{\poolGreedyGreedyUOneMathFiveShiftLarge}{$+1.8\pm0.9$}

\newcommand{\poolGreedyGreedyUTwoMathFiveShiftLarge}{$+0.3\pm0.7$}

\newcommand{\poolGreedyGreedyUFourMathFiveShiftLarge}{$+2.7\pm1.6$}

\newcommand{\poolGreedyGreedyUOneMathFiveShiftSmall}{$-0.1\pm0.8$}

\newcommand{\poolGreedyGreedyUTwoMathFiveShiftSmall}{$+7.7\pm4.7$}

\newcommand{\poolGreedyGreedyUFourMathFiveShiftSmall}{$+1.5\pm1.4$}

\newcommand{\poolToneMeanUOneGsmShiftLargeMathTrain}{$+1.2\pm0.7$}
\newcommand{\poolToneMeanUOneMathFiveShiftLargeMathTrain}{$+1.2\pm0.5$}
\newcommand{\poolToneMeanUTwoGsmShiftLargeMathTrain}{$+2.5\pm1.3$}
\newcommand{\poolToneMeanUTwoMathFiveShiftLargeMathTrain}{$+2.0\pm0.9$}
\newcommand{\poolToneMeanUFourGsmShiftLargeMathTrain}{$+1.4\pm1.2$}
\newcommand{\poolToneMeanUFourMathFiveShiftLargeMathTrain}{$+1.6\pm0.9$}

\newcommand{\poolGreedyGreedyUOneGsmShiftLargeMathTrain}{$+0.9\pm1.1$}
\newcommand{\poolGreedyGreedyUOneMathFiveShiftLargeMathTrain}{$+1.0\pm1.1$}
\newcommand{\poolGreedyGreedyUTwoGsmShiftLargeMathTrain}{$+1.8\pm2.2$}
\newcommand{\poolGreedyGreedyUTwoMathFiveShiftLargeMathTrain}{$-0.5\pm0.9$}
\newcommand{\poolGreedyGreedyUFourGsmShiftLargeMathTrain}{$+2.1\pm2.8$}
\newcommand{\poolGreedyGreedyUFourMathFiveShiftLargeMathTrain}{$+2.8\pm1.0$}

\newcommand{\poolToneMeanUOneGsmShiftSmallMathTrain}{$+0.1\pm0.3$}
\newcommand{\poolToneMeanUOneMathFiveShiftSmallMathTrain}{$+0.1\pm0.3$}
\newcommand{\poolToneMeanUTwoGsmShiftSmallMathTrain}{$+0.1\pm0.3$}
\newcommand{\poolToneMeanUTwoMathFiveShiftSmallMathTrain}{$-0.5\pm0.2$}
\newcommand{\poolToneMeanUFourGsmShiftSmallMathTrain}{$-0.3\pm0.3$}
\newcommand{\poolToneMeanUFourMathFiveShiftSmallMathTrain}{$+0.3\pm0.2$}

\newcommand{\poolGreedyGreedyUOneGsmShiftSmallMathTrain}{$+8.9\pm4.9$}
\newcommand{\poolGreedyGreedyUOneMathFiveShiftSmallMathTrain}{$+2.0\pm1.6$}
\newcommand{\poolGreedyGreedyUTwoGsmShiftSmallMathTrain}{$+12.0\pm6.1$}
\newcommand{\poolGreedyGreedyUTwoMathFiveShiftSmallMathTrain}{$+7.3\pm3.7$}
\newcommand{\poolGreedyGreedyUFourGsmShiftSmallMathTrain}{$+6.5\pm6.6$}
\newcommand{\poolGreedyGreedyUFourMathFiveShiftSmallMathTrain}{$+7.1\pm4.4$}

\newcommand{\poolToneMeanUSixtyFourMathFiveLargeU}{$-1.6\pm2.3$}

\newcommand{\poolToneMeanUTwoFiftySixMathFiveLargeU}{$-4.7\pm2.2$}

\newcommand{\poolGreedyGreedyUSixtyFourMathFiveLargeU}{$-2.9\pm1.3$}

\newcommand{\poolGreedyGreedyUTwoFiftySixMathFiveLargeU}{$-0.8\pm1.6$}

\newcommand{\poolToneMeanUSixtyFourMathFiveLargeUSmall}{$-2.4\pm0.6$}

\newcommand{\poolToneMeanUTwoFiftySixMathFiveLargeUSmall}{$-9.7\pm0.7$}

\newcommand{\poolGreedyGreedyUSixtyFourMathFiveLargeUSmall}{$-0.4\pm1.4$}

\newcommand{\poolGreedyGreedyUTwoFiftySixMathFiveLargeUSmall}{$-4.6\pm1.3$}

\newcommand{\shiftDeltaToneMeanUTwoGsmLarge}{$-2.3\pm1.9$}
\newcommand{\shiftDeltaToneMeanUTwoMathFiveLarge}{$-1.7\pm1.4$}
\newcommand{\shiftDeltaToneMeanUFourGsmLarge}{$-2.7\pm1.8$}
\newcommand{\shiftDeltaToneMeanUFourMathFiveLarge}{$-1.9\pm1.4$}

\newcommand{\shiftDeltaBFourToneMeanUFourGsmLarge}{$-2.9\pm1.8$}
\newcommand{\shiftDeltaBFourToneMeanUFourMathFiveLarge}{$-1.3\pm1.5$}

\newcommand{\searchBiasTone}{$+0.4\pm0.9$}
\newcommand{\searchBiasBFourTone}{$+3.7\pm1.1$}
\newcommand{\searchBiasModelsTone}{5}
\newcommand{\searchBiasGreedy}{$+1.1\pm0.3$}
\newcommand{\searchBiasBFourGreedy}{$+1.9\pm0.7$}
\newcommand{\searchBiasModelsGreedy}{9}

\newcommand{\worstToneMeanLargeMathFive}{-9.5}

\newcommand{\worstGreedyGreedyLargeMathFive}{-7.8}

\newcommand{\worstToneMeanLargeMathFiveSmall}{-12.5}

\newcommand{\othersTinyBest}{1.9}

\newcommand{\loraGainMeanQwenMathFive}{$+28.5$}

\newcommand{\loraGainMeanInstMathFive}{$+2.9$}

\newcommand{\loraGainMeanMathMathFive}{$+30.4$}

\newcommand{\loraGainGreedyLlamaGsm}{$+18.9$}

\newcommand{\loraGainGreedyLlamaMathFive}{$+5.5$}

\newcommand{\loraGainGreedyMistralGsm}{$+19.4$}

\newcommand{\loraGainGreedyMistralMathFive}{$+5.7$}

\newcommand{\loraGainGreedyDeepseekBaseGsm}{$+2.1$}

\newcommand{\loraGainGreedyDeepseekBaseMathFive}{$+2.4$}

\newcommand{\grpoOverheadSmall}{1.4}
\newcommand{\grpoOverheadLarge}{1.6--10.4}

\newcommand{\grpoOverheadDeepseek}{1.6}
\newcommand{\grpoOverheadQwen}{8.9}
\newcommand{\memExampleFree}{5.8}
\newcommand{\memExampleSeqMiB}{84}
\newcommand{\memExampleSeqs}{71}
\newcommand{\memExampleGrpoPerSeq}{2.2}
\newcommand{\memExampleGrpoSeqs}{2}
\newcommand{\grpoMicroBatch}{4}
\newcommand{\cardSmallGiB}{22.0}
\newcommand{\cardFortyGiB}{39.4}
\newcommand{\cardLargeGiB}{44.4}
\newcommand{\cardEightyGiB}{79.2}
\newcommand{\engineGiB}{2}
\newcommand{\designEffectiveQwen}{2.0--3.1}
\newcommand{\designEffectMax}{3.0}
\newcommand{\designEqualOthers}{83--97}
\newcommand{\designSteps}{20}
\newcommand{\grpoGroup}{4}
\newcommand{\nLargeArch}{4}
\newcommand{\grpoNoFitSmallCard}{3}

\newcommand{\boSequencesSmallCard}{10--71}
\newcommand{\boSequenceRatio}{1.6--29}

\newcommand{\snrSemInstruct}{2.6--3.0}
\newcommand{\snrReplicateSem}{1.0}
\newcommand{\snrTrials}{497}
\newcommand{\grpoBestInstruct}{1.2}
\newcommand{\snrEvalsOne}{61}
\newcommand{\snrEvalsHalf}{246}
\newcommand{\snrBelievedInstUOne}{$+0.8\pm0.2$}
\newcommand{\snrThetaInstUOne}{0.11}

\newcommand{\snrBelievedInstUFour}{$+2.3\pm0.3$}
\newcommand{\snrThetaInstUFour}{0.45}
\newcommand{\snrBelievedInstSmallUOne}{$+0.5\pm0.2$}

\newcommand{\snrBelievedInstSmallUFour}{$+1.7\pm0.3$}

\newcommand{\snrBelievedQwenUOne}{$+4.6\pm0.8$}

\newcommand{\snrBelievedQwenUFour}{$+9.2\pm0.9$}

\newcommand{\mcN}{\mathcal{N}}
\newcommand{\mcD}{\mathcal{D}}
\newcommand{\mbR}{\mathbb{R}}
\newcommand{\mbE}{\mathbb{E}}
\newcommand{\llmbo}{LlMBO}

\definecolor{limboGreenText}{RGB}{35,105,55}
\definecolor{bothGray}{RGB}{247,247,247}

\begin{document}

\twocolumn[
\aistatstitle{How Low Can You Go: Fine-Tuning LLMs \\with Bayesian Optimization (\llmbo{})}
\aistatsauthor{
  Aditya Kulkarni\textsuperscript{1,*} \And
  Davide Sartor\textsuperscript{2,*} \And
  Carlos Soto\textsuperscript{2} \And
  Nathan Wycoff\textsuperscript{2}
}
\aistatsaddress{
  \textsuperscript{1}Department of Physics\\
  \textsuperscript{2}Department of Mathematics and Statistics\\
  University of Massachusetts Amherst
}]
\def\thefootnote{*}\footnotetext{Equal contribution.}\def\thefootnote{\arabic{footnote}}

\begin{abstract}
Recently, Morris et al. introduced TinyLoRA, a strategy for LLM fine-tuning which introduces arbitrarily few parameters, and demonstrated accuracy improvements using less than $10$ parameters rather than the ${\gtrsim 10,000}$ typically demanded by LoRA-based methods.
However, that study continued to use gradient-based local search methods of the kind deployed for training multi-billion parameter LLMs in the first place.
This article is predicated on the observation that, in the low dimensional setting, global optimization methods become tractable, unlocking Large language Model Bayesian Optimization (\llmbo{}).
\llmbo{} is sample-efficient, gradient-free, memory-light, parallelizes across devices without synchronization, and, unlike policy-gradient methods, can optimize greedy-decoding accuracy directly.
We describe numerical experiments comparing \llmbo{} against GRPO on fine-tuning tasks for a variety of models on a math benchmark suite which show that \llmbo{} can be a suitable alternative in this regime, matching or exceeding GRPO's fine tuning gains, while using a fraction of the compute and memory.
\end{abstract}

\section{Introduction}\label{sec:intro}
\begin{figure}[t]
\centering
\includegraphics[width=\linewidth]{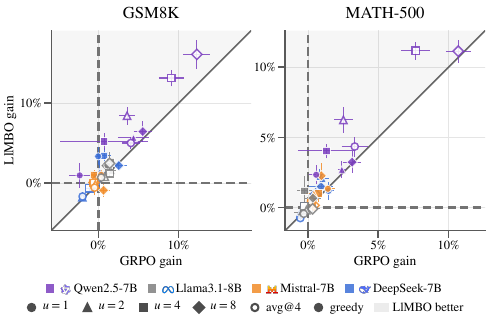}
\caption{\llmbo{} against GRPO, evaluated both in terms of mean performance at temperature 1 on 4 trials (avg@4, open points)
and via greedy decoding (filled).
Gain over the base model (mean $\pm$ s.e. over seeds) for runs with $u\leq8$ trainable parameters (marker) and model family (color).
}
\label{fig:headline}
\end{figure}
Reinforcement learning with verifiable rewards has become the standard recipe for teaching language models to reason. In this setting, we sample solutions to problems with checkable answers, reward the correct ones, and follow the policy gradient \citep{shao2024deepseekmath,zeng2025simplerl}.
Recent work shows how little of the network this recipe has to change:
TinyLoRA \citep{morris2026learning} expresses each weight update in a random subspace of the weight's own top singular vectors and ties the coefficients across all layers; with only a handful of trainable parameters, even down to a single one, it retains much of the gain of full fine-tuning.

This ultra-low parameter scale opens the fine-tuning objective to optimizers other than gradient-based reinforcement learning.
An optimization problem with a handful of variables and an expensive, noisy, black-box objective is the textbook setting for Bayesian optimization \citep[BO,][]{garnett_bayesoptbook_2023}.
Seen from that perspective, a policy gradient is a heavy tool for TinyLoRA. To update just a few scalars, Group Relative Policy Optimization (GRPO) rescores every rollout, backpropagates through billions of weights and holds the optimizer state.
Further, it also constrains the objective.
GRPO requires sampled rollouts, so it optimizes accuracy at the sampling temperature, while the deployed model is often decoded greedily.

We introduce Large language Model Bayesian Optimization (\llmbo{}), replacing GRPO with trust-region BO \citep[TuRBO,][]{eriksson2019turbo} over the adapter parameters.
Each candidate adapter is loaded into an inference engine and scored by the fraction of a random batch of training problems it solves. A Gaussian process (GP) on the logit of that pass rate proposes the subsequent batch of candidates. 
\llmbo{} is thus 1. \textit{sample efficient}, since the GP pools every evaluation, 2. \textit{gradient-free and memory-light}, since scoring is inference only, 3. \textit{parallelizable} across devices with minimal synchronization, since a candidate is $u$ independent numbers and its score is a pass count, and 4. free in its decoding rule, so that, unlike a policy gradient, it can optimize greedy-decoding accuracy directly.

We compare \llmbo{} against GRPO on the same TinyLoRA adapter configurations, with the same SVD bases and random projections seed for seed, the same training data (Hendrycks MATH \citep{hendrycks2021math} and GSM8K \citep{cobbe2021gsm8k}) and a matched number of generated completions. We trace how each approach scales as the adapter shrinks to a single parameter with LoRA and LoRA-XS under GRPO as references at larger scales. The comparison covers \nModels{} models of the Qwen2.5, Llama 3, Mistral and DeepSeek families, in base, instruct and math variants from 1.5B to 8B parameters, comprising \nRuns{} training runs in all, evaluated on 6 math reasoning benchmarks under a common harness.
Our contributions are:
\begin{itemize}
    \item \llmbo{}, a gradient-free fine-tuning method for ultra-low-parameter adapters, with a moment-matched logit observation model for pass rates, a dimension-scaled lengthscale prior inside the trust region, and a search box set by how far GRPO could move each parameter (Section~\ref{sec:method}).
    \item Evidence that \llmbo{} matches or exceeds GRPO's fine-tuning gains at $u \le 4$ trainable parameters on GRPO's own objective and metric. This is reported per model group, with one 7-8B base model per family, the math-specialized models and the Qwen2.5 1.5B models (Section~\ref{sec:results}, Appendix~\ref{app:results}; the Instruct models in Appendix~\ref{app:instruct}). \llmbo{} leads GRPO on the Qwen2.5 7B base and Math models, is level elsewhere, leads pooled over the 7-8B models, and keeps its lead when the models train on MATH alone.
    \item Training directly on greedy-decoding accuracy, which policy gradient cannot achieve (Section~\ref{sec:greedy-objective}). Trained and evaluated greedily, \llmbo{} leads GRPO on greedy GSM8K at every $u \le 4$ in every model group and is level or ahead on MATH-500. This is most clear on the math-specialized models. Seed for seed, training on greedy accuracy rather than on the pass rate at $T = 1$ raises greedy accuracy, a benefit of the free choice of objective afforded \llmbo{}.
    \item Computational efficiency, as \llmbo{} utilizes a fraction of GRPO's compute and memory (Section~\ref{sec:efficiency}). \llmbo{} spends about \boComputeShare{}\% of GRPO's training FLOPs; trained on greedy accuracy, it passes GRPO's final greedy accuracy after \reachPoolGreedyGsm{} completions on average over the 7-8B models, where GRPO spends \grpoCompletions{}. Further, it needs no memory beyond inference, so on a 24\,GB card it keeps \boSequencesSmallCard{} sequences in memory on all 7-8B architectures, while GRPO cannot fit even its \grpoMicroBatch{}-sequence micro-batch on \grpoNoFitSmallCard{} of them.
\end{itemize}
The anonymized code of \llmbo{}, the GRPO baseline, and the evaluation are in the Supplementary Material.

\section{Related Literature}\label{sec:related}

\begin{figure*}[t]
\centering
\includegraphics[width=\linewidth]{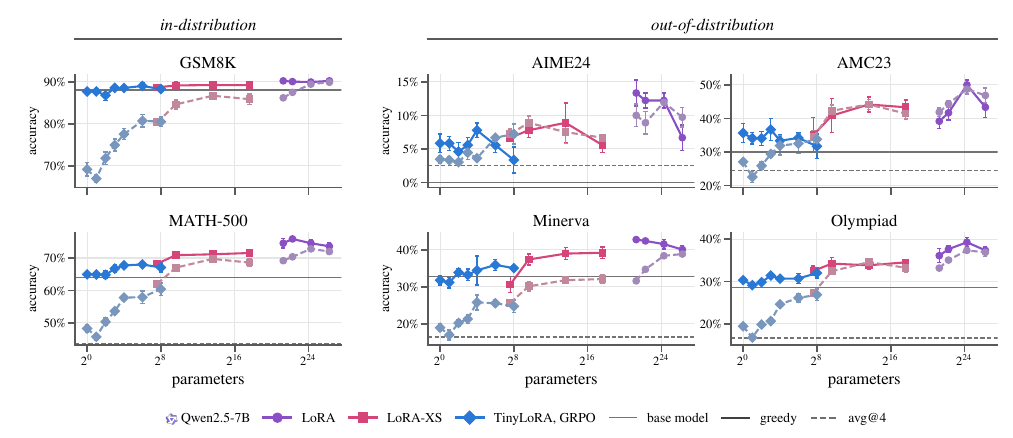}
\caption{\textbf{TinyLoRA scaling laws under GRPO.} Greedy (solid) and avg@4 (dashed) accuracy at the end of training under GRPO (mean $\pm$ s.e.\ over seeds) for TinyLoRA, LoRA-XS and LoRA as a function of trainable parameters. Results shown for Qwen2.5-7B, scaling for other model families can be found in Figure~\ref{fig:scaling-base}.}
\label{fig:scaling-qwen}
\end{figure*}

LoRA \citep[Low-Rank Adaptation,][]{hu2022lora} trains a low-rank update $BA$ per weight matrix. LoRA-XS \citep{balazy2024loraxs} freezes $A$ and $B$ to the weight's truncated singular vectors and trains only an $r \times r$ matrix between them. TinyLoRA \citep{morris2026learning} expresses that matrix as a combination of $u$ many fixed random matrices and ties the $u$ coefficients across layers, thus allowing for a single trainable parameter. Under GRPO it retains most of the reasoning gain of full fine-tuning on Qwen2.5 models (see, e.g., Figure~\ref{fig:scaling-qwen}). The success of such small updates echoes the low intrinsic dimension of fine-tuning \citep{aghajanyan2021intrinsic}. We utilize the TinyLoRA's update, simplify its weight tying to a single scaling axis (Appendix~\ref{app:tying}), and replace its optimizer.

GRPO \citep{shao2024deepseekmath} estimates advantages by normalizing rewards within a group of rollouts for the same prompt, removing PPO's value network \citep{schulman2017proximal}. SimpleRL-Zoo \citep{zeng2025simplerl} applies GRPO to base models of several families. We use training data tiers and prompts of SimpleRL-Zoo.

GP-based BO \citep{williams2006gaussian,garnett_bayesoptbook_2023} is sample efficient in low dimensions but its global surrogate degrades as the dimension grows. TuRBO \citep{eriksson2019turbo} restricts the search to a trust region around the incumbent that expands on success and shrinks on failure. \citet{hvarfner2024vanilla} show that scaling the lengthscale prior with $\sqrt{D}$ lets plain GP BO work in high dimensions. Utilizing BoTorch \citep{balandat2020botorch}, we leverage both. Our observation model, a Gaussian moment-matched to the logit of a Beta posterior, follows the same framework as Dirichlet-based GP classification \citep{milios2018dirichlet}.

MeZO \citep[memory-efficient zeroth-order optimizer,][]{malladi2023mezo} fine-tunes models with zeroth-order SPSA gradient estimates from forward passes only. Evolution strategies have also been scaled to full-parameter reasoning fine-tuning \citep{qiu2025evolution}. Both still take many small steps in millions to billions of dimensions; \llmbo{} instead exploits a search space small enough for a global surrogate model.
\section{Background}\label{sec:background}
\subsection{Gaussian-process Bayesian Optimization}
The main goal in Bayesian optimization (BO) is to optimize an unknown function $f$ via only its evaluations: 
\begin{equation}
    \max_x f(x).
\end{equation}
To this end, BO provides a way to select locations at which to evaluate $f$. 
It is, thus, useful to envision $f$ as being expensive to evaluate so that grid searches become infeasible. Given a set of locations $X=\{x_1,x_2,\dots,x_T\}$ and evaluations $Y=\{f(x_1),f(x_2),\dots,f(x_T)\}$, BO provides a way to select the best next evaluation location. 
In BO, we typically assume that $f$ can be modeled by a Gaussian Process (GP) as $g \sim \mathcal{GP}(0, k)$ where  $k$ is the covariance function and conditioning on noisy observations $y_n = g(\theta_n) + \varepsilon_n$, $\varepsilon_n \sim \mcN(0, \sigma_n^2)$.  With $K = [k(\theta_n, \theta_m)]_{nm}$, $\Lambda = \mathrm{diag}(\sigma_n^2)$ and $\mathbf{k}(\theta) = [k(\theta, \theta_n)]_n$, the posterior is also a GP with \citep{williams2006gaussian}
\begin{align}
    \mu(\theta) &= \mathbf{k}(\theta)^\top (K + \Lambda)^{-1} \mathbf{y}, \nonumber \\
    k'(\theta, \theta') &= k(\theta, \theta') - \mathbf{k}(\theta)^\top (K + \Lambda)^{-1} \mathbf{k}(\theta').
\end{align}

\subsection{LoRA and low-parameter variants}
For a frozen weight $W \in \mbR^{d \times k}$, LoRA \citep{hu2022lora} trains a low-rank update,
\begin{equation}
    W' = W + \frac{\alpha}{r} B A, \qquad B \in \mbR^{d \times r},\ A \in \mbR^{r \times k},
    \label{eq:lora}
\end{equation}
with scale $\alpha$, $A$ initialized at random and $B = 0$, so that training starts from the base model. It trains $r(d + k)$ parameters per weight matrix, which is significantly lower than full fine tuning, achieving comparable performance.
However, for a 7B model this means millions of trainable parameters, even at $r = 1$.

LoRA-XS \citep{balazy2024loraxs} instead takes the rank-$r$ SVD $W \approx U \Sigma V^\top$, freezes both factors and trains only a small matrix $R$ between them,
\begin{equation}
    W' = W + U \Sigma R V^\top, \qquad R \in \mbR^{r \times r},
    \label{eq:loraxs}
\end{equation}
which contributes $r^2$ trainable parameters per weight matrix, independent of $d$ and $k$.
This significantly reduces the trainable parameters (a few thousand at the lowest) while retaining much of LoRA's performance.

TinyLoRA \citep{morris2026learning} goes further and expresses $R$ as a combination of fixed random matrices,
\begin{equation}
    W' = W + U \Sigma \Big( \sum_{i=1}^{u} v_i P_i \Big) V^\top,
    \label{eq:tinylora}
\end{equation}
with frozen random matrix $P_i \in \mbR^{r \times r}$, $P_i \sim \mcN(0, 1)$ entrywise, drawn independently for every module, and a single coefficient vector $\theta = v \in \mbR^u$ shared by every attention and MLP projection of the network (full weight tying; Appendix~\ref{app:tying}). The adapter has $u$ trainable parameters, and $\theta = 0$ is the base model. We use $r = 2$ throughout.

\subsection{GRPO}
For each training prompt $q$, GRPO samples a group of $G$ completions $o_1, \dots, o_G$ from the current policy at temperature $T = 1$, scores each with a binary correctness reward $R_j$, and uses the group-normalized advantage 
\begin{equation}
    A_j = \frac{R_j - \mathrm{mean}(R_j)}{\mathrm{std}(R_j)}
    \label{eq:advantage}
\end{equation}
in a clipped importance-weighted objective,
\begin{equation*}
    \mathcal{J}(\theta) = \mbE_{q, \{o_j\}} \frac{1}{G} \sum_{j=1}^{G} \frac{1}{|o_j|} \sum_{t=1}^{|o_j|} \min\!\big( \rho_{j,t} A_j,\ \hat\rho_{j,t} A_j \big),
\end{equation*}
where $\rho_{j,t} = \pi_\theta(o_{j,t} \mid q, o_{j,<t}) / \pi_{\theta_\mathrm{old}}(o_{j,t} \mid q, o_{j,<t})$ and $\hat\rho_{j,t} = \mathrm{clip}(\rho_{j,t}, 1 - \epsilon, 1 + \epsilon)$ for $\epsilon>0$.
Each step needs the rollouts, a forward pass of the policy for the old log-probabilities, and a bidirectional pass for the gradient.
\section{Proposed Method}
\label{sec:method}

\subsection{The Black-box}

Our goal is to maximize the adapted model's pass rate, however defined, on the training distribution:
\begin{equation}
    f(\theta) = \Pr_{q \sim \mcD,\ o \sim \pi_\theta(\cdot \mid q)}\big[ o \text{ answers } q \text{ correctly} \big],
\end{equation}

where $o$ is either one completion sampled at $T = 1$ (the rollout distribution on which GRPO trains) or the greedy completion (the decoding rule of deployment), and $\mcD$ represents the training data.

One completion per prompt is the most informative use of a fixed budget: completions of the same prompt are correlated, so $G$ of them carry the information of fewer than $G$ independent ones (Appendix~\ref{sec:one-per-prompt}).
In our numerical study, \llmbo{} therefore scores each candidate on one completion per prompt, while GRPO needs a group of $G \ge 2$ completions per prompt to estimate advantages and uses $G = 4$.
Since a policy gradient needs sampled rollouts, GRPO cannot optimize the (piecewise-constant in $\theta$) greedy objective and is always trained on the pass rate at $T = 1$, whereas \llmbo{} is trained on the objective that matches the metric it is evaluated on.

\begin{figure*}[t]
\centering
\includegraphics{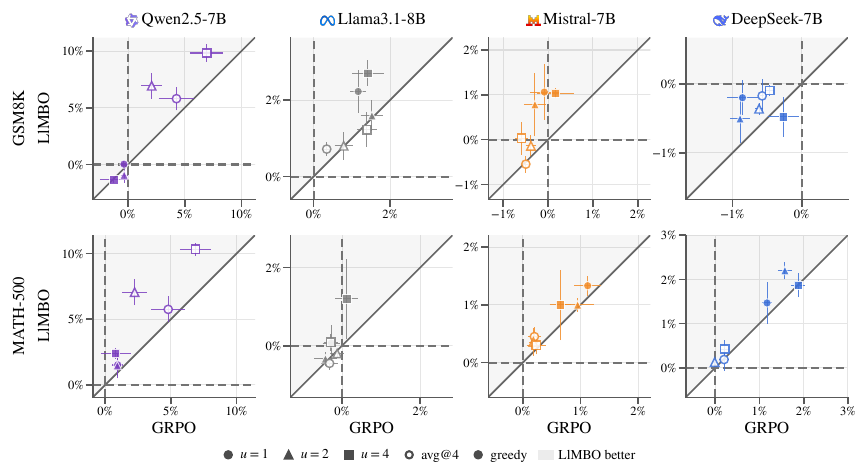}
\caption{Gain over the base model of \llmbo{} against that of GRPO on the in-distribution benchmarks, for the four vanilla base models; avg@4 (open) and greedy accuracy (filled), one point per $u \le 4$ (marker), mean $\pm$ s.e.\ over seeds. \llmbo{} ($b = 1$) is trained on the metric of each point; points in the shaded half favor \llmbo{}. Every model and benchmark in Figure~\ref{fig:scatter-main}.}
\label{fig:scatter-bases}
\end{figure*}

The training objective for a candidate $\theta$ involves returning the number of solutions $s$ out of $n$ attempted prompts, which may be profitably conceived as binomial distribution.
However, inference on stochastic processes with a binomial outcome is considerably harder than Gaussian process inference, which motivates a Gaussian approximation.
But for some models on some benchmarks, pass rates will lie near 0 or 1, making the standard Gaussian approximation poor for smaller $n$.
We therefore deploy at the batch-level a Bayesian Uniform-prior Binomial analysis to estimate the mean and variance associated with a batch.
Under such an analysis, the batch-level posterior takes form $\textrm{Beta}(\alpha,\beta)$ for $\alpha = 1+s$ and $\beta = 1+n-s$.
We transform this posterior via the logit function to achieve a real-valued quantity suitable for Gaussian process interpolation.
The posterior mean and variance in logit space are given by:
\begin{equation}
    \hat\mu = \psi(\alpha) - \psi(\beta), \qquad \hat\sigma^2 = \psi_1(\alpha) + \psi_1(\beta),
\end{equation}
with $\psi$ the digamma and $\psi_1$ the trigamma function.
We add a learned noise floor to the GP for prompt-to-prompt variation not captured by this intra-prompt standard error.
Appendix~\ref{app:obsmodel} compares this scheme with a classical Gaussian approximation.

The $b$ candidates of a batch (one by default) are scored in one call of a vLLM engine \citep{kwon2023vllm}, each attached as its own LoRA adapter.
Subsequently, $\hat\mu$ is then treated as the black-box response $y$ and $\sqrt{\hat\sigma^2}$ as its standard error, both of which are added to the GP training data and the mdoel refit.
The GP runs on CPU while the GPU only runs the inference engine.
A candidate $\theta$ consists merely of $u$ floating point numbers and its integer pass count, so the prompts of a batch, and for $b > 1$ its candidates, can be spread over devices with only these scalars exchanged.

\subsection{BO Strategy}

We fit a GP using a Mat\'ern 5/2 kernel with one lengthscale per dimension on inputs rescaled to the unit cube and standardized outputs following BO best practices \citep{snoek2012practical}.
Following \citet{hvarfner2024vanilla}, we endow the lengthscales with a Lognormal prior $\mathrm{LogNormal}(\sqrt{2} + \log(L\sqrt{D}), \sqrt{3})$ with $D = u$ and $L$ defined in next paragraph.
Hyperparameters are fit by maximization \textit{a posteriori}.

Our acquisition strategy is based on TuRBO-1 \citep{eriksson2019turbo}, which involves searching within a \textit{trust region}.
A trust region is a subset of the search space given by a box centered at the current design point with the highest posterior mean, with each side length given by a common constant $L$ subsequently scaled by that dimension's lengthscale in the Mat\'ern kernel.
After each observation from the black box, we perform the following iteration.
First, we refit the GP, which also updates the size of the trust region.
Then, to form each element of the batch, we perform Thompson sampling over a Sobol set of candidate locations inside the trust region and selects the maximing argument for inclusion in the batch.

Following \citet{eriksson2019turbo}, we update the common trust region length $L$ on the basis of success of the search: if the most recent observation improved on the incumbent value, that is labeled a success, and if not, a failure. 
Three consecutive successes doubles $L$, $\lceil \max(4, D)/b \rceil$ consecutive failures halve it, and when $L$ falls below $0.5^7$ the region restarts from a fresh Sobol design.
However, \citet{eriksson2019turbo} was developed for noiseless simulators, and in our noisy setting the value of the most recent observation is insufficient for determining success or failure.
We therefore instead define a success to be a batch whose posterior mean improves on the posterior mean of previously evaluated design points.
We also use  the trust-region side $L$ in the lognormal prior such that it matches the region that the GP models. 

Finally, we must define our search space over $\theta$ values. 
We do this by considering the effective search space of a policy gradient method, which only a neighborhood of $\theta = 0$ in its budget.
With Adam, each coordinate moves by at most approximately the learning rate $\eta$ per step, so after $S$ steps $|\theta_i| \lesssim \eta S$.
We therefore use the box $[-\eta S, \eta S]^u$ with GRPO's $\eta$ and $S$ as the bound for BO, making the effective search regions commensurable.

\section{Experimental Results}
\subsection{Experimental setup}\label{sec:experiments}
\begin{table*}[t]
\centering
\caption{Paired gap \llmbo{} $-$ GRPO (points) per model (rows) at $u = 1, 2, 4$ and benchmark (columns), mean over seed pairs; green: \llmbo{} ahead. avg@4: accuracy averaged over four completions at $T{=}1$; greedy: accuracy of the greedy completion (Section~\ref{sec:greedy-objective}); \llmbo{} is trained on the metric of each column. One row block per $u$, each listing the four vanilla base models and the two math-specialized ones, each group closed by its \emph{Avg} row. Standard errors and $u \le 16$ in Table~\ref{tab:main}; the Qwen2.5 1.5B and Instruct models in Appendices~\ref{app:small} and~\ref{app:instruct}.}
\label{tab:headline}\vspace{0.3em}
\footnotesize\renewcommand{\arraystretch}{0.9}\setlength{\tabcolsep}{2.5pt}
\begin{tabular}{l l cccc@{\hspace{1.2em}}cccccccc}
\toprule
 & & \multicolumn{4}{c}{in-distribution} & \multicolumn{8}{c}{out-of-distribution} \\
\cmidrule(lr){3-6}\cmidrule(lr){7-14}
 & & \multicolumn{2}{c}{GSM8K} & \multicolumn{2}{c}{MATH-500} & \multicolumn{2}{c}{AIME24} & \multicolumn{2}{c}{AMC23} & \multicolumn{2}{c}{Minerva} & \multicolumn{2}{c}{Olympiad} \\
\cmidrule(lr){3-4}\cmidrule(lr){5-6}\cmidrule(lr){7-8}\cmidrule(lr){9-10}\cmidrule(lr){11-12}\cmidrule(lr){13-14}
$u$ & Model & avg@4 & greedy & avg@4 & greedy & avg@4 & greedy & avg@4 & greedy & avg@4 & greedy & avg@4 & greedy \\
\midrule
\multirow{8}{*}{1} & \raisebox{-0.15em}{\includegraphics[height=0.9em]{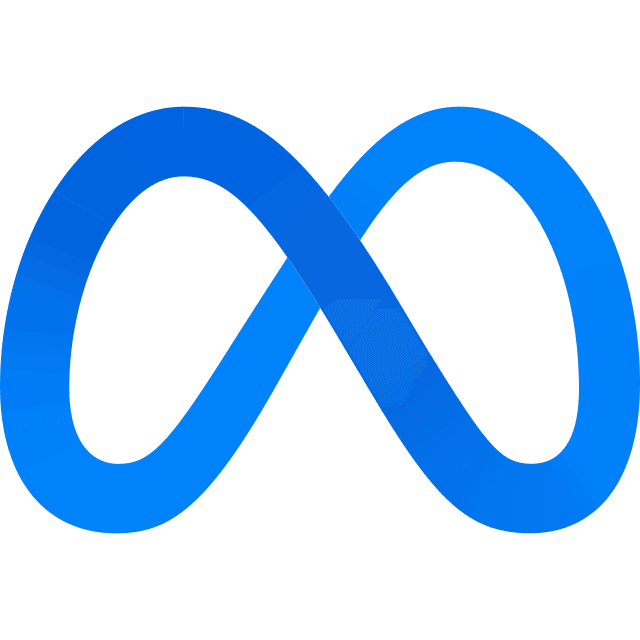}}~Llama3.1-8B & \gapdot{green!5}{$+0.4$} & \gapdot{green!16}{$+1.1$} & \gapdot{red!2}{$-0.1$} & \gapdot{green!1}{$+0.1$} & \gapdot{red!1}{$-0.1$} & \gapdot{none}{$+0.0$} & \gapdot{green!9}{$+0.6$} & \gapdot{green!28}{$+2.5$} & \gapdot{red!3}{$-0.2$} & \gapdot{green!7}{$+0.5$} & \gapdot{none}{$+0.0$} & \gapdot{green!3}{$+0.2$} \\
 & \raisebox{-0.15em}{\includegraphics[height=0.9em]{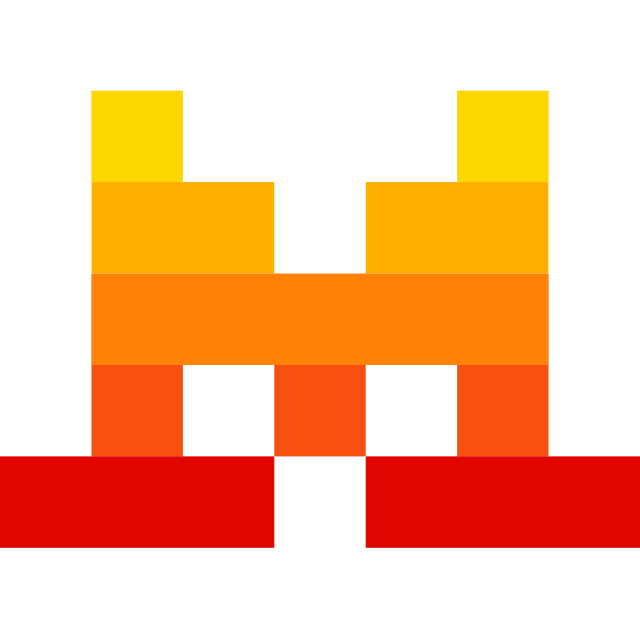}}~Mistral-7B & \gapdot{red!1}{$+0.0$} & \gapdot{green!10}{$+0.7$} & \gapdot{green!4}{$+0.3$} & \gapdot{red!1}{$-0.1$} & \gapdot{red!1}{$-0.1$} & \gapdot{none}{$+0.0$} & \gapdot{red!1}{$-0.1$} & \gapdot{red!12}{$-0.8$} & \gapdot{green!2}{$+0.1$} & \gapdot{red!2}{$-0.1$} & \gapdot{green!2}{$+0.1$} & \gapdot{red!1}{$+0.0$} \\
 & \raisebox{-0.15em}{\includegraphics[height=0.9em]{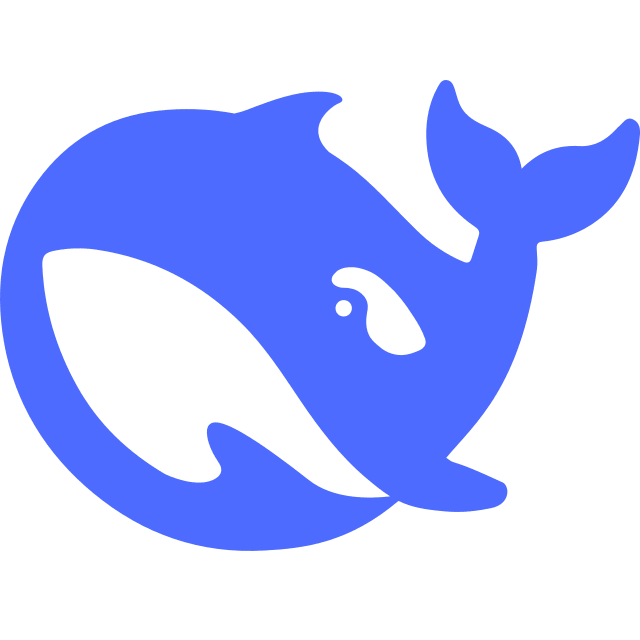}}~DeepSeek-7B & \gapdot{green!6}{$+0.4$} & \gapdot{green!16}{$+1.2$} & \gapdot{none}{$+0.0$} & \gapdot{green!8}{$+0.6$} & \gapdot{none}{$+0.0$} & \gapdot{none}{$+0.0$} & \gapdot{green!20}{$+1.5$} & \gapdot{none}{$+0.0$} & \gapdot{none}{$+0.0$} & \gapdot{green!3}{$+0.2$} & \gapdot{green!5}{$+0.4$} & \gapdot{green!1}{$+0.1$} \\
 & \raisebox{-0.15em}{\includegraphics[height=0.9em]{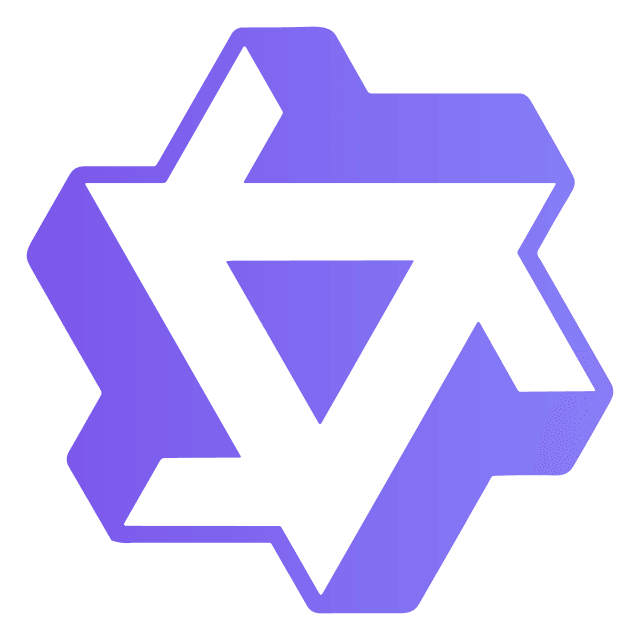}}~Qwen2.5-7B & \gapdot{green!21}{$+1.5$} & \gapdot{green!11}{$+0.8$} & \gapdot{green!13}{$+1.0$} & \gapdot{green!2}{$+0.1$} & \gapdot{green!7}{$+0.5$} & \gapdot{red!28}{$-4.4$} & \gapdot{green!17}{$+1.2$} & \gapdot{green!28}{$+5.8$} & \gapdot{green!12}{$+0.9$} & \gapdot{green!9}{$+0.6$} & \gapdot{red!2}{$-0.1$} & \gapdot{green!14}{$+1.0$} \\
\addlinespace[0.25em]
 & \emph{Avg} & \gapdot{green!8}{$+0.6$} & \gapdot{green!13}{$+0.9$} & \gapdot{green!5}{$+0.3$} & \gapdot{green!2}{$+0.1$} & \gapdot{green!1}{$+0.1$} & \gapdot{red!17}{$-1.2$} & \gapdot{green!10}{$+0.7$} & \gapdot{green!28}{$+2.0$} & \gapdot{green!3}{$+0.2$} & \gapdot{green!4}{$+0.3$} & \gapdot{green!1}{$+0.0$} & \gapdot{green!5}{$+0.3$} \\
\cmidrule(l){2-14}
 & \raisebox{-0.15em}{\includegraphics[height=0.9em]{logos/deepseek.png}}~DeepSeek-Math-7B & \gapdot{green!3}{$+0.2$} & \gapdot{green!28}{$+5.6$} & \gapdot{red!6}{$-0.4$} & \gapdot{red!16}{$-1.1$} & \gapdot{red!1}{$-0.1$} & \gapdot{green!16}{$+1.1$} & \gapdot{red!7}{$-0.5$} & \gapdot{green!28}{$+5.0$} & \gapdot{red!10}{$-0.7$} & \gapdot{green!3}{$+0.2$} & \gapdot{green!1}{$+0.1$} & \gapdot{green!8}{$+0.5$} \\
 & \raisebox{-0.15em}{\includegraphics[height=0.9em]{logos/qwen.png}}~Qwen2.5-Math-7B & \gapdot{green!6}{$+0.4$} & \gapdot{green!28}{$+5.9$} & \gapdot{green!15}{$+1.1$} & \gapdot{green!28}{$+3.4$} & \gapdot{red!28}{$-2.6$} & \gapdot{red!16}{$-1.1$} & \gapdot{red!28}{$-3.0$} & \gapdot{red!12}{$-0.8$} & \gapdot{red!1}{$+0.0$} & \gapdot{green!28}{$+2.3$} & \gapdot{green!7}{$+0.5$} & \gapdot{green!28}{$+5.5$} \\
\addlinespace[0.25em]
 & \emph{Avg} & \gapdot{green!5}{$+0.3$} & \gapdot{green!28}{$+5.8$} & \gapdot{green!5}{$+0.3$} & \gapdot{green!16}{$+1.1$} & \gapdot{red!19}{$-1.4$} & \gapdot{none}{$+0.0$} & \gapdot{red!24}{$-1.7$} & \gapdot{green!28}{$+2.1$} & \gapdot{red!5}{$-0.4$} & \gapdot{green!18}{$+1.3$} & \gapdot{green!4}{$+0.3$} & \gapdot{green!28}{$+3.0$} \\
\addlinespace[0.6em]\midrule[\heavyrulewidth]
\multirow{8}{*}{2} & \raisebox{-0.15em}{\includegraphics[height=0.9em]{logos/llama.png}}~Llama3.1-8B & \gapdot{none}{$+0.0$} & \gapdot{red!1}{$-0.1$} & \gapdot{red!1}{$-0.1$} & \gapdot{green!3}{$+0.2$} & \gapdot{green!3}{$+0.2$} & \gapdot{none}{$+0.0$} & \gapdot{green!8}{$+0.5$} & \gapdot{green!28}{$+2.5$} & \gapdot{green!2}{$+0.1$} & \gapdot{red!3}{$-0.2$} & \gapdot{green!1}{$+0.0$} & \gapdot{red!1}{$-0.1$} \\
 & \raisebox{-0.15em}{\includegraphics[height=0.9em]{logos/mistral.png}}~Mistral-7B & \gapdot{green!3}{$+0.2$} & \gapdot{green!11}{$+0.8$} & \gapdot{green!2}{$+0.2$} & \gapdot{green!8}{$+0.6$} & \gapdot{green!3}{$+0.2$} & \gapdot{none}{$+0.0$} & \gapdot{green!7}{$+0.5$} & \gapdot{green!23}{$+1.7$} & \gapdot{green!6}{$+0.4$} & \gapdot{green!2}{$+0.1$} & \gapdot{red!1}{$+0.0$} & \gapdot{red!3}{$-0.2$} \\
 & \raisebox{-0.15em}{\includegraphics[height=0.9em]{logos/deepseek.png}}~DeepSeek-7B & \gapdot{green!3}{$+0.2$} & \gapdot{green!8}{$+0.6$} & \gapdot{green!2}{$+0.2$} & \gapdot{green!15}{$+1.1$} & \gapdot{green!2}{$+0.1$} & \gapdot{none}{$+0.0$} & \gapdot{red!1}{$-0.1$} & \gapdot{none}{$+0.0$} & \gapdot{none}{$+0.0$} & \gapdot{green!2}{$+0.1$} & \gapdot{green!1}{$+0.1$} & \gapdot{none}{$+0.0$} \\
 & \raisebox{-0.15em}{\includegraphics[height=0.9em]{logos/qwen.png}}~Qwen2.5-7B & \gapdot{green!28}{$+4.9$} & \gapdot{red!15}{$-1.0$} & \gapdot{green!28}{$+4.8$} & \gapdot{green!16}{$+1.1$} & \gapdot{green!4}{$+0.3$} & \gapdot{green!28}{$+5.6$} & \gapdot{green!28}{$+4.6$} & \gapdot{green!23}{$+1.7$} & \gapdot{green!28}{$+2.8$} & \gapdot{red!28}{$-3.3$} & \gapdot{green!28}{$+3.6$} & \gapdot{green!28}{$+2.5$} \\
\addlinespace[0.25em]
 & \emph{Avg} & \gapdot{green!19}{$+1.4$} & \gapdot{green!1}{$+0.1$} & \gapdot{green!18}{$+1.3$} & \gapdot{green!11}{$+0.8$} & \gapdot{green!3}{$+0.2$} & \gapdot{green!19}{$+1.4$} & \gapdot{green!20}{$+1.4$} & \gapdot{green!20}{$+1.5$} & \gapdot{green!12}{$+0.9$} & \gapdot{red!12}{$-0.8$} & \gapdot{green!13}{$+0.9$} & \gapdot{green!7}{$+0.5$} \\
\cmidrule(l){2-14}
 & \raisebox{-0.15em}{\includegraphics[height=0.9em]{logos/deepseek.png}}~DeepSeek-Math-7B & \gapdot{green!4}{$+0.3$} & \gapdot{green!27}{$+1.9$} & \gapdot{green!2}{$+0.1$} & \gapdot{green!13}{$+0.9$} & \gapdot{none}{$+0.0$} & \gapdot{red!16}{$-1.1$} & \gapdot{green!1}{$+0.1$} & \gapdot{green!28}{$+4.2$} & \gapdot{green!5}{$+0.3$} & \gapdot{red!7}{$-0.5$} & \gapdot{none}{$+0.0$} & \gapdot{green!7}{$+0.5$} \\
 & \raisebox{-0.15em}{\includegraphics[height=0.9em]{logos/qwen.png}}~Qwen2.5-Math-7B & \gapdot{green!28}{$+4.9$} & \gapdot{green!28}{$+3.7$} & \gapdot{green!28}{$+2.7$} & \gapdot{red!7}{$-0.5$} & \gapdot{red!6}{$-0.4$} & \gapdot{red!28}{$-2.2$} & \gapdot{green!28}{$+3.7$} & \gapdot{green!28}{$+3.3$} & \gapdot{green!8}{$+0.6$} & \gapdot{green!24}{$+1.7$} & \gapdot{green!21}{$+1.5$} & \gapdot{green!9}{$+0.6$} \\
\addlinespace[0.25em]
 & \emph{Avg} & \gapdot{green!28}{$+2.6$} & \gapdot{green!28}{$+2.8$} & \gapdot{green!20}{$+1.4$} & \gapdot{green!3}{$+0.2$} & \gapdot{red!3}{$-0.2$} & \gapdot{red!23}{$-1.7$} & \gapdot{green!26}{$+1.9$} & \gapdot{green!28}{$+3.8$} & \gapdot{green!6}{$+0.5$} & \gapdot{green!9}{$+0.6$} & \gapdot{green!11}{$+0.8$} & \gapdot{green!8}{$+0.6$} \\
\addlinespace[0.6em]\midrule[\heavyrulewidth]
\multirow{8}{*}{4} & \raisebox{-0.15em}{\includegraphics[height=0.9em]{logos/llama.png}}~Llama3.1-8B & \gapdot{red!2}{$-0.2$} & \gapdot{green!17}{$+1.2$} & \gapdot{green!5}{$+0.4$} & \gapdot{green!21}{$+1.5$} & \gapdot{green!1}{$+0.1$} & \gapdot{none}{$+0.0$} & \gapdot{green!1}{$+0.1$} & \gapdot{none}{$+0.0$} & \gapdot{green!4}{$+0.3$} & \gapdot{none}{$+0.0$} & \gapdot{red!1}{$-0.1$} & \gapdot{red!1}{$+0.0$} \\
 & \raisebox{-0.15em}{\includegraphics[height=0.9em]{logos/mistral.png}}~Mistral-7B & \gapdot{green!9}{$+0.6$} & \gapdot{green!22}{$+1.6$} & \gapdot{green!1}{$+0.1$} & \gapdot{green!4}{$+0.3$} & \gapdot{green!1}{$+0.1$} & \gapdot{none}{$+0.0$} & \gapdot{green!2}{$+0.2$} & \gapdot{none}{$+0.0$} & \gapdot{red!2}{$-0.1$} & \gapdot{green!2}{$+0.1$} & \gapdot{red!1}{$+0.0$} & \gapdot{red!2}{$-0.1$} \\
 & \raisebox{-0.15em}{\includegraphics[height=0.9em]{logos/deepseek.png}}~DeepSeek-7B & \gapdot{green!5}{$+0.4$} & \gapdot{none}{$+0.0$} & \gapdot{green!3}{$+0.2$} & \gapdot{green!4}{$+0.3$} & \gapdot{green!3}{$+0.2$} & \gapdot{none}{$+0.0$} & \gapdot{green!19}{$+1.3$} & \gapdot{none}{$+0.0$} & \gapdot{none}{$+0.0$} & \gapdot{green!2}{$+0.1$} & \gapdot{green!5}{$+0.4$} & \gapdot{red!1}{$-0.1$} \\
 & \raisebox{-0.15em}{\includegraphics[height=0.9em]{logos/qwen.png}}~Qwen2.5-7B & \gapdot{green!28}{$+2.9$} & \gapdot{red!8}{$-0.6$} & \gapdot{green!28}{$+3.5$} & \gapdot{green!28}{$+3.2$} & \gapdot{green!7}{$+0.5$} & \gapdot{red!16}{$-1.1$} & \gapdot{green!28}{$+3.4$} & \gapdot{red!28}{$-2.5$} & \gapdot{green!3}{$+0.2$} & \gapdot{green!2}{$+0.1$} & \gapdot{green!23}{$+1.6$} & \gapdot{green!7}{$+0.5$} \\
\addlinespace[0.25em]
 & \emph{Avg} & \gapdot{green!13}{$+0.9$} & \gapdot{green!8}{$+0.5$} & \gapdot{green!14}{$+1.0$} & \gapdot{green!18}{$+1.3$} & \gapdot{green!3}{$+0.2$} & \gapdot{red!4}{$-0.3$} & \gapdot{green!17}{$+1.2$} & \gapdot{red!9}{$-0.6$} & \gapdot{green!1}{$+0.1$} & \gapdot{green!1}{$+0.1$} & \gapdot{green!7}{$+0.5$} & \gapdot{green!1}{$+0.0$} \\
\cmidrule(l){2-14}
 & \raisebox{-0.15em}{\includegraphics[height=0.9em]{logos/deepseek.png}}~DeepSeek-Math-7B & \gapdot{green!10}{$+0.7$} & \gapdot{green!28}{$+5.5$} & \gapdot{green!7}{$+0.5$} & \gapdot{green!28}{$+2.5$} & \gapdot{red!1}{$-0.1$} & \gapdot{green!28}{$+2.2$} & \gapdot{green!3}{$+0.2$} & \gapdot{none}{$+0.0$} & \gapdot{green!1}{$+0.1$} & \gapdot{red!12}{$-0.9$} & \gapdot{red!2}{$-0.1$} & \gapdot{green!23}{$+1.6$} \\
 & \raisebox{-0.15em}{\includegraphics[height=0.9em]{logos/qwen.png}}~Qwen2.5-Math-7B & \gapdot{green!28}{$+5.2$} & \gapdot{green!28}{$+9.7$} & \gapdot{green!28}{$+3.6$} & \gapdot{green!28}{$+2.3$} & \gapdot{red!10}{$-0.7$} & \gapdot{green!16}{$+1.1$} & \gapdot{green!28}{$+2.9$} & \gapdot{green!28}{$+5.0$} & \gapdot{green!27}{$+1.9$} & \gapdot{green!28}{$+2.6$} & \gapdot{green!25}{$+1.8$} & \gapdot{green!28}{$+5.3$} \\
\addlinespace[0.25em]
 & \emph{Avg} & \gapdot{green!28}{$+3.0$} & \gapdot{green!28}{$+7.6$} & \gapdot{green!28}{$+2.0$} & \gapdot{green!28}{$+2.4$} & \gapdot{red!6}{$-0.4$} & \gapdot{green!23}{$+1.7$} & \gapdot{green!22}{$+1.6$} & \gapdot{green!28}{$+2.5$} & \gapdot{green!14}{$+1.0$} & \gapdot{green!12}{$+0.9$} & \gapdot{green!12}{$+0.8$} & \gapdot{green!28}{$+3.5$} \\
\bottomrule
\end{tabular}

\vspace{6pt}
\end{table*}

We fine-tune \nModels{} models from four families: Qwen2.5 1.5B and 7B \citep{qwen2025qwen25} in their base, Math \citep{yang2024qwen25math} and Instruct variants, Llama3.1-8B \citep{dubey2024llama3}, Mistral-7B \citep{jiang2023mistral}, DeepSeek-7B \citep{bi2024deepseekllm} and DeepSeek-Math-7B \citep{shao2024deepseekmath}, all with the SimpleRL-Zoo prompt templates \citep{zeng2025simplerl}. Training problems come from Hendrycks MATH \citep{hendrycks2021math} and GSM8K \citep{cobbe2021gsm8k}, split into the SimpleRL-Zoo difficulty tiers. The 1.5B models train on the \emph{easy} tier (GSM8K and MATH level 1) and the 7-8B models on the \emph{hard} tier (MATH levels 3-5). We also propose an ablation (Appendix~\ref{app:shift}) where the four Qwen2.5 base and Math models are trained on the full MATH training set, evaluating the effect of distribution shifts introduced by the SimpleRL zoo setup. No training set contains any of the MATH-500 or GSM8K test problems.

TinyLoRA ($r = 2$) runs at $u \in \{1, 2, 4, 8, 16, 64, 256\}$ trainable parameters under GRPO and at $u \le 8$ trainable parameters under \llmbo{}. We, further, include \llmbo{} at $u \in \{16, 64, 256\}$ in Appendix~\ref{sec:large-u}. As references at larger scales, LoRA and LoRA-XS at rank $r \in \{1, 2, 8, 32\}$ are run under GRPO, from hundreds of trainable parameters (LoRA-XS, $r = 1$) to tens of millions (LoRA, $r = 32$). Appendix~\ref{app:scaling} details the parameterization and the weight tying and how both depart from \citet{morris2026learning}. Figure~\ref{fig:scaling-qwen} places TinyLoRA on this scale. 

On Qwen2.5-7B, GRPO with $u = 4$ parameters already gains \gainGrpoMeanQwenUFourMathFive{} points on MATH-500 avg@4 and with $u = 256$ it gains \gainGrpoMeanQwenUTwoFiftySixMathFive{}. Notably this latter gain is over half of LoRA's \loraGainMeanQwenMathFive{} with 81M parameters at $r = 32$. Qwen2.5-Math-7B scales similarly (\gainGrpoMeanMathUFourMathFive{}, \gainGrpoMeanMathUTwoFiftySixMathFive{} and \loraGainMeanMathMathFive{} points), while the Instruct model leaves little to gain at any scale (\loraGainMeanInstMathFive{} under LoRA). Figure~\ref{fig:scaling-base} displays every model (Appendices~\ref{app:scaling-results} and~\ref{app:instruct}).

GRPO takes 64 prompts $\times$ 4 rollouts at $T = 1$ per step, with the binary math-verify correctness reward, no KL penalty, clip parameter $\epsilon = 0.2$ and a dual clip at $\delta = 2$ that keeps the loss bounded (Appendix~\ref{app:dual-clip}).
The optimizer used is AdamW 8-bit with a constant learning rate after 10 warm-up steps ($10^{-3}/(r\sqrt{u})$ for TinyLoRA, $10^{-3}/r$ for LoRA-XS, $5 \cdot 10^{-6}$ for LoRA).
Prompts are capped at 512 tokens and completions at 1024. Each \llmbo{} batch generates the same 256 completions as a GRPO step, one candidate on $n = 256$ prompts ($b = 1$), and \llmbo{} runs as many GP-guided batches as GRPO takes steps, plus its initial and restart designs. It scores candidates on the pass rate under the decoding rule of the evaluation metric: at $T = 1$, GRPO's own objective, for avg@4, and greedily for greedy accuracy (Section~\ref{sec:greedy-objective}; Appendix~\ref{app:objective} compares the two objectives seed for seed). An ablation with $b = 4$ spreads the same 256 completions over four candidates on $n = 64$ prompts each, testing whether evaluating more adapters per completion makes the search more completion-efficient (Appendix~\ref{sec:parallel}).

\llmbo{} seed $k$ uses the SVD bases and random projections of GRPO seed $k$, pinned so that they do not depend on the GPU model (Appendix~\ref{app:svd-signs}). The two runs of a seed thus differ only in the optimizer, and every comparison is paired by seed. Experiments at $u \le 4$ use 8 seeds for GRPO and for \llmbo{} at $b = 1$ and $b = 4$. Every other experiment uses 3 seeds.

We evaluate on MATH-500 \citep{lightman2024verify}, GSM8K, AIME24 \citep{aime24}, AMC23 \citep{amc23}, Minerva Math \citep{lewkowycz2022minerva}, and OlympiadBench \citep{he2024olympiadbench} with one harness, the math-verify grader and the 1024-token completion budget of training for every model. MATH-500 and GSM8K are the in-distribution tasks. MATH-500 is drawn from the distribution of every training set; GSM8K is part of the easy tier and a same-domain, easier task for the 7-8B models. The four competition benchmarks measure out-of-distribution (OOD) generalization. These are small datsets composed of 30-675 problems each and are much harder problems that differ significantly from the ones found in the training set.

We report two metrics 1. \emph{avg@4}, the accuracy averaged over four completions at $T = 1$ and 2. \emph{greedy} accuracy, that of the single completion at $T = 0$. Here avg@4 is the headline metric, it is the expected accuracy at $T=1$ that both optimizers train on. So, objective, rollout distribution and metric match, and the comparison isolates the optimizer. Greedy accuracy is the accuracy of the model as usually deployed, which \llmbo{} can optimize directly (Section~\ref{sec:greedy-objective}). Every number is reported per benchmark and per $u$, as mean $\pm$ standard error over seeds or, for paired gaps {\llmbo{} $-$ GRPO}, over seed pairs. Greedy evaluations at intermediate checkpoints (steps $1, 2, 4, \dots$) trace accuracy against completions generated.

\subsection{Results}
\label{sec:results}
\label{sec:pooled}

Table~\ref{tab:headline} gives the paired gap \llmbo{} $-$ GRPO per model at $u = 1, 2, 4$ on all six benchmarks, in avg@4 and in greedy accuracy, with each model group closed by its average row. This section discusses avg@4, the objective both optimizers train on; Section~\ref{sec:greedy-objective} turns to greedy accuracy. Appendix~\ref{app:results} gives the full per-model analysis: the gains of both optimizers over the base model and their paired gap at every $u$ (Figures~\ref{fig:scatter-main} and~\ref{fig:gap-main}, Table~\ref{tab:main}).

Pooled over the four vanilla base models, one per family (Figure~\ref{fig:scatter-bases}), \llmbo{} leads GRPO at every $u \le 4$: by \poolToneMeanUOneGsmBase{}, \poolToneMeanUTwoGsmBase{}, and \poolToneMeanUFourGsmBase{} points on GSM8K at $u = 1, 2, 4$, and by \poolToneMeanUOneMathFiveBase{}, \poolToneMeanUTwoMathFiveBase{}, and \poolToneMeanUFourMathFiveBase{} on MATH-500. Over the two math-specialized models the lead increases to \poolToneMeanUTwoGsmMathVariants{} and \poolToneMeanUFourGsmMathVariants{} points on GSM8K at $u = 2, 4$ and \poolToneMeanUTwoMathFiveMathVariants{} and \poolToneMeanUFourMathFiveMathVariants{} on MATH-500. Over the two Qwen2.5 1.5B models the optimizers are level (Appendix~\ref{app:small}). These performance gains are with the same objective, rollout distribution, and metric as GRPO, so \llmbo{} is, thus, at least as good an optimizer in the ultra-low-parameter regime, and a better one where the random subspace sampled contains a gain to be had.

This is the case for the two Qwen2.5 7B models (Figure~\ref{fig:scaling-qwen}). On Qwen2.5-7B, \llmbo{} gains \gainToneMeanQwenUOneMathFive{}, \gainToneMeanQwenUTwoMathFive{}, and \gainToneMeanQwenUFourMathFive{} points on MATH-500 at $u = 1, 2, 4$, against GRPO's \gainGrpoMeanQwenUOneMathFive{}, \gainGrpoMeanQwenUTwoMathFive{}, and \gainGrpoMeanQwenUFourMathFive{}, with a paired gap of \gapToneMeanQwenUTwoMathFive{} at $u = 2$. On Qwen2.5-Math-7B its paired gap at $u = 4$ is \gapToneMeanMathUFourGsm{} on GSM8K and \gapToneMeanMathUFourMathFive{} on MATH-500. On both models the lead carries over to most OOD benchmarks (Appendix~\ref{app:results-ood}), and it holds when they train on the full MATH training set instead of the hard tier. Pooled over the two, \llmbo{}'s paired gap on MATH-500 is then \poolToneMeanUTwoMathFiveShiftLargeMathTrain{} and \poolToneMeanUFourMathFiveShiftLargeMathTrain{} points at $u = 2, 4$ (Appendix~\ref{app:shift}). On Llama3.1-8B, Mistral-7B and DeepSeek-7B, whose TinyLoRA subspace holds little, and on the Qwen2.5 1.5B models, \llmbo{} is level with GRPO. That is, where there is little to find, it gives up nothing. On the already RL-tuned Instruct models neither optimizer gains enough to differentiate a gain from noise (Appendix~\ref{app:instruct}).

\llmbo{} leads where a GP can cover the search space within the budget. For $u = 8$, Qwen2.5-7B's MATH-500 lead is gone (\gapToneMeanQwenUEightMathFive{}), while Qwen2.5-Math-7B stays ahead (\gapToneMeanMathUEightGsm{} on GSM8K). At $u \in \{64, 256\}$, with the evaluation budget used at $u \le 16$, \llmbo{} is not yet robust: pooled over the Qwen2.5 7B models its paired gap on MATH-500 is \poolToneMeanUSixtyFourMathFiveLargeU{} and \poolToneMeanUTwoFiftySixMathFiveLargeU{} points, with single seed pairs as low as $\worstToneMeanLargeMathFive{}$ (Appendix~\ref{sec:large-u}).

\begin{figure}[t]
\centering
\includegraphics[width=\linewidth]{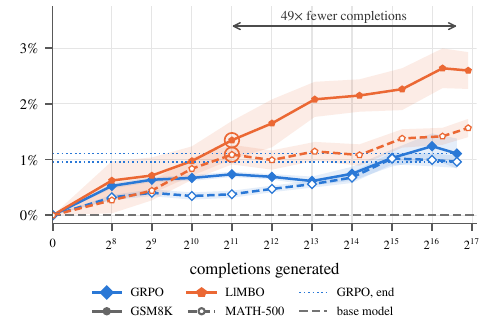}
\caption{\llmbo{} against GRPO, gain over the base model in greedy accuracy as a function of training completions generated, on GSM8K (solid) and MATH-500 (dashed, open markers).
Mean $\pm$ s.e.\ over seeds, averaged over $u \le 4$ and the four 7-8B model families; circles mark where \llmbo{} passes GRPO's final gain (dotted) for good.
Each model and $u$ separately in Figure~\ref{fig:budget}.
}
\label{fig:budget-main}
\end{figure}

\begin{table*}[t]
\centering
\caption{Single-GPU memory feasibility per architecture.
Weights in bf16. Each cell reports the number of simultaneous
full-length sequences that fit on each GPU, in \llmbo{}/GRPO order.
Green: \llmbo{} fits more.
\textcolor{red!70}{\ding{55}}:
the method does not fit.}
\label{tab:gpu-feasibility}
\vspace{4pt}\small\setlength{\tabcolsep}{17.5pt}
\begin{tabular}{@{}lccccc@{}}
\toprule
Architecture & \textbf{L4} 24\,GB & \textbf{L40S} 48\,GB & \textbf{A100} 40\,GB & \textbf{A100} 80\,GB & \textbf{H100} 80\,GB \\
\midrule
\raisebox{-0.15em}{\includegraphics[height=0.9em]{logos/qwen.png}}~Qwen2.5-7B & \fitdot{green!15}{71}\,/\,\textcolor{red!70}{\ding{55}} & \fitdot{green!15}{343}\,/\,12 & \fitdot{green!14}{282}\,/\,10 & \fitdot{green!14}{768}\,/\,27 & \fitdot{green!14}{767}\,/\,27 \\
\raisebox{-0.15em}{\includegraphics[height=0.9em]{logos/llama.png}}~Llama3.1-8B & \fitdot{green!15}{27}\,/\,\textcolor{red!70}{\ding{55}} & \fitdot{green!11}{146}\,/\,12 & \fitdot{green!11}{119}\,/\,10 & \fitdot{green!11}{332}\,/\,28 & \fitdot{green!11}{331}\,/\,28 \\
\raisebox{-0.15em}{\includegraphics[height=0.9em]{logos/mistral.png}}~Mistral-7B & \fitdot{green!15}{34}\,/\,\textcolor{red!70}{\ding{55}} & \fitdot{green!12}{154}\,/\,10 & \fitdot{green!12}{127}\,/\,8 & \fitdot{green!12}{340}\,/\,22 & \fitdot{green!12}{339}\,/\,22 \\
\raisebox{-0.15em}{\includegraphics[height=0.9em]{logos/deepseek.png}}~DeepSeek-7B & \fitdot{green!2}{10}\,/\,6 & \fitdot{green!2}{41}\,/\,26 & \fitdot{green!2}{34}\,/\,22 & \fitdot{green!2}{91}\,/\,58 & \fitdot{green!2}{91}\,/\,58 \\
\bottomrule
\end{tabular}

\vspace{6pt}
\end{table*}

\subsection{Training on greedy decoding directly}
\label{sec:greedy-objective}
\label{sec:greedy}
\label{sec:objective}
\label{sec:ablations}

GRPO needs sampled rollouts, so it optimizes accuracy at the sampling temperature, while a deployed model is often decoded greedily. \llmbo{} only needs objective values, so it can score candidates on greedy accuracy itself, a piecewise-constant function of the adapter that no policy gradient can follow. Here \llmbo{} trains on greedy accuracy, GRPO trains as before, and both are evaluated greedily (greedy columns of Table~\ref{tab:headline}; per model in Appendix~\ref{app:results}).

Pooled over the four vanilla base models, \llmbo{} leads GRPO on greedy MATH-500 at $u = 2, 4$, by \poolGreedyGreedyUTwoMathFiveBase{} and \poolGreedyGreedyUFourMathFiveBase{} points, and on greedy GSM8K at $u = 1, 4$, by \poolGreedyGreedyUOneGsmBase{} and \poolGreedyGreedyUFourGsmBase{}, and is level elsewhere. The lead is largest on the math-specialized models: \poolGreedyGreedyUOneGsmMathVariants{}, \poolGreedyGreedyUTwoGsmMathVariants{}, and \poolGreedyGreedyUFourGsmMathVariants{} points on greedy GSM8K pooled over the two 7B models and \gapGreedyGreedyMathSmallUOneGsm{}, \gapGreedyGreedyMathSmallUTwoGsm{}, and \gapGreedyGreedyMathSmallUFourGsm{} on Qwen2.5-Math-1.5B at $u = 1, 2, 4$ (Appendix~\ref{app:greedy-small}).

Blocking over seeds, \llmbo{} trained on greedy accuracy beats \llmbo{} trained on the pass rate at $T = 1$ in greedy accuracy at $u = 4$, by \objectiveUFourGsmBase{} points on greedy GSM8K over the vanilla models and \objectiveUFourGsmMathVariants{} over the math models (Appendix~\ref{app:objective}). The decoding rule is thus a free choice of objective for \llmbo{}: a model that will be decoded greedily can be trained for greedy decoding directly.

\subsection{Compute and memory efficiency}
\label{sec:efficiency}

\llmbo{} reaches these gains with a fraction of GRPO's compute, completions, and memory (Appendix~\ref{app:efficiency}).

Each completion costs \llmbo{} one forward pass per token, and it fits its GP on the CPU. GRPO also rescores every completion and passes it forward and backward through the model, at least \grpoPasses{} forward passes per token in all, so over a median run \llmbo{} spends about \boComputeShare{}\% of GRPO's training FLOPs.

Averaged over the four 7-8B families and $u \le 4$, \llmbo{} trained on greedy accuracy passes GRPO's final greedy accuracy after \reachPoolGreedyGsm{} completions on GSM8K and \reachPoolGreedyMathFive{} on MATH-500, where a median GRPO run generates \grpoCompletions{} (Figure~\ref{fig:budget-main}). It ends \poolEndGreedyGsm{} and \poolEndGreedyMathFive{} points above the base model, against GRPO's \poolEndGrpoGsm{} and \poolEndGrpoMathFive{}. On Qwen2.5-Math-1.5B at $u = 2$ it passes GRPO's final accuracy with its first GP-guided batch, \reachGreedyMathSmallUTwoGsm{} completions (Figure~\ref{fig:budget}).

\llmbo{} needs no memory beyond inference, while GRPO's backward pass holds \grpoOverheadSmall{}\,GiB on the 1.5B models and \grpoOverheadLarge{}\,GiB on the 7-8B ones for every \grpoMicroBatch{} sequences it trains on. On a 24\,GB card, GRPO cannot fit even its \grpoMicroBatch{}-sequence micro-batch on \grpoNoFitSmallCard{} of the \nLargeArch{} 7-8B architectures, while \llmbo{} runs \boSequencesSmallCard{} sequences concurrently on all of them (Table~\ref{tab:gpu-feasibility}; measurements and estimates in Appendix~\ref{sec:cost}, Table~\ref{tab:memory}).

\section{Discussion}
\label{sec:discussion}
\subsection{Limitations}
\label{sec:limitations}
\llmbo{} targets the ultra-low-parameter end; from $u = 8$ its edge fades, and at $u \in \{64, 256\}$ it trails GRPO and varies strongly across seeds (Appendix~\ref{sec:large-u}), so there a gradient remains the natural choice, and combining the two, e.g.\ BO over a low-dimensional subspace of a larger adapter, is a promising direction. On most models TinyLoRA at $u \le 4$ moves accuracy little under either optimizer (Appendix~\ref{sec:base}), so the strongest results are based on the Qwen2.5 base and Math models; adapters that reach further into other families are also promising.
Our compute comparison counts FLOPs; wall-clock savings also depend on how efficiently the inference engine batches generation.

\subsection{Conclusion}
\label{sec:conclusion}
In the ultra-low-parameter regime, \llmbo{} is a suitable alternative to GRPO: trust-region BO on the pass rate matches or exceeds GRPO's fine-tuning gains across model families, variants, sizes and training data
This effect is most prominent when TinyLoRA exhibits the greatest gains and when we optimize the greedy-decoding accuracy directly.
This accuracy win occurs with a fraction of GRPO's compute and memory.

A candidate is a handful of scalars and its score is a pass count, so \llmbo{} scales across devices with almost no synchronization, and its memory footprint makes fine-tuning possible on inference-only hardware. BO also opens objectives that are awkward for policy gradients: non-differentiable metrics, decoding-rule-specific accuracy, or multi-objective trade-offs between accuracy and length, and its posterior mean offers a natural stopping rule.

\FloatBarrier

\subsection*{AI use statement}
In this work, we used generative AI tools for implementation (code generation and editing), figure creation, paper drafting and editing, and formatting.
We have not used generative AI tools for theoretical model development, mathematical claims, hypothesis refinement or research methodology design.
All AI-assisted code and text were reviewed and verified by the authors, who take responsibility for the final content of this work.

\subsection*{Reproducibility statement}
The anonymized code, with its dependencies, is in the supplementary material: the adapters, \llmbo{}, the GRPO baseline, the evaluation harness and the job scripts of every run. Table~\ref{tab:hparams} lists every hyperparameter of both optimizers and Appendix~\ref{app:harness} the evaluation harness; all tables, figures and quoted numbers are generated from the run directories by the scripts in \texttt{paper/scripts}, also in the supplementary material. Every exported adapter stores its SVD bases, so its search space can be rebuilt exactly on any hardware (Appendix~\ref{app:svd-signs}).

\bibliographystyle{apalike}
\bibliography{references}


\appendix
\onecolumn
\aistatstitle{How Low Can You Go: Fine-Tuning LLMs with Bayesian Optimization (\llmbo{}): \\
Supplementary Materials}
\suppressfloats[t]

\section{Greedy accuracy: pooled gaps and training objective}
\label{app:greedy}
This appendix pools the greedy paired gaps of Section~\ref{sec:greedy-objective} over models (per model in Appendix~\ref{app:results}) and compares \llmbo{}'s two training objectives seed for seed.

\subsection{Pooled over models}
\label{app:greedy-pooled}

\begin{table}[!htbp]
\centering
\caption{Paired gaps (points) per $u \le 4$, pooled over the six 7-8B base and math models and over the two Qwen2.5 1.5B models (row blocks; each model counted once), mean $\pm$ s.e.\ over seed pairs, in avg@4 and greedy accuracy (sub-columns), \llmbo{} trained on the metric of each column. Left: \llmbo{} $-$ GRPO; \emph{train tier}: greedy accuracy on up to 1024 problems of each model's training tier, the quantity \llmbo{} maximizes there (evaluated in greedy only). Right: the $b = 4$ ablation of \llmbo{} against $b = 1$ (Appendix~\ref{sec:parallel}). Tint: sign and size of the gap (green: the first term ahead).}
\label{tab:ablations}
\renewcommand{\arraystretch}{0.85}
\resizebox{\linewidth}{!}{\begin{tabular}{l l ccccc@{\hspace{1.2em}}cccc}
\toprule
 & & \multicolumn{5}{c}{\llmbo{} $-$ GRPO} & \multicolumn{4}{c}{\llmbo{}: $b{=}4$ $-$ $b{=}1$} \\
\cmidrule(lr){3-7}\cmidrule(lr){8-11}
 &  & \multicolumn{2}{c}{GSM8K} & \multicolumn{2}{c}{MATH-500} & train tier & \multicolumn{2}{c}{GSM8K} & \multicolumn{2}{c}{MATH-500} \\
\cmidrule(lr){3-4}\cmidrule(lr){5-6}\cmidrule(lr){8-9}\cmidrule(lr){10-11}\cmidrule(lr){7-7}
Models &  & avg@4 & greedy & avg@4 & greedy & greedy & avg@4 & greedy & avg@4 & greedy \\
\midrule
7-8B models & $u=1$ & \gapchip{green!7}{$+0.5$\,{\scriptsize$\pm$0.3}} & \gapchip{green!28}{$+2.6$\,{\scriptsize$\pm$1.1}} & \gapchip{green!5}{$+0.3$\,{\scriptsize$\pm$0.2}} & \gapchip{green!7}{$+0.5$\,{\scriptsize$\pm$0.5}} & \gapchip{green!4}{$+0.3$\,{\scriptsize$\pm$0.6}} & \gapchip{red!1}{$-0.1$\,{\scriptsize$\pm$0.2}} & \gapchip{red!9}{$-0.7$\,{\scriptsize$\pm$0.3}} & \gapchip{none}{$+0.0$\,{\scriptsize$\pm$0.2}} & \gapchip{red!1}{$-0.1$\,{\scriptsize$\pm$0.3}} \\
 & $u=2$ & \gapchip{green!25}{$+1.8$\,{\scriptsize$\pm$0.6}} & \gapchip{green!14}{$+1.0$\,{\scriptsize$\pm$1.0}} & \gapchip{green!19}{$+1.3$\,{\scriptsize$\pm$0.5}} & \gapchip{green!8}{$+0.6$\,{\scriptsize$\pm$0.3}} & \gapchip{green!18}{$+1.3$\,{\scriptsize$\pm$0.3}} & \gapchip{red!11}{$-0.8$\,{\scriptsize$\pm$0.4}} & \gapchip{green!6}{$+0.5$\,{\scriptsize$\pm$0.7}} & \gapchip{red!4}{$-0.3$\,{\scriptsize$\pm$0.2}} & \gapchip{red!7}{$-0.5$\,{\scriptsize$\pm$0.3}} \\
 & $u=4$ & \gapchip{green!23}{$+1.6$\,{\scriptsize$\pm$0.5}} & \gapchip{green!28}{$+2.9$\,{\scriptsize$\pm$1.9}} & \gapchip{green!19}{$+1.4$\,{\scriptsize$\pm$0.4}} & \gapchip{green!23}{$+1.7$\,{\scriptsize$\pm$0.6}} & \gapchip{green!18}{$+1.3$\,{\scriptsize$\pm$0.9}} & \gapchip{red!7}{$-0.5$\,{\scriptsize$\pm$0.3}} & \gapchip{green!1}{$+0.1$\,{\scriptsize$\pm$0.6}} & \gapchip{red!6}{$-0.4$\,{\scriptsize$\pm$0.2}} & \gapchip{red!10}{$-0.7$\,{\scriptsize$\pm$0.4}} \\
\midrule
1.5B models & $u=1$ & \gapchip{green!3}{$+0.2$\,{\scriptsize$\pm$0.4}} & \gapchip{green!28}{$+4.5$\,{\scriptsize$\pm$2.3}} & \gapchip{red!1}{$-0.1$\,{\scriptsize$\pm$0.4}} & \gapchip{red!1}{$-0.1$\,{\scriptsize$\pm$0.8}} & \gapchip{green!28}{$+2.9$\,{\scriptsize$\pm$2.2}} & \gapchip{red!2}{$-0.2$\,{\scriptsize$\pm$0.4}} & \gapchip{red!2}{$-0.1$\,{\scriptsize$\pm$0.5}} & \gapchip{red!1}{$-0.1$\,{\scriptsize$\pm$0.5}} & \gapchip{red!3}{$-0.2$\,{\scriptsize$\pm$0.4}} \\
 & $u=2$ & \gapchip{green!8}{$+0.6$\,{\scriptsize$\pm$0.4}} & \gapchip{green!28}{$+14.1$\,{\scriptsize$\pm$7.1}} & \gapchip{green!2}{$+0.1$\,{\scriptsize$\pm$0.4}} & \gapchip{green!28}{$+7.7$\,{\scriptsize$\pm$4.7}} & \gapchip{green!28}{$+13.8$\,{\scriptsize$\pm$7.4}} & \gapchip{red!7}{$-0.5$\,{\scriptsize$\pm$0.5}} & \gapchip{green!5}{$+0.4$\,{\scriptsize$\pm$0.2}} & \gapchip{green!1}{$+0.1$\,{\scriptsize$\pm$0.4}} & \gapchip{green!1}{$+0.1$\,{\scriptsize$\pm$0.6}} \\
 & $u=4$ & \gapchip{green!4}{$+0.3$\,{\scriptsize$\pm$0.5}} & \gapchip{green!28}{$+4.2$\,{\scriptsize$\pm$1.8}} & \gapchip{red!10}{$-0.7$\,{\scriptsize$\pm$0.4}} & \gapchip{green!21}{$+1.5$\,{\scriptsize$\pm$1.4}} & \gapchip{green!28}{$+4.5$\,{\scriptsize$\pm$2.2}} & \gapchip{red!1}{$-0.1$\,{\scriptsize$\pm$0.5}} & \gapchip{red!2}{$-0.1$\,{\scriptsize$\pm$0.5}} & \gapchip{none}{$+0.0$\,{\scriptsize$\pm$0.4}} & \gapchip{red!6}{$-0.4$\,{\scriptsize$\pm$0.8}} \\
\bottomrule
\end{tabular}
}
\end{table}

Over the six 7-8B models \llmbo{} leads GRPO in greedy accuracy at every $u$ (Table~\ref{tab:ablations}): on greedy GSM8K by \poolGreedyGreedyUOneGsm{}, \poolGreedyGreedyUTwoGsm{}, and \poolGreedyGreedyUFourGsm{} points at $u = 1, 2, 4$ and on greedy MATH-500 by \poolGreedyGreedyUOneMathFive{}, \poolGreedyGreedyUTwoMathFive{}, and \poolGreedyGreedyUFourMathFive{}. On its own training problems its paired gap is small on the 7-8B models (\trainGapGreedyUOne{}, \trainGapGreedyUTwo{}, and \trainGapGreedyUFour{} points) and large at 1.5B (\trainGapGreedyUTwoSmall{} at $u = 2$; Table~\ref{tab:ablations}), where its held-out gains are largest too: held-out gains appear where \llmbo{} optimizes its objective better.
The lead is largest on the math-specialized models and at 1.5B (Appendices~\ref{sec:math} and~\ref{app:greedy-small}).

\FloatBarrier
\section{Per-model results}
\label{app:results}
\begin{figure}[!htbp]
\centering
\includegraphics[width=\linewidth]{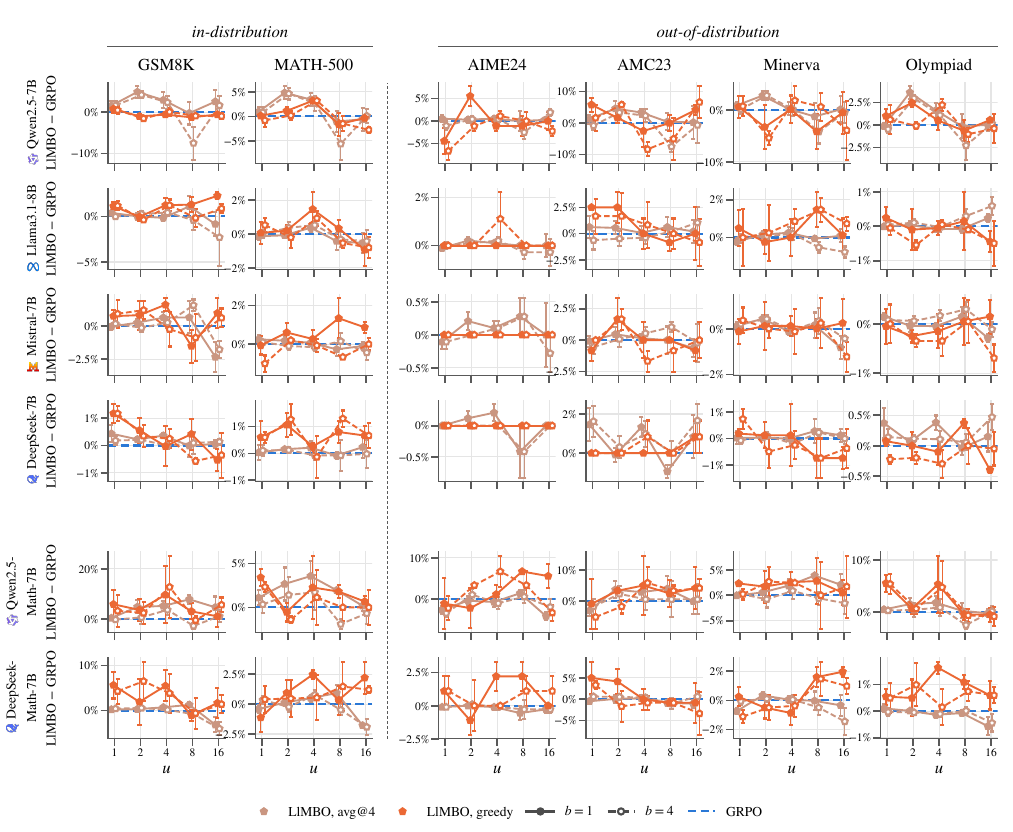}
\caption{Paired gap \llmbo{} $-$ GRPO in avg@4 and in greedy accuracy (color), \llmbo{} trained on each metric, against $u \in \{1, \dots, 16\}$, mean $\pm$ s.e.\ over seed pairs, for the one 7-8B base model per family and the two math-specialized 7B models (rows, in two blocks), one column per benchmark (in-distribution GSM8K and MATH-500 first, then the four out-of-distribution benchmarks), each panel on its own gap axis. Solid, filled: $b = 1$ candidate per batch (the default); dashed, open: the $b = 4$ ablation.}
\label{fig:gap-main}
\end{figure}

This appendix details Sections~\ref{sec:results} and~\ref{sec:greedy-objective} per model, in avg@4 and in greedy accuracy, for one 7-8B base model per family (Section~\ref{sec:base}) and the two math-specialized models (Section~\ref{sec:math}). The 1.5B and Instruct models are in Appendices~\ref{app:small} and~\ref{app:instruct}. The pooled greedy gaps and the comparison of \llmbo{}'s two objectives in Appendix~\ref{app:greedy}. Figure~\ref{fig:scatter-main} compares the two optimizers' gains at $u \le 4$, Figure~\ref{fig:gap-main} traces their paired gap over $u \le 16$, with the $b = 4$ ablation, and Table~\ref{tab:main} lists both on all six benchmarks. In greedy accuracy \llmbo{} is trained on greedy accuracy, so GRPO is evaluated across a shift in decoding rule that \llmbo{} does not face.

\begin{table}[!htbp]
\centering
\caption{Avg@4 (a) and greedy (b) accuracy at the end of training, for the four vanilla base models, then the two math-specialized ones (row blocks), per $u \le 16$ (rows) and benchmark (columns: in-distribution GSM8K and MATH-500, then the four out-of-distribution benchmarks); \llmbo{} is trained on the metric of each subtable (greedy: Section~\ref{sec:greedy-objective}). \emph{base}: accuracy of the base model in that metric (\%). \emph{GRPO}, \emph{\llmbo{}}: gain over the base model (points), mean over seeds (GRPO and avg@4-trained \llmbo{} 8 per cell at $u \le 4$ and 3 at $u = 8, 16$, greedy-trained \llmbo{} 3 per cell); bold: the larger, unless tied. $\Delta$: paired gap \llmbo{} $-$ GRPO, mean $\pm$ s.e.\ over the seed pairs both have, so it can differ from the difference of the two gains; tint: its sign and size (green: \llmbo{} ahead).}
\label{tab:main}
\renewcommand{\arraystretch}{0.85}
{\small (a) avg@4}\\[0.2em]
\resizebox{\linewidth}{!}{\begin{tabular}{c l cccccc@{\hspace{1.2em}}cccccccccccc}
\toprule
 & & \multicolumn{6}{c}{in-distribution} & \multicolumn{12}{c}{out-of-distribution} \\
\cmidrule(lr){3-8}\cmidrule(lr){9-20}
 & & \multicolumn{3}{c}{GSM8K} & \multicolumn{3}{c}{MATH-500} & \multicolumn{3}{c}{AIME24} & \multicolumn{3}{c}{AMC23} & \multicolumn{3}{c}{Minerva} & \multicolumn{3}{c}{Olympiad} \\
\cmidrule(lr){3-5}\cmidrule(lr){6-8}\cmidrule(lr){9-11}\cmidrule(lr){12-14}\cmidrule(lr){15-17}\cmidrule(lr){18-20}
 & $u$ & GRPO & LlMBO & $\Delta$ & GRPO & LlMBO & $\Delta$ & GRPO & LlMBO & $\Delta$ & GRPO & LlMBO & $\Delta$ & GRPO & LlMBO & $\Delta$ & GRPO & LlMBO & $\Delta$ \\
\midrule
\multirow{6}{*}{\rotatebox[origin=c]{90}{\scriptsize\begin{tabular}{@{}c@{}}\raisebox{-0.15em}{\includegraphics[height=0.9em]{logos/llama.png}}~Llama3.1\\ 8B\end{tabular}}} & base & \multicolumn{3}{c}{14.5} & \multicolumn{3}{c}{5.4} & \multicolumn{3}{c}{0.0} & \multicolumn{3}{c}{3.1} & \multicolumn{3}{c}{2.8} & \multicolumn{3}{c}{1.5} \\
 & 1 & $+0.3$ & \textbf{\boldmath $+0.7$} & \gapchip{green!5}{$+0.4$\,{\scriptsize$\pm$0.2}} & \textbf{\boldmath $-0.3$} & $-0.5$ & \gapchip{red!2}{$-0.1$\,{\scriptsize$\pm$0.2}} & \textbf{\boldmath $+0.1$} & $+0.0$ & \gapchip{red!1}{$-0.1$\,{\scriptsize$\pm$0.1}} & $-0.2$ & \textbf{\boldmath $+0.5$} & \gapchip{green!9}{$+0.6$\,{\scriptsize$\pm$1.1}} & \textbf{\boldmath $-0.6$} & $-0.8$ & \gapchip{red!3}{$-0.2$\,{\scriptsize$\pm$0.2}} & $+0.0$ & $+0.0$ & \gapchip{none}{$+0.0$\,{\scriptsize$\pm$0.1}} \\
 & 2 & $+0.8$ & $+0.8$ & \gapchip{none}{$+0.0$\,{\scriptsize$\pm$0.4}} & \textbf{\boldmath $-0.1$} & $-0.2$ & \gapchip{red!1}{$-0.1$\,{\scriptsize$\pm$0.1}} & $+0.0$ & \textbf{\boldmath $+0.2$} & \gapchip{green!3}{$+0.2$\,{\scriptsize$\pm$0.1}} & $+0.2$ & \textbf{\boldmath $+0.8$} & \gapchip{green!8}{$+0.5$\,{\scriptsize$\pm$0.6}} & $-0.8$ & \textbf{\boldmath $-0.6$} & \gapchip{green!2}{$+0.1$\,{\scriptsize$\pm$0.2}} & $+0.0$ & $+0.0$ & \gapchip{green!1}{$+0.0$\,{\scriptsize$\pm$0.1}} \\
 & 4 & \textbf{\boldmath $+1.4$} & $+1.2$ & \gapchip{red!2}{$-0.2$\,{\scriptsize$\pm$0.3}} & $-0.3$ & \textbf{\boldmath $+0.1$} & \gapchip{green!5}{$+0.4$\,{\scriptsize$\pm$0.3}} & $+0.0$ & \textbf{\boldmath $+0.1$} & \gapchip{green!1}{$+0.1$\,{\scriptsize$\pm$0.1}} & $+0.2$ & $+0.2$ & \gapchip{green!1}{$+0.1$\,{\scriptsize$\pm$0.7}} & $-0.6$ & \textbf{\boldmath $-0.4$} & \gapchip{green!4}{$+0.3$\,{\scriptsize$\pm$0.3}} & $+0.1$ & $+0.1$ & \gapchip{red!1}{$-0.1$\,{\scriptsize$\pm$0.1}} \\
 & 8 & $+1.4$ & \textbf{\boldmath $+2.5$} & \gapchip{green!15}{$+1.1$\,{\scriptsize$\pm$1.1}} & \textbf{\boldmath $+0.3$} & $-0.1$ & \gapchip{red!6}{$-0.4$\,{\scriptsize$\pm$0.5}} & $+0.3$ & $+0.3$ & \gapchip{none}{$+0.0$\,{\scriptsize$\pm$0.0}} & $-0.4$ & \textbf{\boldmath $+0.2$} & \gapchip{green!9}{$+0.6$\,{\scriptsize$\pm$1.0}} & $-0.6$ & $-0.6$ & \gapchip{red!1}{$-0.1$\,{\scriptsize$\pm$0.1}} & \textbf{\boldmath $-0.1$} & $-0.2$ & \gapchip{red!1}{$-0.1$\,{\scriptsize$\pm$0.1}} \\
 & 16 & \textbf{\boldmath $+2.8$} & $+1.9$ & \gapchip{red!12}{$-0.9$\,{\scriptsize$\pm$1.3}} & \textbf{\boldmath $+0.5$} & $-0.1$ & \gapchip{red!7}{$-0.5$\,{\scriptsize$\pm$0.6}} & $+0.6$ & $+0.6$ & \gapchip{none}{$+0.0$\,{\scriptsize$\pm$0.5}} & $-0.8$ & \textbf{\boldmath $-0.6$} & \gapchip{green!3}{$+0.2$\,{\scriptsize$\pm$1.1}} & $-0.3$ & \textbf{\boldmath $-0.2$} & \gapchip{green!1}{$+0.1$\,{\scriptsize$\pm$0.3}} & $-0.1$ & \textbf{\boldmath $+0.1$} & \gapchip{green!3}{$+0.2$\,{\scriptsize$\pm$0.1}} \\
\midrule
\multirow{6}{*}{\rotatebox[origin=c]{90}{\scriptsize\begin{tabular}{@{}c@{}}\raisebox{-0.15em}{\includegraphics[height=0.9em]{logos/mistral.png}}~Mistral\\ 7B\end{tabular}}} & base & \multicolumn{3}{c}{8.5} & \multicolumn{3}{c}{3.4} & \multicolumn{3}{c}{0.0} & \multicolumn{3}{c}{1.9} & \multicolumn{3}{c}{2.5} & \multicolumn{3}{c}{1.1} \\
 & 1 & $-0.5$ & $-0.5$ & \gapchip{red!1}{$+0.0$\,{\scriptsize$\pm$0.2}} & $+0.2$ & \textbf{\boldmath $+0.4$} & \gapchip{green!4}{$+0.3$\,{\scriptsize$\pm$0.2}} & \textbf{\boldmath $+0.1$} & $+0.0$ & \gapchip{red!1}{$-0.1$\,{\scriptsize$\pm$0.1}} & $+0.2$ & $+0.2$ & \gapchip{red!1}{$-0.1$\,{\scriptsize$\pm$0.4}} & $-0.9$ & \textbf{\boldmath $-0.7$} & \gapchip{green!2}{$+0.1$\,{\scriptsize$\pm$0.2}} & $-0.1$ & \textbf{\boldmath $+0.0$} & \gapchip{green!2}{$+0.1$\,{\scriptsize$\pm$0.1}} \\
 & 2 & $-0.4$ & \textbf{\boldmath $-0.1$} & \gapchip{green!3}{$+0.2$\,{\scriptsize$\pm$0.3}} & $+0.2$ & \textbf{\boldmath $+0.4$} & \gapchip{green!2}{$+0.2$\,{\scriptsize$\pm$0.2}} & $+0.0$ & \textbf{\boldmath $+0.2$} & \gapchip{green!3}{$+0.2$\,{\scriptsize$\pm$0.1}} & $+0.5$ & \textbf{\boldmath $+1.0$} & \gapchip{green!7}{$+0.5$\,{\scriptsize$\pm$0.6}} & $-1.1$ & \textbf{\boldmath $-0.6$} & \gapchip{green!6}{$+0.4$\,{\scriptsize$\pm$0.2}} & \textbf{\boldmath $+0.0$} & $-0.1$ & \gapchip{red!1}{$+0.0$\,{\scriptsize$\pm$0.1}} \\
 & 4 & $-0.6$ & \textbf{\boldmath $+0.0$} & \gapchip{green!9}{$+0.6$\,{\scriptsize$\pm$0.4}} & $+0.2$ & \textbf{\boldmath $+0.3$} & \gapchip{green!1}{$+0.1$\,{\scriptsize$\pm$0.2}} & $+0.0$ & \textbf{\boldmath $+0.1$} & \gapchip{green!1}{$+0.1$\,{\scriptsize$\pm$0.1}} & $-0.2$ & \textbf{\boldmath $+0.0$} & \gapchip{green!2}{$+0.2$\,{\scriptsize$\pm$0.3}} & \textbf{\boldmath $-0.6$} & $-0.8$ & \gapchip{red!2}{$-0.1$\,{\scriptsize$\pm$0.2}} & $-0.1$ & $-0.1$ & \gapchip{red!1}{$+0.0$\,{\scriptsize$\pm$0.1}} \\
 & 8 & $-0.5$ & \textbf{\boldmath $+0.1$} & \gapchip{green!9}{$+0.6$\,{\scriptsize$\pm$1.2}} & \textbf{\boldmath $+0.4$} & $+0.2$ & \gapchip{red!4}{$-0.3$\,{\scriptsize$\pm$0.3}} & $+0.0$ & \textbf{\boldmath $+0.3$} & \gapchip{green!4}{$+0.3$\,{\scriptsize$\pm$0.3}} & $+0.2$ & $+0.2$ & \gapchip{none}{$+0.0$\,{\scriptsize$\pm$1.1}} & $-0.6$ & \textbf{\boldmath $-0.5$} & \gapchip{green!2}{$+0.2$\,{\scriptsize$\pm$0.6}} & $-0.2$ & \textbf{\boldmath $-0.1$} & \gapchip{green!2}{$+0.1$\,{\scriptsize$\pm$0.2}} \\
 & 16 & \textbf{\boldmath $-0.3$} & $-2.6$ & \gapchip{red!28}{$-2.3$\,{\scriptsize$\pm$1.1}} & \textbf{\boldmath $+0.1$} & $+0.0$ & \gapchip{red!1}{$-0.1$\,{\scriptsize$\pm$0.2}} & $+0.3$ & $+0.3$ & \gapchip{none}{$+0.0$\,{\scriptsize$\pm$0.5}} & \textbf{\boldmath $+0.6$} & $+0.0$ & \gapchip{red!9}{$-0.6$\,{\scriptsize$\pm$0.7}} & \textbf{\boldmath $-0.5$} & $-1.3$ & \gapchip{red!12}{$-0.8$\,{\scriptsize$\pm$0.5}} & \textbf{\boldmath $+0.0$} & $-0.3$ & \gapchip{red!4}{$-0.3$\,{\scriptsize$\pm$0.1}} \\
\midrule
\multirow{6}{*}{\rotatebox[origin=c]{90}{\scriptsize\begin{tabular}{@{}c@{}}\raisebox{-0.15em}{\includegraphics[height=0.9em]{logos/deepseek.png}}~DeepSeek\\ 7B\end{tabular}}} & base & \multicolumn{3}{c}{1.5} & \multicolumn{3}{c}{1.1} & \multicolumn{3}{c}{0.0} & \multicolumn{3}{c}{0.6} & \multicolumn{3}{c}{0.3} & \multicolumn{3}{c}{0.6} \\
 & 1 & $-0.6$ & \textbf{\boldmath $-0.2$} & \gapchip{green!6}{$+0.4$\,{\scriptsize$\pm$0.3}} & $+0.2$ & $+0.2$ & \gapchip{none}{$+0.0$\,{\scriptsize$\pm$0.3}} & $+0.0$ & $+0.0$ & \gapchip{none}{$+0.0$\,{\scriptsize$\pm$0.0}} & $+0.0$ & \textbf{\boldmath $+1.6$} & \gapchip{green!20}{$+1.5$\,{\scriptsize$\pm$0.8}} & \textbf{\boldmath $+0.4$} & $+0.3$ & \gapchip{none}{$+0.0$\,{\scriptsize$\pm$0.2}} & $-0.1$ & \textbf{\boldmath $+0.1$} & \gapchip{green!5}{$+0.4$\,{\scriptsize$\pm$0.3}} \\
 & 2 & $-0.6$ & \textbf{\boldmath $-0.4$} & \gapchip{green!3}{$+0.2$\,{\scriptsize$\pm$0.1}} & $+0.0$ & \textbf{\boldmath $+0.1$} & \gapchip{green!2}{$+0.2$\,{\scriptsize$\pm$0.1}} & $+0.0$ & \textbf{\boldmath $+0.1$} & \gapchip{green!2}{$+0.1$\,{\scriptsize$\pm$0.1}} & \textbf{\boldmath $+0.5$} & $+0.4$ & \gapchip{red!1}{$-0.1$\,{\scriptsize$\pm$0.4}} & \textbf{\boldmath $+0.3$} & $+0.2$ & \gapchip{none}{$+0.0$\,{\scriptsize$\pm$0.0}} & $-0.1$ & $-0.1$ & \gapchip{green!1}{$+0.1$\,{\scriptsize$\pm$0.1}} \\
 & 4 & $-0.5$ & \textbf{\boldmath $-0.1$} & \gapchip{green!5}{$+0.4$\,{\scriptsize$\pm$0.1}} & $+0.2$ & \textbf{\boldmath $+0.4$} & \gapchip{green!3}{$+0.2$\,{\scriptsize$\pm$0.3}} & $+0.0$ & \textbf{\boldmath $+0.2$} & \gapchip{green!3}{$+0.2$\,{\scriptsize$\pm$0.1}} & $+0.2$ & \textbf{\boldmath $+1.5$} & \gapchip{green!19}{$+1.3$\,{\scriptsize$\pm$0.5}} & $+0.4$ & $+0.4$ & \gapchip{none}{$+0.0$\,{\scriptsize$\pm$0.1}} & $-0.1$ & \textbf{\boldmath $+0.3$} & \gapchip{green!5}{$+0.4$\,{\scriptsize$\pm$0.1}} \\
 & 8 & $-0.6$ & $-0.6$ & \gapchip{none}{$+0.0$\,{\scriptsize$\pm$0.4}} & $+0.2$ & $+0.2$ & \gapchip{red!1}{$-0.1$\,{\scriptsize$\pm$0.1}} & \textbf{\boldmath $+0.4$} & $+0.0$ & \gapchip{red!6}{$-0.4$\,{\scriptsize$\pm$0.4}} & \textbf{\boldmath $+0.3$} & $-0.4$ & \gapchip{red!13}{$-0.9$\,{\scriptsize$\pm$0.3}} & $+0.1$ & \textbf{\boldmath $+0.3$} & \gapchip{green!4}{$+0.3$\,{\scriptsize$\pm$0.5}} & $-0.1$ & $-0.1$ & \gapchip{red!1}{$+0.0$\,{\scriptsize$\pm$0.1}} \\
 & 16 & $-0.3$ & $-0.3$ & \gapchip{green!1}{$+0.1$\,{\scriptsize$\pm$0.1}} & $+0.4$ & \textbf{\boldmath $+0.6$} & \gapchip{green!3}{$+0.2$\,{\scriptsize$\pm$0.4}} & $+0.0$ & $+0.0$ & \gapchip{none}{$+0.0$\,{\scriptsize$\pm$0.0}} & $+0.4$ & \textbf{\boldmath $+1.2$} & \gapchip{green!12}{$+0.8$\,{\scriptsize$\pm$1.1}} & $+0.3$ & \textbf{\boldmath $+0.4$} & \gapchip{green!2}{$+0.1$\,{\scriptsize$\pm$0.2}} & $-0.1$ & \textbf{\boldmath $+0.0$} & \gapchip{green!2}{$+0.1$\,{\scriptsize$\pm$0.2}} \\
\midrule
\multirow{6}{*}{\rotatebox[origin=c]{90}{\scriptsize\begin{tabular}{@{}c@{}}\raisebox{-0.15em}{\includegraphics[height=0.9em]{logos/qwen.png}}~Qwen2.5\\ 7B\end{tabular}}} & base & \multicolumn{3}{c}{64.9} & \multicolumn{3}{c}{43.5} & \multicolumn{3}{c}{2.5} & \multicolumn{3}{c}{24.4} & \multicolumn{3}{c}{16.4} & \multicolumn{3}{c}{16.7} \\
 & 1 & $+4.3$ & \textbf{\boldmath $+5.8$} & \gapchip{green!21}{$+1.5$\,{\scriptsize$\pm$1.1}} & $+4.8$ & \textbf{\boldmath $+5.8$} & \gapchip{green!13}{$+1.0$\,{\scriptsize$\pm$0.9}} & $+0.9$ & \textbf{\boldmath $+1.5$} & \gapchip{green!7}{$+0.5$\,{\scriptsize$\pm$0.8}} & $+2.7$ & \textbf{\boldmath $+3.9$} & \gapchip{green!17}{$+1.2$\,{\scriptsize$\pm$1.4}} & $+2.6$ & \textbf{\boldmath $+3.4$} & \gapchip{green!12}{$+0.9$\,{\scriptsize$\pm$0.6}} & \textbf{\boldmath $+2.8$} & $+2.6$ & \gapchip{red!2}{$-0.1$\,{\scriptsize$\pm$0.4}} \\
 & 2 & $+2.1$ & \textbf{\boldmath $+6.9$} & \gapchip{green!28}{$+4.9$\,{\scriptsize$\pm$1.4}} & $+2.2$ & \textbf{\boldmath $+7.1$} & \gapchip{green!28}{$+4.8$\,{\scriptsize$\pm$1.4}} & $+0.8$ & \textbf{\boldmath $+1.1$} & \gapchip{green!4}{$+0.3$\,{\scriptsize$\pm$0.7}} & $-1.9$ & \textbf{\boldmath $+2.7$} & \gapchip{green!28}{$+4.6$\,{\scriptsize$\pm$2.4}} & $+0.6$ & \textbf{\boldmath $+3.4$} & \gapchip{green!28}{$+2.8$\,{\scriptsize$\pm$0.9}} & $+0.1$ & \textbf{\boldmath $+3.7$} & \gapchip{green!28}{$+3.6$\,{\scriptsize$\pm$0.7}} \\
 & 4 & $+6.9$ & \textbf{\boldmath $+9.8$} & \gapchip{green!28}{$+2.9$\,{\scriptsize$\pm$1.8}} & $+6.9$ & \textbf{\boldmath $+10.3$} & \gapchip{green!28}{$+3.5$\,{\scriptsize$\pm$1.4}} & $+0.5$ & \textbf{\boldmath $+1.0$} & \gapchip{green!7}{$+0.5$\,{\scriptsize$\pm$0.6}} & $+1.5$ & \textbf{\boldmath $+4.8$} & \gapchip{green!28}{$+3.4$\,{\scriptsize$\pm$1.3}} & $+3.8$ & \textbf{\boldmath $+4.1$} & \gapchip{green!3}{$+0.2$\,{\scriptsize$\pm$1.3}} & $+3.1$ & \textbf{\boldmath $+4.8$} & \gapchip{green!23}{$+1.6$\,{\scriptsize$\pm$0.8}} \\
 & 8 & \textbf{\boldmath $+10.1$} & $+9.8$ & \gapchip{red!4}{$-0.3$\,{\scriptsize$\pm$2.7}} & \textbf{\boldmath $+10.1$} & $+9.1$ & \gapchip{red!15}{$-1.0$\,{\scriptsize$\pm$2.2}} & $+1.9$ & $+1.9$ & \gapchip{none}{$+0.0$\,{\scriptsize$\pm$2.2}} & $+5.0$ & \textbf{\boldmath $+5.4$} & \gapchip{green!6}{$+0.4$\,{\scriptsize$\pm$2.7}} & \textbf{\boldmath $+4.9$} & $+3.6$ & \gapchip{red!18}{$-1.3$\,{\scriptsize$\pm$2.6}} & \textbf{\boldmath $+4.0$} & $+3.8$ & \gapchip{red!2}{$-0.1$\,{\scriptsize$\pm$1.3}} \\
 & 16 & $+12.7$ & \textbf{\boldmath $+15.2$} & \gapchip{green!28}{$+2.5$\,{\scriptsize$\pm$2.8}} & \textbf{\boldmath $+14.3$} & $+13.7$ & \gapchip{red!8}{$-0.6$\,{\scriptsize$\pm$2.3}} & $+1.1$ & \textbf{\boldmath $+1.4$} & \gapchip{green!4}{$+0.3$\,{\scriptsize$\pm$0.6}} & $+7.5$ & \textbf{\boldmath $+7.7$} & \gapchip{green!3}{$+0.2$\,{\scriptsize$\pm$5.2}} & $+9.4$ & \textbf{\boldmath $+9.5$} & \gapchip{green!2}{$+0.1$\,{\scriptsize$\pm$2.5}} & $+7.9$ & $+7.9$ & \gapchip{none}{$+0.0$\,{\scriptsize$\pm$0.7}} \\
\midrule[\heavyrulewidth]
\multirow{6}{*}{\rotatebox[origin=c]{90}{\scriptsize\begin{tabular}{@{}c@{}}\raisebox{-0.15em}{\includegraphics[height=0.9em]{logos/deepseek.png}}~DeepSeek\\ Math-7B\end{tabular}}} & base & \multicolumn{3}{c}{18.2} & \multicolumn{3}{c}{10.4} & \multicolumn{3}{c}{0.0} & \multicolumn{3}{c}{1.9} & \multicolumn{3}{c}{6.5} & \multicolumn{3}{c}{1.7} \\
 & 1 & $-3.4$ & \textbf{\boldmath $-3.2$} & \gapchip{green!3}{$+0.2$\,{\scriptsize$\pm$0.3}} & \textbf{\boldmath $-1.3$} & $-1.7$ & \gapchip{red!6}{$-0.4$\,{\scriptsize$\pm$0.3}} & \textbf{\boldmath $+0.1$} & $+0.0$ & \gapchip{red!1}{$-0.1$\,{\scriptsize$\pm$0.1}} & \textbf{\boldmath $+0.4$} & $-0.1$ & \gapchip{red!7}{$-0.5$\,{\scriptsize$\pm$0.4}} & \textbf{\boldmath $-2.0$} & $-2.7$ & \gapchip{red!10}{$-0.7$\,{\scriptsize$\pm$0.2}} & $-0.1$ & \textbf{\boldmath $+0.0$} & \gapchip{green!1}{$+0.1$\,{\scriptsize$\pm$0.1}} \\
 & 2 & $-3.4$ & \textbf{\boldmath $-3.1$} & \gapchip{green!4}{$+0.3$\,{\scriptsize$\pm$0.3}} & $-1.0$ & \textbf{\boldmath $-0.9$} & \gapchip{green!2}{$+0.1$\,{\scriptsize$\pm$0.3}} & $+0.1$ & $+0.1$ & \gapchip{none}{$+0.0$\,{\scriptsize$\pm$0.2}} & $+0.4$ & \textbf{\boldmath $+0.5$} & \gapchip{green!1}{$+0.1$\,{\scriptsize$\pm$1.0}} & $-2.1$ & \textbf{\boldmath $-1.7$} & \gapchip{green!5}{$+0.3$\,{\scriptsize$\pm$0.2}} & $+0.0$ & $+0.0$ & \gapchip{none}{$+0.0$\,{\scriptsize$\pm$0.1}} \\
 & 4 & $-1.8$ & \textbf{\boldmath $-1.1$} & \gapchip{green!10}{$+0.7$\,{\scriptsize$\pm$0.1}} & $-0.8$ & \textbf{\boldmath $-0.3$} & \gapchip{green!7}{$+0.5$\,{\scriptsize$\pm$0.3}} & \textbf{\boldmath $+0.2$} & $+0.1$ & \gapchip{red!1}{$-0.1$\,{\scriptsize$\pm$0.2}} & $+0.6$ & \textbf{\boldmath $+0.9$} & \gapchip{green!3}{$+0.2$\,{\scriptsize$\pm$0.7}} & $-1.8$ & $-1.8$ & \gapchip{green!1}{$+0.1$\,{\scriptsize$\pm$0.3}} & \textbf{\boldmath $+0.2$} & $+0.1$ & \gapchip{red!2}{$-0.1$\,{\scriptsize$\pm$0.1}} \\
 & 8 & $+1.1$ & \textbf{\boldmath $+2.3$} & \gapchip{green!18}{$+1.3$\,{\scriptsize$\pm$0.5}} & $+0.9$ & \textbf{\boldmath $+1.8$} & \gapchip{green!13}{$+0.9$\,{\scriptsize$\pm$0.7}} & \textbf{\boldmath $+0.6$} & $+0.0$ & \gapchip{red!8}{$-0.6$\,{\scriptsize$\pm$0.6}} & \textbf{\boldmath $+2.3$} & $+1.5$ & \gapchip{red!12}{$-0.8$\,{\scriptsize$\pm$0.8}} & \textbf{\boldmath $-0.9$} & $-1.0$ & \gapchip{none}{$+0.0$\,{\scriptsize$\pm$0.6}} & \textbf{\boldmath $+0.4$} & $+0.3$ & \gapchip{red!1}{$-0.1$\,{\scriptsize$\pm$0.1}} \\
 & 16 & \textbf{\boldmath $+3.6$} & $+0.5$ & \gapchip{red!28}{$-3.1$\,{\scriptsize$\pm$1.6}} & \textbf{\boldmath $+2.2$} & $+0.4$ & \gapchip{red!25}{$-1.8$\,{\scriptsize$\pm$0.1}} & \textbf{\boldmath $+0.3$} & $+0.0$ & \gapchip{red!4}{$-0.3$\,{\scriptsize$\pm$0.3}} & \textbf{\boldmath $+1.2$} & $+0.2$ & \gapchip{red!15}{$-1.0$\,{\scriptsize$\pm$0.9}} & \textbf{\boldmath $+0.1$} & $-0.3$ & \gapchip{red!5}{$-0.3$\,{\scriptsize$\pm$0.7}} & \textbf{\boldmath $+0.7$} & $+0.1$ & \gapchip{red!8}{$-0.6$\,{\scriptsize$\pm$0.3}} \\
\midrule
\multirow{6}{*}{\rotatebox[origin=c]{90}{\scriptsize\begin{tabular}{@{}c@{}}\raisebox{-0.15em}{\includegraphics[height=0.9em]{logos/qwen.png}}~Qwen2.5\\ Math-7B\end{tabular}}} & base & \multicolumn{3}{c}{46.4} & \multicolumn{3}{c}{44.9} & \multicolumn{3}{c}{4.2} & \multicolumn{3}{c}{30.6} & \multicolumn{3}{c}{17.9} & \multicolumn{3}{c}{19.9} \\
 & 1 & $+3.8$ & \textbf{\boldmath $+4.3$} & \gapchip{green!6}{$+0.4$\,{\scriptsize$\pm$0.9}} & $+1.8$ & \textbf{\boldmath $+2.9$} & \gapchip{green!15}{$+1.1$\,{\scriptsize$\pm$0.8}} & \textbf{\boldmath $+6.6$} & $+4.0$ & \gapchip{red!28}{$-2.6$\,{\scriptsize$\pm$0.9}} & \textbf{\boldmath $+2.0$} & $-1.0$ & \gapchip{red!28}{$-3.0$\,{\scriptsize$\pm$1.7}} & \textbf{\boldmath $+1.5$} & $+1.4$ & \gapchip{red!1}{$+0.0$\,{\scriptsize$\pm$0.4}} & $+1.2$ & \textbf{\boldmath $+1.7$} & \gapchip{green!7}{$+0.5$\,{\scriptsize$\pm$0.4}} \\
 & 2 & $+5.0$ & \textbf{\boldmath $+9.9$} & \gapchip{green!28}{$+4.9$\,{\scriptsize$\pm$2.6}} & $+2.8$ & \textbf{\boldmath $+5.4$} & \gapchip{green!28}{$+2.7$\,{\scriptsize$\pm$1.9}} & \textbf{\boldmath $+4.9$} & $+4.5$ & \gapchip{red!6}{$-0.4$\,{\scriptsize$\pm$1.1}} & $+0.6$ & \textbf{\boldmath $+4.3$} & \gapchip{green!28}{$+3.7$\,{\scriptsize$\pm$2.7}} & $+2.4$ & \textbf{\boldmath $+3.0$} & \gapchip{green!8}{$+0.6$\,{\scriptsize$\pm$1.5}} & $+2.3$ & \textbf{\boldmath $+3.8$} & \gapchip{green!21}{$+1.5$\,{\scriptsize$\pm$1.2}} \\
 & 4 & $+11.2$ & \textbf{\boldmath $+16.5$} & \gapchip{green!28}{$+5.2$\,{\scriptsize$\pm$2.1}} & $+8.5$ & \textbf{\boldmath $+12.0$} & \gapchip{green!28}{$+3.6$\,{\scriptsize$\pm$1.7}} & \textbf{\boldmath $+6.5$} & $+5.7$ & \gapchip{red!10}{$-0.7$\,{\scriptsize$\pm$1.3}} & $+5.5$ & \textbf{\boldmath $+8.4$} & \gapchip{green!28}{$+2.9$\,{\scriptsize$\pm$2.4}} & $+5.1$ & \textbf{\boldmath $+7.0$} & \gapchip{green!27}{$+1.9$\,{\scriptsize$\pm$1.3}} & $+5.0$ & \textbf{\boldmath $+6.8$} & \gapchip{green!25}{$+1.8$\,{\scriptsize$\pm$1.2}} \\
 & 8 & $+14.5$ & \textbf{\boldmath $+22.4$} & \gapchip{green!28}{$+7.9$\,{\scriptsize$\pm$1.7}} & $+11.3$ & \textbf{\boldmath $+13.2$} & \gapchip{green!27}{$+1.9$\,{\scriptsize$\pm$0.2}} & $+5.8$ & \textbf{\boldmath $+7.2$} & \gapchip{green!19}{$+1.4$\,{\scriptsize$\pm$0.7}} & $+5.8$ & \textbf{\boldmath $+9.8$} & \gapchip{green!28}{$+4.0$\,{\scriptsize$\pm$4.0}} & $+5.4$ & \textbf{\boldmath $+9.3$} & \gapchip{green!28}{$+3.9$\,{\scriptsize$\pm$0.3}} & $+7.2$ & \textbf{\boldmath $+7.4$} & \gapchip{green!3}{$+0.2$\,{\scriptsize$\pm$1.4}} \\
 & 16 & $+18.4$ & \textbf{\boldmath $+23.1$} & \gapchip{green!28}{$+4.7$\,{\scriptsize$\pm$4.3}} & $+14.0$ & \textbf{\boldmath $+14.2$} & \gapchip{green!3}{$+0.2$\,{\scriptsize$\pm$0.6}} & \textbf{\boldmath $+7.8$} & $+3.6$ & \gapchip{red!28}{$-4.2$\,{\scriptsize$\pm$1.0}} & $+11.2$ & $+11.2$ & \gapchip{none}{$+0.0$\,{\scriptsize$\pm$4.2}} & $+8.5$ & \textbf{\boldmath $+10.6$} & \gapchip{green!28}{$+2.1$\,{\scriptsize$\pm$2.1}} & \textbf{\boldmath $+8.5$} & $+8.2$ & \gapchip{red!3}{$-0.2$\,{\scriptsize$\pm$0.8}} \\
\bottomrule
\end{tabular}
}\\[0.8em]
{\small (b) greedy}\\[0.2em]
\resizebox{\linewidth}{!}{\begin{tabular}{c l cccccc@{\hspace{1.2em}}cccccccccccc}
\toprule
 & & \multicolumn{6}{c}{in-distribution} & \multicolumn{12}{c}{out-of-distribution} \\
\cmidrule(lr){3-8}\cmidrule(lr){9-20}
 & & \multicolumn{3}{c}{GSM8K} & \multicolumn{3}{c}{MATH-500} & \multicolumn{3}{c}{AIME24} & \multicolumn{3}{c}{AMC23} & \multicolumn{3}{c}{Minerva} & \multicolumn{3}{c}{Olympiad} \\
\cmidrule(lr){3-5}\cmidrule(lr){6-8}\cmidrule(lr){9-11}\cmidrule(lr){12-14}\cmidrule(lr){15-17}\cmidrule(lr){18-20}
 & $u$ & GRPO & LlMBO & $\Delta$ & GRPO & LlMBO & $\Delta$ & GRPO & LlMBO & $\Delta$ & GRPO & LlMBO & $\Delta$ & GRPO & LlMBO & $\Delta$ & GRPO & LlMBO & $\Delta$ \\
\midrule
\multirow{6}{*}{\rotatebox[origin=c]{90}{\scriptsize\begin{tabular}{@{}c@{}}\raisebox{-0.15em}{\includegraphics[height=0.9em]{logos/llama.png}}~Llama3.1\\ 8B\end{tabular}}} & base & \multicolumn{3}{c}{43.9} & \multicolumn{3}{c}{16.4} & \multicolumn{3}{c}{0.0} & \multicolumn{3}{c}{5.0} & \multicolumn{3}{c}{7.4} & \multicolumn{3}{c}{2.8} \\
 & 1 & $+1.2$ & \textbf{\boldmath $+2.2$} & \gapchip{green!16}{$+1.1$\,{\scriptsize$\pm$0.5}} & $-0.3$ & \textbf{\boldmath $+0.1$} & \gapchip{green!1}{$+0.1$\,{\scriptsize$\pm$0.7}} & $+0.0$ & $+0.0$ & \gapchip{none}{$+0.0$\,{\scriptsize$\pm$0.0}} & $-1.6$ & \textbf{\boldmath $+0.0$} & \gapchip{green!28}{$+2.5$\,{\scriptsize$\pm$0.0}} & $-1.2$ & \textbf{\boldmath $-0.7$} & \gapchip{green!7}{$+0.5$\,{\scriptsize$\pm$1.0}} & $-0.4$ & \textbf{\boldmath $+0.0$} & \gapchip{green!3}{$+0.2$\,{\scriptsize$\pm$0.3}} \\
 & 2 & $+1.5$ & \textbf{\boldmath $+1.6$} & \gapchip{red!1}{$-0.1$\,{\scriptsize$\pm$0.5}} & $-0.4$ & \textbf{\boldmath $-0.3$} & \gapchip{green!3}{$+0.2$\,{\scriptsize$\pm$0.3}} & $+0.0$ & $+0.0$ & \gapchip{none}{$+0.0$\,{\scriptsize$\pm$0.0}} & $-2.2$ & \textbf{\boldmath $+0.0$} & \gapchip{green!28}{$+2.5$\,{\scriptsize$\pm$1.4}} & \textbf{\boldmath $-1.2$} & $-2.0$ & \gapchip{red!3}{$-0.2$\,{\scriptsize$\pm$0.6}} & \textbf{\boldmath $-0.4$} & $-0.5$ & \gapchip{red!1}{$-0.1$\,{\scriptsize$\pm$0.6}} \\
 & 4 & $+1.4$ & \textbf{\boldmath $+2.7$} & \gapchip{green!17}{$+1.2$\,{\scriptsize$\pm$0.8}} & $+0.1$ & \textbf{\boldmath $+1.2$} & \gapchip{green!21}{$+1.5$\,{\scriptsize$\pm$1.0}} & $+0.0$ & $+0.0$ & \gapchip{none}{$+0.0$\,{\scriptsize$\pm$0.0}} & \textbf{\boldmath $-1.6$} & $-1.7$ & \gapchip{none}{$+0.0$\,{\scriptsize$\pm$1.4}} & \textbf{\boldmath $-1.5$} & $-2.0$ & \gapchip{none}{$+0.0$\,{\scriptsize$\pm$1.0}} & $-0.4$ & $-0.4$ & \gapchip{red!1}{$+0.0$\,{\scriptsize$\pm$0.2}} \\
 & 8 & $+0.9$ & \textbf{\boldmath $+2.2$} & \gapchip{green!18}{$+1.3$\,{\scriptsize$\pm$0.9}} & $+0.3$ & \textbf{\boldmath $+0.7$} & \gapchip{green!5}{$+0.3$\,{\scriptsize$\pm$0.5}} & $+0.0$ & $+0.0$ & \gapchip{none}{$+0.0$\,{\scriptsize$\pm$0.0}} & \textbf{\boldmath $-2.5$} & $-3.3$ & \gapchip{red!12}{$-0.8$\,{\scriptsize$\pm$0.8}} & $-2.7$ & \textbf{\boldmath $-1.2$} & \gapchip{green!21}{$+1.5$\,{\scriptsize$\pm$0.9}} & $-0.3$ & \textbf{\boldmath $-0.2$} & \gapchip{green!2}{$+0.1$\,{\scriptsize$\pm$0.8}} \\
 & 16 & $+0.7$ & \textbf{\boldmath $+2.9$} & \gapchip{green!28}{$+2.2$\,{\scriptsize$\pm$0.4}} & \textbf{\boldmath $+0.1$} & $-0.8$ & \gapchip{red!13}{$-0.9$\,{\scriptsize$\pm$0.9}} & $+0.0$ & $+0.0$ & \gapchip{none}{$+0.0$\,{\scriptsize$\pm$0.0}} & $-1.7$ & $-1.7$ & \gapchip{none}{$+0.0$\,{\scriptsize$\pm$2.5}} & $-1.6$ & \textbf{\boldmath $-1.5$} & \gapchip{green!2}{$+0.1$\,{\scriptsize$\pm$0.9}} & \textbf{\boldmath $+0.0$} & $-0.4$ & \gapchip{red!6}{$-0.4$\,{\scriptsize$\pm$0.3}} \\
\midrule
\multirow{6}{*}{\rotatebox[origin=c]{90}{\scriptsize\begin{tabular}{@{}c@{}}\raisebox{-0.15em}{\includegraphics[height=0.9em]{logos/mistral.png}}~Mistral\\ 7B\end{tabular}}} & base & \multicolumn{3}{c}{22.4} & \multicolumn{3}{c}{7.4} & \multicolumn{3}{c}{0.0} & \multicolumn{3}{c}{0.0} & \multicolumn{3}{c}{5.5} & \multicolumn{3}{c}{1.5} \\
 & 1 & $-0.1$ & \textbf{\boldmath $+1.1$} & \gapchip{green!10}{$+0.7$\,{\scriptsize$\pm$0.5}} & $+1.1$ & \textbf{\boldmath $+1.3$} & \gapchip{red!1}{$-0.1$\,{\scriptsize$\pm$0.5}} & $+0.0$ & $+0.0$ & \gapchip{none}{$+0.0$\,{\scriptsize$\pm$0.0}} & \textbf{\boldmath $+0.9$} & $+0.0$ & \gapchip{red!12}{$-0.8$\,{\scriptsize$\pm$0.8}} & \textbf{\boldmath $-0.9$} & $-1.3$ & \gapchip{red!2}{$-0.1$\,{\scriptsize$\pm$0.5}} & \textbf{\boldmath $+0.1$} & $+0.0$ & \gapchip{red!1}{$+0.0$\,{\scriptsize$\pm$0.4}} \\
 & 2 & $-0.3$ & \textbf{\boldmath $+0.8$} & \gapchip{green!11}{$+0.8$\,{\scriptsize$\pm$1.1}} & $+1.0$ & $+1.0$ & \gapchip{green!8}{$+0.6$\,{\scriptsize$\pm$0.5}} & $+0.0$ & $+0.0$ & \gapchip{none}{$+0.0$\,{\scriptsize$\pm$0.0}} & $+0.9$ & \textbf{\boldmath $+3.3$} & \gapchip{green!23}{$+1.7$\,{\scriptsize$\pm$1.7}} & \textbf{\boldmath $-1.3$} & $-1.7$ & \gapchip{green!2}{$+0.1$\,{\scriptsize$\pm$1.0}} & $+0.1$ & $+0.1$ & \gapchip{red!3}{$-0.2$\,{\scriptsize$\pm$0.1}} \\
 & 4 & $+0.2$ & \textbf{\boldmath $+1.0$} & \gapchip{green!22}{$+1.6$\,{\scriptsize$\pm$0.5}} & $+0.7$ & \textbf{\boldmath $+1.0$} & \gapchip{green!4}{$+0.3$\,{\scriptsize$\pm$0.5}} & $+0.0$ & $+0.0$ & \gapchip{none}{$+0.0$\,{\scriptsize$\pm$0.0}} & \textbf{\boldmath $+2.2$} & $+1.7$ & \gapchip{none}{$+0.0$\,{\scriptsize$\pm$1.4}} & $-1.5$ & \textbf{\boldmath $-1.2$} & \gapchip{green!2}{$+0.1$\,{\scriptsize$\pm$0.4}} & $+0.2$ & $+0.2$ & \gapchip{red!2}{$-0.1$\,{\scriptsize$\pm$0.3}} \\
 & 8 & \textbf{\boldmath $+0.6$} & $-0.9$ & \gapchip{red!21}{$-1.5$\,{\scriptsize$\pm$1.2}} & $+0.9$ & \textbf{\boldmath $+2.3$} & \gapchip{green!19}{$+1.3$\,{\scriptsize$\pm$1.0}} & $+0.0$ & $+0.0$ & \gapchip{none}{$+0.0$\,{\scriptsize$\pm$0.0}} & $+1.7$ & $+1.7$ & \gapchip{none}{$+0.0$\,{\scriptsize$\pm$0.0}} & $-1.5$ & $-1.5$ & \gapchip{none}{$+0.0$\,{\scriptsize$\pm$0.4}} & $-0.1$ & \textbf{\boldmath $+0.0$} & \gapchip{green!1}{$+0.0$\,{\scriptsize$\pm$0.5}} \\
 & 16 & $-0.9$ & \textbf{\boldmath $+0.1$} & \gapchip{green!13}{$+1.0$\,{\scriptsize$\pm$1.1}} & $+1.1$ & \textbf{\boldmath $+2.0$} & \gapchip{green!12}{$+0.9$\,{\scriptsize$\pm$0.3}} & $+0.0$ & $+0.0$ & \gapchip{none}{$+0.0$\,{\scriptsize$\pm$0.0}} & \textbf{\boldmath $+1.7$} & $+0.8$ & \gapchip{red!12}{$-0.8$\,{\scriptsize$\pm$0.8}} & $-0.5$ & \textbf{\boldmath $-0.2$} & \gapchip{green!3}{$+0.2$\,{\scriptsize$\pm$1.0}} & $+0.3$ & \textbf{\boldmath $+0.5$} & \gapchip{green!2}{$+0.1$\,{\scriptsize$\pm$0.3}} \\
\midrule
\multirow{6}{*}{\rotatebox[origin=c]{90}{\scriptsize\begin{tabular}{@{}c@{}}\raisebox{-0.15em}{\includegraphics[height=0.9em]{logos/deepseek.png}}~DeepSeek\\ 7B\end{tabular}}} & base & \multicolumn{3}{c}{2.7} & \multicolumn{3}{c}{3.0} & \multicolumn{3}{c}{0.0} & \multicolumn{3}{c}{0.0} & \multicolumn{3}{c}{1.5} & \multicolumn{3}{c}{0.7} \\
 & 1 & $-0.9$ & \textbf{\boldmath $-0.2$} & \gapchip{green!16}{$+1.2$\,{\scriptsize$\pm$0.3}} & $+1.2$ & \textbf{\boldmath $+1.5$} & \gapchip{green!8}{$+0.6$\,{\scriptsize$\pm$0.6}} & $+0.0$ & $+0.0$ & \gapchip{none}{$+0.0$\,{\scriptsize$\pm$0.0}} & $+0.0$ & $+0.0$ & \gapchip{none}{$+0.0$\,{\scriptsize$\pm$0.0}} & $+0.8$ & \textbf{\boldmath $+0.9$} & \gapchip{green!3}{$+0.2$\,{\scriptsize$\pm$0.2}} & $+0.1$ & $+0.1$ & \gapchip{green!1}{$+0.1$\,{\scriptsize$\pm$0.1}} \\
 & 2 & $-0.9$ & \textbf{\boldmath $-0.5$} & \gapchip{green!8}{$+0.6$\,{\scriptsize$\pm$0.5}} & $+1.6$ & \textbf{\boldmath $+2.2$} & \gapchip{green!15}{$+1.1$\,{\scriptsize$\pm$0.4}} & $+0.0$ & $+0.0$ & \gapchip{none}{$+0.0$\,{\scriptsize$\pm$0.0}} & $+0.0$ & $+0.0$ & \gapchip{none}{$+0.0$\,{\scriptsize$\pm$0.0}} & $+0.7$ & $+0.7$ & \gapchip{green!2}{$+0.1$\,{\scriptsize$\pm$0.5}} & \textbf{\boldmath $+0.3$} & $+0.2$ & \gapchip{none}{$+0.0$\,{\scriptsize$\pm$0.2}} \\
 & 4 & \textbf{\boldmath $-0.3$} & $-0.5$ & \gapchip{none}{$+0.0$\,{\scriptsize$\pm$0.7}} & $+1.9$ & $+1.9$ & \gapchip{green!4}{$+0.3$\,{\scriptsize$\pm$0.1}} & $+0.0$ & $+0.0$ & \gapchip{none}{$+0.0$\,{\scriptsize$\pm$0.0}} & \textbf{\boldmath $+0.3$} & $+0.0$ & \gapchip{none}{$+0.0$\,{\scriptsize$\pm$0.0}} & $+1.0$ & \textbf{\boldmath $+1.2$} & \gapchip{green!2}{$+0.1$\,{\scriptsize$\pm$1.2}} & \textbf{\boldmath $+0.1$} & $+0.0$ & \gapchip{red!1}{$-0.1$\,{\scriptsize$\pm$0.2}} \\
 & 8 & $-0.4$ & \textbf{\boldmath $-0.3$} & \gapchip{green!6}{$+0.4$\,{\scriptsize$\pm$0.6}} & $+1.3$ & \textbf{\boldmath $+2.4$} & \gapchip{green!11}{$+0.8$\,{\scriptsize$\pm$0.4}} & $+0.0$ & $+0.0$ & \gapchip{none}{$+0.0$\,{\scriptsize$\pm$0.0}} & $+0.0$ & $+0.0$ & \gapchip{none}{$+0.0$\,{\scriptsize$\pm$0.0}} & \textbf{\boldmath $+1.3$} & $+0.2$ & \gapchip{red!10}{$-0.7$\,{\scriptsize$\pm$0.7}} & $+0.1$ & \textbf{\boldmath $+0.3$} & \gapchip{green!5}{$+0.4$\,{\scriptsize$\pm$0.1}} \\
 & 16 & \textbf{\boldmath $-0.1$} & $-0.7$ & \gapchip{red!7}{$-0.5$\,{\scriptsize$\pm$0.4}} & $+1.7$ & \textbf{\boldmath $+2.4$} & \gapchip{green!9}{$+0.7$\,{\scriptsize$\pm$0.4}} & $+0.0$ & $+0.0$ & \gapchip{none}{$+0.0$\,{\scriptsize$\pm$0.0}} & $+0.0$ & \textbf{\boldmath $+0.8$} & \gapchip{green!12}{$+0.8$\,{\scriptsize$\pm$0.8}} & \textbf{\boldmath $+0.9$} & $+0.1$ & \gapchip{red!10}{$-0.7$\,{\scriptsize$\pm$0.4}} & \textbf{\boldmath $+0.2$} & $-0.2$ & \gapchip{red!6}{$-0.4$\,{\scriptsize$\pm$0.0}} \\
\midrule
\multirow{6}{*}{\rotatebox[origin=c]{90}{\scriptsize\begin{tabular}{@{}c@{}}\raisebox{-0.15em}{\includegraphics[height=0.9em]{logos/qwen.png}}~Qwen2.5\\ 7B\end{tabular}}} & base & \multicolumn{3}{c}{88.0} & \multicolumn{3}{c}{64.0} & \multicolumn{3}{c}{0.0} & \multicolumn{3}{c}{30.0} & \multicolumn{3}{c}{32.7} & \multicolumn{3}{c}{28.6} \\
 & 1 & $-0.4$ & \textbf{\boldmath $+0.0$} & \gapchip{green!11}{$+0.8$\,{\scriptsize$\pm$0.9}} & $+1.0$ & \textbf{\boldmath $+1.5$} & \gapchip{green!2}{$+0.1$\,{\scriptsize$\pm$0.3}} & \textbf{\boldmath $+5.8$} & $+4.4$ & \gapchip{red!28}{$-4.4$\,{\scriptsize$\pm$2.9}} & $+5.6$ & \textbf{\boldmath $+6.7$} & \gapchip{green!28}{$+5.8$\,{\scriptsize$\pm$2.2}} & $-1.0$ & \textbf{\boldmath $-0.6$} & \gapchip{green!9}{$+0.6$\,{\scriptsize$\pm$2.1}} & $+1.7$ & \textbf{\boldmath $+2.1$} & \gapchip{green!14}{$+1.0$\,{\scriptsize$\pm$0.8}} \\
 & 2 & \textbf{\boldmath $-0.3$} & $-0.9$ & \gapchip{red!15}{$-1.0$\,{\scriptsize$\pm$0.6}} & $+0.9$ & \textbf{\boldmath $+1.5$} & \gapchip{green!16}{$+1.1$\,{\scriptsize$\pm$0.7}} & $+5.8$ & \textbf{\boldmath $+11.1$} & \gapchip{green!28}{$+5.6$\,{\scriptsize$\pm$2.2}} & \textbf{\boldmath $+4.1$} & $+3.3$ & \gapchip{green!23}{$+1.7$\,{\scriptsize$\pm$1.7}} & \textbf{\boldmath $-1.6$} & $-2.6$ & \gapchip{red!28}{$-3.3$\,{\scriptsize$\pm$4.3}} & $+0.5$ & \textbf{\boldmath $+2.6$} & \gapchip{green!28}{$+2.5$\,{\scriptsize$\pm$0.4}} \\
 & 4 & \textbf{\boldmath $-1.3$} & $-1.4$ & \gapchip{red!8}{$-0.6$\,{\scriptsize$\pm$0.6}} & $+0.8$ & \textbf{\boldmath $+2.4$} & \gapchip{green!28}{$+3.2$\,{\scriptsize$\pm$0.4}} & \textbf{\boldmath $+4.6$} & $+3.3$ & \gapchip{red!16}{$-1.1$\,{\scriptsize$\pm$1.1}} & $+4.1$ & \textbf{\boldmath $+6.7$} & \gapchip{red!28}{$-2.5$\,{\scriptsize$\pm$5.2}} & \textbf{\boldmath $+1.1$} & $-1.6$ & \gapchip{green!2}{$+0.1$\,{\scriptsize$\pm$2.7}} & $+1.2$ & \textbf{\boldmath $+1.6$} & \gapchip{green!7}{$+0.5$\,{\scriptsize$\pm$1.1}} \\
 & 8 & \textbf{\boldmath $+0.5$} & $-0.8$ & \gapchip{red!18}{$-1.3$\,{\scriptsize$\pm$0.4}} & \textbf{\boldmath $+2.7$} & $+1.1$ & \gapchip{red!22}{$-1.6$\,{\scriptsize$\pm$2.3}} & \textbf{\boldmath $+5.6$} & $+4.4$ & \gapchip{red!16}{$-1.1$\,{\scriptsize$\pm$2.2}} & $+6.7$ & $+6.7$ & \gapchip{none}{$+0.0$\,{\scriptsize$\pm$2.5}} & \textbf{\boldmath $+0.5$} & $-3.7$ & \gapchip{red!28}{$-4.2$\,{\scriptsize$\pm$3.3}} & \textbf{\boldmath $+2.8$} & $+2.1$ & \gapchip{red!9}{$-0.6$\,{\scriptsize$\pm$1.2}} \\
 & 16 & \textbf{\boldmath $+0.5$} & $-0.1$ & \gapchip{red!8}{$-0.6$\,{\scriptsize$\pm$0.4}} & \textbf{\boldmath $+3.7$} & $+3.4$ & \gapchip{red!5}{$-0.3$\,{\scriptsize$\pm$0.6}} & $+7.8$ & $+7.8$ & \gapchip{none}{$+0.0$\,{\scriptsize$\pm$1.9}} & $+3.3$ & \textbf{\boldmath $+7.5$} & \gapchip{green!28}{$+4.2$\,{\scriptsize$\pm$1.7}} & \textbf{\boldmath $+1.7$} & $+1.2$ & \gapchip{red!7}{$-0.5$\,{\scriptsize$\pm$4.1}} & $+2.0$ & \textbf{\boldmath $+2.6$} & \gapchip{green!8}{$+0.5$\,{\scriptsize$\pm$0.2}} \\
\midrule[\heavyrulewidth]
\multirow{6}{*}{\rotatebox[origin=c]{90}{\scriptsize\begin{tabular}{@{}c@{}}\raisebox{-0.15em}{\includegraphics[height=0.9em]{logos/deepseek.png}}~DeepSeek\\ Math-7B\end{tabular}}} & base & \multicolumn{3}{c}{28.7} & \multicolumn{3}{c}{21.2} & \multicolumn{3}{c}{0.0} & \multicolumn{3}{c}{15.0} & \multicolumn{3}{c}{9.6} & \multicolumn{3}{c}{5.9} \\
 & 1 & $+2.7$ & \textbf{\boldmath $+6.8$} & \gapchip{green!28}{$+5.6$\,{\scriptsize$\pm$3.0}} & \textbf{\boldmath $+1.4$} & $+0.7$ & \gapchip{red!16}{$-1.1$\,{\scriptsize$\pm$1.2}} & $+0.4$ & \textbf{\boldmath $+1.1$} & \gapchip{green!16}{$+1.1$\,{\scriptsize$\pm$1.1}} & $-7.2$ & \textbf{\boldmath $-5.8$} & \gapchip{green!28}{$+5.0$\,{\scriptsize$\pm$3.8}} & $+3.4$ & \textbf{\boldmath $+4.3$} & \gapchip{green!3}{$+0.2$\,{\scriptsize$\pm$0.6}} & $+0.7$ & \textbf{\boldmath $+1.1$} & \gapchip{green!8}{$+0.5$\,{\scriptsize$\pm$0.3}} \\
 & 2 & $+2.1$ & \textbf{\boldmath $+3.2$} & \gapchip{green!27}{$+1.9$\,{\scriptsize$\pm$5.2}} & $+0.6$ & \textbf{\boldmath $+1.9$} & \gapchip{green!13}{$+0.9$\,{\scriptsize$\pm$1.1}} & \textbf{\boldmath $+0.4$} & $+0.0$ & \gapchip{red!16}{$-1.1$\,{\scriptsize$\pm$1.1}} & $-9.4$ & \textbf{\boldmath $-4.2$} & \gapchip{green!28}{$+4.2$\,{\scriptsize$\pm$3.0}} & \textbf{\boldmath $+3.9$} & $+2.7$ & \gapchip{red!7}{$-0.5$\,{\scriptsize$\pm$0.7}} & $+0.0$ & \textbf{\boldmath $+0.9$} & \gapchip{green!7}{$+0.5$\,{\scriptsize$\pm$0.7}} \\
 & 4 & $+1.7$ & \textbf{\boldmath $+7.4$} & \gapchip{green!28}{$+5.5$\,{\scriptsize$\pm$3.4}} & $-0.1$ & \textbf{\boldmath $+2.5$} & \gapchip{green!28}{$+2.5$\,{\scriptsize$\pm$0.4}} & $+0.0$ & \textbf{\boldmath $+2.2$} & \gapchip{green!28}{$+2.2$\,{\scriptsize$\pm$1.1}} & $-6.9$ & \textbf{\boldmath $-6.7$} & \gapchip{none}{$+0.0$\,{\scriptsize$\pm$2.5}} & \textbf{\boldmath $+3.5$} & $+2.7$ & \gapchip{red!12}{$-0.9$\,{\scriptsize$\pm$0.8}} & $-0.5$ & \textbf{\boldmath $+1.2$} & \gapchip{green!23}{$+1.6$\,{\scriptsize$\pm$0.2}} \\
 & 8 & \textbf{\boldmath $+5.5$} & $+4.4$ & \gapchip{red!15}{$-1.1$\,{\scriptsize$\pm$1.2}} & $+0.5$ & \textbf{\boldmath $+0.9$} & \gapchip{green!5}{$+0.3$\,{\scriptsize$\pm$1.2}} & $+0.0$ & \textbf{\boldmath $+2.2$} & \gapchip{green!28}{$+2.2$\,{\scriptsize$\pm$1.1}} & \textbf{\boldmath $-8.3$} & $-9.2$ & \gapchip{red!12}{$-0.8$\,{\scriptsize$\pm$0.8}} & $+2.8$ & \textbf{\boldmath $+4.4$} & \gapchip{green!22}{$+1.6$\,{\scriptsize$\pm$0.7}} & $-0.4$ & \textbf{\boldmath $+0.6$} & \gapchip{green!15}{$+1.1$\,{\scriptsize$\pm$0.3}} \\
 & 16 & $+4.7$ & \textbf{\boldmath $+6.3$} & \gapchip{green!23}{$+1.7$\,{\scriptsize$\pm$3.3}} & $-0.1$ & \textbf{\boldmath $+2.1$} & \gapchip{green!28}{$+2.2$\,{\scriptsize$\pm$1.3}} & $+0.0$ & $+0.0$ & \gapchip{none}{$+0.0$\,{\scriptsize$\pm$0.0}} & \textbf{\boldmath $-7.5$} & $-8.3$ & \gapchip{red!12}{$-0.8$\,{\scriptsize$\pm$4.6}} & $+1.5$ & \textbf{\boldmath $+3.4$} & \gapchip{green!27}{$+2.0$\,{\scriptsize$\pm$0.3}} & $-0.2$ & \textbf{\boldmath $+0.3$} & \gapchip{green!8}{$+0.6$\,{\scriptsize$\pm$0.5}} \\
\midrule
\multirow{6}{*}{\rotatebox[origin=c]{90}{\scriptsize\begin{tabular}{@{}c@{}}\raisebox{-0.15em}{\includegraphics[height=0.9em]{logos/qwen.png}}~Qwen2.5\\ Math-7B\end{tabular}}} & base & \multicolumn{3}{c}{68.8} & \multicolumn{3}{c}{64.2} & \multicolumn{3}{c}{10.0} & \multicolumn{3}{c}{37.5} & \multicolumn{3}{c}{21.7} & \multicolumn{3}{c}{26.4} \\
 & 1 & \textbf{\boldmath $+4.0$} & $+1.9$ & \gapchip{green!28}{$+5.9$\,{\scriptsize$\pm$5.6}} & $+2.1$ & \textbf{\boldmath $+3.1$} & \gapchip{green!28}{$+3.4$\,{\scriptsize$\pm$1.0}} & \textbf{\boldmath $-0.8$} & $-1.1$ & \gapchip{red!16}{$-1.1$\,{\scriptsize$\pm$6.2}} & \textbf{\boldmath $+9.1$} & $+2.5$ & \gapchip{red!12}{$-0.8$\,{\scriptsize$\pm$7.9}} & $+4.4$ & $+4.4$ & \gapchip{green!28}{$+2.3$\,{\scriptsize$\pm$0.1}} & $+3.2$ & \textbf{\boldmath $+5.1$} & \gapchip{green!28}{$+5.5$\,{\scriptsize$\pm$1.7}} \\
 & 2 & $+4.4$ & \textbf{\boldmath $+12.3$} & \gapchip{green!28}{$+3.7$\,{\scriptsize$\pm$4.4}} & $+3.0$ & \textbf{\boldmath $+3.9$} & \gapchip{red!7}{$-0.5$\,{\scriptsize$\pm$1.1}} & \textbf{\boldmath $+2.9$} & $+1.1$ & \gapchip{red!28}{$-2.2$\,{\scriptsize$\pm$4.8}} & $+2.2$ & \textbf{\boldmath $+10.8$} & \gapchip{green!28}{$+3.3$\,{\scriptsize$\pm$0.8}} & $+3.2$ & \textbf{\boldmath $+6.6$} & \gapchip{green!24}{$+1.7$\,{\scriptsize$\pm$3.1}} & $+3.0$ & \textbf{\boldmath $+7.0$} & \gapchip{green!9}{$+0.6$\,{\scriptsize$\pm$1.1}} \\
 & 4 & $+7.9$ & \textbf{\boldmath $+11.9$} & \gapchip{green!28}{$+9.7$\,{\scriptsize$\pm$11.3}} & $+4.4$ & \textbf{\boldmath $+5.7$} & \gapchip{green!28}{$+2.3$\,{\scriptsize$\pm$3.6}} & \textbf{\boldmath $+3.3$} & $+0.0$ & \gapchip{green!16}{$+1.1$\,{\scriptsize$\pm$2.2}} & $+8.8$ & \textbf{\boldmath $+10.0$} & \gapchip{green!28}{$+5.0$\,{\scriptsize$\pm$9.5}} & $+7.5$ & \textbf{\boldmath $+7.6$} & \gapchip{green!28}{$+2.6$\,{\scriptsize$\pm$2.0}} & $+5.7$ & \textbf{\boldmath $+6.7$} & \gapchip{green!28}{$+5.3$\,{\scriptsize$\pm$5.5}} \\
 & 8 & $+10.6$ & \textbf{\boldmath $+13.7$} & \gapchip{green!28}{$+3.1$\,{\scriptsize$\pm$3.2}} & $+3.5$ & \textbf{\boldmath $+5.3$} & \gapchip{green!25}{$+1.8$\,{\scriptsize$\pm$0.8}} & $+0.0$ & \textbf{\boldmath $+6.7$} & \gapchip{green!28}{$+6.7$\,{\scriptsize$\pm$0.0}} & $+10.8$ & \textbf{\boldmath $+13.3$} & \gapchip{green!28}{$+2.5$\,{\scriptsize$\pm$4.3}} & $+7.5$ & \textbf{\boldmath $+10.3$} & \gapchip{green!28}{$+2.8$\,{\scriptsize$\pm$4.6}} & \textbf{\boldmath $+7.5$} & $+7.1$ & \gapchip{red!6}{$-0.4$\,{\scriptsize$\pm$1.0}} \\
 & 16 & $+7.1$ & \textbf{\boldmath $+8.2$} & \gapchip{green!16}{$+1.1$\,{\scriptsize$\pm$4.2}} & $+4.3$ & \textbf{\boldmath $+4.9$} & \gapchip{green!9}{$+0.7$\,{\scriptsize$\pm$0.6}} & $+1.1$ & \textbf{\boldmath $+6.7$} & \gapchip{green!28}{$+5.6$\,{\scriptsize$\pm$2.9}} & $+3.3$ & \textbf{\boldmath $+7.5$} & \gapchip{green!28}{$+4.2$\,{\scriptsize$\pm$6.7}} & $+5.0$ & \textbf{\boldmath $+5.5$} & \gapchip{green!7}{$+0.5$\,{\scriptsize$\pm$2.5}} & \textbf{\boldmath $+7.3$} & $+6.8$ & \gapchip{red!7}{$-0.5$\,{\scriptsize$\pm$1.3}} \\
\bottomrule
\end{tabular}
}
\end{table}

\begin{figure}[!htbp]
\centering
\includegraphics[width=\linewidth]{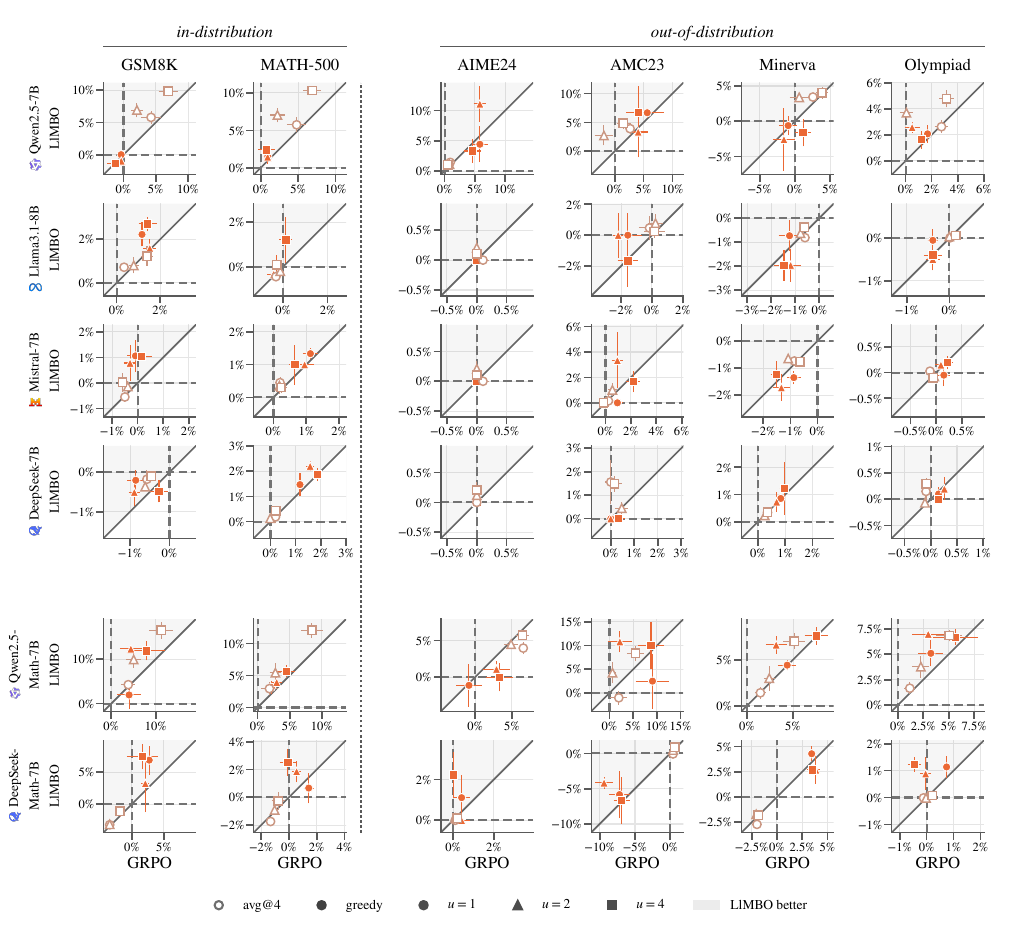}
\caption{Gain over the base model of \llmbo{} against GRPO's, in avg@4 and in greedy accuracy (color; greedy filled, avg@4 open), \llmbo{} trained on the metric of each point, for the one 7-8B base model per family and the two math-specialized 7B models (rows, in two blocks) on each benchmark (columns), each panel on its own axes. Training uses GSM8K and MATH: the left two columns are in-distribution, the four benchmarks right of the dotted line out-of-distribution. One point per $u \le 4$ (marker), mean $\pm$ s.e.\ over seeds. Points in the shaded half above the diagonal favor \llmbo{}; dashed: no gain.}
\label{fig:scatter-main}
\end{figure}

\subsection{Base models across families}
\label{sec:base}
\label{sec:scale}
\label{sec:gains}
\label{sec:per-model-base}

Which models TinyLoRA can move at all is set by its scaling under GRPO (Appendix~\ref{app:scaling-results}): Qwen2.5-7B a large amount, Llama3.1-8B, Mistral-7B, and DeepSeek-7B by at most \othersTinyBest{} points at $u \le 4$. On these three the comparison checks that \llmbo{} is as safe as GRPO.

First consider Qwen2.5-7B. On the one base model TinyLoRA moves, \llmbo{} lies above the diagonal on both in-distribution tasks in avg@4 at every $u \le 4$ (Section~\ref{sec:results}); on GSM8K avg@4 it gains \gainToneMeanQwenUTwoGsm{} points at $u = 2$ against GRPO's \gainGrpoMeanQwenUTwoGsm{}. Greedy GSM8K barely moves under either optimizer, while on greedy MATH-500 \llmbo{} gains \gainGreedyGreedyQwenUOneMathFive{}, \gainGreedyGreedyQwenUTwoMathFive{}, and \gainGreedyGreedyQwenUFourMathFive{} points at $u = 1, 2, 4$ against GRPO's \gainGrpoGreedyQwenUOneMathFive{}, \gainGrpoGreedyQwenUTwoMathFive{}, and \gainGrpoGreedyQwenUFourMathFive{}, a paired gap that grows to \gapGreedyGreedyQwenUFourMathFive{} at $u = 4$. Starting with $u = 8$ the MATH-500 lead is gone in both metrics (\gapToneMeanQwenUEightMathFive{} in avg@4, \gapGreedyGreedyQwenUEightMathFive{} in greedy accuracy; Figure~\ref{fig:gap-main}).

Regarding Llama, Mistral, and DeepSeek we have the following.
The points of both optimizers sit close to the origin and to the diagonal: in avg@4 \llmbo{} is level with GRPO at every $u \le 4$, within a point on both tasks (e.g.\ \gapToneMeanLlamaUOneGsm{}, \gapToneMeanMistralUFourGsm{} and \gapToneMeanDeepseekBaseUFourGsm{} on GSM8K for Llama3.1-8B at $u = 1$, Mistral-7B and DeepSeek-7B at $u = 4$). Where TinyLoRA's subspace holds little, \llmbo{} gives up nothing to GRPO. In greedy accuracy it stays level with or slightly ahead of GRPO on GSM8K on Llama3.1-8B and Mistral-7B at every $u \le 4$ (Llama \gapGreedyGreedyLlamaUOneGsm{}, \gapGreedyGreedyLlamaUTwoGsm{}, \gapGreedyGreedyLlamaUFourGsm{}; Mistral \gapGreedyGreedyMistralUOneGsm{}, \gapGreedyGreedyMistralUTwoGsm{}, \gapGreedyGreedyMistralUFourGsm{}), and up to $u = 16$ (\gapGreedyGreedyLlamaUSixteenGsm{}, \gapGreedyGreedyMistralUSixteenGsm{}). Pooled over the four base models it is level with or ahead of GRPO in greedy accuracy at every $u \le 4$ on both tasks (Section~\ref{sec:greedy-objective}): optimizing the deployed decoding rule directly finds some gain even here.

Beyond $u = 4$ the advantage is specific to the low-parameter end, where a GP can cover the search space within the budget: at $u = 16$ the avg@4 gap turns negative on GSM8K for Mistral (\gapToneMeanMistralUSixteenGsm{}), and at $u \in \{64, 256\}$ \llmbo{} is not yet robust across seeds (Section~\ref{sec:large-u}).

\FloatBarrier
\subsection{Math variants}
\label{sec:math}

The math-specialized models are continued-pretrained on mathematical text; Qwen2.5-Math-7B and DeepSeek-Math-7B are the two at 7B. As between the base models, TinyLoRA moves the Qwen model a lot and the DeepSeek model a little (Appendix~\ref{app:scaling}).

For Qwen2.5-Math-7B both optimizers gain several points, and \llmbo{}'s paired gap is positive at every $u \le 4$ and largest at $u = 4$ (Figure~\ref{fig:scatter-main}). It gains \gainToneMeanMathUFourGsm{} points on GSM8K avg@4 against GRPO's \gainGrpoMeanMathUFourGsm{} and \gainToneMeanMathUFourMathFive{} against \gainGrpoMeanMathUFourMathFive{} on MATH-500 avg@4 with paired gaps of \gapToneMeanMathUFourGsm{} and \gapToneMeanMathUFourMathFive{}. The gaps vary widely across seeds, since single searches find or miss a several-point gain. In greedy accuracy \llmbo{} gains \gainGreedyGreedyMathUTwoGsm{} and \gainGreedyGreedyMathUFourGsm{} points on GSM8K at $u = 2, 4$ against GRPO's \gainGrpoGreedyMathUTwoGsm{} and \gainGrpoGreedyMathUFourGsm{}, and it gains \gainGreedyGreedyMathUFourMathFive{} against \gainGrpoGreedyMathUFourMathFive{} on MATH-500 at $u = 4$. Unlike Qwen2.5-7B, the Math model stays ahead on both primary tasks at $u = 8$ in both metrics (avg@4: \gapToneMeanMathUEightGsm{} on GSM8K, \gapToneMeanMathUEightMathFive{} on MATH-500; greedy: \gapGreedyGreedyMathUEightGsm{} and \gapGreedyGreedyMathUEightMathFive{}; Figure~\ref{fig:gap-main}).

Next for DeepSeek-Math-7B we find that in avg@4 \llmbo{} is level with GRPO at every $u \le 4$ (\gapToneMeanDeepseekUOneGsm{}, \gapToneMeanDeepseekUTwoGsm{} and \gapToneMeanDeepseekUFourGsm{} on GSM8K at $u = 1, 2, 4$), stays level at $u = 8$ and falls behind at $u = 16$ (\gapToneMeanDeepseekUSixteenGsm{} on GSM8K). Training on greedy accuracy pulls it ahead: it leads on greedy GSM8K at every $u \le 4$ (\gapGreedyGreedyDeepseekUOneGsm{}, \gapGreedyGreedyDeepseekUTwoGsm{} and \gapGreedyGreedyDeepseekUFourGsm{} points) and on greedy MATH-500 at $u = 4$ (\gapGreedyGreedyDeepseekUFourMathFive{}). When we pool over the two Math models \llmbo{}'s lead grows with $u$ up to $u = 8$, where it is \poolToneMeanUEightGsmMathVariants{} points on GSM8K avg@4 and \poolToneMeanUEightMathFiveMathVariants{} on MATH-500 (Table~\ref{tab:main}a).

\FloatBarrier
\subsection{Larger adapters}
\label{sec:large-u}

\begin{table}[!htbp]
\centering
\caption{Avg@4 (a) and greedy (b) accuracy at the end of training, for the \nLargeModels{} Qwen2.5 7B models (row blocks), per $u \in \{64, 256\}$ (rows) and benchmark (columns: in-distribution GSM8K and MATH-500, then the four out-of-distribution benchmarks); \llmbo{} is trained on the metric of each subtable (greedy: Section~\ref{sec:greedy-objective}). \emph{base}: accuracy of the base model in that metric (\%). \emph{GRPO}, \emph{\llmbo{}}: gain over the base model (points), mean over seeds (3 per cell); bold: the larger, unless tied. $\Delta$: paired gap \llmbo{} $-$ GRPO, mean $\pm$ s.e.\ over the seed pairs both have, so it can differ from the difference of the two gains; tint: its sign and size (green: \llmbo{} ahead).}
\label{tab:large-u}
{\small (a) avg@4}\\[0.2em]
\resizebox{\linewidth}{!}{\begin{tabular}{c l cccccc@{\hspace{1.2em}}cccccccccccc}
\toprule
 & & \multicolumn{6}{c}{in-distribution} & \multicolumn{12}{c}{out-of-distribution} \\
\cmidrule(lr){3-8}\cmidrule(lr){9-20}
 & & \multicolumn{3}{c}{GSM8K} & \multicolumn{3}{c}{MATH-500} & \multicolumn{3}{c}{AIME24} & \multicolumn{3}{c}{AMC23} & \multicolumn{3}{c}{Minerva} & \multicolumn{3}{c}{Olympiad} \\
\cmidrule(lr){3-5}\cmidrule(lr){6-8}\cmidrule(lr){9-11}\cmidrule(lr){12-14}\cmidrule(lr){15-17}\cmidrule(lr){18-20}
 & $u$ & GRPO & LlMBO & $\Delta$ & GRPO & LlMBO & $\Delta$ & GRPO & LlMBO & $\Delta$ & GRPO & LlMBO & $\Delta$ & GRPO & LlMBO & $\Delta$ & GRPO & LlMBO & $\Delta$ \\
\midrule
\raisebox{-0.15em}{\includegraphics[height=0.9em]{logos/qwen.png}}~Qwen2.5-7B & base & \multicolumn{3}{c}{64.9} & \multicolumn{3}{c}{43.5} & \multicolumn{3}{c}{2.5} & \multicolumn{3}{c}{24.4} & \multicolumn{3}{c}{16.4} & \multicolumn{3}{c}{16.7} \\
 & 64 & \textbf{\boldmath $+15.8$} & $+13.0$ & \gapchip{red!28}{$-2.8$\,{\scriptsize$\pm$3.7}} & \textbf{\boldmath $+14.4$} & $+12.9$ & \gapchip{red!22}{$-1.6$\,{\scriptsize$\pm$3.7}} & \textbf{\boldmath $+4.2$} & $+1.9$ & \gapchip{red!28}{$-2.2$\,{\scriptsize$\pm$1.2}} & \textbf{\boldmath $+8.1$} & $+6.9$ & \gapchip{red!18}{$-1.3$\,{\scriptsize$\pm$1.3}} & \textbf{\boldmath $+9.2$} & $+6.5$ & \gapchip{red!28}{$-2.7$\,{\scriptsize$\pm$2.4}} & \textbf{\boldmath $+9.5$} & $+7.6$ & \gapchip{red!25}{$-1.8$\,{\scriptsize$\pm$1.8}} \\
 & 256 & $+15.6$ & \textbf{\boldmath $+16.2$} & \gapchip{green!8}{$+0.5$\,{\scriptsize$\pm$1.6}} & \textbf{\boldmath $+16.9$} & $+15.8$ & \gapchip{red!16}{$-1.2$\,{\scriptsize$\pm$3.3}} & \textbf{\boldmath $+4.7$} & $+0.6$ & \gapchip{red!28}{$-4.2$\,{\scriptsize$\pm$2.7}} & \textbf{\boldmath $+9.4$} & $+9.2$ & \gapchip{red!3}{$-0.2$\,{\scriptsize$\pm$1.2}} & $+8.4$ & \textbf{\boldmath $+11.0$} & \gapchip{green!28}{$+2.6$\,{\scriptsize$\pm$0.5}} & \textbf{\boldmath $+10.2$} & $+8.6$ & \gapchip{red!23}{$-1.6$\,{\scriptsize$\pm$3.0}} \\
\midrule
\raisebox{-0.15em}{\includegraphics[height=0.9em]{logos/qwen.png}}~Qwen2.5-Math-7B & base & \multicolumn{3}{c}{46.4} & \multicolumn{3}{c}{44.9} & \multicolumn{3}{c}{4.2} & \multicolumn{3}{c}{30.6} & \multicolumn{3}{c}{17.9} & \multicolumn{3}{c}{19.9} \\
 & 64 & $+21.3$ & \textbf{\boldmath $+25.9$} & \gapchip{green!28}{$+4.6$\,{\scriptsize$\pm$4.1}} & \textbf{\boldmath $+18.0$} & $+16.4$ & \gapchip{red!22}{$-1.5$\,{\scriptsize$\pm$3.7}} & \textbf{\boldmath $+10.0$} & $+6.9$ & \gapchip{red!28}{$-3.1$\,{\scriptsize$\pm$3.2}} & \textbf{\boldmath $+15.8$} & $+12.9$ & \gapchip{red!28}{$-2.9$\,{\scriptsize$\pm$6.0}} & \textbf{\boldmath $+11.1$} & $+10.6$ & \gapchip{red!7}{$-0.5$\,{\scriptsize$\pm$3.2}} & \textbf{\boldmath $+11.9$} & $+8.9$ & \gapchip{red!28}{$-3.0$\,{\scriptsize$\pm$2.7}} \\
 & 256 & \textbf{\boldmath $+30.5$} & $+20.1$ & \gapchip{red!28}{$-10.4$\,{\scriptsize$\pm$2.9}} & \textbf{\boldmath $+21.7$} & $+13.5$ & \gapchip{red!28}{$-8.2$\,{\scriptsize$\pm$1.2}} & \textbf{\boldmath $+11.9$} & $+5.3$ & \gapchip{red!28}{$-6.7$\,{\scriptsize$\pm$2.7}} & \textbf{\boldmath $+18.1$} & $+12.9$ & \gapchip{red!28}{$-5.2$\,{\scriptsize$\pm$5.8}} & \textbf{\boldmath $+14.3$} & $+9.1$ & \gapchip{red!28}{$-5.2$\,{\scriptsize$\pm$1.9}} & \textbf{\boldmath $+14.0$} & $+7.6$ & \gapchip{red!28}{$-6.3$\,{\scriptsize$\pm$0.7}} \\
\bottomrule
\end{tabular}
}\\[0.8em]
{\small (b) greedy}\\[0.2em]
\resizebox{\linewidth}{!}{\begin{tabular}{c l cccccc@{\hspace{1.2em}}cccccccccccc}
\toprule
 & & \multicolumn{6}{c}{in-distribution} & \multicolumn{12}{c}{out-of-distribution} \\
\cmidrule(lr){3-8}\cmidrule(lr){9-20}
 & & \multicolumn{3}{c}{GSM8K} & \multicolumn{3}{c}{MATH-500} & \multicolumn{3}{c}{AIME24} & \multicolumn{3}{c}{AMC23} & \multicolumn{3}{c}{Minerva} & \multicolumn{3}{c}{Olympiad} \\
\cmidrule(lr){3-5}\cmidrule(lr){6-8}\cmidrule(lr){9-11}\cmidrule(lr){12-14}\cmidrule(lr){15-17}\cmidrule(lr){18-20}
 & $u$ & GRPO & LlMBO & $\Delta$ & GRPO & LlMBO & $\Delta$ & GRPO & LlMBO & $\Delta$ & GRPO & LlMBO & $\Delta$ & GRPO & LlMBO & $\Delta$ & GRPO & LlMBO & $\Delta$ \\
\midrule
\raisebox{-0.15em}{\includegraphics[height=0.9em]{logos/qwen.png}}~Qwen2.5-7B & base & \multicolumn{3}{c}{88.0} & \multicolumn{3}{c}{64.0} & \multicolumn{3}{c}{0.0} & \multicolumn{3}{c}{30.0} & \multicolumn{3}{c}{32.7} & \multicolumn{3}{c}{28.6} \\
 & 64 & \textbf{\boldmath $+1.0$} & $-3.5$ & \gapchip{red!28}{$-4.4$\,{\scriptsize$\pm$1.8}} & \textbf{\boldmath $+4.0$} & $+1.5$ & \gapchip{red!28}{$-2.5$\,{\scriptsize$\pm$2.7}} & $+5.6$ & $+5.6$ & \gapchip{none}{$+0.0$\,{\scriptsize$\pm$1.9}} & $+4.2$ & \textbf{\boldmath $+6.7$} & \gapchip{green!28}{$+2.5$\,{\scriptsize$\pm$1.4}} & \textbf{\boldmath $+3.2$} & $-5.3$ & \gapchip{red!28}{$-8.5$\,{\scriptsize$\pm$0.6}} & \textbf{\boldmath $+2.1$} & $+1.2$ & \gapchip{red!12}{$-0.8$\,{\scriptsize$\pm$1.8}} \\
 & 256 & \textbf{\boldmath $+0.2$} & $-1.4$ & \gapchip{red!23}{$-1.7$\,{\scriptsize$\pm$1.3}} & $+3.1$ & \textbf{\boldmath $+3.9$} & \gapchip{green!12}{$+0.9$\,{\scriptsize$\pm$1.3}} & $+3.3$ & \textbf{\boldmath $+6.7$} & \gapchip{green!28}{$+3.3$\,{\scriptsize$\pm$1.9}} & $+1.7$ & \textbf{\boldmath $+10.0$} & \gapchip{green!28}{$+8.3$\,{\scriptsize$\pm$3.0}} & \textbf{\boldmath $+2.3$} & $-2.9$ & \gapchip{red!28}{$-5.3$\,{\scriptsize$\pm$2.9}} & \textbf{\boldmath $+3.4$} & $+2.8$ & \gapchip{red!8}{$-0.5$\,{\scriptsize$\pm$1.2}} \\
\midrule
\raisebox{-0.15em}{\includegraphics[height=0.9em]{logos/qwen.png}}~Qwen2.5-Math-7B & base & \multicolumn{3}{c}{68.8} & \multicolumn{3}{c}{64.2} & \multicolumn{3}{c}{10.0} & \multicolumn{3}{c}{37.5} & \multicolumn{3}{c}{21.7} & \multicolumn{3}{c}{26.4} \\
 & 64 & $+11.7$ & \textbf{\boldmath $+13.6$} & \gapchip{green!28}{$+2.0$\,{\scriptsize$\pm$1.1}} & \textbf{\boldmath $+7.3$} & $+3.9$ & \gapchip{red!28}{$-3.3$\,{\scriptsize$\pm$0.9}} & $+2.2$ & \textbf{\boldmath $+4.4$} & \gapchip{green!28}{$+2.2$\,{\scriptsize$\pm$1.1}} & $+9.2$ & \textbf{\boldmath $+10.0$} & \gapchip{green!12}{$+0.8$\,{\scriptsize$\pm$3.6}} & $+6.3$ & \textbf{\boldmath $+11.0$} & \gapchip{green!28}{$+4.8$\,{\scriptsize$\pm$4.9}} & $+6.7$ & \textbf{\boldmath $+6.9$} & \gapchip{green!3}{$+0.2$\,{\scriptsize$\pm$1.1}} \\
 & 256 & \textbf{\boldmath $+11.3$} & $+7.2$ & \gapchip{red!28}{$-4.1$\,{\scriptsize$\pm$6.2}} & \textbf{\boldmath $+4.5$} & $+2.0$ & \gapchip{red!28}{$-2.5$\,{\scriptsize$\pm$2.8}} & \textbf{\boldmath $+6.7$} & $-2.2$ & \gapchip{red!28}{$-8.9$\,{\scriptsize$\pm$2.9}} & \textbf{\boldmath $+11.7$} & $+9.2$ & \gapchip{red!28}{$-2.5$\,{\scriptsize$\pm$6.3}} & \textbf{\boldmath $+6.7$} & $+5.3$ & \gapchip{red!21}{$-1.5$\,{\scriptsize$\pm$3.6}} & \textbf{\boldmath $+8.1$} & $+4.6$ & \gapchip{red!28}{$-3.4$\,{\scriptsize$\pm$3.7}} \\
\bottomrule
\end{tabular}
}
\end{table}

We also ran \llmbo{} at $u \in \{64, 256\}$ on the Qwen2.5 7B models with the settings and evaluation budget used at $u \le 16$ (Table~\ref{tab:large-u}). Here the method is not yet robust. Pooled over seed pairs, \llmbo{} trails GRPO on MATH-500 avg@4 by \poolToneMeanUSixtyFourMathFiveLargeU{} points at $u = 64$ and \poolToneMeanUTwoFiftySixMathFiveLargeU{} at $u = 256$. The mean hides the larger problem, the spread across seeds: single pairs fall up to \worstToneMeanLargeMathFive{} points behind GRPO on MATH-500. Where a search succeeds it remains competitive. Greedy accuracy shows the same: \llmbo{} trails GRPO on greedy MATH-500 by \poolGreedyGreedyUSixtyFourMathFiveLargeU{} and \poolGreedyGreedyUTwoFiftySixMathFiveLargeU{} points at $u = 64, 256$, with single pairs down to \worstGreedyGreedyLargeMathFive{}.
At 1.5B the picture is the same: from $u = 64$ \llmbo{} falls behind on the base and Math models (\gapToneMeanQwenSmallUSixtyFourGsm{} and \gapToneMeanMathSmallUSixtyFourGsm{} on GSM8K avg@4), with pooled MATH-500 avg@4 gaps of \poolToneMeanUSixtyFourMathFiveLargeUSmall{} and \poolToneMeanUTwoFiftySixMathFiveLargeUSmall{} at $u = 64, 256$ and single pairs down to \worstToneMeanLargeMathFiveSmall{}, and greedy MATH-500 gaps of \poolGreedyGreedyUSixtyFourMathFiveLargeUSmall{} and \poolGreedyGreedyUTwoFiftySixMathFiveLargeUSmall{}.

We attribute this to the search budget. The budget that lets a GP and its trust region cover 1-16 dimensions is thin in 64-256: the initial design and the $\lceil \max(4, D)/b \rceil$ failures a trust region tolerates before it shrinks grow with $u$, so a run spends a larger share of its batches on exploration and restarts less often, and a search can settle on a poor region. Its final pick then rests on few, noisy pass-rate estimates in a large space, where the winner's curse is strongest (Appendix~\ref{app:instruct-snr}). Budgets that grow with $u$, dimension-aware priors and trust regions, or searching a low-dimensional subspace of the adapter, with BO over the leading directions and the rest fixed, are natural remedies.

\FloatBarrier
\subsection{Out-of-distribution benchmarks}
\label{app:results-ood}

The four competition benchmarks are harder than the training data, further from it and small, from 30 problems (AIME24) to 675 (OlympiadBench), so their paired gaps are noisier than on the two in-distribution tasks, and on AIME24 a single seed can swing a gap by several points. Their gaps follow the in-distribution ones rather than trading off against them (Table~\ref{tab:main}, Figure~\ref{fig:gap-main}). On Qwen2.5-7B \llmbo{}'s paired gap at $u = 2$ is positive on all four benchmarks, and on AMC23, Minerva Math and OlympiadBench of the size of its in-distribution lead. On Qwen2.5-Math-7B it is positive on AMC23, Minerva Math and OlympiadBench at $u = 2, 4$ and on all four at $u = 8$, with AIME24, the smallest benchmark, the exception at $u \le 4$. On Llama3.1-8B, Mistral-7B and DeepSeek-7B, where the in-distribution gaps are within a point, so are the out-of-distribution ones at $u \le 4$. Where \llmbo{}'s pick is better on the training distribution, it is thus better beyond it too: its gains are not overfit to the training tier.

\FloatBarrier
\section{Qwen2.5 1.5B models}
\label{app:small}
\label{sec:small}
\label{sec:sizes}
\label{app:greedy-small}
\begin{figure}[!htbp]
\centering
\includegraphics[width=\linewidth]{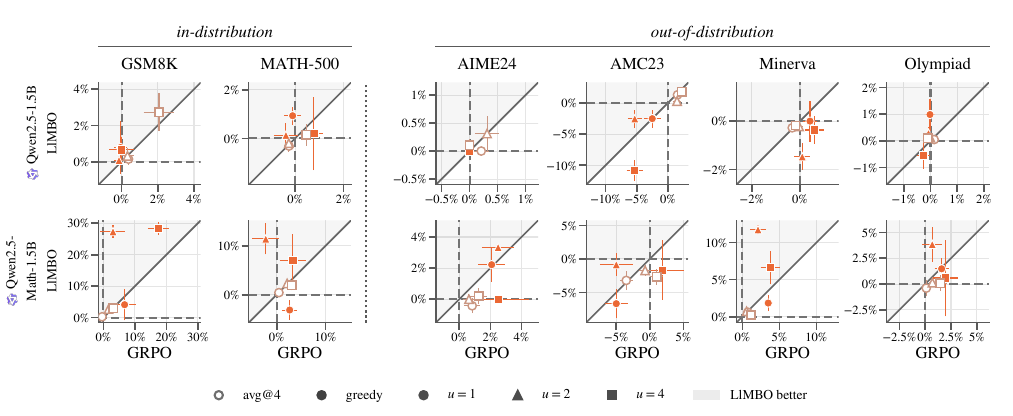}
\caption{Gain over the base model of \llmbo{} against GRPO's, in avg@4 and in greedy accuracy (color; greedy filled, avg@4 open), \llmbo{} trained on the metric of each point, for Qwen2.5-1.5B and Qwen2.5-Math-1.5B (rows) on each benchmark (columns), each panel on its own axes. Training uses GSM8K and MATH: the left two columns are in-distribution, the four benchmarks right of the dotted line out-of-distribution. One point per $u \le 4$ (marker), mean $\pm$ s.e.\ over seeds. Points in the shaded half above the diagonal favor \llmbo{}; dashed: no gain.}
\label{fig:scatter-small}
\end{figure}

\begin{figure}[!htbp]
\centering
\includegraphics[width=\linewidth]{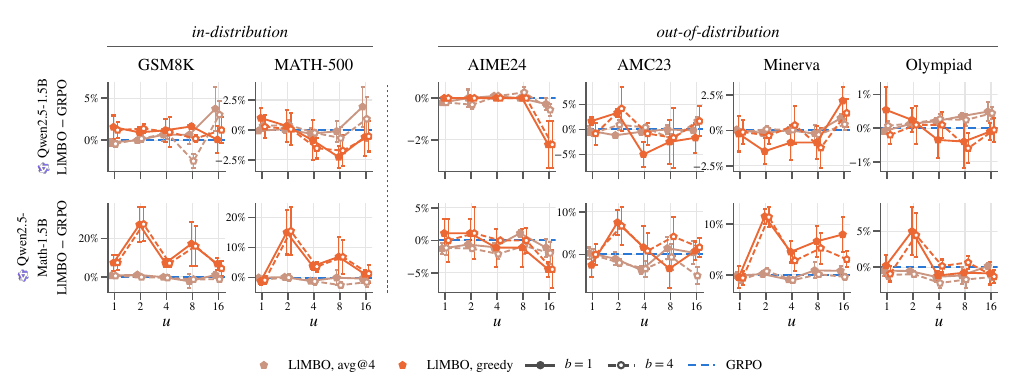}
\caption{Paired gap \llmbo{} $-$ GRPO in avg@4 and in greedy accuracy (color), \llmbo{} trained on each metric, against $u \in \{1, \dots, 16\}$, mean $\pm$ s.e.\ over seed pairs, for Qwen2.5-1.5B and Qwen2.5-Math-1.5B (rows), one column per benchmark (in-distribution GSM8K and MATH-500 first, then the four out-of-distribution benchmarks), each panel on its own gap axis. Solid, filled: $b = 1$ candidate per batch (the default); dashed, open: the $b = 4$ ablation.}
\label{fig:gap-small}
\end{figure}

The Qwen2.5 1.5B base and Math models train on the easy tier (Section~\ref{sec:experiments}). This appendix repeats the per-model analysis of Appendix~\ref{app:results} for them, in both metrics: each optimizer's gain and the paired gap (Table~\ref{tab:small}; Figures~\ref{fig:scatter-small} and~\ref{fig:gap-small}); the pooled gaps are in Table~\ref{tab:ablations}.

\begin{table}[!htbp]
\centering
\caption{Avg@4 (a) and greedy (b) accuracy at the end of training, for Qwen2.5-1.5B and Qwen2.5-Math-1.5B (row blocks), per $u \le 16$ (rows) and benchmark (columns: in-distribution GSM8K and MATH-500, then the four out-of-distribution benchmarks); \llmbo{} is trained on the metric of each subtable (greedy: Section~\ref{sec:greedy-objective}). \emph{base}: accuracy of the base model in that metric (\%). \emph{GRPO}, \emph{\llmbo{}}: gain over the base model (points), mean over seeds (GRPO and avg@4-trained \llmbo{} 8 per cell at $u \le 4$ and 3 at $u = 8, 16$, greedy-trained \llmbo{} 3 per cell); bold: the larger, unless tied. $\Delta$: paired gap \llmbo{} $-$ GRPO, mean $\pm$ s.e.\ over the seed pairs both have, so it can differ from the difference of the two gains; tint: its sign and size (green: \llmbo{} ahead).}
\label{tab:small}
{\small (a) avg@4}\\[0.2em]
\resizebox{\linewidth}{!}{\begin{tabular}{c l cccccc@{\hspace{1.2em}}cccccccccccc}
\toprule
 & & \multicolumn{6}{c}{in-distribution} & \multicolumn{12}{c}{out-of-distribution} \\
\cmidrule(lr){3-8}\cmidrule(lr){9-20}
 & & \multicolumn{3}{c}{GSM8K} & \multicolumn{3}{c}{MATH-500} & \multicolumn{3}{c}{AIME24} & \multicolumn{3}{c}{AMC23} & \multicolumn{3}{c}{Minerva} & \multicolumn{3}{c}{Olympiad} \\
\cmidrule(lr){3-5}\cmidrule(lr){6-8}\cmidrule(lr){9-11}\cmidrule(lr){12-14}\cmidrule(lr){15-17}\cmidrule(lr){18-20}
 & $u$ & GRPO & LlMBO & $\Delta$ & GRPO & LlMBO & $\Delta$ & GRPO & LlMBO & $\Delta$ & GRPO & LlMBO & $\Delta$ & GRPO & LlMBO & $\Delta$ & GRPO & LlMBO & $\Delta$ \\
\midrule
\multirow{6}{*}{\rotatebox[origin=c]{90}{\scriptsize\begin{tabular}{@{}c@{}}\raisebox{-0.15em}{\includegraphics[height=0.9em]{logos/qwen.png}}~Qwen2.5\\ 1.5B\end{tabular}}} & base & \multicolumn{3}{c}{22.6} & \multicolumn{3}{c}{8.9} & \multicolumn{3}{c}{0.0} & \multicolumn{3}{c}{1.9} & \multicolumn{3}{c}{2.3} & \multicolumn{3}{c}{1.6} \\
 & 1 & \textbf{\boldmath $+0.4$} & $+0.1$ & \gapchip{red!3}{$-0.2$\,{\scriptsize$\pm$0.4}} & \textbf{\boldmath $-0.2$} & $-0.3$ & \gapchip{red!1}{$-0.1$\,{\scriptsize$\pm$0.2}} & \textbf{\boldmath $+0.2$} & $+0.0$ & \gapchip{red!3}{$-0.2$\,{\scriptsize$\pm$0.1}} & \textbf{\boldmath $+1.6$} & $+1.3$ & \gapchip{red!3}{$-0.2$\,{\scriptsize$\pm$0.9}} & $-0.3$ & $-0.3$ & \gapchip{green!1}{$+0.0$\,{\scriptsize$\pm$0.2}} & $+0.1$ & $+0.1$ & \gapchip{red!1}{$-0.1$\,{\scriptsize$\pm$0.1}} \\
 & 2 & $+0.3$ & $+0.3$ & \gapchip{none}{$+0.0$\,{\scriptsize$\pm$0.3}} & $-0.3$ & \textbf{\boldmath $-0.2$} & \gapchip{green!1}{$+0.1$\,{\scriptsize$\pm$0.3}} & $+0.3$ & $+0.3$ & \gapchip{none}{$+0.0$\,{\scriptsize$\pm$0.4}} & \textbf{\boldmath $+1.5$} & $+0.3$ & \gapchip{red!16}{$-1.2$\,{\scriptsize$\pm$0.5}} & \textbf{\boldmath $+0.0$} & $-0.2$ & \gapchip{red!3}{$-0.2$\,{\scriptsize$\pm$0.2}} & $+0.0$ & \textbf{\boldmath $+0.1$} & \gapchip{green!2}{$+0.1$\,{\scriptsize$\pm$0.1}} \\
 & 4 & $+2.1$ & \textbf{\boldmath $+2.7$} & \gapchip{green!9}{$+0.7$\,{\scriptsize$\pm$0.7}} & \textbf{\boldmath $+0.5$} & $+0.2$ & \gapchip{red!4}{$-0.3$\,{\scriptsize$\pm$0.5}} & $+0.0$ & \textbf{\boldmath $+0.1$} & \gapchip{green!1}{$+0.1$\,{\scriptsize$\pm$0.1}} & \textbf{\boldmath $+2.3$} & $+1.8$ & \gapchip{red!7}{$-0.5$\,{\scriptsize$\pm$0.7}} & $-0.2$ & $-0.2$ & \gapchip{none}{$+0.0$\,{\scriptsize$\pm$0.3}} & $-0.1$ & \textbf{\boldmath $+0.1$} & \gapchip{green!3}{$+0.2$\,{\scriptsize$\pm$0.1}} \\
 & 8 & $+3.9$ & \textbf{\boldmath $+4.6$} & \gapchip{green!9}{$+0.7$\,{\scriptsize$\pm$1.1}} & \textbf{\boldmath $+1.2$} & $+1.1$ & \gapchip{red!2}{$-0.1$\,{\scriptsize$\pm$0.9}} & $+0.0$ & $+0.0$ & \gapchip{none}{$+0.0$\,{\scriptsize$\pm$0.0}} & $+2.7$ & $+2.7$ & \gapchip{none}{$+0.0$\,{\scriptsize$\pm$1.0}} & \textbf{\boldmath $+0.2$} & $+0.0$ & \gapchip{red!3}{$-0.2$\,{\scriptsize$\pm$0.6}} & $+0.0$ & \textbf{\boldmath $+0.4$} & \gapchip{green!5}{$+0.4$\,{\scriptsize$\pm$0.1}} \\
 & 16 & $+6.2$ & \textbf{\boldmath $+9.9$} & \gapchip{green!28}{$+3.7$\,{\scriptsize$\pm$2.6}} & $+1.5$ & \textbf{\boldmath $+3.4$} & \gapchip{green!27}{$+1.9$\,{\scriptsize$\pm$1.7}} & \textbf{\boldmath $+0.6$} & $+0.3$ & \gapchip{red!4}{$-0.3$\,{\scriptsize$\pm$0.3}} & \textbf{\boldmath $+1.5$} & $+1.2$ & \gapchip{red!3}{$-0.2$\,{\scriptsize$\pm$0.6}} & $-0.2$ & \textbf{\boldmath $+0.7$} & \gapchip{green!12}{$+0.9$\,{\scriptsize$\pm$0.3}} & $+0.2$ & \textbf{\boldmath $+0.7$} & \gapchip{green!7}{$+0.5$\,{\scriptsize$\pm$0.3}} \\
\midrule
\multirow{6}{*}{\rotatebox[origin=c]{90}{\scriptsize\begin{tabular}{@{}c@{}}\raisebox{-0.15em}{\includegraphics[height=0.9em]{logos/qwen.png}}~Qwen2.5\\ Math-1.5B\end{tabular}}} & base & \multicolumn{3}{c}{34.6} & \multicolumn{3}{c}{32.5} & \multicolumn{3}{c}{2.5} & \multicolumn{3}{c}{20.0} & \multicolumn{3}{c}{10.2} & \multicolumn{3}{c}{16.2} \\
 & 1 & $-0.4$ & \textbf{\boldmath $+0.3$} & \gapchip{green!10}{$+0.7$\,{\scriptsize$\pm$0.7}} & \textbf{\boldmath $+0.5$} & $+0.4$ & \gapchip{red!1}{$+0.0$\,{\scriptsize$\pm$0.7}} & \textbf{\boldmath $+0.8$} & $-0.4$ & \gapchip{red!18}{$-1.2$\,{\scriptsize$\pm$0.8}} & $-3.5$ & \textbf{\boldmath $-3.2$} & \gapchip{green!4}{$+0.3$\,{\scriptsize$\pm$1.3}} & \textbf{\boldmath $+0.7$} & $+0.6$ & \gapchip{red!2}{$-0.1$\,{\scriptsize$\pm$0.6}} & \textbf{\boldmath $+0.1$} & $-0.4$ & \gapchip{red!7}{$-0.5$\,{\scriptsize$\pm$0.7}} \\
 & 2 & $+1.6$ & \textbf{\boldmath $+2.7$} & \gapchip{green!16}{$+1.1$\,{\scriptsize$\pm$0.7}} & $+2.1$ & \textbf{\boldmath $+2.3$} & \gapchip{green!2}{$+0.2$\,{\scriptsize$\pm$0.8}} & \textbf{\boldmath $+0.6$} & $+0.0$ & \gapchip{red!9}{$-0.6$\,{\scriptsize$\pm$0.6}} & \textbf{\boldmath $-0.7$} & $-1.7$ & \gapchip{red!14}{$-1.0$\,{\scriptsize$\pm$1.5}} & $+0.6$ & \textbf{\boldmath $+0.7$} & \gapchip{green!2}{$+0.2$\,{\scriptsize$\pm$0.5}} & \textbf{\boldmath $+0.7$} & $+0.1$ & \gapchip{red!8}{$-0.6$\,{\scriptsize$\pm$0.3}} \\
 & 4 & \textbf{\boldmath $+3.1$} & $+3.0$ & \gapchip{red!1}{$-0.1$\,{\scriptsize$\pm$0.7}} & \textbf{\boldmath $+3.1$} & $+2.0$ & \gapchip{red!16}{$-1.1$\,{\scriptsize$\pm$0.7}} & \textbf{\boldmath $+1.2$} & $+0.2$ & \gapchip{red!15}{$-1.0$\,{\scriptsize$\pm$0.6}} & \textbf{\boldmath $+1.0$} & $-2.7$ & \gapchip{red!28}{$-3.7$\,{\scriptsize$\pm$1.0}} & \textbf{\boldmath $+1.2$} & $+0.2$ & \gapchip{red!13}{$-0.9$\,{\scriptsize$\pm$0.3}} & \textbf{\boldmath $+1.5$} & $+0.1$ & \gapchip{red!19}{$-1.4$\,{\scriptsize$\pm$0.7}} \\
 & 8 & \textbf{\boldmath $+6.7$} & $+4.5$ & \gapchip{red!28}{$-2.3$\,{\scriptsize$\pm$3.7}} & \textbf{\boldmath $+4.6$} & $+4.5$ & \gapchip{none}{$+0.0$\,{\scriptsize$\pm$2.5}} & $-1.1$ & \textbf{\boldmath $+0.0$} & \gapchip{green!16}{$+1.1$\,{\scriptsize$\pm$0.6}} & $-1.7$ & \textbf{\boldmath $-0.2$} & \gapchip{green!20}{$+1.5$\,{\scriptsize$\pm$2.6}} & $+1.8$ & \textbf{\boldmath $+2.7$} & \gapchip{green!12}{$+0.9$\,{\scriptsize$\pm$1.6}} & \textbf{\boldmath $+1.5$} & $+0.3$ & \gapchip{red!16}{$-1.2$\,{\scriptsize$\pm$1.2}} \\
 & 16 & $+14.1$ & \textbf{\boldmath $+15.2$} & \gapchip{green!15}{$+1.0$\,{\scriptsize$\pm$1.7}} & \textbf{\boldmath $+10.5$} & $+10.2$ & \gapchip{red!5}{$-0.3$\,{\scriptsize$\pm$1.4}} & \textbf{\boldmath $+1.1$} & $+0.0$ & \gapchip{red!16}{$-1.1$\,{\scriptsize$\pm$1.4}} & $+3.7$ & \textbf{\boldmath $+4.2$} & \gapchip{green!6}{$+0.4$\,{\scriptsize$\pm$2.7}} & $+2.5$ & \textbf{\boldmath $+3.3$} & \gapchip{green!11}{$+0.8$\,{\scriptsize$\pm$1.0}} & $+3.4$ & \textbf{\boldmath $+3.5$} & \gapchip{green!2}{$+0.1$\,{\scriptsize$\pm$0.4}} \\
\bottomrule
\end{tabular}
}\\[0.8em]
{\small (b) greedy}\\[0.2em]
\resizebox{\linewidth}{!}{\begin{tabular}{c l cccccc@{\hspace{1.2em}}cccccccccccc}
\toprule
 & & \multicolumn{6}{c}{in-distribution} & \multicolumn{12}{c}{out-of-distribution} \\
\cmidrule(lr){3-8}\cmidrule(lr){9-20}
 & & \multicolumn{3}{c}{GSM8K} & \multicolumn{3}{c}{MATH-500} & \multicolumn{3}{c}{AIME24} & \multicolumn{3}{c}{AMC23} & \multicolumn{3}{c}{Minerva} & \multicolumn{3}{c}{Olympiad} \\
\cmidrule(lr){3-5}\cmidrule(lr){6-8}\cmidrule(lr){9-11}\cmidrule(lr){12-14}\cmidrule(lr){15-17}\cmidrule(lr){18-20}
 & $u$ & GRPO & LlMBO & $\Delta$ & GRPO & LlMBO & $\Delta$ & GRPO & LlMBO & $\Delta$ & GRPO & LlMBO & $\Delta$ & GRPO & LlMBO & $\Delta$ & GRPO & LlMBO & $\Delta$ \\
\midrule
\multirow{6}{*}{\rotatebox[origin=c]{90}{\scriptsize\begin{tabular}{@{}c@{}}\raisebox{-0.15em}{\includegraphics[height=0.9em]{logos/qwen.png}}~Qwen2.5\\ 1.5B\end{tabular}}} & base & \multicolumn{3}{c}{56.3} & \multicolumn{3}{c}{28.4} & \multicolumn{3}{c}{0.0} & \multicolumn{3}{c}{12.5} & \multicolumn{3}{c}{8.5} & \multicolumn{3}{c}{6.5} \\
 & 1 & $-0.1$ & \textbf{\boldmath $+0.8$} & \gapchip{green!22}{$+1.6$\,{\scriptsize$\pm$1.4}} & $-0.1$ & \textbf{\boldmath $+0.9$} & \gapchip{green!14}{$+1.0$\,{\scriptsize$\pm$0.8}} & $+0.0$ & $+0.0$ & \gapchip{none}{$+0.0$\,{\scriptsize$\pm$0.0}} & $-2.5$ & $-2.5$ & \gapchip{green!23}{$+1.7$\,{\scriptsize$\pm$0.8}} & \textbf{\boldmath $+0.4$} & $+0.0$ & \gapchip{red!3}{$-0.2$\,{\scriptsize$\pm$1.2}} & $+0.0$ & \textbf{\boldmath $+1.0$} & \gapchip{green!8}{$+0.5$\,{\scriptsize$\pm$0.7}} \\
 & 2 & $-0.1$ & \textbf{\boldmath $+0.1$} & \gapchip{green!13}{$+0.9$\,{\scriptsize$\pm$0.7}} & $-0.4$ & \textbf{\boldmath $+0.1$} & \gapchip{green!5}{$+0.3$\,{\scriptsize$\pm$1.3}} & $+0.0$ & $+0.0$ & \gapchip{none}{$+0.0$\,{\scriptsize$\pm$0.0}} & $-5.3$ & \textbf{\boldmath $-2.5$} & \gapchip{green!28}{$+3.3$\,{\scriptsize$\pm$0.8}} & \textbf{\boldmath $+0.1$} & $-1.5$ & \gapchip{red!21}{$-1.5$\,{\scriptsize$\pm$0.9}} & $-0.1$ & \textbf{\boldmath $+0.3$} & \gapchip{green!3}{$+0.2$\,{\scriptsize$\pm$0.4}} \\
 & 4 & $+0.0$ & \textbf{\boldmath $+0.7$} & \gapchip{green!17}{$+1.2$\,{\scriptsize$\pm$1.4}} & \textbf{\boldmath $+0.8$} & $+0.2$ & \gapchip{red!12}{$-0.9$\,{\scriptsize$\pm$1.7}} & $+0.0$ & $+0.0$ & \gapchip{none}{$+0.0$\,{\scriptsize$\pm$0.0}} & \textbf{\boldmath $-5.3$} & $-10.8$ & \gapchip{red!28}{$-5.0$\,{\scriptsize$\pm$2.5}} & \textbf{\boldmath $+0.6$} & $-0.4$ & \gapchip{red!12}{$-0.9$\,{\scriptsize$\pm$1.2}} & \textbf{\boldmath $-0.3$} & $-0.5$ & \gapchip{red!5}{$-0.3$\,{\scriptsize$\pm$0.5}} \\
 & 8 & $+1.1$ & \textbf{\boldmath $+2.8$} & \gapchip{green!23}{$+1.6$\,{\scriptsize$\pm$0.3}} & \textbf{\boldmath $+2.7$} & $+0.4$ & \gapchip{red!28}{$-2.3$\,{\scriptsize$\pm$0.9}} & $+0.0$ & $+0.0$ & \gapchip{none}{$+0.0$\,{\scriptsize$\pm$0.0}} & \textbf{\boldmath $-1.7$} & $-4.2$ & \gapchip{red!28}{$-2.5$\,{\scriptsize$\pm$4.3}} & \textbf{\boldmath $+1.6$} & $+0.7$ & \gapchip{red!12}{$-0.9$\,{\scriptsize$\pm$1.2}} & \textbf{\boldmath $+0.2$} & $-0.2$ & \gapchip{red!6}{$-0.4$\,{\scriptsize$\pm$0.8}} \\
 & 16 & $+1.7$ & \textbf{\boldmath $+1.8$} & \gapchip{green!1}{$+0.1$\,{\scriptsize$\pm$1.6}} & \textbf{\boldmath $+1.3$} & $+0.7$ & \gapchip{red!8}{$-0.6$\,{\scriptsize$\pm$1.6}} & \textbf{\boldmath $+2.2$} & $+0.0$ & \gapchip{red!28}{$-2.2$\,{\scriptsize$\pm$1.1}} & \textbf{\boldmath $-5.8$} & $-7.5$ & \gapchip{red!23}{$-1.7$\,{\scriptsize$\pm$3.0}} & $-1.7$ & \textbf{\boldmath $+0.4$} & \gapchip{green!28}{$+2.1$\,{\scriptsize$\pm$1.0}} & \textbf{\boldmath $-0.6$} & $-0.7$ & \gapchip{red!1}{$-0.1$\,{\scriptsize$\pm$0.3}} \\
\midrule
\multirow{6}{*}{\rotatebox[origin=c]{90}{\scriptsize\begin{tabular}{@{}c@{}}\raisebox{-0.15em}{\includegraphics[height=0.9em]{logos/qwen.png}}~Qwen2.5\\ Math-1.5B\end{tabular}}} & base & \multicolumn{3}{c}{41.5} & \multicolumn{3}{c}{47.2} & \multicolumn{3}{c}{3.3} & \multicolumn{3}{c}{32.5} & \multicolumn{3}{c}{8.5} & \multicolumn{3}{c}{23.4} \\
 & 1 & \textbf{\boldmath $+6.7$} & $+4.1$ & \gapchip{green!28}{$+7.4$\,{\scriptsize$\pm$4.1}} & \textbf{\boldmath $+2.7$} & $-3.1$ & \gapchip{red!17}{$-1.2$\,{\scriptsize$\pm$1.2}} & $+2.1$ & \textbf{\boldmath $+2.2$} & \gapchip{green!16}{$+1.1$\,{\scriptsize$\pm$2.2}} & \textbf{\boldmath $-5.0$} & $-6.7$ & \gapchip{red!28}{$-2.5$\,{\scriptsize$\pm$2.9}} & \textbf{\boldmath $+3.5$} & $+1.8$ & \gapchip{red!7}{$-0.5$\,{\scriptsize$\pm$2.2}} & \textbf{\boldmath $+1.6$} & $+1.5$ & \gapchip{green!3}{$+0.2$\,{\scriptsize$\pm$1.5}} \\
 & 2 & $+3.0$ & \textbf{\boldmath $+27.3$} & \gapchip{green!28}{$+27.3$\,{\scriptsize$\pm$9.0}} & $-2.2$ & \textbf{\boldmath $+11.4$} & \gapchip{green!28}{$+15.1$\,{\scriptsize$\pm$7.4}} & $+2.5$ & \textbf{\boldmath $+3.3$} & \gapchip{green!16}{$+1.1$\,{\scriptsize$\pm$2.2}} & $-5.0$ & \textbf{\boldmath $-0.8$} & \gapchip{green!28}{$+7.5$\,{\scriptsize$\pm$2.9}} & $+2.1$ & \textbf{\boldmath $+11.8$} & \gapchip{green!28}{$+11.5$\,{\scriptsize$\pm$1.4}} & $+0.7$ & \textbf{\boldmath $+3.9$} & \gapchip{green!28}{$+5.0$\,{\scriptsize$\pm$3.4}} \\
 & 4 & $+17.5$ & \textbf{\boldmath $+28.3$} & \gapchip{green!28}{$+7.2$\,{\scriptsize$\pm$2.2}} & $+3.3$ & \textbf{\boldmath $+7.0$} & \gapchip{green!28}{$+3.8$\,{\scriptsize$\pm$1.1}} & \textbf{\boldmath $+2.5$} & $+0.0$ & \gapchip{red!16}{$-1.1$\,{\scriptsize$\pm$2.9}} & \textbf{\boldmath $+1.9$} & $-1.7$ & \gapchip{green!23}{$+1.7$\,{\scriptsize$\pm$5.1}} & $+3.8$ & \textbf{\boldmath $+6.6$} & \gapchip{green!28}{$+4.5$\,{\scriptsize$\pm$2.1}} & \textbf{\boldmath $+1.9$} & $+0.6$ & \gapchip{red!17}{$-1.2$\,{\scriptsize$\pm$0.7}} \\
 & 8 & $+16.2$ & \textbf{\boldmath $+33.5$} & \gapchip{green!28}{$+17.3$\,{\scriptsize$\pm$11.4}} & $+6.3$ & \textbf{\boldmath $+13.2$} & \gapchip{green!28}{$+6.9$\,{\scriptsize$\pm$6.6}} & \textbf{\boldmath $+0.0$} & $-1.1$ & \gapchip{red!16}{$-1.1$\,{\scriptsize$\pm$2.9}} & \textbf{\boldmath $+7.5$} & $+4.2$ & \gapchip{red!28}{$-3.3$\,{\scriptsize$\pm$4.6}} & $+5.5$ & \textbf{\boldmath $+12.1$} & \gapchip{green!28}{$+6.6$\,{\scriptsize$\pm$3.1}} & \textbf{\boldmath $+2.8$} & $+2.0$ & \gapchip{red!11}{$-0.8$\,{\scriptsize$\pm$0.3}} \\
 & 16 & $+28.0$ & \textbf{\boldmath $+34.8$} & \gapchip{green!28}{$+6.8$\,{\scriptsize$\pm$3.0}} & $+11.7$ & \textbf{\boldmath $+12.9$} & \gapchip{green!17}{$+1.2$\,{\scriptsize$\pm$1.4}} & \textbf{\boldmath $+4.4$} & $+0.0$ & \gapchip{red!28}{$-4.4$\,{\scriptsize$\pm$1.1}} & $+0.8$ & \textbf{\boldmath $+1.7$} & \gapchip{green!12}{$+0.8$\,{\scriptsize$\pm$2.2}} & $+6.5$ & \textbf{\boldmath $+14.5$} & \gapchip{green!28}{$+8.0$\,{\scriptsize$\pm$3.4}} & \textbf{\boldmath $+3.5$} & $+2.6$ & \gapchip{red!12}{$-0.9$\,{\scriptsize$\pm$1.5}} \\
\bottomrule
\end{tabular}
}
\end{table}

For Qwen2.5-1.5B In avg@4 \llmbo{} is level with GRPO at $u \le 4$ (GSM8K \gapToneMeanQwenSmallUOneGsm{}, \gapToneMeanQwenSmallUTwoGsm{}, \gapToneMeanQwenSmallUFourGsm{}) and pulls ahead at $u = 16$ (\gapToneMeanQwenSmallUSixteenGsm{} on GSM8K, \gapToneMeanQwenSmallUSixteenMathFive{} on MATH-500; Figure~\ref{fig:gap-small}). In greedy accuracy it keeps a small lead on GSM8K at $u \le 4$ (\gapGreedyGreedyQwenSmallUOneGsm{}, \gapGreedyGreedyQwenSmallUTwoGsm{}, \gapGreedyGreedyQwenSmallUFourGsm{}).

Whereas for Qwen2.5-Math-1.5B in avg@4 \llmbo{} is level with GRPO at $u \le 4$ (GSM8K \gapToneMeanMathSmallUOneGsm{}, \gapToneMeanMathSmallUTwoGsm{}, \gapToneMeanMathSmallUFourGsm{}). In greedy accuracy its lead is the largest of any model: its points lie far above the diagonal (Figure~\ref{fig:scatter-small}), it leads GRPO on GSM8K by \gapGreedyGreedyMathSmallUOneGsm{}, \gapGreedyGreedyMathSmallUTwoGsm{} and \gapGreedyGreedyMathSmallUFourGsm{} points at $u = 1, 2, 4$ and on MATH-500 by \gapGreedyGreedyMathSmallUTwoMathFive{} at $u = 2$, stays ahead up to $u = 16$ (\gapGreedyGreedyMathSmallUEightGsm{} and \gapGreedyGreedyMathSmallUSixteenGsm{} on GSM8K at $u = 8, 16$) for both batch sizes alike, and is still level at $u = 64$ (\gapGreedyGreedyMathSmallUSixtyFourGsm{}; Figure~\ref{fig:gap-small}). At $u = 2$ it gains \gainGreedyGreedyMathSmallUTwoGsm{} points on GSM8K and \gainGreedyGreedyMathSmallUTwoMathFive{} on MATH-500, against GRPO's \gainGrpoGreedyMathSmallUTwoGsm{} and \gainGrpoGreedyMathSmallUTwoMathFive{}; on the OOD benchmarks it is level or ahead, most on Minerva Math. Part of this gain comes from suppressing a repetition loop that ends a large share of the base model's greedy GSM8K answers: a failure the greedy objective sees directly and a single TinyLoRA direction can switch off.

When pooled over the two 1.5B models \llmbo{} is level with GRPO in avg@4 at every $u$, within a point on GSM8K (\poolToneMeanUOneGsmSmall{}, \poolToneMeanUTwoGsmSmall{} and \poolToneMeanUFourGsmSmall{} points) and on MATH-500 (\poolToneMeanUOneMathFiveSmall{}, \poolToneMeanUTwoMathFiveSmall{}, \poolToneMeanUFourMathFiveSmall{}), and leads on greedy GSM8K at every $u$ (\poolGreedyGreedyUOneGsmSmall{}, \poolGreedyGreedyUTwoGsmSmall{} and \poolGreedyGreedyUFourGsmSmall{}; Table~\ref{tab:ablations}), driven by Qwen2.5-Math-1.5B. At $u \in \{64, 256\}$ both metrics fall behind (Appendix~\ref{sec:large-u}).

\FloatBarrier
\section{Instruct models: no gain to find}
\label{app:instruct}
Qwen2.5-7B-Instruct and Qwen2.5-1.5B-Instruct are already RL-tuned for math reasoning, and TinyLoRA at $u \le 4$ leaves them almost nothing to gain. This appendix gives their detailed results and argues why, in this regime, \llmbo{} trails GRPO slightly: the pass rate carries too little signal for a search to establish that the base model is the best it can do, and gradient descent from $\theta = 0$ has an inductive bias towards staying there that BO lacks.

\subsection{Results}
\label{app:instruct-results}

\begin{table}[!htbp]
\centering
\caption{Avg@4 (a) and greedy (b) accuracy at the end of training, for Qwen2.5-7B-Instruct and Qwen2.5-1.5B-Instruct (row blocks), per $u \le 16$ (rows) and benchmark (columns: in-distribution GSM8K and MATH-500, then the four out-of-distribution benchmarks); \llmbo{} is trained on the metric of each subtable (greedy: Section~\ref{sec:greedy-objective}). \emph{base}: accuracy of the base model in that metric (\%). \emph{GRPO}, \emph{\llmbo{}}: gain over the base model (points), mean over seeds (GRPO and avg@4-trained \llmbo{} 8 per cell at $u \le 4$ and 3 at $u = 8, 16$, greedy-trained \llmbo{} 3 per cell); bold: the larger, unless tied. $\Delta$: paired gap \llmbo{} $-$ GRPO, mean $\pm$ s.e.\ over the seed pairs both have, so it can differ from the difference of the two gains; tint: its sign and size (green: \llmbo{} ahead).}
\label{tab:instruct}
{\small (a) avg@4}\\[0.2em]
\resizebox{\linewidth}{!}{\begin{tabular}{c l cccccc@{\hspace{1.2em}}cccccccccccc}
\toprule
 & & \multicolumn{6}{c}{in-distribution} & \multicolumn{12}{c}{out-of-distribution} \\
\cmidrule(lr){3-8}\cmidrule(lr){9-20}
 & & \multicolumn{3}{c}{GSM8K} & \multicolumn{3}{c}{MATH-500} & \multicolumn{3}{c}{AIME24} & \multicolumn{3}{c}{AMC23} & \multicolumn{3}{c}{Minerva} & \multicolumn{3}{c}{Olympiad} \\
\cmidrule(lr){3-5}\cmidrule(lr){6-8}\cmidrule(lr){9-11}\cmidrule(lr){12-14}\cmidrule(lr){15-17}\cmidrule(lr){18-20}
 & $u$ & GRPO & LlMBO & $\Delta$ & GRPO & LlMBO & $\Delta$ & GRPO & LlMBO & $\Delta$ & GRPO & LlMBO & $\Delta$ & GRPO & LlMBO & $\Delta$ & GRPO & LlMBO & $\Delta$ \\
\midrule
\multirow{6}{*}{\rotatebox[origin=c]{90}{\scriptsize\begin{tabular}{@{}c@{}}\raisebox{-0.15em}{\includegraphics[height=0.9em]{logos/qwen.png}}~Qwen2.5\\ 7B-Inst.\end{tabular}}} & base & \multicolumn{3}{c}{90.9} & \multicolumn{3}{c}{72.1} & \multicolumn{3}{c}{9.2} & \multicolumn{3}{c}{42.5} & \multicolumn{3}{c}{40.1} & \multicolumn{3}{c}{35.4} \\
 & 1 & \textbf{\boldmath $+0.3$} & $+0.1$ & \gapchip{red!3}{$-0.2$\,{\scriptsize$\pm$0.2}} & \textbf{\boldmath $+0.7$} & $+0.1$ & \gapchip{red!7}{$-0.5$\,{\scriptsize$\pm$0.8}} & \textbf{\boldmath $-0.6$} & $-0.8$ & \gapchip{red!3}{$-0.2$\,{\scriptsize$\pm$1.2}} & \textbf{\boldmath $+2.8$} & $+1.7$ & \gapchip{red!15}{$-1.1$\,{\scriptsize$\pm$2.1}} & \textbf{\boldmath $-0.1$} & $-1.0$ & \gapchip{red!13}{$-0.9$\,{\scriptsize$\pm$0.6}} & \textbf{\boldmath $+1.2$} & $+0.3$ & \gapchip{red!13}{$-0.9$\,{\scriptsize$\pm$0.8}} \\
 & 2 & \textbf{\boldmath $+0.3$} & $+0.0$ & \gapchip{red!4}{$-0.3$\,{\scriptsize$\pm$0.2}} & $-0.2$ & \textbf{\boldmath $+0.0$} & \gapchip{green!3}{$+0.2$\,{\scriptsize$\pm$0.2}} & $-2.1$ & \textbf{\boldmath $-1.2$} & \gapchip{green!12}{$+0.8$\,{\scriptsize$\pm$0.8}} & $+1.3$ & \textbf{\boldmath $+3.1$} & \gapchip{green!25}{$+1.8$\,{\scriptsize$\pm$1.6}} & \textbf{\boldmath $-0.9$} & $-1.3$ & \gapchip{red!5}{$-0.4$\,{\scriptsize$\pm$0.4}} & \textbf{\boldmath $+0.1$} & $-0.2$ & \gapchip{red!3}{$-0.2$\,{\scriptsize$\pm$0.3}} \\
 & 4 & \textbf{\boldmath $+0.4$} & $-0.4$ & \gapchip{red!11}{$-0.8$\,{\scriptsize$\pm$0.2}} & \textbf{\boldmath $+0.2$} & $-1.0$ & \gapchip{red!16}{$-1.1$\,{\scriptsize$\pm$0.6}} & $-2.1$ & \textbf{\boldmath $-1.7$} & \gapchip{green!6}{$+0.4$\,{\scriptsize$\pm$0.9}} & \textbf{\boldmath $+4.9$} & $+0.4$ & \gapchip{red!28}{$-4.5$\,{\scriptsize$\pm$2.4}} & \textbf{\boldmath $-0.8$} & $-1.6$ & \gapchip{red!11}{$-0.8$\,{\scriptsize$\pm$0.7}} & \textbf{\boldmath $+0.4$} & $-0.9$ & \gapchip{red!18}{$-1.3$\,{\scriptsize$\pm$0.6}} \\
 & 8 & \textbf{\boldmath $+0.2$} & $-0.6$ & \gapchip{red!11}{$-0.8$\,{\scriptsize$\pm$0.2}} & \textbf{\boldmath $+0.1$} & $-0.8$ & \gapchip{red!12}{$-0.9$\,{\scriptsize$\pm$0.8}} & $-2.8$ & \textbf{\boldmath $-1.4$} & \gapchip{green!19}{$+1.4$\,{\scriptsize$\pm$1.4}} & \textbf{\boldmath $+3.3$} & $+0.8$ & \gapchip{red!28}{$-2.5$\,{\scriptsize$\pm$0.6}} & \textbf{\boldmath $-0.2$} & $-1.6$ & \gapchip{red!21}{$-1.5$\,{\scriptsize$\pm$0.5}} & \textbf{\boldmath $-0.3$} & $-1.1$ & \gapchip{red!11}{$-0.8$\,{\scriptsize$\pm$0.7}} \\
 & 16 & \textbf{\boldmath $+0.1$} & $-0.6$ & \gapchip{red!10}{$-0.7$\,{\scriptsize$\pm$0.2}} & \textbf{\boldmath $+0.1$} & $-1.4$ & \gapchip{red!22}{$-1.5$\,{\scriptsize$\pm$0.7}} & $-2.8$ & $-2.8$ & \gapchip{none}{$+0.0$\,{\scriptsize$\pm$1.4}} & \textbf{\boldmath $+3.3$} & $-1.2$ & \gapchip{red!28}{$-4.6$\,{\scriptsize$\pm$2.5}} & \textbf{\boldmath $-1.6$} & $-3.2$ & \gapchip{red!22}{$-1.6$\,{\scriptsize$\pm$0.8}} & \textbf{\boldmath $+0.2$} & $-2.5$ & \gapchip{red!28}{$-2.7$\,{\scriptsize$\pm$0.4}} \\
\midrule
\multirow{6}{*}{\rotatebox[origin=c]{90}{\scriptsize\begin{tabular}{@{}c@{}}\raisebox{-0.15em}{\includegraphics[height=0.9em]{logos/qwen.png}}~Qwen2.5\\ 1.5B-Inst.\end{tabular}}} & base & \multicolumn{3}{c}{63.9} & \multicolumn{3}{c}{45.2} & \multicolumn{3}{c}{0.8} & \multicolumn{3}{c}{14.4} & \multicolumn{3}{c}{13.2} & \multicolumn{3}{c}{14.6} \\
 & 1 & $+1.1$ & \textbf{\boldmath $+1.3$} & \gapchip{green!3}{$+0.2$\,{\scriptsize$\pm$0.2}} & \textbf{\boldmath $+0.2$} & $-1.0$ & \gapchip{red!17}{$-1.2$\,{\scriptsize$\pm$0.4}} & \textbf{\boldmath $+0.9$} & $+0.6$ & \gapchip{red!4}{$-0.3$\,{\scriptsize$\pm$0.4}} & \textbf{\boldmath $+4.5$} & $+3.4$ & \gapchip{red!16}{$-1.2$\,{\scriptsize$\pm$1.4}} & $+0.3$ & \textbf{\boldmath $+0.7$} & \gapchip{green!6}{$+0.4$\,{\scriptsize$\pm$0.5}} & \textbf{\boldmath $+0.0$} & $-0.2$ & \gapchip{red!3}{$-0.2$\,{\scriptsize$\pm$0.1}} \\
 & 2 & \textbf{\boldmath $+1.2$} & $+0.8$ & \gapchip{red!5}{$-0.4$\,{\scriptsize$\pm$0.1}} & \textbf{\boldmath $+0.0$} & $-0.4$ & \gapchip{red!5}{$-0.4$\,{\scriptsize$\pm$0.5}} & $+0.2$ & \textbf{\boldmath $+0.4$} & \gapchip{green!3}{$+0.2$\,{\scriptsize$\pm$0.2}} & $+4.8$ & \textbf{\boldmath $+5.9$} & \gapchip{green!15}{$+1.1$\,{\scriptsize$\pm$1.5}} & $+1.1$ & \textbf{\boldmath $+1.6$} & \gapchip{green!7}{$+0.5$\,{\scriptsize$\pm$0.2}} & $-0.2$ & \textbf{\boldmath $+0.0$} & \gapchip{green!2}{$+0.1$\,{\scriptsize$\pm$0.2}} \\
 & 4 & \textbf{\boldmath $+1.2$} & $+0.6$ & \gapchip{red!9}{$-0.7$\,{\scriptsize$\pm$0.2}} & \textbf{\boldmath $+0.1$} & $-0.5$ & \gapchip{red!9}{$-0.6$\,{\scriptsize$\pm$0.5}} & \textbf{\boldmath $+0.8$} & $+0.4$ & \gapchip{red!6}{$-0.4$\,{\scriptsize$\pm$0.5}} & \textbf{\boldmath $+5.5$} & $+5.4$ & \gapchip{red!1}{$-0.1$\,{\scriptsize$\pm$1.4}} & \textbf{\boldmath $+1.2$} & $+0.6$ & \gapchip{red!8}{$-0.6$\,{\scriptsize$\pm$0.5}} & \textbf{\boldmath $+0.1$} & $-0.4$ & \gapchip{red!7}{$-0.5$\,{\scriptsize$\pm$0.2}} \\
 & 8 & \textbf{\boldmath $+1.6$} & $+0.7$ & \gapchip{red!13}{$-0.9$\,{\scriptsize$\pm$0.4}} & \textbf{\boldmath $+0.4$} & $-0.5$ & \gapchip{red!13}{$-0.9$\,{\scriptsize$\pm$0.3}} & \textbf{\boldmath $+1.1$} & $+0.3$ & \gapchip{red!12}{$-0.8$\,{\scriptsize$\pm$1.3}} & $+4.6$ & \textbf{\boldmath $+4.8$} & \gapchip{green!3}{$+0.2$\,{\scriptsize$\pm$0.6}} & \textbf{\boldmath $+0.8$} & $+0.6$ & \gapchip{red!3}{$-0.2$\,{\scriptsize$\pm$0.6}} & $+0.3$ & $+0.3$ & \gapchip{green!1}{$+0.1$\,{\scriptsize$\pm$0.7}} \\
 & 16 & \textbf{\boldmath $+2.5$} & $+0.4$ & \gapchip{red!28}{$-2.1$\,{\scriptsize$\pm$1.2}} & \textbf{\boldmath $+0.4$} & $-1.5$ & \gapchip{red!27}{$-1.9$\,{\scriptsize$\pm$0.5}} & \textbf{\boldmath $+0.6$} & $+0.0$ & \gapchip{red!8}{$-0.6$\,{\scriptsize$\pm$0.3}} & \textbf{\boldmath $+5.0$} & $+4.8$ & \gapchip{red!3}{$-0.2$\,{\scriptsize$\pm$4.3}} & \textbf{\boldmath $+1.0$} & $+0.4$ & \gapchip{red!8}{$-0.6$\,{\scriptsize$\pm$1.0}} & \textbf{\boldmath $+0.2$} & $-0.7$ & \gapchip{red!13}{$-0.9$\,{\scriptsize$\pm$0.7}} \\
\bottomrule
\end{tabular}
}\\[0.8em]
{\small (b) greedy}\\[0.2em]
\resizebox{\linewidth}{!}{\begin{tabular}{c l cccccc@{\hspace{1.2em}}cccccccccccc}
\toprule
 & & \multicolumn{6}{c}{in-distribution} & \multicolumn{12}{c}{out-of-distribution} \\
\cmidrule(lr){3-8}\cmidrule(lr){9-20}
 & & \multicolumn{3}{c}{GSM8K} & \multicolumn{3}{c}{MATH-500} & \multicolumn{3}{c}{AIME24} & \multicolumn{3}{c}{AMC23} & \multicolumn{3}{c}{Minerva} & \multicolumn{3}{c}{Olympiad} \\
\cmidrule(lr){3-5}\cmidrule(lr){6-8}\cmidrule(lr){9-11}\cmidrule(lr){12-14}\cmidrule(lr){15-17}\cmidrule(lr){18-20}
 & $u$ & GRPO & LlMBO & $\Delta$ & GRPO & LlMBO & $\Delta$ & GRPO & LlMBO & $\Delta$ & GRPO & LlMBO & $\Delta$ & GRPO & LlMBO & $\Delta$ & GRPO & LlMBO & $\Delta$ \\
\midrule
\multirow{6}{*}{\rotatebox[origin=c]{90}{\scriptsize\begin{tabular}{@{}c@{}}\raisebox{-0.15em}{\includegraphics[height=0.9em]{logos/qwen.png}}~Qwen2.5\\ 7B-Inst.\end{tabular}}} & base & \multicolumn{3}{c}{91.7} & \multicolumn{3}{c}{75.4} & \multicolumn{3}{c}{13.3} & \multicolumn{3}{c}{50.0} & \multicolumn{3}{c}{41.2} & \multicolumn{3}{c}{36.8} \\
 & 1 & \textbf{\boldmath $+0.1$} & $-0.2$ & \gapchip{red!5}{$-0.4$\,{\scriptsize$\pm$0.4}} & $-0.9$ & $-0.9$ & \gapchip{green!4}{$+0.3$\,{\scriptsize$\pm$0.2}} & \textbf{\boldmath $-4.2$} & $-5.6$ & \gapchip{green!16}{$+1.1$\,{\scriptsize$\pm$1.1}} & $-1.2$ & \textbf{\boldmath $-0.8$} & \gapchip{red!28}{$-4.2$\,{\scriptsize$\pm$0.8}} & \textbf{\boldmath $-0.2$} & $-0.9$ & \gapchip{red!9}{$-0.6$\,{\scriptsize$\pm$0.7}} & $+0.0$ & $+0.0$ & \gapchip{green!4}{$+0.3$\,{\scriptsize$\pm$0.6}} \\
 & 2 & \textbf{\boldmath $+0.1$} & $-0.1$ & \gapchip{red!1}{$-0.1$\,{\scriptsize$\pm$0.2}} & $-1.1$ & $-1.1$ & \gapchip{red!2}{$-0.1$\,{\scriptsize$\pm$0.5}} & \textbf{\boldmath $-4.6$} & $-7.8$ & \gapchip{red!28}{$-3.3$\,{\scriptsize$\pm$0.0}} & \textbf{\boldmath $-0.9$} & $-3.3$ & \gapchip{red!12}{$-0.8$\,{\scriptsize$\pm$2.2}} & $-0.6$ & \textbf{\boldmath $+0.1$} & \gapchip{red!3}{$-0.2$\,{\scriptsize$\pm$0.9}} & $-0.1$ & \textbf{\boldmath $+0.2$} & \gapchip{green!10}{$+0.7$\,{\scriptsize$\pm$0.5}} \\
 & 4 & \textbf{\boldmath $+0.0$} & $-0.5$ & \gapchip{red!5}{$-0.3$\,{\scriptsize$\pm$0.3}} & \textbf{\boldmath $-0.5$} & $-2.7$ & \gapchip{red!22}{$-1.6$\,{\scriptsize$\pm$0.4}} & $-5.8$ & \textbf{\boldmath $-2.2$} & \gapchip{green!28}{$+4.4$\,{\scriptsize$\pm$2.2}} & $-3.8$ & \textbf{\boldmath $-0.8$} & \gapchip{green!28}{$+3.3$\,{\scriptsize$\pm$2.2}} & \textbf{\boldmath $-0.6$} & $-1.1$ & \gapchip{red!10}{$-0.7$\,{\scriptsize$\pm$0.9}} & \textbf{\boldmath $-0.2$} & $-0.7$ & \gapchip{red!4}{$-0.3$\,{\scriptsize$\pm$0.6}} \\
 & 8 & $-0.1$ & \textbf{\boldmath $+0.2$} & \gapchip{green!4}{$+0.3$\,{\scriptsize$\pm$0.3}} & \textbf{\boldmath $-1.6$} & $-1.9$ & \gapchip{red!4}{$-0.3$\,{\scriptsize$\pm$0.5}} & $-5.6$ & $-5.6$ & \gapchip{none}{$+0.0$\,{\scriptsize$\pm$1.9}} & \textbf{\boldmath $-3.3$} & $-10.0$ & \gapchip{red!28}{$-6.7$\,{\scriptsize$\pm$4.4}} & \textbf{\boldmath $+0.6$} & $-1.0$ & \gapchip{red!22}{$-1.6$\,{\scriptsize$\pm$0.2}} & \textbf{\boldmath $-0.5$} & $-0.8$ & \gapchip{red!4}{$-0.3$\,{\scriptsize$\pm$0.6}} \\
 & 16 & \textbf{\boldmath $-0.1$} & $-1.0$ & \gapchip{red!13}{$-0.9$\,{\scriptsize$\pm$0.0}} & \textbf{\boldmath $-1.0$} & $-4.0$ & \gapchip{red!28}{$-3.0$\,{\scriptsize$\pm$0.4}} & $-5.6$ & $-5.6$ & \gapchip{none}{$+0.0$\,{\scriptsize$\pm$3.3}} & \textbf{\boldmath $+0.8$} & $+0.0$ & \gapchip{red!12}{$-0.8$\,{\scriptsize$\pm$3.6}} & $-1.1$ & \textbf{\boldmath $-0.9$} & \gapchip{green!3}{$+0.2$\,{\scriptsize$\pm$0.9}} & \textbf{\boldmath $+0.1$} & $-2.4$ & \gapchip{red!28}{$-2.5$\,{\scriptsize$\pm$0.6}} \\
\midrule
\multirow{6}{*}{\rotatebox[origin=c]{90}{\scriptsize\begin{tabular}{@{}c@{}}\raisebox{-0.15em}{\includegraphics[height=0.9em]{logos/qwen.png}}~Qwen2.5\\ 1.5B-Inst.\end{tabular}}} & base & \multicolumn{3}{c}{73.9} & \multicolumn{3}{c}{55.8} & \multicolumn{3}{c}{0.0} & \multicolumn{3}{c}{17.5} & \multicolumn{3}{c}{22.8} & \multicolumn{3}{c}{20.5} \\
 & 1 & $-0.3$ & $-0.3$ & \gapchip{red!1}{$-0.1$\,{\scriptsize$\pm$0.2}} & \textbf{\boldmath $-1.1$} & $-2.1$ & \gapchip{red!23}{$-1.7$\,{\scriptsize$\pm$1.1}} & \textbf{\boldmath $+2.5$} & $+2.2$ & \gapchip{none}{$+0.0$\,{\scriptsize$\pm$3.3}} & \textbf{\boldmath $+8.4$} & $+7.5$ & \gapchip{red!23}{$-1.7$\,{\scriptsize$\pm$3.0}} & $-0.1$ & \textbf{\boldmath $+0.7$} & \gapchip{red!2}{$-0.1$\,{\scriptsize$\pm$1.8}} & \textbf{\boldmath $+0.6$} & $-0.4$ & \gapchip{red!22}{$-1.6$\,{\scriptsize$\pm$0.7}} \\
 & 2 & $-0.2$ & \textbf{\boldmath $-0.1$} & \gapchip{green!8}{$+0.6$\,{\scriptsize$\pm$0.3}} & $-1.2$ & \textbf{\boldmath $-0.5$} & \gapchip{red!12}{$-0.9$\,{\scriptsize$\pm$1.5}} & $+1.7$ & \textbf{\boldmath $+2.2$} & \gapchip{red!16}{$-1.1$\,{\scriptsize$\pm$1.1}} & \textbf{\boldmath $+6.9$} & $+6.7$ & \gapchip{red!28}{$-2.5$\,{\scriptsize$\pm$3.8}} & $-0.6$ & \textbf{\boldmath $+0.9$} & \gapchip{green!26}{$+1.8$\,{\scriptsize$\pm$2.2}} & \textbf{\boldmath $-0.5$} & $-0.6$ & \gapchip{green!1}{$+0.1$\,{\scriptsize$\pm$1.3}} \\
 & 4 & $-0.1$ & \textbf{\boldmath $+0.2$} & \gapchip{green!4}{$+0.3$\,{\scriptsize$\pm$0.6}} & \textbf{\boldmath $-0.8$} & $-1.0$ & \gapchip{red!7}{$-0.5$\,{\scriptsize$\pm$0.9}} & \textbf{\boldmath $+2.1$} & $+1.1$ & \gapchip{red!16}{$-1.1$\,{\scriptsize$\pm$1.1}} & \textbf{\boldmath $+8.4$} & $+2.5$ & \gapchip{red!28}{$-8.3$\,{\scriptsize$\pm$3.3}} & $-0.9$ & \textbf{\boldmath $+0.4$} & \gapchip{green!26}{$+1.8$\,{\scriptsize$\pm$0.4}} & \textbf{\boldmath $-0.4$} & $-1.5$ & \gapchip{red!17}{$-1.2$\,{\scriptsize$\pm$1.3}} \\
 & 8 & $-0.5$ & \textbf{\boldmath $+0.9$} & \gapchip{green!20}{$+1.4$\,{\scriptsize$\pm$0.6}} & \textbf{\boldmath $-0.7$} & $-2.5$ & \gapchip{red!26}{$-1.9$\,{\scriptsize$\pm$0.7}} & \textbf{\boldmath $+3.3$} & $+2.2$ & \gapchip{red!16}{$-1.1$\,{\scriptsize$\pm$1.1}} & \textbf{\boldmath $+10.0$} & $+4.2$ & \gapchip{red!28}{$-5.8$\,{\scriptsize$\pm$3.0}} & $-0.2$ & \textbf{\boldmath $+0.2$} & \gapchip{green!7}{$+0.5$\,{\scriptsize$\pm$0.5}} & $-0.5$ & \textbf{\boldmath $+0.0$} & \gapchip{green!8}{$+0.5$\,{\scriptsize$\pm$1.1}} \\
 & 16 & \textbf{\boldmath $-0.4$} & $-0.6$ & \gapchip{red!2}{$-0.1$\,{\scriptsize$\pm$1.0}} & \textbf{\boldmath $-2.2$} & $-3.2$ & \gapchip{red!14}{$-1.0$\,{\scriptsize$\pm$1.8}} & \textbf{\boldmath $+3.3$} & $+2.2$ & \gapchip{red!16}{$-1.1$\,{\scriptsize$\pm$4.0}} & \textbf{\boldmath $+10.8$} & $+5.8$ & \gapchip{red!28}{$-5.0$\,{\scriptsize$\pm$1.4}} & \textbf{\boldmath $-1.5$} & $-2.2$ & \gapchip{red!10}{$-0.7$\,{\scriptsize$\pm$1.3}} & \textbf{\boldmath $-0.8$} & $-0.9$ & \gapchip{red!1}{$-0.1$\,{\scriptsize$\pm$0.1}} \\
\bottomrule
\end{tabular}
}
\end{table}

\begin{figure}[!htbp]
\centering
\includegraphics[width=\linewidth]{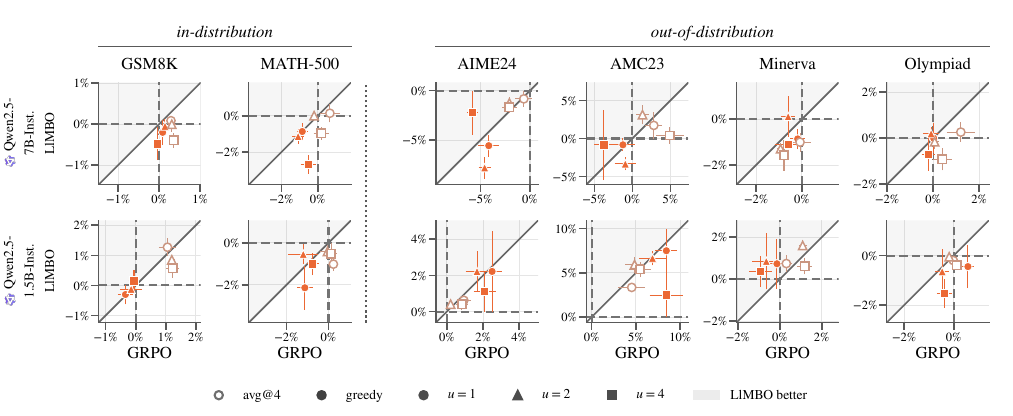}
\caption{Gain over the base model of \llmbo{} against GRPO's, in avg@4 and in greedy accuracy (color; greedy filled, avg@4 open), \llmbo{} trained on the metric of each point, for Qwen2.5-7B-Instruct and Qwen2.5-1.5B-Instruct (rows) on each benchmark (columns), each panel on its own axes. Training uses GSM8K and MATH: the left two columns are in-distribution, the four benchmarks right of the dotted line out-of-distribution. One point per $u \le 4$ (marker), mean $\pm$ s.e.\ over seeds. Points in the shaded half above the diagonal favor \llmbo{}; dashed: no gain.}
\label{fig:scatter-instruct}
\end{figure}

\begin{figure}[!htbp]
\centering
\includegraphics[width=\linewidth]{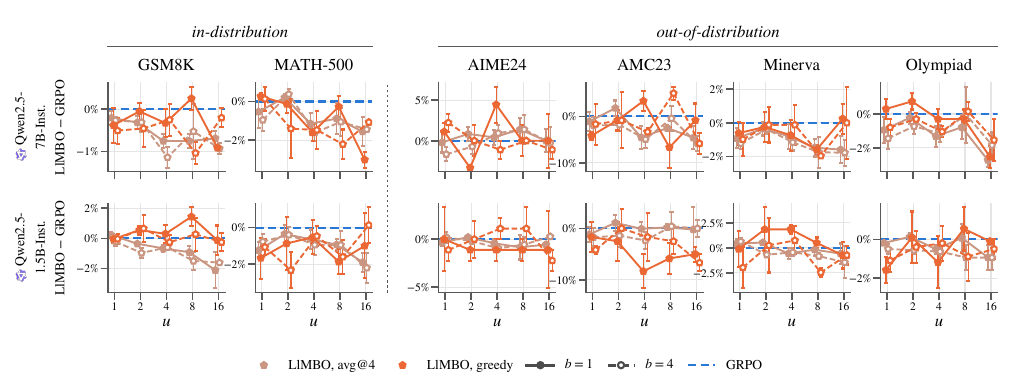}
\caption{Paired gap \llmbo{} $-$ GRPO in avg@4 and in greedy accuracy (color), \llmbo{} trained on each metric, against $u \in \{1, \dots, 16\}$, mean $\pm$ s.e.\ over seed pairs, for Qwen2.5-7B-Instruct and Qwen2.5-1.5B-Instruct (rows), one column per benchmark (in-distribution GSM8K and MATH-500 first, then the four out-of-distribution benchmarks), each panel on its own gap axis. Solid, filled: $b = 1$ candidate per batch (the default); dashed, open: the $b = 4$ ablation.}
\label{fig:gap-instruct}
\end{figure}

Under GRPO, avg@4 on GSM8K and MATH-500 moves by at most \grpoBestInstruct{} points at any $u \le 4$ (Table~\ref{tab:instruct}a), and the points of both optimizers cluster around zero gain (Figure~\ref{fig:scatter-instruct}). \llmbo{} is level with GRPO on GSM8K avg@4 on Qwen2.5-1.5B-Instruct (\gapToneMeanInstSmallUOneGsm{}, \gapToneMeanInstSmallUTwoGsm{}, \gapToneMeanInstSmallUFourGsm{} at $u = 1, 2, 4$) and trails slightly on MATH-500 avg@4: on Qwen2.5-7B-Instruct \gainToneMeanInstUOneMathFive{} and \gainToneMeanInstUFourMathFive{} points at $u = 1, 4$, against GRPO's \gainGrpoMeanInstUOneMathFive{} and \gainGrpoMeanInstUFourMathFive{}, and on Qwen2.5-1.5B-Instruct \gainToneMeanInstSmallUOneMathFive{} against \gainGrpoMeanInstSmallUOneMathFive{} at $u = 1$. In greedy accuracy (Table~\ref{tab:instruct}b, Figures~\ref{fig:scatter-instruct} and~\ref{fig:gap-instruct}) \llmbo{} stays within about half a point of GRPO on both models, except on Qwen2.5-7B-Instruct's MATH-500 at $u = 4$ (\gapGreedyGreedyInstUFourMathFive{}). GRPO's small gains are no larger than the noise of an evaluation, so the Instruct models only test whether an optimizer does harm when there is nothing to find. GRPO does essentially none; \llmbo{} does a little.

\subsection{Too little signal to certify $\theta = 0$}
\label{app:instruct-snr}

\begin{table}[!htbp]
\centering
\caption{What the search sees and what it delivers, per \llmbo{} run at $u \le 4$ ($b = 1$, trained on the pass rate at $T = 1$), mean $\pm$ s.e.\ over seeds; Qwen2.5-7B, which TinyLoRA moves, for contrast. \emph{Training distribution}: the base model's sampled pass rate on the training tier, estimated from the eight $\theta = 0$ replicates of the search design; the standard error of one candidate's evaluation (one completion on each of $n = 256$ prompts), in points; \emph{GP gain}: the final GP's posterior mean at the returned adapter minus its posterior mean at $\theta = 0$, in pass-rate points, the improvement the search believes it found; $\|\theta\|_\infty / R$: how far out in the search box $[-R, R]^u$ the returned adapter lies. \emph{Held-out gain}: avg@4 accuracy on GSM8K and MATH-500 minus the base model's (points).}
\label{tab:snr}
\begin{tabular}{l c cccc cc}
\toprule
 & & \multicolumn{4}{c}{training distribution} & \multicolumn{2}{c}{held-out gain} \\
\cmidrule(lr){3-6}\cmidrule(lr){7-8}
Model & $u$ & base (\%) & s.e./eval. & GP gain & $\|\theta\|_\infty / R$ & GSM8K & MATH-500 \\
\midrule
\raisebox{-0.15em}{\includegraphics[height=0.9em]{logos/qwen.png}}~Qwen2.5-7B-Inst. & 1 & $65.1$\,{\scriptsize$\pm$0.2} & $3.0$ & $+0.8$\,{\scriptsize$\pm$0.2} & $0.11$\,{\scriptsize$\pm$0.02} & $+0.1$\,{\scriptsize$\pm$0.1} & $+0.1$\,{\scriptsize$\pm$0.5} \\
 & 2 & $64.9$\,{\scriptsize$\pm$0.3} & $3.0$ & $+1.6$\,{\scriptsize$\pm$0.2} & $0.20$\,{\scriptsize$\pm$0.05} & $+0.0$\,{\scriptsize$\pm$0.2} & $+0.0$\,{\scriptsize$\pm$0.2} \\
 & 4 & $65.2$\,{\scriptsize$\pm$0.3} & $2.9$ & $+2.3$\,{\scriptsize$\pm$0.3} & $0.45$\,{\scriptsize$\pm$0.09} & $-0.4$\,{\scriptsize$\pm$0.2} & $-1.0$\,{\scriptsize$\pm$0.4} \\
\midrule
\raisebox{-0.15em}{\includegraphics[height=0.9em]{logos/qwen.png}}~Qwen2.5-1.5B-Inst. & 1 & $77.4$\,{\scriptsize$\pm$0.2} & $2.6$ & $+0.5$\,{\scriptsize$\pm$0.2} & $0.16$\,{\scriptsize$\pm$0.03} & $+1.3$\,{\scriptsize$\pm$0.3} & $-1.0$\,{\scriptsize$\pm$0.3} \\
 & 2 & $77.2$\,{\scriptsize$\pm$0.2} & $2.6$ & $+1.2$\,{\scriptsize$\pm$0.3} & $0.18$\,{\scriptsize$\pm$0.03} & $+0.8$\,{\scriptsize$\pm$0.1} & $-0.4$\,{\scriptsize$\pm$0.3} \\
 & 4 & $77.2$\,{\scriptsize$\pm$0.1} & $2.5$ & $+1.7$\,{\scriptsize$\pm$0.3} & $0.36$\,{\scriptsize$\pm$0.04} & $+0.6$\,{\scriptsize$\pm$0.3} & $-0.5$\,{\scriptsize$\pm$0.5} \\
\midrule
\raisebox{-0.15em}{\includegraphics[height=0.9em]{logos/qwen.png}}~Qwen2.5-7B & 1 & $42.1$\,{\scriptsize$\pm$0.6} & $3.1$ & $+4.6$\,{\scriptsize$\pm$0.8} & $0.22$\,{\scriptsize$\pm$0.03} & $+5.8$\,{\scriptsize$\pm$1.0} & $+5.8$\,{\scriptsize$\pm$1.0} \\
 & 2 & $42.2$\,{\scriptsize$\pm$0.6} & $3.1$ & $+6.3$\,{\scriptsize$\pm$1.0} & $0.32$\,{\scriptsize$\pm$0.04} & $+6.9$\,{\scriptsize$\pm$1.1} & $+7.1$\,{\scriptsize$\pm$1.0} \\
 & 4 & $41.8$\,{\scriptsize$\pm$0.7} & $3.1$ & $+9.2$\,{\scriptsize$\pm$0.9} & $0.64$\,{\scriptsize$\pm$0.08} & $+9.8$\,{\scriptsize$\pm$0.8} & $+10.3$\,{\scriptsize$\pm$0.5} \\
\bottomrule
\end{tabular}

\end{table}

The noise in a single \llmbo{} evaluation can exceed the performance gain we seek to detect since it scores a candidate by one sampled completion on each of $n = 256$ training prompts. At the Instruct models' training pass rates the standard error of that estimate is \snrSemInstruct{} points (Table~\ref{tab:snr}); even the base model's rate, measured eight times in the initial design, is known only to $\pm$\snrReplicateSem{} points. Telling two adapters apart by $\delta$ at two standard errors of their difference takes about $8\sigma^2/\delta^2$ evaluations of each: \snrEvalsOne{} for $\delta = 1$ point and \snrEvalsHalf{} for $\delta = 0.5$. A search has about \snrTrials{} evaluations in all, spread over the search box. It therefore cannot establish that no adapter beats $\theta = 0$, the correct answer here: with a flat objective it can only return the trial whose posterior mean happens to be highest.

The search can fit evaluation noise by picking the best of a few hundred noisy evaluations is optimistic by construction (the winner's curse), and the GP removes only part of that bias. Its posterior believes the returned adapter improves the training pass rate in every row of Table~\ref{tab:snr}, by \snrBelievedInstUOne{} points on Qwen2.5-7B-Instruct at $u = 1$, rising to \snrBelievedInstUFour{} at $u = 4$ (\snrBelievedInstSmallUOne{} to \snrBelievedInstSmallUFour{} on Qwen2.5-1.5B-Instruct). On held-out MATH-500, the in-distribution benchmark of every training set, the same runs lose accuracy instead. The optimism grows with $u$: more dimensions hold more noise to fit, and the returned adapter moves further from the origin, from \snrThetaInstUOne{} of the box half-width at $u = 1$ to \snrThetaInstUFour{} at $u = 4$ on Qwen2.5-7B-Instruct, as held-out accuracy falls. The GP is not miscalibrated in general: on Qwen2.5-7B, whose pass rate TinyLoRA does move, its believed gains (\snrBelievedQwenUOne{} to \snrBelievedQwenUFour{} points at $u = 1, 4$) track the held-out gains of the same runs. The failure is specific to a regime where the true gain is below what the evaluation budget can resolve.

GRPO starts at $\theta = 0$ and moves only as far as its gradient consistently pushes it, favoring solutions near the initialization, whereas BO can explore candidates farther from $\theta = 0$. With Adam, a coordinate whose gradient keeps its sign moves about $\eta$ per step, up to $\eta S = R$ after $S$ steps, which is how we set the search box (Section~\ref{sec:method}). A gradient that is pure noise instead gives updates of random sign, correlated only over the momentum horizon, and for $\beta_1 = 0.9$ these add up to a random walk of about $\eta\sqrt{S}$, a fraction $1/\sqrt{S} \approx 0.05$ of the box after $S \approx 400$ steps. Without signal, GRPO thus stays close to the base model, as if under a prior concentrated at $\theta = 0$ with width shrinking as $1/\sqrt{S}$, and its held-out accuracy barely moves. \llmbo{} has no such prior: its Sobol designs and trust-region restarts cover the box uniformly, its GP has a constant prior mean, and $\theta = 0$ competes with every other evaluated point on equal terms, so on a flat objective the pick lands wherever the noise peaks. Where there is a gain to find, as on the models of Appendix~\ref{app:results}, this freedom is what lets \llmbo{} find it within the budget. Where there is none, it costs a fraction of a point.

The bias is cheap to add to \llmbo{} by returning $\theta = 0$ unless the posterior probability that the pick beats it is high, re-evaluate the final candidates against fresh $\theta = 0$ replicates before committing, or place a GP prior mean or a trust-region schedule that favors the origin. Each trades a little of the search's reach for safety on models with no headroom.

\FloatBarrier
\subsection{Training objective: greedy against $T = 1$}
\label{app:objective}

Everywhere else each metric is reported for \llmbo{} trained on it. Here we compare the two objectives directly. For each seed, \llmbo{} trained on greedy accuracy and \llmbo{} trained on the pass rate at $T = 1$ search the same TinyLoRA subspace (the SVD bases and random projections of the same GRPO seed), and both are evaluated greedily (Table~\ref{tab:objective}).
At $u = 4$ training on greedy accuracy raises greedy accuracy in every model group: on greedy GSM8K by \objectiveUFourGsmBase{} points over the four vanilla base models, \objectiveUFourGsmMathVariants{} over the two math models and \objectiveUFourGsmSmall{} over the two 1.5B models, and on greedy MATH-500 by \objectiveUFourMathFiveBase{}, \objectiveUFourMathFiveMathVariants{} and \objectiveUFourMathFiveSmall{}. The difference is largest on Qwen2.5-Math-1.5B (\objectiveMathSmallUTwoGsm{} points on greedy GSM8K at $u = 2$), whose base model's repetition loop only the greedy objective sees (Appendix~\ref{app:greedy-small}). On Qwen2.5-7B the two objectives are level (\objectiveQwenUFourGsm{} on GSM8K, \objectiveQwenUFourMathFive{} on MATH-500 at $u = 4$).

\begin{table}[!htbp]
\centering
\caption{Greedy accuracy at the end of training of \llmbo{} trained on the pass rate at $T{=}1$ (\emph{avg@4}) and on greedy accuracy (\emph{greedy}), the same 3 seeds per cell searching the same subspaces, for the four vanilla base models, the two math-specialized 7B models and the two Qwen2.5 1.5B models (row blocks), per $u \le 16$ (rows) and benchmark (columns: in-distribution GSM8K and MATH-500, then the four out-of-distribution benchmarks). \emph{base}: greedy accuracy of the base model (\%). \emph{avg@4}, \emph{greedy}: gain over the base model (points), mean over seeds; bold: the larger, unless tied. $\Delta$: paired gap of the greedy objective over the avg@4 one, mean $\pm$ s.e.\ over seed pairs; tint: its sign and size (green: the greedy objective ahead).}
\label{tab:objective}
\renewcommand{\arraystretch}{0.85}
\resizebox{\linewidth}{!}{\begin{tabular}{c l cccccc@{\hspace{1.2em}}cccccccccccc}
\toprule
 & & \multicolumn{6}{c}{in-distribution} & \multicolumn{12}{c}{out-of-distribution} \\
\cmidrule(lr){3-8}\cmidrule(lr){9-20}
 & & \multicolumn{3}{c}{GSM8K} & \multicolumn{3}{c}{MATH-500} & \multicolumn{3}{c}{AIME24} & \multicolumn{3}{c}{AMC23} & \multicolumn{3}{c}{Minerva} & \multicolumn{3}{c}{Olympiad} \\
\cmidrule(lr){3-5}\cmidrule(lr){6-8}\cmidrule(lr){9-11}\cmidrule(lr){12-14}\cmidrule(lr){15-17}\cmidrule(lr){18-20}
 & $u$ & avg@4 & greedy & $\Delta$ & avg@4 & greedy & $\Delta$ & avg@4 & greedy & $\Delta$ & avg@4 & greedy & $\Delta$ & avg@4 & greedy & $\Delta$ & avg@4 & greedy & $\Delta$ \\
\midrule
\multirow{6}{*}{\rotatebox[origin=c]{90}{\scriptsize\begin{tabular}{@{}c@{}}\raisebox{-0.15em}{\includegraphics[height=0.9em]{logos/llama.png}}~Llama3.1\\ 8B\end{tabular}}} & base & \multicolumn{3}{c}{43.9} & \multicolumn{3}{c}{16.4} & \multicolumn{3}{c}{0.0} & \multicolumn{3}{c}{5.0} & \multicolumn{3}{c}{7.4} & \multicolumn{3}{c}{2.8} \\
 & 1 & $+1.1$ & \textbf{\boldmath $+2.2$} & \gapchip{green!2}{$+0.2$\,{\scriptsize$\pm$0.6}} & $-0.3$ & \textbf{\boldmath $+0.1$} & \gapchip{red!3}{$-0.2$\,{\scriptsize$\pm$0.4}} & $+0.0$ & $+0.0$ & \gapchip{none}{$+0.0$\,{\scriptsize$\pm$0.0}} & $-1.6$ & \textbf{\boldmath $+0.0$} & \gapchip{green!12}{$+0.8$\,{\scriptsize$\pm$1.7}} & $-1.3$ & \textbf{\boldmath $-0.7$} & \gapchip{green!3}{$+0.2$\,{\scriptsize$\pm$0.3}} & $-0.2$ & \textbf{\boldmath $+0.0$} & \gapchip{green!1}{$+0.0$\,{\scriptsize$\pm$0.2}} \\
 & 2 & $-0.3$ & \textbf{\boldmath $+1.6$} & \gapchip{green!28}{$+3.8$\,{\scriptsize$\pm$0.3}} & $-1.7$ & \textbf{\boldmath $-0.3$} & \gapchip{green!28}{$+2.1$\,{\scriptsize$\pm$0.6}} & $+0.0$ & $+0.0$ & \gapchip{none}{$+0.0$\,{\scriptsize$\pm$0.0}} & $-2.5$ & \textbf{\boldmath $+0.0$} & \gapchip{green!28}{$+3.3$\,{\scriptsize$\pm$1.7}} & $-2.6$ & \textbf{\boldmath $-2.0$} & \gapchip{green!5}{$+0.4$\,{\scriptsize$\pm$1.0}} & \textbf{\boldmath $-0.3$} & $-0.5$ & \gapchip{green!1}{$+0.1$\,{\scriptsize$\pm$0.7}} \\
 & 4 & $-1.6$ & \textbf{\boldmath $+2.7$} & \gapchip{green!28}{$+4.3$\,{\scriptsize$\pm$3.9}} & $-2.1$ & \textbf{\boldmath $+1.2$} & \gapchip{green!28}{$+3.5$\,{\scriptsize$\pm$1.3}} & $+0.0$ & $+0.0$ & \gapchip{none}{$+0.0$\,{\scriptsize$\pm$0.0}} & \textbf{\boldmath $-1.6$} & $-1.7$ & \gapchip{none}{$+0.0$\,{\scriptsize$\pm$1.4}} & $-2.3$ & \textbf{\boldmath $-2.0$} & \gapchip{green!15}{$+1.1$\,{\scriptsize$\pm$1.2}} & $-0.4$ & $-0.4$ & \gapchip{green!1}{$+0.0$\,{\scriptsize$\pm$0.3}} \\
 & 8 & $-5.1$ & \textbf{\boldmath $+2.2$} & \gapchip{green!28}{$+7.3$\,{\scriptsize$\pm$2.3}} & $-3.6$ & \textbf{\boldmath $+0.7$} & \gapchip{green!28}{$+4.3$\,{\scriptsize$\pm$1.2}} & $+0.0$ & $+0.0$ & \gapchip{none}{$+0.0$\,{\scriptsize$\pm$0.0}} & $-3.3$ & $-3.3$ & \gapchip{none}{$+0.0$\,{\scriptsize$\pm$0.0}} & $-3.1$ & \textbf{\boldmath $-1.2$} & \gapchip{green!26}{$+1.8$\,{\scriptsize$\pm$0.9}} & $-0.8$ & \textbf{\boldmath $-0.2$} & \gapchip{green!9}{$+0.6$\,{\scriptsize$\pm$0.7}} \\
 & 16 & $-7.0$ & \textbf{\boldmath $+2.9$} & \gapchip{green!28}{$+10.0$\,{\scriptsize$\pm$0.5}} & $-4.3$ & \textbf{\boldmath $-0.8$} & \gapchip{green!28}{$+3.5$\,{\scriptsize$\pm$0.2}} & $+0.0$ & $+0.0$ & \gapchip{none}{$+0.0$\,{\scriptsize$\pm$0.0}} & $-5.0$ & \textbf{\boldmath $-1.7$} & \gapchip{green!28}{$+3.3$\,{\scriptsize$\pm$0.8}} & $-2.7$ & \textbf{\boldmath $-1.5$} & \gapchip{green!17}{$+1.2$\,{\scriptsize$\pm$0.7}} & $-0.9$ & \textbf{\boldmath $-0.4$} & \gapchip{green!6}{$+0.4$\,{\scriptsize$\pm$0.8}} \\
\midrule
\multirow{6}{*}{\rotatebox[origin=c]{90}{\scriptsize\begin{tabular}{@{}c@{}}\raisebox{-0.15em}{\includegraphics[height=0.9em]{logos/mistral.png}}~Mistral\\ 7B\end{tabular}}} & base & \multicolumn{3}{c}{22.4} & \multicolumn{3}{c}{7.4} & \multicolumn{3}{c}{0.0} & \multicolumn{3}{c}{0.0} & \multicolumn{3}{c}{5.5} & \multicolumn{3}{c}{1.5} \\
 & 1 & $-1.0$ & \textbf{\boldmath $+1.1$} & \gapchip{red!2}{$-0.2$\,{\scriptsize$\pm$0.2}} & $+0.4$ & \textbf{\boldmath $+1.3$} & \gapchip{green!8}{$+0.6$\,{\scriptsize$\pm$0.5}} & $+0.0$ & $+0.0$ & \gapchip{none}{$+0.0$\,{\scriptsize$\pm$0.0}} & \textbf{\boldmath $+1.2$} & $+0.0$ & \gapchip{red!23}{$-1.7$\,{\scriptsize$\pm$0.8}} & \textbf{\boldmath $-0.9$} & $-1.3$ & \gapchip{green!7}{$+0.5$\,{\scriptsize$\pm$0.7}} & $-0.1$ & \textbf{\boldmath $+0.0$} & \gapchip{red!2}{$-0.1$\,{\scriptsize$\pm$0.1}} \\
 & 2 & $-1.5$ & \textbf{\boldmath $+0.8$} & \gapchip{green!28}{$+3.2$\,{\scriptsize$\pm$1.9}} & \textbf{\boldmath $+1.3$} & $+1.0$ & \gapchip{green!14}{$+1.0$\,{\scriptsize$\pm$0.3}} & $+0.0$ & $+0.0$ & \gapchip{none}{$+0.0$\,{\scriptsize$\pm$0.0}} & $+0.6$ & \textbf{\boldmath $+3.3$} & \gapchip{green!28}{$+2.5$\,{\scriptsize$\pm$2.9}} & $-2.0$ & \textbf{\boldmath $-1.7$} & \gapchip{green!7}{$+0.5$\,{\scriptsize$\pm$0.3}} & $-0.1$ & \textbf{\boldmath $+0.1$} & \gapchip{green!3}{$+0.2$\,{\scriptsize$\pm$0.2}} \\
 & 4 & $-1.4$ & \textbf{\boldmath $+1.0$} & \gapchip{green!28}{$+7.1$\,{\scriptsize$\pm$2.9}} & $+0.7$ & \textbf{\boldmath $+1.0$} & \gapchip{green!10}{$+0.7$\,{\scriptsize$\pm$1.3}} & \textbf{\boldmath $+0.4$} & $+0.0$ & \gapchip{none}{$+0.0$\,{\scriptsize$\pm$0.0}} & $+0.6$ & \textbf{\boldmath $+1.7$} & \gapchip{green!23}{$+1.7$\,{\scriptsize$\pm$0.8}} & \textbf{\boldmath $-0.9$} & $-1.2$ & \gapchip{red!5}{$-0.4$\,{\scriptsize$\pm$0.4}} & $+0.1$ & \textbf{\boldmath $+0.2$} & \gapchip{none}{$+0.0$\,{\scriptsize$\pm$0.3}} \\
 & 8 & $-1.9$ & \textbf{\boldmath $-0.9$} & \gapchip{green!15}{$+1.1$\,{\scriptsize$\pm$2.1}} & $-0.5$ & \textbf{\boldmath $+2.3$} & \gapchip{green!28}{$+2.7$\,{\scriptsize$\pm$1.0}} & $+0.0$ & $+0.0$ & \gapchip{none}{$+0.0$\,{\scriptsize$\pm$0.0}} & \textbf{\boldmath $+2.5$} & $+1.7$ & \gapchip{red!12}{$-0.8$\,{\scriptsize$\pm$1.7}} & $-3.2$ & \textbf{\boldmath $-1.5$} & \gapchip{green!24}{$+1.7$\,{\scriptsize$\pm$0.5}} & $-0.3$ & \textbf{\boldmath $+0.0$} & \gapchip{green!4}{$+0.3$\,{\scriptsize$\pm$0.5}} \\
 & 16 & $-11.9$ & \textbf{\boldmath $+0.1$} & \gapchip{green!28}{$+12.0$\,{\scriptsize$\pm$3.1}} & $-2.2$ & \textbf{\boldmath $+2.0$} & \gapchip{green!28}{$+4.2$\,{\scriptsize$\pm$0.9}} & $+0.0$ & $+0.0$ & \gapchip{none}{$+0.0$\,{\scriptsize$\pm$0.0}} & \textbf{\boldmath $+1.7$} & $+0.8$ & \gapchip{red!12}{$-0.8$\,{\scriptsize$\pm$1.7}} & $-2.6$ & \textbf{\boldmath $-0.2$} & \gapchip{green!28}{$+2.3$\,{\scriptsize$\pm$1.2}} & $-0.2$ & \textbf{\boldmath $+0.5$} & \gapchip{green!10}{$+0.7$\,{\scriptsize$\pm$0.5}} \\
\midrule
\multirow{6}{*}{\rotatebox[origin=c]{90}{\scriptsize\begin{tabular}{@{}c@{}}\raisebox{-0.15em}{\includegraphics[height=0.9em]{logos/deepseek.png}}~DeepSeek\\ 7B\end{tabular}}} & base & \multicolumn{3}{c}{2.7} & \multicolumn{3}{c}{3.0} & \multicolumn{3}{c}{0.0} & \multicolumn{3}{c}{0.0} & \multicolumn{3}{c}{1.5} & \multicolumn{3}{c}{0.7} \\
 & 1 & $-1.5$ & \textbf{\boldmath $-0.2$} & \gapchip{green!14}{$+1.0$\,{\scriptsize$\pm$0.0}} & $+0.5$ & \textbf{\boldmath $+1.5$} & \gapchip{red!11}{$-0.8$\,{\scriptsize$\pm$0.0}} & $+0.0$ & $+0.0$ & \gapchip{none}{$+0.0$\,{\scriptsize$\pm$0.0}} & $+0.0$ & $+0.0$ & \gapchip{none}{$+0.0$\,{\scriptsize$\pm$0.0}} & $-0.3$ & \textbf{\boldmath $+0.9$} & \gapchip{green!26}{$+1.8$\,{\scriptsize$\pm$0.0}} & $-0.1$ & \textbf{\boldmath $+0.1$} & \gapchip{red!2}{$-0.1$\,{\scriptsize$\pm$0.0}} \\
 & 2 & $-1.3$ & \textbf{\boldmath $-0.5$} & \gapchip{green!9}{$+0.7$\,{\scriptsize$\pm$0.3}} & $+0.2$ & \textbf{\boldmath $+2.2$} & \gapchip{green!23}{$+1.7$\,{\scriptsize$\pm$0.5}} & $+0.0$ & $+0.0$ & \gapchip{none}{$+0.0$\,{\scriptsize$\pm$0.0}} & $+0.0$ & $+0.0$ & \gapchip{none}{$+0.0$\,{\scriptsize$\pm$0.0}} & $-0.2$ & \textbf{\boldmath $+0.7$} & \gapchip{green!17}{$+1.2$\,{\scriptsize$\pm$0.6}} & $-0.2$ & \textbf{\boldmath $+0.2$} & \gapchip{green!3}{$+0.2$\,{\scriptsize$\pm$0.2}} \\
 & 4 & $-2.2$ & \textbf{\boldmath $-0.5$} & \gapchip{green!28}{$+2.0$\,{\scriptsize$\pm$0.4}} & $-1.7$ & \textbf{\boldmath $+1.9$} & \gapchip{green!28}{$+3.9$\,{\scriptsize$\pm$0.3}} & $+0.0$ & $+0.0$ & \gapchip{none}{$+0.0$\,{\scriptsize$\pm$0.0}} & $+0.0$ & $+0.0$ & \gapchip{none}{$+0.0$\,{\scriptsize$\pm$0.0}} & $-1.2$ & \textbf{\boldmath $+1.2$} & \gapchip{green!28}{$+2.5$\,{\scriptsize$\pm$1.2}} & $-0.5$ & \textbf{\boldmath $+0.0$} & \gapchip{green!8}{$+0.5$\,{\scriptsize$\pm$0.1}} \\
 & 8 & $-0.7$ & \textbf{\boldmath $-0.3$} & \gapchip{green!6}{$+0.4$\,{\scriptsize$\pm$0.8}} & $+0.5$ & \textbf{\boldmath $+2.4$} & \gapchip{green!27}{$+1.9$\,{\scriptsize$\pm$0.6}} & $+0.0$ & $+0.0$ & \gapchip{none}{$+0.0$\,{\scriptsize$\pm$0.0}} & \textbf{\boldmath $+0.8$} & $+0.0$ & \gapchip{red!12}{$-0.8$\,{\scriptsize$\pm$0.8}} & $-0.1$ & \textbf{\boldmath $+0.2$} & \gapchip{green!5}{$+0.4$\,{\scriptsize$\pm$0.4}} & $+0.0$ & \textbf{\boldmath $+0.3$} & \gapchip{green!5}{$+0.3$\,{\scriptsize$\pm$0.3}} \\
 & 16 & $-2.4$ & \textbf{\boldmath $-0.7$} & \gapchip{green!25}{$+1.8$\,{\scriptsize$\pm$0.5}} & $-0.5$ & \textbf{\boldmath $+2.4$} & \gapchip{green!28}{$+2.9$\,{\scriptsize$\pm$1.5}} & $+0.0$ & $+0.0$ & \gapchip{none}{$+0.0$\,{\scriptsize$\pm$0.0}} & $+0.0$ & \textbf{\boldmath $+0.8$} & \gapchip{green!12}{$+0.8$\,{\scriptsize$\pm$0.8}} & $-1.0$ & \textbf{\boldmath $+0.1$} & \gapchip{green!15}{$+1.1$\,{\scriptsize$\pm$0.8}} & $-0.4$ & \textbf{\boldmath $-0.2$} & \gapchip{green!3}{$+0.2$\,{\scriptsize$\pm$0.2}} \\
\midrule
\multirow{6}{*}{\rotatebox[origin=c]{90}{\scriptsize\begin{tabular}{@{}c@{}}\raisebox{-0.15em}{\includegraphics[height=0.9em]{logos/qwen.png}}~Qwen2.5\\ 7B\end{tabular}}} & base & \multicolumn{3}{c}{88.0} & \multicolumn{3}{c}{64.0} & \multicolumn{3}{c}{0.0} & \multicolumn{3}{c}{30.0} & \multicolumn{3}{c}{32.7} & \multicolumn{3}{c}{28.6} \\
 & 1 & $-0.1$ & \textbf{\boldmath $+0.0$} & \gapchip{green!5}{$+0.4$\,{\scriptsize$\pm$0.4}} & \textbf{\boldmath $+2.0$} & $+1.5$ & \gapchip{red!6}{$-0.4$\,{\scriptsize$\pm$1.5}} & $+4.2$ & \textbf{\boldmath $+4.4$} & \gapchip{red!28}{$-2.2$\,{\scriptsize$\pm$1.1}} & $+5.3$ & \textbf{\boldmath $+6.7$} & \gapchip{green!23}{$+1.7$\,{\scriptsize$\pm$1.7}} & \textbf{\boldmath $-0.3$} & $-0.6$ & \gapchip{green!9}{$+0.6$\,{\scriptsize$\pm$1.5}} & $+0.7$ & \textbf{\boldmath $+2.1$} & \gapchip{green!28}{$+2.2$\,{\scriptsize$\pm$0.8}} \\
 & 2 & \textbf{\boldmath $-0.4$} & $-0.9$ & \gapchip{red!8}{$-0.6$\,{\scriptsize$\pm$0.4}} & \textbf{\boldmath $+2.3$} & $+1.5$ & \gapchip{red!15}{$-1.1$\,{\scriptsize$\pm$1.5}} & $+4.2$ & \textbf{\boldmath $+11.1$} & \gapchip{green!28}{$+7.8$\,{\scriptsize$\pm$2.9}} & \textbf{\boldmath $+5.3$} & $+3.3$ & \gapchip{red!23}{$-1.7$\,{\scriptsize$\pm$1.7}} & \textbf{\boldmath $+0.6$} & $-2.6$ & \gapchip{red!28}{$-3.9$\,{\scriptsize$\pm$2.5}} & $+1.7$ & \textbf{\boldmath $+2.6$} & \gapchip{green!15}{$+1.1$\,{\scriptsize$\pm$1.0}} \\
 & 4 & \textbf{\boldmath $-1.1$} & $-1.4$ & \gapchip{red!6}{$-0.4$\,{\scriptsize$\pm$0.7}} & $+2.2$ & \textbf{\boldmath $+2.4$} & \gapchip{red!11}{$-0.8$\,{\scriptsize$\pm$1.6}} & \textbf{\boldmath $+4.2$} & $+3.3$ & \gapchip{red!28}{$-2.2$\,{\scriptsize$\pm$2.9}} & $+3.4$ & \textbf{\boldmath $+6.7$} & \gapchip{green!28}{$+5.0$\,{\scriptsize$\pm$7.6}} & \textbf{\boldmath $+0.7$} & $-1.6$ & \gapchip{red!22}{$-1.6$\,{\scriptsize$\pm$1.8}} & $+1.1$ & \textbf{\boldmath $+1.6$} & \gapchip{green!17}{$+1.2$\,{\scriptsize$\pm$0.8}} \\
 & 8 & $-1.4$ & \textbf{\boldmath $-0.8$} & \gapchip{green!10}{$+0.7$\,{\scriptsize$\pm$0.4}} & \textbf{\boldmath $+1.9$} & $+1.1$ & \gapchip{red!11}{$-0.8$\,{\scriptsize$\pm$1.6}} & $+4.4$ & $+4.4$ & \gapchip{none}{$+0.0$\,{\scriptsize$\pm$1.9}} & $+5.0$ & \textbf{\boldmath $+6.7$} & \gapchip{green!23}{$+1.7$\,{\scriptsize$\pm$2.2}} & \textbf{\boldmath $-1.0$} & $-3.7$ & \gapchip{red!28}{$-2.7$\,{\scriptsize$\pm$3.2}} & $+0.1$ & \textbf{\boldmath $+2.1$} & \gapchip{green!28}{$+2.0$\,{\scriptsize$\pm$2.1}} \\
 & 16 & $-0.1$ & $-0.1$ & \gapchip{none}{$+0.0$\,{\scriptsize$\pm$0.5}} & $+2.4$ & \textbf{\boldmath $+3.4$} & \gapchip{green!14}{$+1.0$\,{\scriptsize$\pm$1.3}} & $+1.1$ & \textbf{\boldmath $+7.8$} & \gapchip{green!28}{$+6.7$\,{\scriptsize$\pm$1.9}} & $+0.8$ & \textbf{\boldmath $+7.5$} & \gapchip{green!28}{$+6.7$\,{\scriptsize$\pm$1.7}} & $+0.6$ & \textbf{\boldmath $+1.2$} & \gapchip{green!9}{$+0.6$\,{\scriptsize$\pm$1.8}} & $+2.6$ & $+2.6$ & \gapchip{red!1}{$+0.0$\,{\scriptsize$\pm$0.8}} \\
\midrule[\heavyrulewidth]
\multirow{6}{*}{\rotatebox[origin=c]{90}{\scriptsize\begin{tabular}{@{}c@{}}\raisebox{-0.15em}{\includegraphics[height=0.9em]{logos/deepseek.png}}~DeepSeek\\ Math-7B\end{tabular}}} & base & \multicolumn{3}{c}{28.7} & \multicolumn{3}{c}{21.2} & \multicolumn{3}{c}{0.0} & \multicolumn{3}{c}{15.0} & \multicolumn{3}{c}{9.6} & \multicolumn{3}{c}{5.9} \\
 & 1 & $+3.9$ & \textbf{\boldmath $+6.8$} & \gapchip{green!28}{$+2.5$\,{\scriptsize$\pm$1.7}} & \textbf{\boldmath $+1.7$} & $+0.7$ & \gapchip{red!23}{$-1.7$\,{\scriptsize$\pm$0.9}} & $+0.8$ & \textbf{\boldmath $+1.1$} & \gapchip{green!16}{$+1.1$\,{\scriptsize$\pm$1.1}} & $-6.2$ & \textbf{\boldmath $-5.8$} & \gapchip{green!12}{$+0.8$\,{\scriptsize$\pm$0.8}} & $+3.4$ & \textbf{\boldmath $+4.3$} & \gapchip{green!27}{$+2.0$\,{\scriptsize$\pm$0.9}} & $+0.5$ & \textbf{\boldmath $+1.1$} & \gapchip{green!7}{$+0.5$\,{\scriptsize$\pm$0.2}} \\
 & 2 & \textbf{\boldmath $+3.6$} & $+3.2$ & \gapchip{red!14}{$-1.0$\,{\scriptsize$\pm$4.1}} & $+1.8$ & \textbf{\boldmath $+1.9$} & \gapchip{red!7}{$-0.5$\,{\scriptsize$\pm$1.3}} & \textbf{\boldmath $+0.8$} & $+0.0$ & \gapchip{red!28}{$-2.2$\,{\scriptsize$\pm$1.1}} & $-7.8$ & \textbf{\boldmath $-4.2$} & \gapchip{green!23}{$+1.7$\,{\scriptsize$\pm$2.2}} & \textbf{\boldmath $+4.1$} & $+2.7$ & \gapchip{red!9}{$-0.6$\,{\scriptsize$\pm$2.0}} & $+0.2$ & \textbf{\boldmath $+0.9$} & \gapchip{green!13}{$+0.9$\,{\scriptsize$\pm$0.6}} \\
 & 4 & $+4.9$ & \textbf{\boldmath $+7.4$} & \gapchip{green!28}{$+4.5$\,{\scriptsize$\pm$2.4}} & $+0.3$ & \textbf{\boldmath $+2.5$} & \gapchip{green!28}{$+3.3$\,{\scriptsize$\pm$2.4}} & $+0.4$ & \textbf{\boldmath $+2.2$} & \gapchip{green!28}{$+2.2$\,{\scriptsize$\pm$1.1}} & $-9.4$ & \textbf{\boldmath $-6.7$} & \gapchip{green!23}{$+1.7$\,{\scriptsize$\pm$1.7}} & \textbf{\boldmath $+3.5$} & $+2.7$ & \gapchip{green!7}{$+0.5$\,{\scriptsize$\pm$1.0}} & $-0.2$ & \textbf{\boldmath $+1.2$} & \gapchip{green!28}{$+2.0$\,{\scriptsize$\pm$1.0}} \\
 & 8 & \textbf{\boldmath $+6.2$} & $+4.4$ & \gapchip{red!25}{$-1.8$\,{\scriptsize$\pm$0.7}} & $+0.0$ & \textbf{\boldmath $+0.9$} & \gapchip{green!12}{$+0.9$\,{\scriptsize$\pm$0.5}} & $+0.0$ & \textbf{\boldmath $+2.2$} & \gapchip{green!28}{$+2.2$\,{\scriptsize$\pm$1.1}} & $-9.2$ & $-9.2$ & \gapchip{none}{$+0.0$\,{\scriptsize$\pm$1.4}} & $+3.4$ & \textbf{\boldmath $+4.4$} & \gapchip{green!14}{$+1.0$\,{\scriptsize$\pm$0.2}} & $-0.3$ & \textbf{\boldmath $+0.6$} & \gapchip{green!13}{$+0.9$\,{\scriptsize$\pm$0.7}} \\
 & 16 & $+5.2$ & \textbf{\boldmath $+6.3$} & \gapchip{green!16}{$+1.1$\,{\scriptsize$\pm$1.2}} & $+0.7$ & \textbf{\boldmath $+2.1$} & \gapchip{green!20}{$+1.4$\,{\scriptsize$\pm$1.3}} & \textbf{\boldmath $+1.1$} & $+0.0$ & \gapchip{red!16}{$-1.1$\,{\scriptsize$\pm$1.1}} & \textbf{\boldmath $-7.5$} & $-8.3$ & \gapchip{red!12}{$-0.8$\,{\scriptsize$\pm$5.1}} & \textbf{\boldmath $+5.0$} & $+3.4$ & \gapchip{red!22}{$-1.6$\,{\scriptsize$\pm$0.5}} & $+0.0$ & \textbf{\boldmath $+0.3$} & \gapchip{green!5}{$+0.3$\,{\scriptsize$\pm$1.0}} \\
\midrule
\multirow{6}{*}{\rotatebox[origin=c]{90}{\scriptsize\begin{tabular}{@{}c@{}}\raisebox{-0.15em}{\includegraphics[height=0.9em]{logos/qwen.png}}~Qwen2.5\\ Math-7B\end{tabular}}} & base & \multicolumn{3}{c}{68.8} & \multicolumn{3}{c}{64.2} & \multicolumn{3}{c}{10.0} & \multicolumn{3}{c}{37.5} & \multicolumn{3}{c}{21.7} & \multicolumn{3}{c}{26.4} \\
 & 1 & \textbf{\boldmath $+4.5$} & $+1.9$ & \gapchip{green!28}{$+3.0$\,{\scriptsize$\pm$6.7}} & $+2.6$ & \textbf{\boldmath $+3.1$} & \gapchip{green!28}{$+2.9$\,{\scriptsize$\pm$0.8}} & \textbf{\boldmath $+0.8$} & $-1.1$ & \gapchip{red!16}{$-1.1$\,{\scriptsize$\pm$1.1}} & \textbf{\boldmath $+7.5$} & $+2.5$ & \gapchip{red!28}{$-2.5$\,{\scriptsize$\pm$7.6}} & \textbf{\boldmath $+5.0$} & $+4.4$ & \gapchip{green!14}{$+1.0$\,{\scriptsize$\pm$1.4}} & $+3.6$ & \textbf{\boldmath $+5.1$} & \gapchip{green!28}{$+4.7$\,{\scriptsize$\pm$2.3}} \\
 & 2 & $+6.9$ & \textbf{\boldmath $+12.3$} & \gapchip{green!28}{$+3.3$\,{\scriptsize$\pm$3.0}} & $+0.9$ & \textbf{\boldmath $+3.9$} & \gapchip{green!28}{$+3.1$\,{\scriptsize$\pm$2.7}} & \textbf{\boldmath $+1.3$} & $+1.1$ & \gapchip{none}{$+0.0$\,{\scriptsize$\pm$1.9}} & $+9.7$ & \textbf{\boldmath $+10.8$} & \gapchip{green!28}{$+2.5$\,{\scriptsize$\pm$4.3}} & $+4.9$ & \textbf{\boldmath $+6.6$} & \gapchip{green!28}{$+2.6$\,{\scriptsize$\pm$3.7}} & $+4.7$ & \textbf{\boldmath $+7.0$} & \gapchip{green!25}{$+1.8$\,{\scriptsize$\pm$2.2}} \\
 & 4 & $+11.2$ & \textbf{\boldmath $+11.9$} & \gapchip{green!28}{$+3.9$\,{\scriptsize$\pm$5.6}} & $+3.4$ & \textbf{\boldmath $+5.7$} & \gapchip{green!28}{$+2.3$\,{\scriptsize$\pm$1.1}} & \textbf{\boldmath $+0.8$} & $+0.0$ & \gapchip{green!28}{$+2.2$\,{\scriptsize$\pm$2.9}} & $+7.5$ & \textbf{\boldmath $+10.0$} & \gapchip{green!28}{$+3.3$\,{\scriptsize$\pm$10.4}} & \textbf{\boldmath $+9.4$} & $+7.6$ & \gapchip{red!28}{$-2.2$\,{\scriptsize$\pm$0.8}} & \textbf{\boldmath $+6.8$} & $+6.7$ & \gapchip{red!6}{$-0.4$\,{\scriptsize$\pm$1.9}} \\
 & 8 & $+11.1$ & \textbf{\boldmath $+13.7$} & \gapchip{green!28}{$+2.6$\,{\scriptsize$\pm$4.9}} & $+0.2$ & \textbf{\boldmath $+5.3$} & \gapchip{green!28}{$+5.1$\,{\scriptsize$\pm$1.5}} & $+6.7$ & $+6.7$ & \gapchip{none}{$+0.0$\,{\scriptsize$\pm$1.9}} & $+11.7$ & \textbf{\boldmath $+13.3$} & \gapchip{green!23}{$+1.7$\,{\scriptsize$\pm$3.0}} & \textbf{\boldmath $+10.8$} & $+10.3$ & \gapchip{red!7}{$-0.5$\,{\scriptsize$\pm$4.3}} & $+5.5$ & \textbf{\boldmath $+7.1$} & \gapchip{green!21}{$+1.5$\,{\scriptsize$\pm$0.7}} \\
 & 16 & \textbf{\boldmath $+12.7$} & $+8.2$ & \gapchip{red!28}{$-4.4$\,{\scriptsize$\pm$5.0}} & $+3.7$ & \textbf{\boldmath $+4.9$} & \gapchip{green!18}{$+1.3$\,{\scriptsize$\pm$0.4}} & $+0.0$ & \textbf{\boldmath $+6.7$} & \gapchip{green!28}{$+6.7$\,{\scriptsize$\pm$0.0}} & $+4.2$ & \textbf{\boldmath $+7.5$} & \gapchip{green!28}{$+3.3$\,{\scriptsize$\pm$2.2}} & \textbf{\boldmath $+9.1$} & $+5.5$ & \gapchip{red!28}{$-3.6$\,{\scriptsize$\pm$2.4}} & $+6.2$ & \textbf{\boldmath $+6.8$} & \gapchip{green!8}{$+0.6$\,{\scriptsize$\pm$0.1}} \\
\midrule[\heavyrulewidth]
\multirow{6}{*}{\rotatebox[origin=c]{90}{\scriptsize\begin{tabular}{@{}c@{}}\raisebox{-0.15em}{\includegraphics[height=0.9em]{logos/qwen.png}}~Qwen2.5\\ 1.5B\end{tabular}}} & base & \multicolumn{3}{c}{56.3} & \multicolumn{3}{c}{28.4} & \multicolumn{3}{c}{0.0} & \multicolumn{3}{c}{12.5} & \multicolumn{3}{c}{8.5} & \multicolumn{3}{c}{6.5} \\
 & 1 & $+0.0$ & \textbf{\boldmath $+0.8$} & \gapchip{green!11}{$+0.8$\,{\scriptsize$\pm$1.7}} & $+0.2$ & \textbf{\boldmath $+0.9$} & \gapchip{green!9}{$+0.7$\,{\scriptsize$\pm$0.3}} & $+0.0$ & $+0.0$ & \gapchip{none}{$+0.0$\,{\scriptsize$\pm$0.0}} & \textbf{\boldmath $-2.2$} & $-2.5$ & \gapchip{none}{$+0.0$\,{\scriptsize$\pm$2.5}} & \textbf{\boldmath $+0.1$} & $+0.0$ & \gapchip{red!3}{$-0.2$\,{\scriptsize$\pm$0.9}} & $+0.0$ & \textbf{\boldmath $+1.0$} & \gapchip{green!6}{$+0.4$\,{\scriptsize$\pm$0.5}} \\
 & 2 & $-0.4$ & \textbf{\boldmath $+0.1$} & \gapchip{green!13}{$+0.9$\,{\scriptsize$\pm$0.2}} & \textbf{\boldmath $+0.4$} & $+0.1$ & \gapchip{green!11}{$+0.8$\,{\scriptsize$\pm$0.6}} & $+0.0$ & $+0.0$ & \gapchip{none}{$+0.0$\,{\scriptsize$\pm$0.0}} & $-5.6$ & \textbf{\boldmath $-2.5$} & \gapchip{green!28}{$+2.5$\,{\scriptsize$\pm$1.4}} & \textbf{\boldmath $+0.5$} & $-1.5$ & \gapchip{red!19}{$-1.3$\,{\scriptsize$\pm$1.0}} & $+0.0$ & \textbf{\boldmath $+0.3$} & \gapchip{green!6}{$+0.4$\,{\scriptsize$\pm$0.8}} \\
 & 4 & $-0.5$ & \textbf{\boldmath $+0.7$} & \gapchip{green!28}{$+3.6$\,{\scriptsize$\pm$3.8}} & $-0.9$ & \textbf{\boldmath $+0.2$} & \gapchip{red!1}{$-0.1$\,{\scriptsize$\pm$1.2}} & \textbf{\boldmath $+0.4$} & $+0.0$ & \gapchip{none}{$+0.0$\,{\scriptsize$\pm$0.0}} & \textbf{\boldmath $-4.7$} & $-10.8$ & \gapchip{red!28}{$-5.8$\,{\scriptsize$\pm$3.0}} & $-0.5$ & \textbf{\boldmath $-0.4$} & \gapchip{red!5}{$-0.4$\,{\scriptsize$\pm$0.4}} & $-1.0$ & \textbf{\boldmath $-0.5$} & \gapchip{red!1}{$+0.0$\,{\scriptsize$\pm$0.5}} \\
 & 8 & $-1.1$ & \textbf{\boldmath $+2.8$} & \gapchip{green!28}{$+3.9$\,{\scriptsize$\pm$1.4}} & $+0.0$ & \textbf{\boldmath $+0.4$} & \gapchip{green!6}{$+0.4$\,{\scriptsize$\pm$1.0}} & $+0.0$ & $+0.0$ & \gapchip{none}{$+0.0$\,{\scriptsize$\pm$0.0}} & \textbf{\boldmath $-3.3$} & $-4.2$ & \gapchip{red!12}{$-0.8$\,{\scriptsize$\pm$3.0}} & $-0.2$ & \textbf{\boldmath $+0.7$} & \gapchip{green!14}{$+1.0$\,{\scriptsize$\pm$0.7}} & $-0.7$ & \textbf{\boldmath $-0.2$} & \gapchip{green!7}{$+0.5$\,{\scriptsize$\pm$1.0}} \\
 & 16 & $+1.8$ & $+1.8$ & \gapchip{red!1}{$-0.1$\,{\scriptsize$\pm$1.2}} & $+0.7$ & $+0.7$ & \gapchip{none}{$+0.0$\,{\scriptsize$\pm$1.3}} & $+0.0$ & $+0.0$ & \gapchip{none}{$+0.0$\,{\scriptsize$\pm$0.0}} & \textbf{\boldmath $-0.8$} & $-7.5$ & \gapchip{red!28}{$-6.7$\,{\scriptsize$\pm$1.7}} & $+0.2$ & \textbf{\boldmath $+0.4$} & \gapchip{green!2}{$+0.1$\,{\scriptsize$\pm$1.0}} & \textbf{\boldmath $-0.2$} & $-0.7$ & \gapchip{red!8}{$-0.5$\,{\scriptsize$\pm$0.7}} \\
\midrule
\multirow{6}{*}{\rotatebox[origin=c]{90}{\scriptsize\begin{tabular}{@{}c@{}}\raisebox{-0.15em}{\includegraphics[height=0.9em]{logos/qwen.png}}~Qwen2.5\\ Math-1.5B\end{tabular}}} & base & \multicolumn{3}{c}{41.5} & \multicolumn{3}{c}{47.2} & \multicolumn{3}{c}{3.3} & \multicolumn{3}{c}{32.5} & \multicolumn{3}{c}{8.5} & \multicolumn{3}{c}{23.4} \\
 & 1 & \textbf{\boldmath $+9.2$} & $+4.1$ & \gapchip{green!28}{$+7.8$\,{\scriptsize$\pm$5.4}} & \textbf{\boldmath $+3.5$} & $-3.1$ & \gapchip{red!23}{$-1.7$\,{\scriptsize$\pm$1.6}} & \textbf{\boldmath $+2.9$} & $+2.2$ & \gapchip{red!16}{$-1.1$\,{\scriptsize$\pm$1.1}} & \textbf{\boldmath $-2.5$} & $-6.7$ & \gapchip{green!23}{$+1.7$\,{\scriptsize$\pm$2.2}} & \textbf{\boldmath $+4.0$} & $+1.8$ & \gapchip{green!7}{$+0.5$\,{\scriptsize$\pm$1.0}} & $+1.5$ & $+1.5$ & \gapchip{green!19}{$+1.3$\,{\scriptsize$\pm$1.4}} \\
 & 2 & $+0.0$ & \textbf{\boldmath $+27.3$} & \gapchip{green!28}{$+29.4$\,{\scriptsize$\pm$15.6}} & $-3.8$ & \textbf{\boldmath $+11.4$} & \gapchip{green!28}{$+16.3$\,{\scriptsize$\pm$11.7}} & $+0.8$ & \textbf{\boldmath $+3.3$} & \gapchip{green!28}{$+3.3$\,{\scriptsize$\pm$1.9}} & $-7.8$ & \textbf{\boldmath $-0.8$} & \gapchip{green!28}{$+14.2$\,{\scriptsize$\pm$7.4}} & $+2.0$ & \textbf{\boldmath $+11.8$} & \gapchip{green!28}{$+10.4$\,{\scriptsize$\pm$3.9}} & $-1.3$ & \textbf{\boldmath $+3.9$} & \gapchip{green!28}{$+6.4$\,{\scriptsize$\pm$4.4}} \\
 & 4 & $+9.5$ & \textbf{\boldmath $+28.3$} & \gapchip{green!28}{$+8.9$\,{\scriptsize$\pm$5.2}} & $-2.8$ & \textbf{\boldmath $+7.0$} & \gapchip{green!28}{$+6.7$\,{\scriptsize$\pm$1.8}} & \textbf{\boldmath $+1.7$} & $+0.0$ & \gapchip{red!28}{$-3.3$\,{\scriptsize$\pm$1.9}} & $-4.4$ & \textbf{\boldmath $-1.7$} & \gapchip{green!28}{$+2.5$\,{\scriptsize$\pm$3.8}} & $+3.4$ & \textbf{\boldmath $+6.6$} & \gapchip{green!28}{$+3.3$\,{\scriptsize$\pm$1.4}} & $-0.6$ & \textbf{\boldmath $+0.6$} & \gapchip{none}{$+0.0$\,{\scriptsize$\pm$0.5}} \\
 & 8 & $-2.9$ & \textbf{\boldmath $+33.5$} & \gapchip{green!28}{$+36.4$\,{\scriptsize$\pm$13.7}} & $-3.1$ & \textbf{\boldmath $+13.2$} & \gapchip{green!28}{$+16.3$\,{\scriptsize$\pm$7.6}} & \textbf{\boldmath $+0.0$} & $-1.1$ & \gapchip{red!16}{$-1.1$\,{\scriptsize$\pm$1.1}} & $-6.7$ & \textbf{\boldmath $+4.2$} & \gapchip{green!28}{$+10.8$\,{\scriptsize$\pm$3.6}} & $+2.3$ & \textbf{\boldmath $+12.1$} & \gapchip{green!28}{$+9.8$\,{\scriptsize$\pm$5.5}} & $-2.4$ & \textbf{\boldmath $+2.0$} & \gapchip{green!28}{$+4.4$\,{\scriptsize$\pm$2.9}} \\
 & 16 & $+29.3$ & \textbf{\boldmath $+34.8$} & \gapchip{green!28}{$+5.5$\,{\scriptsize$\pm$3.4}} & $+12.6$ & \textbf{\boldmath $+12.9$} & \gapchip{green!4}{$+0.3$\,{\scriptsize$\pm$3.0}} & \textbf{\boldmath $+5.6$} & $+0.0$ & \gapchip{red!28}{$-5.6$\,{\scriptsize$\pm$2.2}} & \textbf{\boldmath $+7.5$} & $+1.7$ & \gapchip{red!28}{$-5.8$\,{\scriptsize$\pm$5.5}} & $+11.2$ & \textbf{\boldmath $+14.5$} & \gapchip{green!28}{$+3.3$\,{\scriptsize$\pm$4.0}} & \textbf{\boldmath $+3.5$} & $+2.6$ & \gapchip{red!12}{$-0.9$\,{\scriptsize$\pm$2.0}} \\
\bottomrule
\end{tabular}
}
\end{table}

\FloatBarrier
\section{Sample efficiency, compute and memory}
\label{app:efficiency}
This appendix details the efficiency comparison of Section~\ref{sec:efficiency}: completions, compute, and memory.

\subsection{Sample efficiency}
\label{sec:budget}

\begin{figure}[!htbp]
\centering
\includegraphics[width=\linewidth]{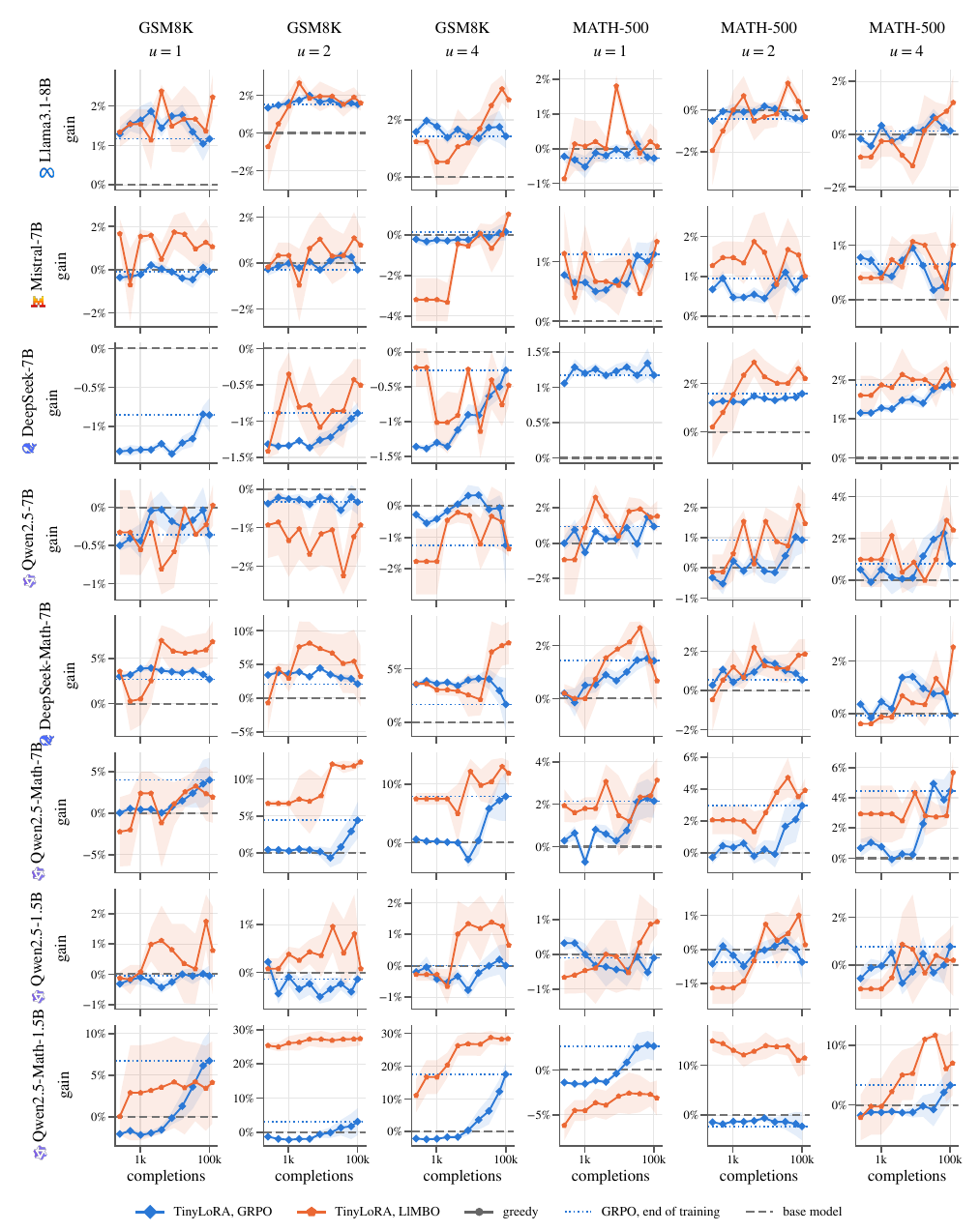}
\caption{Gain of the current adapter over the base model, in greedy accuracy against training completions generated so far, on the two in-distribution tasks at $u \in \{1, 2, 4\}$ (columns), for every model (rows) trained on the SimpleRL-Zoo tiers, mean $\pm$ s.e. \llmbo{} ($b = 1$) is trained on greedy accuracy (Section~\ref{sec:greedy-objective}); its adapter at each point is its evaluated trial of highest posterior mean; its completions are counted from the end of its initial design. Dotted: GRPO at the end of training; gray dashed: the base model.}
\label{fig:budget}
\end{figure}

We measure effort in generated training completions, independent of hardware: a median GRPO run generates \grpoCompletions{}, a median \llmbo{} run \boCompletions{} after its initial design (restart designs included).
Figure~\ref{fig:budget} tracks the greedy gain over the base model along the way, against where GRPO ends.
On the math-specialized models \llmbo{} gets there almost immediately: on Qwen2.5-Math-1.5B at $u = 2$ it passes GRPO's final accuracy with its first GP-guided batch, \reachGreedyMathSmallUTwoGsm{} completions, on both tasks, and ends \budgetEndGreedyMathSmallUTwoGsm{} (GSM8K) and \budgetEndGreedyMathSmallUTwoMathFive{} (MATH-500) points above it; on Qwen2.5-Math-7B at $u = 4$ it does the same on GSM8K and ends \budgetEndGreedyMathUFourGsm{} above. GRPO reaches its final accuracy only at the end of its run. Elsewhere the curves mostly stay within a couple of points of GRPO's final accuracy throughout. Trained on MATH alone (Figure~\ref{fig:budget-mathtrain}), \llmbo{} on Qwen2.5-Math-1.5B is ahead of GRPO's final accuracy from its first batch, \reachGreedyMathSmallUOneGsmMathTrain{} completions, in all six columns, and ends \budgetEndGreedyMathSmallUFourGsmMathTrain{} to \budgetEndGreedyMathSmallUTwoGsmMathTrain{} points above it on GSM8K and \budgetEndGreedyMathSmallUOneMathFiveMathTrain{} to \budgetEndGreedyMathSmallUTwoMathFiveMathTrain{} on MATH-500.
\subsection{One completion per prompt}
\label{sec:one-per-prompt}

\begin{table}[!htbp]
\centering
\caption{Correlation between completions of the same prompt, per model, from GRPO's first \designSteps{} steps at $u \le 4$, close to the base model, at $T = 1$ on the training tier. \emph{Equal-reward groups}: groups of \grpoGroup{} completions of one prompt that all get the same reward, as observed and as expected if the completions were independent at the same pass rate; such groups give GRPO no gradient. $\rho$: intraclass correlation of the beta-binomial matching pass rate and observed equal-reward groups; $1 + 3\rho$: variance inflation of a pass-rate estimate from groups of \grpoGroup{} over one completion per prompt; $k_\mathrm{eff} = 4 / (1 + 3\rho)$: independent completions a group is worth.}
\label{tab:design}
\begin{tabular}{l c cc c c c}
\toprule
 & & \multicolumn{2}{c}{equal-reward groups (\%)} & & & \\
\cmidrule(lr){3-4}
Model & pass rate (\%) & observed & independent & $\rho$ & $1 + 3\rho$ & $k_\mathrm{eff}$ \\
\midrule
\raisebox{-0.15em}{\includegraphics[height=0.9em]{logos/llama.png}}~Llama3.1-8B & 3 & 90 & 88 & 0.09 & 1.27 & 3.2 \\
\raisebox{-0.15em}{\includegraphics[height=0.9em]{logos/mistral.png}}~Mistral-7B & 2 & 93 & 92 & 0.06 & 1.19 & 3.4 \\
\raisebox{-0.15em}{\includegraphics[height=0.9em]{logos/deepseek.png}}~DeepSeek-7B & 1 & 97 & 97 & 0.03 & 1.09 & 3.7 \\
\raisebox{-0.15em}{\includegraphics[height=0.9em]{logos/qwen.png}}~Qwen2.5-7B & 42 & 43 & 14 & 0.35 & 2.04 & 2.0 \\
\midrule
\raisebox{-0.15em}{\includegraphics[height=0.9em]{logos/qwen.png}}~Qwen2.5-Math-7B & 46 & 35 & 13 & 0.27 & 1.81 & 2.2 \\
\raisebox{-0.15em}{\includegraphics[height=0.9em]{logos/deepseek.png}}~DeepSeek-Math-7B & 6 & 83 & 79 & 0.14 & 1.43 & 2.8 \\
\midrule
\raisebox{-0.15em}{\includegraphics[height=0.9em]{logos/qwen.png}}~Qwen2.5-1.5B & 31 & 36 & 23 & 0.17 & 1.51 & 2.6 \\
\raisebox{-0.15em}{\includegraphics[height=0.9em]{logos/qwen.png}}~Qwen2.5-Math-1.5B & 43 & 22 & 14 & 0.10 & 1.31 & 3.1 \\
\midrule
\raisebox{-0.15em}{\includegraphics[height=0.9em]{logos/qwen.png}}~Qwen2.5-7B-Inst. & 66 & 73 & 20 & 0.67 & 3.00 & 1.3 \\
\raisebox{-0.15em}{\includegraphics[height=0.9em]{logos/qwen.png}}~Qwen2.5-1.5B-Inst. & 77 & 59 & 35 & 0.37 & 2.10 & 1.9 \\
\bottomrule
\end{tabular}

\end{table}

\llmbo{} scores a candidate on one completion per prompt; GRPO samples a group of \grpoGroup{} per prompt for its baseline. For a fixed budget of $N$ completions, one per prompt gives the more precise estimate of the pass rate $p$. With $k$ completions on each of $N/k$ prompts, the estimate has variance
\[
\mathrm{Var}(\hat p) = \frac{p(1 - p)}{N}\,\bigl(1 + (k - 1)\rho\bigr), \qquad \rho = \frac{\mathrm{Var}_q(p_q)}{p(1 - p)},
\]
where $p_q$ is the pass rate on prompt $q$ and $\rho$ the correlation between two completions of the same prompt: the more prompts differ in difficulty, the less a second completion of the same prompt adds. The variance is smallest at $k = 1$, the binomial variance $p(1 - p)/N$ of independent completions; $k$ completions per prompt are worth $k / (1 + (k - 1)\rho)$ independent ones. One completion per prompt costs nothing extra here, since the training tiers hold thousands of prompts and \llmbo{} draws fresh ones for every candidate; repeating a prompt saves only its prefill, at most 512 of the 1536 tokens of a sequence.

GRPO's logs measure $\rho$ for its groups of \grpoGroup{} (Table~\ref{tab:design}). Completions of the same prompt agree much more often than independent ones would, and most strongly on the models TinyLoRA moves: on the Qwen2.5 base and Math models a group of \grpoGroup{} is worth only \designEffectiveQwen{} independent completions, and on the Instruct models the variance inflation reaches \designEffectMax{}. On Llama3.1-8B, Mistral-7B and the DeepSeek models the pass rate on the training tier is low and the correlation weak, but there \designEqualOthers{}\% of GRPO's groups get the same reward for all \grpoGroup{} completions, so their advantages vanish and they give no gradient, at the full cost of generating and back-propagating them. \llmbo{} uses every completion: each one is on a fresh prompt and counts toward the pass rate.

\subsection{Compute}
\label{sec:compute}
GRPO's \grpoPasses{} forward passes per token (Section~\ref{sec:efficiency}) are generation, rescoring for the old log-probabilities, and the forward and backward pass for the gradient, the backward pass counted as one since with frozen weights it only propagates activation gradients. The \boComputeShare{}\% counts \llmbo{}'s full run of \boCompletions{} completions against GRPO's \grpoCompletions{}.

\subsection{Memory}
\label{sec:cost}

\begin{table}[!htbp]
\centering
\caption{Memory per architecture, pooling its models (DeepSeek-7B: the DeepSeek-Math-7B runs, same architecture); weights in bf16. \emph{Layers $\times$ KV heads}: the attention shape that sets the KV cache (head dimension 128 throughout); \emph{KV/seq.}: KV cache of one full-length sequence, 512 prompt + 1024 completion tokens. \emph{Beyond vLLM}: the working set $M$ a method allocates beyond the vLLM engine's reserve, median over runs (for GRPO, one micro-batch of \grpoMicroBatch{} sequences). \emph{Sequences in flight}: estimated full-length sequences a method keeps in flight on each card (\cardSmallGiB{}, \cardFortyGiB{}, \cardLargeGiB{} and \cardEightyGiB{}\,GiB usable). Bold: the better method; \ding{55}: fewer than \grpoMicroBatch{} sequences fit. Derivation and a worked example in the text.}
\label{tab:memory}
\footnotesize\setlength{\tabcolsep}{3pt}
\resizebox{\textwidth}{!}{\begin{tabular}{l c cc cc cc cc cc cc}
\toprule
 & & & & \multicolumn{2}{c}{beyond vLLM} & \multicolumn{8}{c}{sequences in flight} \\
\cmidrule(lr){5-6}\cmidrule(lr){7-14}
 & layers $\times$ & weights & KV/seq. & \multicolumn{2}{c}{(GiB)} & \multicolumn{2}{c}{\begin{tabular}{@{}c@{}}L4\\24 GB\end{tabular}} & \multicolumn{2}{c}{\begin{tabular}{@{}c@{}}A100\\40 GB\end{tabular}} & \multicolumn{2}{c}{\begin{tabular}{@{}c@{}}L40S\\48 GB\end{tabular}} & \multicolumn{2}{c}{\begin{tabular}{@{}c@{}}H100\\80 GB\end{tabular}} \\
\cmidrule(lr){7-8}\cmidrule(lr){9-10}\cmidrule(lr){11-12}\cmidrule(lr){13-14}
Architecture & KV heads & (GiB) & (MiB) & GRPO & LlMBO & GRPO & LlMBO & GRPO & LlMBO & GRPO & LlMBO & GRPO & LlMBO \\
\midrule
\raisebox{-0.15em}{\includegraphics[height=0.9em]{logos/qwen.png}}~Qwen2.5-1.5B & 28\,$\times$\,2 & 2.9 & 42 & 1.4 & \textbf{\boldmath 0.0} & 45 & \textbf{\boldmath 418} & 90 & \textbf{\boldmath 841} & 103 & \textbf{\boldmath 963} & 195 & \textbf{\boldmath 1812} \\
\raisebox{-0.15em}{\includegraphics[height=0.9em]{logos/qwen.png}}~Qwen2.5-7B & 28\,$\times$\,4 & 14.2 & 84 & 8.9 & \textbf{\boldmath 0.0} & \ding{55} & \textbf{\boldmath 71} & 10 & \textbf{\boldmath 282} & 12 & \textbf{\boldmath 343} & 27 & \textbf{\boldmath 767} \\
\raisebox{-0.15em}{\includegraphics[height=0.9em]{logos/llama.png}}~Llama3.1-8B & 32\,$\times$\,8 & 15.0 & 192 & 8.0 & \textbf{\boldmath 0.0} & \ding{55} & \textbf{\boldmath 27} & 10 & \textbf{\boldmath 119} & 12 & \textbf{\boldmath 146} & 28 & \textbf{\boldmath 331} \\
\raisebox{-0.15em}{\includegraphics[height=0.9em]{logos/mistral.png}}~Mistral-7B & 32\,$\times$\,8 & 13.5 & 192 & 10.4 & \textbf{\boldmath 0.0} & \ding{55} & \textbf{\boldmath 34} & 8 & \textbf{\boldmath 127} & 10 & \textbf{\boldmath 154} & 22 & \textbf{\boldmath 339} \\
\raisebox{-0.15em}{\includegraphics[height=0.9em]{logos/deepseek.png}}~DeepSeek-7B & 30\,$\times$\,32 & 12.9 & 720 & 1.6 & \textbf{\boldmath 0.0} & 6 & \textbf{\boldmath 10} & 22 & \textbf{\boldmath 34} & 26 & \textbf{\boldmath 41} & 58 & \textbf{\boldmath 91} \\
\bottomrule
\end{tabular}
}
\end{table}

Both methods hold one copy of the weights, shared with the colocated inference engine, and need KV cache for the completions they generate. GRPO additionally holds logits, activations, gradients and optimizer state for its backward pass, which it runs on \grpoMicroBatch{} sequences at a time: \grpoOverheadSmall{}\,GiB on the 1.5B models and \grpoOverheadLarge{}\,GiB on the 7-8B ones (Table~\ref{tab:memory}). \llmbo{} allocates nothing measurable beyond inference; its GP lives on the CPU.

The working-set columns of Table~\ref{tab:memory} derive from each run's peak allocation. Raw peaks are not comparable across methods: the colocated vLLM engine reserves a fixed share $s$ of a card of capacity $C$ up front, $s = 0.85$ for \llmbo{}, which has no training state, and $s = 0.5$ for GRPO ($0.45$ on 80\,GB cards), so \llmbo{}'s peak is the larger by construction. We therefore report the working set $M = \text{peak} - sC$, what a method allocates beyond the engine's reserve, which for a given model is nearly constant across card types. For \llmbo{}, $M$ comes out slightly negative, since the peak allocation misses the engine's CUDA graphs, and we report 0; the same omission makes GRPO's $M$ a slight underestimate, so the comparison favors GRPO.

GRPO's $M$ is that of its micro-batch: it passes a step's completions forward and backward \grpoMicroBatch{} at a time (one prompt's group), accumulating gradients over the step's 64 prompts, so $M$ is the working set of \grpoMicroBatch{} sequences, the same on every card. The sequence columns are an estimate of how many sequences each method could keep in flight, assuming the working set grows linearly with the number of sequences, $M/\grpoMicroBatch{}$ per sequence (activations and logits are per sequence; the adapter's optimizer state is negligible). A card then holds
\[
n = \left\lfloor \frac{C - W - E}{L\,k + M/\grpoMicroBatch{}} \right\rfloor
\]
full-length sequences, with $W$ the weights, $E \approx \engineGiB{}$\,GiB the engine's activation and CUDA-graph overhead, $L = 1536$ tokens and $k$ the KV cache per token, $2 \times \text{layers} \times \text{KV heads} \times \text{head dim} \times 2$ bytes; for \llmbo{}, $M = 0$ and only the KV cache counts. For Qwen2.5-7B on a 24\,GB card ($C = \cardSmallGiB{}$\,GiB usable), $C - W - E = \memExampleFree{}$\,GiB and one sequence's KV cache takes $L\,k = \memExampleSeqMiB{}$\,MiB, so \llmbo{} holds \memExampleSeqs{} sequences; each GRPO sequence also needs $\grpoOverheadQwen{}/\grpoMicroBatch{} \approx \memExampleGrpoPerSeq{}$\,GiB of working set, so GRPO holds \memExampleGrpoSeqs{}, fewer than its micro-batch; we count fewer than \grpoMicroBatch{} sequences as not fitting. These counts assume the engine is given all the memory the method leaves free, and vLLM's paged cache fits more sequences when completions end early.

Beyond the 24\,GB case of Section~\ref{sec:efficiency}, on a 48\,GB card \llmbo{} keeps \boSequenceRatio{} times as many sequences in flight as GRPO, that is, larger generation batches. DeepSeek-7B is the low end of both ranges: its GRPO working set is only \grpoOverheadDeepseek{}\,GiB, while its multi-head attention needs four times Llama's KV cache per token, so few sequences fit under either method.

\FloatBarrier
\section{Batch size and parallelism}
\label{sec:parallel}
\begin{figure}[!htbp]
\centering
\includegraphics[width=\linewidth]{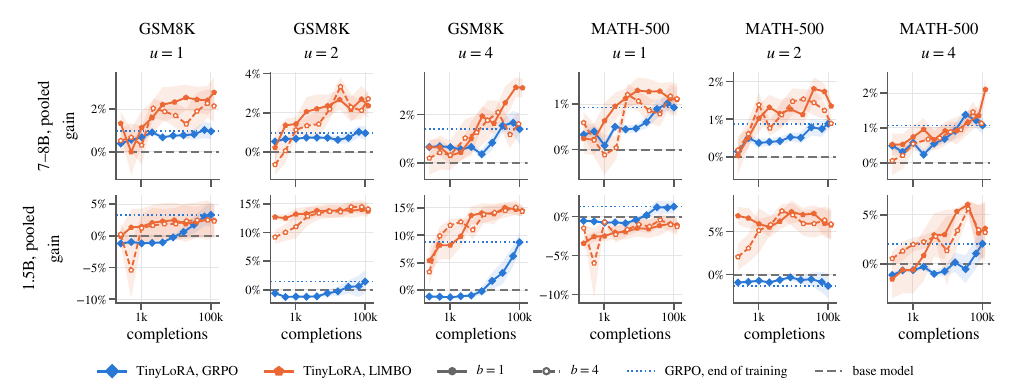}
\caption{Gain of the current adapter over the base model, in greedy accuracy against training completions generated so far, on the two in-distribution tasks at $u \in \{1, 2, 4\}$ (columns), pooled over the four 7-8B families and over the two Qwen2.5 1.5B models (rows; each model counted once), trained on the SimpleRL-Zoo tiers, mean $\pm$ s.e.: GRPO and \llmbo{} trained on greedy accuracy (Section~\ref{sec:greedy-objective}) with $b = 1$ (solid, filled) and $b = 4$ (dashed, open) candidates per batch. \llmbo{}'s adapter at each point is its evaluated trial of highest posterior mean, its completions counted from the end of its initial design. Dotted: GRPO at the end of training; gray dashed: the base model.}
\label{fig:budget-batch}
\end{figure}

A candidate is $u$ numbers and its score a pass count, so the completions that score it can be generated on any number of devices with no synchronization beyond collecting the counts. The $b = 4$ ablation asks whether the same completions go further when spread over more candidates, four adapters on 64 prompts each instead of one on 256, which also makes the candidates of a batch independent jobs. At $u \le 4$ it costs at most about a point on either primary task (Table~\ref{tab:ablations}; at $u = 2$ \llmbo{} trained on avg@4 \batchFourToneMeanUTwoGsm{} on GSM8K avg@4 and \batchFourToneMeanUTwoMathFive{} on MATH-500 avg@4, \llmbo{} trained on greedy accuracy \batchFourGreedyGreedyUTwoGsm{} and \batchFourGreedyGreedyUTwoMathFive{} on greedy accuracy), and for \llmbo{} trained on avg@4 more from $u = 8$, where at $b = 4$ it trails GRPO by \gapBFourToneMeanQwenUEightGsm{} points on Qwen2.5-7B's GSM8K avg@4 (Figure~\ref{fig:gap-main}): the GP gains more from one precise score than from four noisy ones. The training-set evaluations show the noise directly: at $u \le 4$ the in-search score of \llmbo{}'s final pick exceeds the pick's accuracy on the whole training set by \searchBiasBFourTone{} points at $b = 4$ and \searchBiasTone{} at $b = 1$, over the \searchBiasModelsTone{} models evaluated at both (trained on greedy accuracy: \searchBiasBFourGreedy{} and \searchBiasGreedy{} over \searchBiasModelsGreedy{}), so a pick scored on 64 prompts is more often one that drew an easy minibatch. Nor do the extra candidates buy completion efficiency: along the way the $b = 4$ and $b = 1$ curves mostly cross back and forth within their errors, with no consistent edge for $b = 4$, and on the 7-8B models at $u = 4$ $b = 1$ ends ahead on both tasks (Figure~\ref{fig:budget-batch}). One candidate per batch is thus the completion-efficient choice, and $b = 4$ remains an option on hardware that favors independent jobs.

\FloatBarrier
\section{Training data: SimpleRL-Zoo tiers against MATH alone}
\label{app:shift}
The SimpleRL-Zoo tiers do not match the distribution of MATH-500: the hard tier, on which the 7-8B models train, holds MATH levels 3-5 only, and the easy tier, on which the 1.5B models train, GSM8K and MATH level 1, while MATH-500 tests all five levels. The full MATH training set has the level and subject mix of MATH-500. Retraining the four Qwen2.5 base and Math models on it removes the shift and tests whether the gains of either optimizer, and \llmbo{}'s advantage over GRPO, depend on it. Figure~\ref{fig:shift} overlays the two training sets; Table~\ref{tab:shift} gives the MATH-only cells (the tier cells are in Appendices~\ref{app:results} and~\ref{app:small}). The cells on MATH alone carry as many seeds as their tier counterparts, 8 for GRPO and for \llmbo{} trained on avg@4 at $u \le 4$ (3 at $u = 8, 16$) and 3 for \llmbo{} trained on greedy accuracy, each \llmbo{} seed paired with the GRPO seed that shares its subspace.

\subsection{avg@4}
\label{app:shift-mean}

\begin{figure}[!htbp]
\centering
\includegraphics[width=\linewidth]{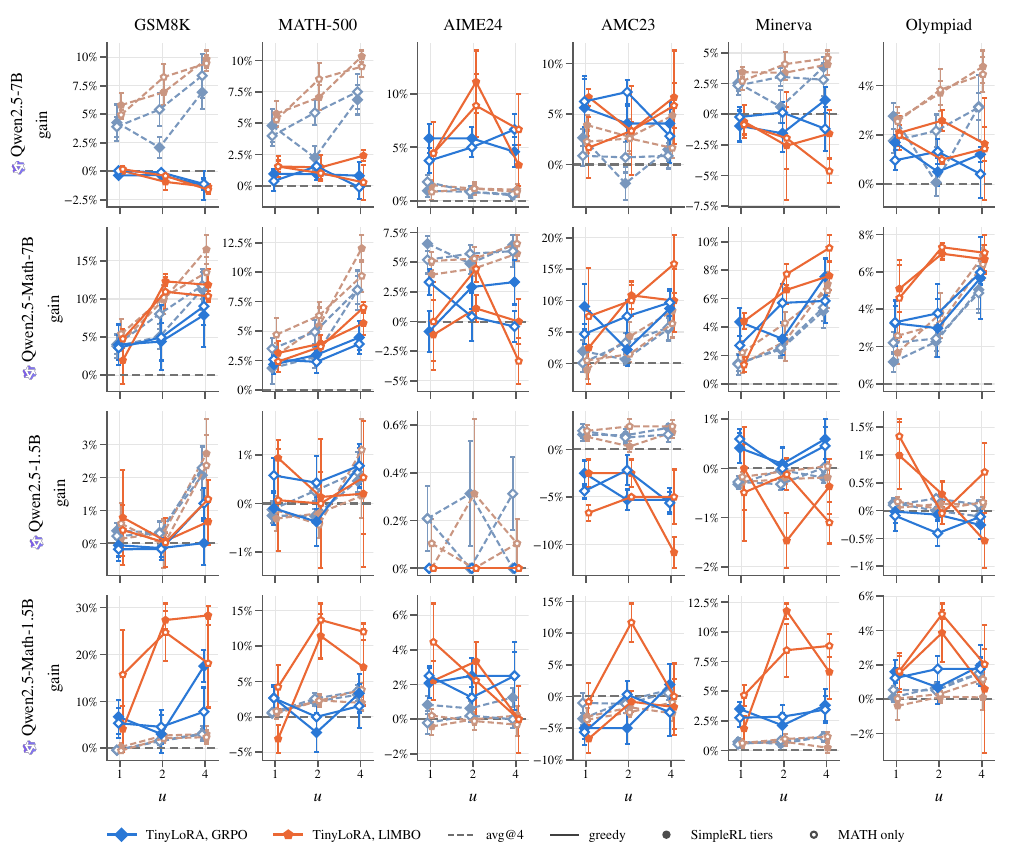}
\caption{Gain over the base model in avg@4 (dashed) and in greedy accuracy (solid) against $u \le 4$, training on the SimpleRL-Zoo tiers (filled; hard tier at 7B, easy tier at 1.5B) and on MATH alone (open), the four Qwen2.5 base and Math models (rows), one column per benchmark (the two in-distribution tasks, then the four OOD benchmarks), each panel on its own scale; GRPO and \llmbo{} trained on the metric of each curve, mean $\pm$ s.e.\ over seeds.}
\label{fig:shift}
\end{figure}

\begin{table}[!htbp]
\centering
\caption{Avg@4 (a) and greedy (b) accuracy at the end of training on MATH alone, for the four Qwen2.5 base and Math models, 7B then 1.5B (row blocks), per $u \le 16$ (rows) and benchmark (columns: in-distribution GSM8K and MATH-500, then the four out-of-distribution benchmarks); \llmbo{} is trained on the metric of each subtable (greedy: Section~\ref{sec:greedy-objective}). \emph{base}: accuracy of the base model in that metric (\%). \emph{GRPO}, \emph{\llmbo{}}: gain over the base model (points), mean over seeds (GRPO and avg@4-trained \llmbo{} 8 per cell at $u \le 4$ and 3 at $u = 8, 16$, greedy-trained \llmbo{} 3 per cell); bold: the larger, unless tied. $\Delta$: paired gap \llmbo{} $-$ GRPO, mean $\pm$ s.e.\ over the seed pairs both have, so it can differ from the difference of the two gains; tint: its sign and size (green: \llmbo{} ahead).}
\label{tab:shift}
{\small (a) avg@4}\\[0.2em]
\resizebox{\linewidth}{!}{\begin{tabular}{c l cccccc@{\hspace{1.2em}}cccccccccccc}
\toprule
 & & \multicolumn{6}{c}{in-distribution} & \multicolumn{12}{c}{out-of-distribution} \\
\cmidrule(lr){3-8}\cmidrule(lr){9-20}
 & & \multicolumn{3}{c}{GSM8K} & \multicolumn{3}{c}{MATH-500} & \multicolumn{3}{c}{AIME24} & \multicolumn{3}{c}{AMC23} & \multicolumn{3}{c}{Minerva} & \multicolumn{3}{c}{Olympiad} \\
\cmidrule(lr){3-5}\cmidrule(lr){6-8}\cmidrule(lr){9-11}\cmidrule(lr){12-14}\cmidrule(lr){15-17}\cmidrule(lr){18-20}
 & $u$ & GRPO & LlMBO & $\Delta$ & GRPO & LlMBO & $\Delta$ & GRPO & LlMBO & $\Delta$ & GRPO & LlMBO & $\Delta$ & GRPO & LlMBO & $\Delta$ & GRPO & LlMBO & $\Delta$ \\
\midrule
\multirow{6}{*}{\rotatebox[origin=c]{90}{\scriptsize\begin{tabular}{@{}c@{}}\raisebox{-0.15em}{\includegraphics[height=0.9em]{logos/qwen.png}}~Qwen2.5\\ 7B\end{tabular}}} & base & \multicolumn{3}{c}{64.9} & \multicolumn{3}{c}{43.5} & \multicolumn{3}{c}{2.5} & \multicolumn{3}{c}{24.4} & \multicolumn{3}{c}{16.4} & \multicolumn{3}{c}{16.7} \\
 & 1 & $+3.9$ & \textbf{\boldmath $+4.9$} & \gapchip{green!15}{$+1.0$\,{\scriptsize$\pm$0.6}} & $+4.0$ & \textbf{\boldmath $+5.3$} & \gapchip{green!18}{$+1.3$\,{\scriptsize$\pm$0.7}} & \textbf{\boldmath $+1.8$} & $+0.8$ & \gapchip{red!13}{$-0.9$\,{\scriptsize$\pm$0.4}} & $+0.9$ & \textbf{\boldmath $+2.3$} & \gapchip{green!20}{$+1.4$\,{\scriptsize$\pm$0.9}} & $+2.3$ & \textbf{\boldmath $+2.7$} & \gapchip{green!5}{$+0.4$\,{\scriptsize$\pm$0.8}} & $+1.7$ & \textbf{\boldmath $+2.5$} & \gapchip{green!11}{$+0.8$\,{\scriptsize$\pm$0.6}} \\
 & 2 & $+5.4$ & \textbf{\boldmath $+8.3$} & \gapchip{green!28}{$+2.8$\,{\scriptsize$\pm$1.4}} & $+5.8$ & \textbf{\boldmath $+8.5$} & \gapchip{green!28}{$+2.7$\,{\scriptsize$\pm$1.2}} & $+0.8$ & \textbf{\boldmath $+1.1$} & \gapchip{green!4}{$+0.3$\,{\scriptsize$\pm$0.8}} & $+0.7$ & \textbf{\boldmath $+2.6$} & \gapchip{green!26}{$+1.9$\,{\scriptsize$\pm$1.4}} & $+3.1$ & \textbf{\boldmath $+4.1$} & \gapchip{green!14}{$+1.0$\,{\scriptsize$\pm$1.1}} & $+2.2$ & \textbf{\boldmath $+3.8$} & \gapchip{green!23}{$+1.7$\,{\scriptsize$\pm$0.8}} \\
 & 4 & $+8.4$ & \textbf{\boldmath $+9.5$} & \gapchip{green!15}{$+1.1$\,{\scriptsize$\pm$2.2}} & $+7.5$ & \textbf{\boldmath $+9.5$} & \gapchip{green!28}{$+2.0$\,{\scriptsize$\pm$1.7}} & $+0.6$ & \textbf{\boldmath $+0.8$} & \gapchip{green!3}{$+0.2$\,{\scriptsize$\pm$0.4}} & $+0.9$ & \textbf{\boldmath $+1.6$} & \gapchip{green!11}{$+0.8$\,{\scriptsize$\pm$1.4}} & $+2.8$ & \textbf{\boldmath $+4.6$} & \gapchip{green!25}{$+1.8$\,{\scriptsize$\pm$1.3}} & $+3.1$ & \textbf{\boldmath $+4.4$} & \gapchip{green!19}{$+1.3$\,{\scriptsize$\pm$1.5}} \\
 & 8 & \textbf{\boldmath $+9.6$} & $+8.7$ & \gapchip{red!13}{$-0.9$\,{\scriptsize$\pm$3.2}} & $+9.1$ & \textbf{\boldmath $+9.8$} & \gapchip{green!9}{$+0.7$\,{\scriptsize$\pm$1.8}} & \textbf{\boldmath $+0.6$} & $+0.3$ & \gapchip{red!4}{$-0.3$\,{\scriptsize$\pm$1.9}} & \textbf{\boldmath $+5.2$} & $+2.7$ & \gapchip{red!28}{$-2.5$\,{\scriptsize$\pm$5.6}} & $+5.0$ & \textbf{\boldmath $+5.8$} & \gapchip{green!12}{$+0.9$\,{\scriptsize$\pm$1.0}} & $+3.9$ & \textbf{\boldmath $+4.3$} & \gapchip{green!5}{$+0.4$\,{\scriptsize$\pm$1.7}} \\
 & 16 & \textbf{\boldmath $+14.8$} & $+12.1$ & \gapchip{red!28}{$-2.7$\,{\scriptsize$\pm$1.5}} & \textbf{\boldmath $+14.8$} & $+12.5$ & \gapchip{red!28}{$-2.4$\,{\scriptsize$\pm$1.0}} & \textbf{\boldmath $+2.5$} & $+1.7$ & \gapchip{red!12}{$-0.8$\,{\scriptsize$\pm$1.7}} & \textbf{\boldmath $+11.2$} & $+7.7$ & \gapchip{red!28}{$-3.5$\,{\scriptsize$\pm$2.7}} & \textbf{\boldmath $+10.1$} & $+5.8$ & \gapchip{red!28}{$-4.4$\,{\scriptsize$\pm$1.0}} & \textbf{\boldmath $+8.1$} & $+7.6$ & \gapchip{red!7}{$-0.5$\,{\scriptsize$\pm$1.5}} \\
\midrule
\multirow{6}{*}{\rotatebox[origin=c]{90}{\scriptsize\begin{tabular}{@{}c@{}}\raisebox{-0.15em}{\includegraphics[height=0.9em]{logos/qwen.png}}~Qwen2.5\\ Math-7B\end{tabular}}} & base & \multicolumn{3}{c}{46.4} & \multicolumn{3}{c}{44.9} & \multicolumn{3}{c}{4.2} & \multicolumn{3}{c}{30.6} & \multicolumn{3}{c}{17.9} & \multicolumn{3}{c}{19.9} \\
 & 1 & $+4.0$ & \textbf{\boldmath $+5.4$} & \gapchip{green!19}{$+1.4$\,{\scriptsize$\pm$1.3}} & $+3.5$ & \textbf{\boldmath $+4.7$} & \gapchip{green!17}{$+1.2$\,{\scriptsize$\pm$0.9}} & \textbf{\boldmath $+5.2$} & $+5.1$ & \gapchip{red!1}{$-0.1$\,{\scriptsize$\pm$1.4}} & $+0.1$ & \textbf{\boldmath $+0.5$} & \gapchip{green!5}{$+0.4$\,{\scriptsize$\pm$1.8}} & $+1.4$ & \textbf{\boldmath $+2.2$} & \gapchip{green!11}{$+0.8$\,{\scriptsize$\pm$1.1}} & $+2.2$ & \textbf{\boldmath $+2.6$} & \gapchip{green!5}{$+0.4$\,{\scriptsize$\pm$0.7}} \\
 & 2 & $+8.0$ & \textbf{\boldmath $+10.2$} & \gapchip{green!28}{$+2.2$\,{\scriptsize$\pm$2.3}} & $+4.9$ & \textbf{\boldmath $+6.3$} & \gapchip{green!19}{$+1.4$\,{\scriptsize$\pm$1.3}} & \textbf{\boldmath $+5.7$} & $+5.3$ & \gapchip{red!6}{$-0.4$\,{\scriptsize$\pm$1.2}} & \textbf{\boldmath $+3.3$} & $+1.5$ & \gapchip{red!25}{$-1.8$\,{\scriptsize$\pm$2.2}} & $+2.5$ & \textbf{\boldmath $+4.3$} & \gapchip{green!24}{$+1.7$\,{\scriptsize$\pm$1.0}} & $+2.4$ & \textbf{\boldmath $+3.2$} & \gapchip{green!11}{$+0.8$\,{\scriptsize$\pm$1.1}} \\
 & 4 & $+11.8$ & \textbf{\boldmath $+13.5$} & \gapchip{green!24}{$+1.7$\,{\scriptsize$\pm$1.1}} & $+8.5$ & \textbf{\boldmath $+9.7$} & \gapchip{green!17}{$+1.2$\,{\scriptsize$\pm$0.8}} & $+5.9$ & \textbf{\boldmath $+6.6$} & \gapchip{green!9}{$+0.6$\,{\scriptsize$\pm$1.7}} & $+5.7$ & \textbf{\boldmath $+6.3$} & \gapchip{green!9}{$+0.6$\,{\scriptsize$\pm$2.3}} & $+5.4$ & \textbf{\boldmath $+6.6$} & \gapchip{green!17}{$+1.2$\,{\scriptsize$\pm$0.8}} & $+4.9$ & \textbf{\boldmath $+6.4$} & \gapchip{green!22}{$+1.5$\,{\scriptsize$\pm$0.8}} \\
 & 8 & $+15.1$ & \textbf{\boldmath $+20.9$} & \gapchip{green!28}{$+5.9$\,{\scriptsize$\pm$2.3}} & $+11.7$ & \textbf{\boldmath $+12.8$} & \gapchip{green!15}{$+1.1$\,{\scriptsize$\pm$0.7}} & \textbf{\boldmath $+5.6$} & $+5.0$ & \gapchip{red!8}{$-0.6$\,{\scriptsize$\pm$0.7}} & $+7.3$ & \textbf{\boldmath $+8.1$} & \gapchip{green!12}{$+0.8$\,{\scriptsize$\pm$3.9}} & $+5.2$ & \textbf{\boldmath $+7.8$} & \gapchip{green!28}{$+2.6$\,{\scriptsize$\pm$1.3}} & $+6.8$ & \textbf{\boldmath $+7.5$} & \gapchip{green!10}{$+0.7$\,{\scriptsize$\pm$1.4}} \\
 & 16 & $+19.6$ & \textbf{\boldmath $+24.3$} & \gapchip{green!28}{$+4.7$\,{\scriptsize$\pm$2.6}} & $+14.5$ & \textbf{\boldmath $+15.1$} & \gapchip{green!8}{$+0.6$\,{\scriptsize$\pm$1.1}} & \textbf{\boldmath $+5.3$} & $+2.5$ & \gapchip{red!28}{$-2.8$\,{\scriptsize$\pm$1.5}} & \textbf{\boldmath $+11.0$} & $+10.4$ & \gapchip{red!9}{$-0.6$\,{\scriptsize$\pm$2.5}} & $+8.1$ & \textbf{\boldmath $+9.3$} & \gapchip{green!18}{$+1.3$\,{\scriptsize$\pm$0.9}} & $+8.7$ & $+8.7$ & \gapchip{none}{$+0.0$\,{\scriptsize$\pm$1.2}} \\
\midrule[\heavyrulewidth]
\multirow{6}{*}{\rotatebox[origin=c]{90}{\scriptsize\begin{tabular}{@{}c@{}}\raisebox{-0.15em}{\includegraphics[height=0.9em]{logos/qwen.png}}~Qwen2.5\\ 1.5B\end{tabular}}} & base & \multicolumn{3}{c}{22.6} & \multicolumn{3}{c}{8.9} & \multicolumn{3}{c}{0.0} & \multicolumn{3}{c}{1.9} & \multicolumn{3}{c}{2.3} & \multicolumn{3}{c}{1.6} \\
 & 1 & $+0.2$ & \textbf{\boldmath $+0.6$} & \gapchip{green!5}{$+0.4$\,{\scriptsize$\pm$0.5}} & $-0.1$ & $-0.1$ & \gapchip{none}{$+0.0$\,{\scriptsize$\pm$0.3}} & \textbf{\boldmath $+0.2$} & $+0.1$ & \gapchip{red!1}{$-0.1$\,{\scriptsize$\pm$0.2}} & $+2.0$ & $+2.0$ & \gapchip{none}{$+0.0$\,{\scriptsize$\pm$1.1}} & $-0.3$ & \textbf{\boldmath $-0.2$} & \gapchip{green!1}{$+0.1$\,{\scriptsize$\pm$0.2}} & $+0.1$ & \textbf{\boldmath $+0.2$} & \gapchip{green!1}{$+0.1$\,{\scriptsize$\pm$0.2}} \\
 & 2 & \textbf{\boldmath $+0.3$} & $+0.0$ & \gapchip{red!5}{$-0.4$\,{\scriptsize$\pm$0.5}} & \textbf{\boldmath $+0.3$} & $-0.4$ & \gapchip{red!10}{$-0.7$\,{\scriptsize$\pm$0.3}} & $+0.0$ & $+0.0$ & \gapchip{none}{$+0.0$\,{\scriptsize$\pm$0.0}} & $+1.2$ & \textbf{\boldmath $+2.4$} & \gapchip{green!16}{$+1.2$\,{\scriptsize$\pm$1.2}} & $-0.3$ & \textbf{\boldmath $-0.1$} & \gapchip{green!4}{$+0.3$\,{\scriptsize$\pm$0.2}} & \textbf{\boldmath $+0.2$} & $+0.0$ & \gapchip{red!3}{$-0.2$\,{\scriptsize$\pm$0.1}} \\
 & 4 & $+2.3$ & \textbf{\boldmath $+2.4$} & \gapchip{green!1}{$+0.1$\,{\scriptsize$\pm$0.4}} & $+0.7$ & \textbf{\boldmath $+1.1$} & \gapchip{green!6}{$+0.5$\,{\scriptsize$\pm$0.3}} & \textbf{\boldmath $+0.3$} & $+0.1$ & \gapchip{red!3}{$-0.2$\,{\scriptsize$\pm$0.2}} & $+1.6$ & \textbf{\boldmath $+2.4$} & \gapchip{green!12}{$+0.9$\,{\scriptsize$\pm$1.0}} & $+0.0$ & $+0.0$ & \gapchip{none}{$+0.0$\,{\scriptsize$\pm$0.1}} & $+0.1$ & $+0.1$ & \gapchip{green!1}{$+0.0$\,{\scriptsize$\pm$0.1}} \\
 & 8 & $+2.4$ & \textbf{\boldmath $+2.6$} & \gapchip{green!2}{$+0.1$\,{\scriptsize$\pm$0.8}} & \textbf{\boldmath $+1.0$} & $+0.0$ & \gapchip{red!13}{$-0.9$\,{\scriptsize$\pm$1.3}} & \textbf{\boldmath $+0.3$} & $+0.0$ & \gapchip{red!4}{$-0.3$\,{\scriptsize$\pm$0.3}} & $+0.4$ & \textbf{\boldmath $+1.7$} & \gapchip{green!18}{$+1.2$\,{\scriptsize$\pm$1.6}} & \textbf{\boldmath $-0.3$} & $-0.5$ & \gapchip{red!2}{$-0.2$\,{\scriptsize$\pm$0.3}} & \textbf{\boldmath $+0.3$} & $+0.0$ & \gapchip{red!4}{$-0.3$\,{\scriptsize$\pm$0.1}} \\
 & 16 & \textbf{\boldmath $+4.8$} & $+3.9$ & \gapchip{red!13}{$-0.9$\,{\scriptsize$\pm$0.8}} & $+1.1$ & \textbf{\boldmath $+1.3$} & \gapchip{green!3}{$+0.2$\,{\scriptsize$\pm$0.7}} & $+0.0$ & $+0.0$ & \gapchip{none}{$+0.0$\,{\scriptsize$\pm$0.0}} & $+1.5$ & \textbf{\boldmath $+3.5$} & \gapchip{green!28}{$+2.1$\,{\scriptsize$\pm$0.6}} & \textbf{\boldmath $+0.4$} & $-0.6$ & \gapchip{red!13}{$-0.9$\,{\scriptsize$\pm$0.3}} & $+0.4$ & $+0.4$ & \gapchip{green!1}{$+0.1$\,{\scriptsize$\pm$0.2}} \\
\midrule
\multirow{6}{*}{\rotatebox[origin=c]{90}{\scriptsize\begin{tabular}{@{}c@{}}\raisebox{-0.15em}{\includegraphics[height=0.9em]{logos/qwen.png}}~Qwen2.5\\ Math-1.5B\end{tabular}}} & base & \multicolumn{3}{c}{34.6} & \multicolumn{3}{c}{32.5} & \multicolumn{3}{c}{2.5} & \multicolumn{3}{c}{20.0} & \multicolumn{3}{c}{10.2} & \multicolumn{3}{c}{16.2} \\
 & 1 & \textbf{\boldmath $-0.4$} & $-0.5$ & \gapchip{red!1}{$-0.1$\,{\scriptsize$\pm$0.4}} & $+0.6$ & \textbf{\boldmath $+0.8$} & \gapchip{green!2}{$+0.2$\,{\scriptsize$\pm$0.4}} & $-0.2$ & \textbf{\boldmath $+0.2$} & \gapchip{green!6}{$+0.4$\,{\scriptsize$\pm$1.0}} & \textbf{\boldmath $-1.0$} & $-3.7$ & \gapchip{red!28}{$-2.7$\,{\scriptsize$\pm$1.8}} & $+0.5$ & \textbf{\boldmath $+0.6$} & \gapchip{green!1}{$+0.1$\,{\scriptsize$\pm$0.5}} & \textbf{\boldmath $+0.5$} & $+0.0$ & \gapchip{red!7}{$-0.5$\,{\scriptsize$\pm$0.3}} \\
 & 2 & $+1.5$ & \textbf{\boldmath $+2.1$} & \gapchip{green!9}{$+0.6$\,{\scriptsize$\pm$0.4}} & \textbf{\boldmath $+2.6$} & $+2.3$ & \gapchip{red!4}{$-0.3$\,{\scriptsize$\pm$0.4}} & \textbf{\boldmath $+0.2$} & $-0.1$ & \gapchip{red!4}{$-0.3$\,{\scriptsize$\pm$0.7}} & \textbf{\boldmath $-1.2$} & $-2.5$ & \gapchip{red!19}{$-1.3$\,{\scriptsize$\pm$1.1}} & $+0.8$ & \textbf{\boldmath $+1.0$} & \gapchip{green!2}{$+0.1$\,{\scriptsize$\pm$0.4}} & \textbf{\boldmath $+0.5$} & $+0.3$ & \gapchip{red!3}{$-0.2$\,{\scriptsize$\pm$0.3}} \\
 & 4 & \textbf{\boldmath $+2.9$} & $+2.4$ & \gapchip{red!8}{$-0.6$\,{\scriptsize$\pm$0.5}} & $+3.6$ & \textbf{\boldmath $+3.8$} & \gapchip{green!3}{$+0.2$\,{\scriptsize$\pm$0.3}} & \textbf{\boldmath $+0.0$} & $-0.3$ & \gapchip{red!4}{$-0.3$\,{\scriptsize$\pm$1.0}} & \textbf{\boldmath $+0.5$} & $-0.9$ & \gapchip{red!21}{$-1.5$\,{\scriptsize$\pm$0.6}} & $+1.1$ & \textbf{\boldmath $+1.2$} & \gapchip{green!1}{$+0.0$\,{\scriptsize$\pm$0.3}} & \textbf{\boldmath $+1.5$} & $+1.2$ & \gapchip{red!5}{$-0.3$\,{\scriptsize$\pm$0.3}} \\
 & 8 & $+4.9$ & \textbf{\boldmath $+6.6$} & \gapchip{green!23}{$+1.6$\,{\scriptsize$\pm$1.6}} & $+4.6$ & \textbf{\boldmath $+6.8$} & \gapchip{green!28}{$+2.2$\,{\scriptsize$\pm$1.4}} & $+0.0$ & \textbf{\boldmath $+0.8$} & \gapchip{green!12}{$+0.8$\,{\scriptsize$\pm$0.5}} & \textbf{\boldmath $+0.6$} & $+0.2$ & \gapchip{red!6}{$-0.4$\,{\scriptsize$\pm$1.7}} & $+1.3$ & $+1.3$ & \gapchip{none}{$+0.0$\,{\scriptsize$\pm$1.4}} & \textbf{\boldmath $+2.1$} & $+1.8$ & \gapchip{red!4}{$-0.3$\,{\scriptsize$\pm$0.2}} \\
 & 16 & \textbf{\boldmath $+9.0$} & $+8.6$ & \gapchip{red!5}{$-0.4$\,{\scriptsize$\pm$3.5}} & \textbf{\boldmath $+9.0$} & $+8.8$ & \gapchip{red!4}{$-0.3$\,{\scriptsize$\pm$2.3}} & \textbf{\boldmath $+2.2$} & $+1.4$ & \gapchip{red!12}{$-0.8$\,{\scriptsize$\pm$1.0}} & $-0.8$ & \textbf{\boldmath $+2.1$} & \gapchip{green!28}{$+2.9$\,{\scriptsize$\pm$2.2}} & \textbf{\boldmath $+3.4$} & $+2.6$ & \gapchip{red!11}{$-0.8$\,{\scriptsize$\pm$0.9}} & \textbf{\boldmath $+2.8$} & $+2.3$ & \gapchip{red!6}{$-0.5$\,{\scriptsize$\pm$1.6}} \\
\bottomrule
\end{tabular}
}\\[0.8em]
{\small (b) greedy}\\[0.2em]
\resizebox{\linewidth}{!}{\begin{tabular}{c l cccccc@{\hspace{1.2em}}cccccccccccc}
\toprule
 & & \multicolumn{6}{c}{in-distribution} & \multicolumn{12}{c}{out-of-distribution} \\
\cmidrule(lr){3-8}\cmidrule(lr){9-20}
 & & \multicolumn{3}{c}{GSM8K} & \multicolumn{3}{c}{MATH-500} & \multicolumn{3}{c}{AIME24} & \multicolumn{3}{c}{AMC23} & \multicolumn{3}{c}{Minerva} & \multicolumn{3}{c}{Olympiad} \\
\cmidrule(lr){3-5}\cmidrule(lr){6-8}\cmidrule(lr){9-11}\cmidrule(lr){12-14}\cmidrule(lr){15-17}\cmidrule(lr){18-20}
 & $u$ & GRPO & LlMBO & $\Delta$ & GRPO & LlMBO & $\Delta$ & GRPO & LlMBO & $\Delta$ & GRPO & LlMBO & $\Delta$ & GRPO & LlMBO & $\Delta$ & GRPO & LlMBO & $\Delta$ \\
\midrule
\multirow{6}{*}{\rotatebox[origin=c]{90}{\scriptsize\begin{tabular}{@{}c@{}}\raisebox{-0.15em}{\includegraphics[height=0.9em]{logos/qwen.png}}~Qwen2.5\\ 7B\end{tabular}}} & base & \multicolumn{3}{c}{88.0} & \multicolumn{3}{c}{64.0} & \multicolumn{3}{c}{0.0} & \multicolumn{3}{c}{30.0} & \multicolumn{3}{c}{32.7} & \multicolumn{3}{c}{28.6} \\
 & 1 & $+0.0$ & \textbf{\boldmath $+0.2$} & \gapchip{green!4}{$+0.3$\,{\scriptsize$\pm$0.5}} & $+0.4$ & \textbf{\boldmath $+1.5$} & \gapchip{green!28}{$+2.3$\,{\scriptsize$\pm$2.0}} & $+3.8$ & \textbf{\boldmath $+4.4$} & \gapchip{red!28}{$-2.2$\,{\scriptsize$\pm$4.8}} & \textbf{\boldmath $+6.2$} & $+1.7$ & \gapchip{red!12}{$-0.8$\,{\scriptsize$\pm$3.6}} & \textbf{\boldmath $-0.2$} & $-0.9$ & \gapchip{red!15}{$-1.1$\,{\scriptsize$\pm$1.2}} & $+1.0$ & \textbf{\boldmath $+2.0$} & \gapchip{green!28}{$+2.1$\,{\scriptsize$\pm$1.0}} \\
 & 2 & \textbf{\boldmath $-0.1$} & $-0.4$ & \gapchip{red!5}{$-0.4$\,{\scriptsize$\pm$0.3}} & \textbf{\boldmath $+1.6$} & $+1.0$ & \gapchip{red!10}{$-0.7$\,{\scriptsize$\pm$1.6}} & $+5.0$ & \textbf{\boldmath $+8.9$} & \gapchip{green!28}{$+4.4$\,{\scriptsize$\pm$4.0}} & \textbf{\boldmath $+7.2$} & $+3.3$ & \gapchip{red!28}{$-4.2$\,{\scriptsize$\pm$0.8}} & \textbf{\boldmath $+0.1$} & $-2.0$ & \gapchip{red!28}{$-2.2$\,{\scriptsize$\pm$4.1}} & \textbf{\boldmath $+1.3$} & $+1.0$ & \gapchip{green!10}{$+0.7$\,{\scriptsize$\pm$1.1}} \\
 & 4 & \textbf{\boldmath $-1.1$} & $-1.6$ & \gapchip{red!10}{$-0.7$\,{\scriptsize$\pm$0.8}} & $-0.1$ & \textbf{\boldmath $+0.3$} & \gapchip{green!15}{$+1.1$\,{\scriptsize$\pm$1.1}} & $+6.7$ & $+6.7$ & \gapchip{green!16}{$+1.1$\,{\scriptsize$\pm$2.9}} & $+2.8$ & \textbf{\boldmath $+5.8$} & \gapchip{red!28}{$-4.2$\,{\scriptsize$\pm$4.4}} & \textbf{\boldmath $-1.2$} & $-4.7$ & \gapchip{red!28}{$-6.9$\,{\scriptsize$\pm$1.4}} & $+0.4$ & \textbf{\boldmath $+1.4$} & \gapchip{green!12}{$+0.9$\,{\scriptsize$\pm$2.4}} \\
 & 8 & \textbf{\boldmath $+0.3$} & $-0.3$ & \gapchip{red!8}{$-0.6$\,{\scriptsize$\pm$0.7}} & \textbf{\boldmath $+3.2$} & $+0.3$ & \gapchip{red!28}{$-2.9$\,{\scriptsize$\pm$3.3}} & $+4.4$ & \textbf{\boldmath $+7.8$} & \gapchip{green!28}{$+3.3$\,{\scriptsize$\pm$1.9}} & $+5.8$ & \textbf{\boldmath $+9.2$} & \gapchip{green!28}{$+3.3$\,{\scriptsize$\pm$3.0}} & \textbf{\boldmath $+2.6$} & $-2.9$ & \gapchip{red!28}{$-5.5$\,{\scriptsize$\pm$2.4}} & $+1.7$ & \textbf{\boldmath $+2.5$} & \gapchip{green!12}{$+0.8$\,{\scriptsize$\pm$0.8}} \\
 & 16 & \textbf{\boldmath $+0.7$} & $-0.8$ & \gapchip{red!20}{$-1.4$\,{\scriptsize$\pm$0.5}} & \textbf{\boldmath $+4.6$} & $+0.9$ & \gapchip{red!28}{$-3.7$\,{\scriptsize$\pm$1.0}} & \textbf{\boldmath $+8.9$} & $+5.6$ & \gapchip{red!28}{$-3.3$\,{\scriptsize$\pm$1.9}} & \textbf{\boldmath $+10.0$} & $+2.5$ & \gapchip{red!28}{$-7.5$\,{\scriptsize$\pm$2.5}} & \textbf{\boldmath $+4.4$} & $-3.3$ & \gapchip{red!28}{$-7.7$\,{\scriptsize$\pm$1.4}} & \textbf{\boldmath $+4.4$} & $+0.0$ & \gapchip{red!28}{$-4.4$\,{\scriptsize$\pm$0.1}} \\
\midrule
\multirow{6}{*}{\rotatebox[origin=c]{90}{\scriptsize\begin{tabular}{@{}c@{}}\raisebox{-0.15em}{\includegraphics[height=0.9em]{logos/qwen.png}}~Qwen2.5\\ Math-7B\end{tabular}}} & base & \multicolumn{3}{c}{68.8} & \multicolumn{3}{c}{64.2} & \multicolumn{3}{c}{10.0} & \multicolumn{3}{c}{37.5} & \multicolumn{3}{c}{21.7} & \multicolumn{3}{c}{26.4} \\
 & 1 & $+3.6$ & \textbf{\boldmath $+4.8$} & \gapchip{green!22}{$+1.6$\,{\scriptsize$\pm$2.3}} & \textbf{\boldmath $+2.6$} & $+2.4$ & \gapchip{red!3}{$-0.2$\,{\scriptsize$\pm$0.6}} & \textbf{\boldmath $+3.3$} & $+0.0$ & \gapchip{red!28}{$-2.2$\,{\scriptsize$\pm$4.0}} & $+4.7$ & \textbf{\boldmath $+7.5$} & \gapchip{green!23}{$+1.7$\,{\scriptsize$\pm$7.1}} & \textbf{\boldmath $+2.7$} & $+1.3$ & \gapchip{green!2}{$+0.1$\,{\scriptsize$\pm$0.9}} & $+3.3$ & \textbf{\boldmath $+4.6$} & \gapchip{green!26}{$+1.9$\,{\scriptsize$\pm$0.6}} \\
 & 2 & $+5.0$ & \textbf{\boldmath $+10.9$} & \gapchip{green!28}{$+4.0$\,{\scriptsize$\pm$4.5}} & $+2.4$ & \textbf{\boldmath $+3.7$} & \gapchip{red!4}{$-0.3$\,{\scriptsize$\pm$1.3}} & $+0.4$ & \textbf{\boldmath $+4.4$} & \gapchip{green!28}{$+5.6$\,{\scriptsize$\pm$2.2}} & $+7.5$ & \textbf{\boldmath $+10.0$} & \gapchip{green!12}{$+0.8$\,{\scriptsize$\pm$2.2}} & $+5.7$ & \textbf{\boldmath $+7.7$} & \gapchip{green!24}{$+1.7$\,{\scriptsize$\pm$3.3}} & $+3.8$ & \textbf{\boldmath $+7.3$} & \gapchip{green!28}{$+3.2$\,{\scriptsize$\pm$0.6}} \\
 & 4 & $+9.1$ & \textbf{\boldmath $+10.4$} & \gapchip{green!28}{$+5.0$\,{\scriptsize$\pm$5.4}} & $+3.9$ & \textbf{\boldmath $+7.0$} & \gapchip{green!28}{$+4.5$\,{\scriptsize$\pm$0.9}} & \textbf{\boldmath $-0.4$} & $-3.3$ & \gapchip{red!28}{$-3.3$\,{\scriptsize$\pm$1.9}} & $+9.7$ & \textbf{\boldmath $+15.8$} & \gapchip{green!28}{$+10.8$\,{\scriptsize$\pm$3.3}} & $+5.8$ & \textbf{\boldmath $+9.6$} & \gapchip{green!28}{$+5.8$\,{\scriptsize$\pm$4.4}} & $+6.0$ & \textbf{\boldmath $+7.0$} & \gapchip{green!28}{$+2.2$\,{\scriptsize$\pm$0.7}} \\
 & 8 & $+9.6$ & $+9.6$ & \gapchip{red!1}{$-0.1$\,{\scriptsize$\pm$1.9}} & $+3.4$ & \textbf{\boldmath $+3.7$} & \gapchip{green!4}{$+0.3$\,{\scriptsize$\pm$1.7}} & \textbf{\boldmath $+6.7$} & $+3.3$ & \gapchip{red!28}{$-3.3$\,{\scriptsize$\pm$3.3}} & \textbf{\boldmath $+11.7$} & $+8.3$ & \gapchip{red!28}{$-3.3$\,{\scriptsize$\pm$2.2}} & $+6.5$ & \textbf{\boldmath $+8.1$} & \gapchip{green!22}{$+1.6$\,{\scriptsize$\pm$3.3}} & $+7.0$ & \textbf{\boldmath $+7.8$} & \gapchip{green!12}{$+0.8$\,{\scriptsize$\pm$2.2}} \\
 & 16 & \textbf{\boldmath $+14.2$} & $+7.1$ & \gapchip{red!28}{$-7.1$\,{\scriptsize$\pm$3.8}} & $+4.1$ & \textbf{\boldmath $+5.1$} & \gapchip{green!13}{$+0.9$\,{\scriptsize$\pm$0.8}} & \textbf{\boldmath $+3.3$} & $-1.1$ & \gapchip{red!28}{$-4.4$\,{\scriptsize$\pm$5.6}} & $+10.0$ & \textbf{\boldmath $+10.8$} & \gapchip{green!12}{$+0.8$\,{\scriptsize$\pm$3.3}} & \textbf{\boldmath $+9.7$} & $+3.3$ & \gapchip{red!28}{$-6.4$\,{\scriptsize$\pm$2.5}} & \textbf{\boldmath $+8.3$} & $+5.9$ & \gapchip{red!28}{$-2.4$\,{\scriptsize$\pm$1.6}} \\
\midrule[\heavyrulewidth]
\multirow{6}{*}{\rotatebox[origin=c]{90}{\scriptsize\begin{tabular}{@{}c@{}}\raisebox{-0.15em}{\includegraphics[height=0.9em]{logos/qwen.png}}~Qwen2.5\\ 1.5B\end{tabular}}} & base & \multicolumn{3}{c}{56.3} & \multicolumn{3}{c}{28.4} & \multicolumn{3}{c}{0.0} & \multicolumn{3}{c}{12.5} & \multicolumn{3}{c}{8.5} & \multicolumn{3}{c}{6.5} \\
 & 1 & $-0.2$ & \textbf{\boldmath $+0.4$} & \gapchip{green!16}{$+1.1$\,{\scriptsize$\pm$0.7}} & \textbf{\boldmath $+0.6$} & $+0.1$ & \gapchip{red!12}{$-0.9$\,{\scriptsize$\pm$1.3}} & $+0.0$ & $+0.0$ & \gapchip{none}{$+0.0$\,{\scriptsize$\pm$0.0}} & \textbf{\boldmath $-4.4$} & $-6.7$ & \gapchip{red!23}{$-1.7$\,{\scriptsize$\pm$0.8}} & \textbf{\boldmath $+0.6$} & $-0.5$ & \gapchip{red!12}{$-0.9$\,{\scriptsize$\pm$1.0}} & $-0.1$ & \textbf{\boldmath $+1.3$} & \gapchip{green!12}{$+0.9$\,{\scriptsize$\pm$0.7}} \\
 & 2 & $-0.2$ & \textbf{\boldmath $+0.0$} & \gapchip{green!11}{$+0.8$\,{\scriptsize$\pm$1.1}} & \textbf{\boldmath $+0.4$} & $+0.0$ & \gapchip{green!7}{$+0.5$\,{\scriptsize$\pm$1.0}} & $+0.0$ & $+0.0$ & \gapchip{none}{$+0.0$\,{\scriptsize$\pm$0.0}} & \textbf{\boldmath $-2.2$} & $-5.0$ & \gapchip{red!28}{$-5.8$\,{\scriptsize$\pm$1.7}} & \textbf{\boldmath $+0.0$} & $-0.1$ & \gapchip{green!5}{$+0.4$\,{\scriptsize$\pm$0.6}} & $-0.4$ & \textbf{\boldmath $+0.0$} & \gapchip{green!1}{$+0.0$\,{\scriptsize$\pm$0.3}} \\
 & 4 & $+1.2$ & \textbf{\boldmath $+1.3$} & \gapchip{red!3}{$-0.2$\,{\scriptsize$\pm$1.2}} & \textbf{\boldmath $+0.8$} & $+0.5$ & \gapchip{green!1}{$+0.1$\,{\scriptsize$\pm$1.2}} & $+0.0$ & $+0.0$ & \gapchip{none}{$+0.0$\,{\scriptsize$\pm$0.0}} & $-5.6$ & \textbf{\boldmath $-5.0$} & \gapchip{green!23}{$+1.7$\,{\scriptsize$\pm$4.6}} & \textbf{\boldmath $+0.5$} & $-1.1$ & \gapchip{red!28}{$-2.2$\,{\scriptsize$\pm$0.4}} & $-0.1$ & \textbf{\boldmath $+0.7$} & \gapchip{green!14}{$+1.0$\,{\scriptsize$\pm$0.8}} \\
 & 8 & \textbf{\boldmath $+0.6$} & $-1.1$ & \gapchip{red!24}{$-1.7$\,{\scriptsize$\pm$0.7}} & $+0.4$ & \textbf{\boldmath $+1.3$} & \gapchip{green!13}{$+0.9$\,{\scriptsize$\pm$1.3}} & $+0.0$ & $+0.0$ & \gapchip{none}{$+0.0$\,{\scriptsize$\pm$0.0}} & $-4.2$ & $-4.2$ & \gapchip{none}{$+0.0$\,{\scriptsize$\pm$1.4}} & $+0.5$ & \textbf{\boldmath $+0.7$} & \gapchip{green!3}{$+0.2$\,{\scriptsize$\pm$1.6}} & \textbf{\boldmath $+0.0$} & $-0.2$ & \gapchip{red!3}{$-0.2$\,{\scriptsize$\pm$0.3}} \\
 & 16 & $+1.3$ & \textbf{\boldmath $+1.7$} & \gapchip{green!5}{$+0.3$\,{\scriptsize$\pm$1.0}} & $+2.1$ & \textbf{\boldmath $+2.3$} & \gapchip{green!3}{$+0.2$\,{\scriptsize$\pm$0.6}} & $+0.0$ & $+0.0$ & \gapchip{none}{$+0.0$\,{\scriptsize$\pm$0.0}} & \textbf{\boldmath $-5.8$} & $-6.7$ & \gapchip{red!12}{$-0.8$\,{\scriptsize$\pm$3.3}} & $-0.1$ & \textbf{\boldmath $+0.5$} & \gapchip{green!9}{$+0.6$\,{\scriptsize$\pm$0.9}} & $+0.5$ & \textbf{\boldmath $+0.8$} & \gapchip{green!4}{$+0.3$\,{\scriptsize$\pm$0.4}} \\
\midrule
\multirow{6}{*}{\rotatebox[origin=c]{90}{\scriptsize\begin{tabular}{@{}c@{}}\raisebox{-0.15em}{\includegraphics[height=0.9em]{logos/qwen.png}}~Qwen2.5\\ Math-1.5B\end{tabular}}} & base & \multicolumn{3}{c}{41.5} & \multicolumn{3}{c}{47.2} & \multicolumn{3}{c}{3.3} & \multicolumn{3}{c}{32.5} & \multicolumn{3}{c}{8.5} & \multicolumn{3}{c}{23.4} \\
 & 1 & $+5.3$ & \textbf{\boldmath $+15.7$} & \gapchip{green!28}{$+16.7$\,{\scriptsize$\pm$7.8}} & $+2.6$ & \textbf{\boldmath $+4.2$} & \gapchip{green!28}{$+4.9$\,{\scriptsize$\pm$1.8}} & $+2.5$ & \textbf{\boldmath $+4.4$} & \gapchip{green!16}{$+1.1$\,{\scriptsize$\pm$2.2}} & $-5.6$ & \textbf{\boldmath $-0.8$} & \gapchip{green!28}{$+7.5$\,{\scriptsize$\pm$2.5}} & $+2.8$ & \textbf{\boldmath $+4.7$} & \gapchip{green!28}{$+3.6$\,{\scriptsize$\pm$0.3}} & $+1.2$ & \textbf{\boldmath $+1.6$} & \gapchip{green!8}{$+0.6$\,{\scriptsize$\pm$0.7}} \\
 & 2 & $+4.6$ & \textbf{\boldmath $+24.7$} & \gapchip{green!28}{$+23.2$\,{\scriptsize$\pm$7.8}} & $+0.0$ & \textbf{\boldmath $+13.7$} & \gapchip{green!28}{$+14.1$\,{\scriptsize$\pm$4.6}} & $+1.2$ & \textbf{\boldmath $+2.2$} & \gapchip{green!16}{$+1.1$\,{\scriptsize$\pm$2.9}} & $+0.3$ & \textbf{\boldmath $+11.7$} & \gapchip{green!28}{$+11.7$\,{\scriptsize$\pm$4.6}} & $+2.8$ & \textbf{\boldmath $+8.5$} & \gapchip{green!28}{$+6.6$\,{\scriptsize$\pm$2.2}} & $+1.8$ & \textbf{\boldmath $+4.9$} & \gapchip{green!28}{$+3.7$\,{\scriptsize$\pm$2.1}} \\
 & 4 & $+7.8$ & \textbf{\boldmath $+18.1$} & \gapchip{green!28}{$+13.2$\,{\scriptsize$\pm$13.0}} & $+1.5$ & \textbf{\boldmath $+12.0$} & \gapchip{green!28}{$+14.1$\,{\scriptsize$\pm$6.9}} & \textbf{\boldmath $+2.5$} & $+0.0$ & \gapchip{none}{$+0.0$\,{\scriptsize$\pm$1.9}} & $-2.5$ & \textbf{\boldmath $+0.0$} & \gapchip{green!28}{$+8.3$\,{\scriptsize$\pm$12.0}} & $+3.4$ & \textbf{\boldmath $+8.8$} & \gapchip{green!28}{$+6.7$\,{\scriptsize$\pm$1.2}} & $+1.8$ & \textbf{\boldmath $+2.0$} & \gapchip{green!24}{$+1.7$\,{\scriptsize$\pm$1.8}} \\
 & 8 & $+17.7$ & \textbf{\boldmath $+27.3$} & \gapchip{green!28}{$+9.6$\,{\scriptsize$\pm$6.2}} & $+8.5$ & \textbf{\boldmath $+13.5$} & \gapchip{green!28}{$+4.9$\,{\scriptsize$\pm$3.2}} & \textbf{\boldmath $+4.4$} & $+2.2$ & \gapchip{red!28}{$-2.2$\,{\scriptsize$\pm$2.9}} & $+7.5$ & $+7.5$ & \gapchip{none}{$+0.0$\,{\scriptsize$\pm$2.9}} & $+3.8$ & \textbf{\boldmath $+8.7$} & \gapchip{green!28}{$+4.9$\,{\scriptsize$\pm$3.2}} & $+4.1$ & \textbf{\boldmath $+4.5$} & \gapchip{green!5}{$+0.3$\,{\scriptsize$\pm$1.1}} \\
 & 16 & $+17.1$ & \textbf{\boldmath $+27.9$} & \gapchip{green!28}{$+10.8$\,{\scriptsize$\pm$8.9}} & $+7.6$ & \textbf{\boldmath $+15.5$} & \gapchip{green!28}{$+7.9$\,{\scriptsize$\pm$7.0}} & $+0.0$ & \textbf{\boldmath $+5.6$} & \gapchip{green!28}{$+5.6$\,{\scriptsize$\pm$1.1}} & $+5.8$ & \textbf{\boldmath $+15.0$} & \gapchip{green!28}{$+9.2$\,{\scriptsize$\pm$9.6}} & $+7.5$ & \textbf{\boldmath $+11.5$} & \gapchip{green!28}{$+4.0$\,{\scriptsize$\pm$2.0}} & $+3.2$ & \textbf{\boldmath $+4.9$} & \gapchip{green!24}{$+1.7$\,{\scriptsize$\pm$0.9}} \\
\bottomrule
\end{tabular}
}
\end{table}

Removing the shift moves neither optimizer much on MATH-500 (Figure~\ref{fig:shift}): on Qwen2.5-7B at $u = 4$ GRPO gains \gainGrpoMeanQwenUFourMathFive{} points on the tiers and \gainGrpoMeanQwenUFourMathFiveMathTrain{} on MATH alone, \llmbo{} \gainToneMeanQwenUFourMathFive{} and \gainToneMeanQwenUFourMathFiveMathTrain{}; elsewhere the two training sets mostly stay within each other's seed spread.
Pooled over Qwen2.5-7B and Qwen2.5-Math-7B, \llmbo{}'s paired gap on MATH-500 avg@4 stays positive at every $u$ on MATH alone (\poolToneMeanUOneMathFiveShiftLargeMathTrain{}, \poolToneMeanUTwoMathFiveShiftLargeMathTrain{}, \poolToneMeanUFourMathFiveShiftLargeMathTrain{} at $u = 1, 2, 4$), as on the tiers (\poolToneMeanUOneMathFiveShiftLarge{}, \poolToneMeanUTwoMathFiveShiftLarge{}, \poolToneMeanUFourMathFiveShiftLarge{}), and so does its gap on GSM8K avg@4 (\poolToneMeanUOneGsmShiftLargeMathTrain{}, \poolToneMeanUTwoGsmShiftLargeMathTrain{}, \poolToneMeanUFourGsmShiftLargeMathTrain{}, against \poolToneMeanUOneGsmShiftLarge{}, \poolToneMeanUTwoGsmShiftLarge{}, \poolToneMeanUFourGsmShiftLarge{} on the tiers).
Every difference between the two training sets is within two standard errors (at $u = 2, 4$: \shiftDeltaToneMeanUTwoMathFiveLarge{} and \shiftDeltaToneMeanUFourMathFiveLarge{} points on MATH-500, \shiftDeltaToneMeanUTwoGsmLarge{} and \shiftDeltaToneMeanUFourGsmLarge{} on GSM8K); the $b = 4$ ablation shows drops of the same size (\shiftDeltaBFourToneMeanUFourMathFiveLarge{} on MATH-500 and \shiftDeltaBFourToneMeanUFourGsmLarge{} on GSM8K at $u = 4$).
Over Qwen2.5-1.5B and Qwen2.5-Math-1.5B, \llmbo{} is level with GRPO on both training sets: on MATH alone its paired gap on MATH-500 avg@4 is \poolToneMeanUOneMathFiveShiftSmallMathTrain{}, \poolToneMeanUTwoMathFiveShiftSmallMathTrain{} and \poolToneMeanUFourMathFiveShiftSmallMathTrain{} at $u = 1, 2, 4$ and on GSM8K avg@4, which this training set no longer contains, \poolToneMeanUOneGsmShiftSmallMathTrain{}, \poolToneMeanUTwoGsmShiftSmallMathTrain{} and \poolToneMeanUFourGsmShiftSmallMathTrain{}.

\FloatBarrier
\subsection{Greedy accuracy}
\label{app:shift-greedy}

\begin{figure}[!htbp]
\centering
\includegraphics[width=\linewidth]{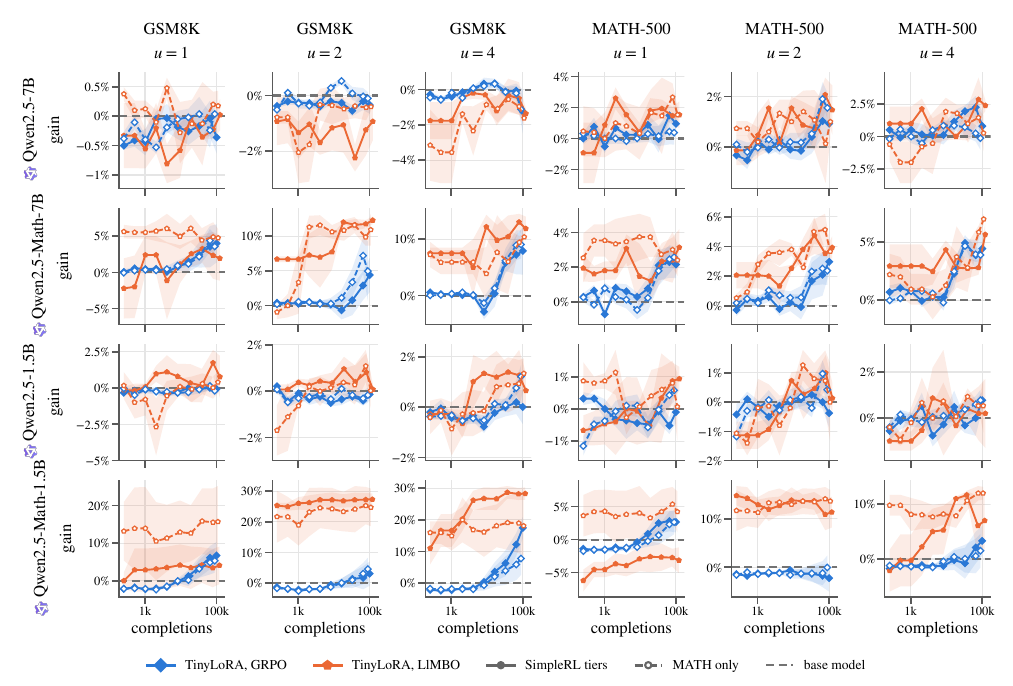}
\caption{Gain of the current adapter over the base model, in greedy accuracy against training completions generated so far, on the two in-distribution tasks at $u \in \{1, 2, 4\}$ (columns), for the four Qwen2.5 base and Math models (rows) trained on the SimpleRL-Zoo tiers (solid, filled) and on MATH alone (dashed, open), mean $\pm$ s.e., for GRPO and for \llmbo{} ($b = 1$) trained on greedy accuracy (Section~\ref{sec:greedy-objective}). \llmbo{}'s adapter at each point is its evaluated trial of highest posterior mean, its completions counted from the end of its initial design. Gray dashed: the base model.}
\label{fig:budget-mathtrain}
\end{figure}

Trained on MATH alone, \llmbo{}'s paired gap on greedy MATH-500, pooled over Qwen2.5-7B and Qwen2.5-Math-7B, is \poolGreedyGreedyUOneMathFiveShiftLargeMathTrain{}, \poolGreedyGreedyUTwoMathFiveShiftLargeMathTrain{} and \poolGreedyGreedyUFourMathFiveShiftLargeMathTrain{} at $u = 1, 2, 4$, against \poolGreedyGreedyUOneMathFiveShiftLarge{}, \poolGreedyGreedyUTwoMathFiveShiftLarge{} and \poolGreedyGreedyUFourMathFiveShiftLarge{} on the tiers, and on greedy GSM8K it stays positive at every $u$ (\poolGreedyGreedyUOneGsmShiftLargeMathTrain{}, \poolGreedyGreedyUTwoGsmShiftLargeMathTrain{}, \poolGreedyGreedyUFourGsmShiftLargeMathTrain{}).
At 1.5B its lead on greedy MATH-500 holds or grows, to \poolGreedyGreedyUOneMathFiveShiftSmallMathTrain{}, \poolGreedyGreedyUTwoMathFiveShiftSmallMathTrain{} and \poolGreedyGreedyUFourMathFiveShiftSmallMathTrain{} points at $u = 1, 2, 4$ over Qwen2.5-1.5B and Qwen2.5-Math-1.5B, from \poolGreedyGreedyUOneMathFiveShiftSmall{}, \poolGreedyGreedyUTwoMathFiveShiftSmall{} and \poolGreedyGreedyUFourMathFiveShiftSmall{} on the tiers, and on greedy GSM8K it leads by \poolGreedyGreedyUOneGsmShiftSmallMathTrain{}, \poolGreedyGreedyUTwoGsmShiftSmallMathTrain{} and \poolGreedyGreedyUFourGsmShiftSmallMathTrain{}. The completion budget of \llmbo{} on MATH alone is in Figure~\ref{fig:budget-mathtrain}.

\FloatBarrier
The level shift therefore does not explain \llmbo{}'s advantage: with the training set matched to MATH-500, both optimizers gain about as much as on the tiers, and \llmbo{}'s lead on the 7B models persists at every $u$, within seed noise of its lead on the tiers.

\FloatBarrier
\section{TinyLoRA under GRPO across the parameter scale}
\label{app:scaling}
This appendix details how we parameterize and scale TinyLoRA, where this departs from \citet{morris2026learning} and why the departure leaves the comparison of \llmbo{} with GRPO unaffected, and how TinyLoRA trained with GRPO compares with LoRA and LoRA-XS from one trainable parameter to tens of millions.

\subsection{Parameterization}
\label{app:param}
We adapt $M$ modules, every attention, and MLP projection (query, key, value, output, gate, up, down) of every layer. We have $M = 196$ on the Qwen2.5 models, $210$ on the DeepSeek models, and $224$ on Llama3.1-8B and Mistral-7B. Module $m$ has a frozen weight $W_m$ with rank-$r$ truncated SVD $U_m \Sigma_m V_m^\top$, computed in fp32 from the bf16 weight, and receives the LoRA-XS update \citep{balazy2024loraxs}
\begin{equation}
    \Delta W_m = U_m \Sigma_m R_m V_m^\top, \qquad R_m \in \mbR^{r \times r}.
\end{equation}
LoRA-XS trains every $R_m$ directly, $M r^2$ parameters in all. TinyLoRA instead writes each $R_m$ as a combination of $u$ fixed random matrices,
\begin{equation}
    R_m = \sum_{i=1}^{u} v_i P^{(m)}_i, \qquad P^{(m)}_i \in \mbR^{r \times r}, \qquad \big[P^{(m)}_i\big]_{jk} \overset{\text{iid}}{\sim} \mcN(0, 1),
    \label{eq:app-tinylora}
\end{equation}
with one coefficient vector $v \in \mbR^u$ shared by all $M$ modules, so the adapter has exactly $u$ trainable parameters and we search $\theta = v$. Throughout, $r = 2$, the rank \citet{morris2026learning} use for their main experiments.

The projections are drawn independently for every module and every $i$, from a generator seeded with the run's seed, as one $u \times r \times r$ tensor per module in network order (layer by layer, the seven projections in the order above). \llmbo{} seed $k$ rebuilds exactly the SVD bases and projections of GRPO seed $k$, so both optimizers search the same subspace in the same coordinates. Different seeds draw independent subspaces; different $u$ draw unrelated ones, since every module's tensor changes shape: the subspace at $u = 2$ does not contain the one at $u = 1$. \citet{morris2026learning} specify neither the distribution of the projections nor whether they are shared across modules; independence across modules is what makes $u > r^2$ useful under full tying (Section~\ref{app:tying}).

Here $v$ starts at zero, so $\Delta W_m = 0$ and $\theta = 0$ is the base model; \citet{morris2026learning} do not state an initialization. Instead of their learning-rate sweep, we fix the learning rate by equal update norm: Adam moves each coordinate of $v$ by about $\eta$ per step, and $\|P^{(m)}_i\|_F \approx r$, so a step changes each $R_m$ by $\|\Delta R_m\|_F \approx \eta r \sqrt{u}$. We set $\eta = 10^{-3} / (r \sqrt{u})$, a fixed $\|\Delta R_m\|_F \approx 10^{-3}$ per step at every $u$, and LoRA-XS, whose coordinates are the entries of $R_m$, gets $\eta = 10^{-3}/r$ by the same rule. Since TinyLoRA and LoRA-XS share the frozen $U_m$, $\Sigma_m$, $V_m$, equal $\|\Delta R_m\|_F$ is equal speed in weight space. The same $\eta$ fixes \llmbo{}'s search box $[-\eta S, \eta S]^u$ (Section~\ref{sec:method}).

LoRA \citep{hu2022lora} trains $r(d_\mathrm{in} + d_\mathrm{out})$ parameters per module, about 2.5M at $r = 1$ and 81M at $r = 32$ on Qwen2.5-7B, with $\alpha = 2r$, the PEFT initialization ($A$ random, $B = 0$) and $\eta = 5 \cdot 10^{-6}$. LoRA-XS trains $M r^2$ parameters, 196 at $r = 1$ to about 200k at $r = 32$ on Qwen2.5-7B, with $R_m = 0$ at the start. All three adapters train under the same GRPO recipe (Section~\ref{sec:experiments}).

\subsection{Weight tying: the original scheme and ours}
\label{app:tying}

The original scheme
\citet{morris2026learning} gives each module, or each group of $n_\mathrm{tie}$ modules, its own vector $v_g \in \mbR^u$, for $M u / n_\mathrm{tie}$ trainable parameters in all. Groups are either \emph{tiled}, nearby modules of similar depth regardless of type, or \emph{structured}, nearby modules of the same type, and tiling works better. A parameter count is reached by trading the projection dimension $u$ against the tying factor $n_\mathrm{tie}$, and their guideline is to lower $u$ (down to 1) before raising $n_\mathrm{tie}$. Scaling TinyLoRA thus spans a two-dimensional grid of configurations, with a grouping rule on top.

Both schemes are linear subspaces of LoRA-XS. Stack the per-module matrices into $\rho = (\mathrm{vec}\, R_1, \dots, \mathrm{vec}\, R_M) \in \mbR^{M r^2}$, the parameter space of LoRA-XS. Every TinyLoRA configuration with $N$ trainable parameters is a fixed linear map
\begin{equation}
    \rho = \Phi \theta, \qquad \Phi \in \mbR^{M r^2 \times N}, \qquad \theta \in \mbR^N.
\end{equation}
With tying in groups, $\theta = (v_1, \dots, v_{M/n_\mathrm{tie}})$ and $\Phi$ is block-sparse: the rows of module $m$ are nonzero only in the $u$ columns of its group, which hold its projections $\mathrm{vec}\, P^{(m)}_i$. With full tying ($n_\mathrm{tie} = M$), $N = u$ and $\Phi$ is dense, every entry an independent standard normal. In either case, as long as $\Phi$ has full column rank, which holds almost surely for $N \le M r^2$ (and $u \le n_\mathrm{tie} r^2$ under grouping), the adapters TinyLoRA can express form an $N$-dimensional linear subspace of the LoRA-XS space, and the two schemes differ only in \emph{which} $N$-dimensional subspace: one aligned with a grouping of the modules, or the column space of a Gaussian matrix, a uniformly random subspace in the spirit of intrinsic-dimension fine-tuning \citep{aghajanyan2021intrinsic}. With enough parameters the distinction vanishes: at $N = M r^2$ both span the whole LoRA-XS space, and untied TinyLoRA with $u = r^2$ is LoRA-XS with a random basis for each $R_m$. Below that, full tying already reaches every count $N \le M r^2$ through $u$ alone, so the second axis, $n_\mathrm{tie}$, and the grouping rule only choose among subspaces of equal dimension. Full tying also needs projections drawn independently per module: with the same $P_i$ in every module, all $R_m$ would coincide and the subspace would stop growing at $u = r^2$.

Our scheme it to fix full tying and scale TinyLoRA along the single axis $u \in \{1, 2, 4, 8, 16, 64, 256\}$, with $N = u$. This is the original scheme's $n_\mathrm{tie} = M$ setting, extended past $u = r^2$ by independent per-module projections, and it stays well inside the regime of distinct directions: $u \le 256 < 784 \le M r^2$ on every model. It is also well conditioned: each column of $\Phi$ has norm about $r\sqrt{M}$, and two columns have cosine of order $1/(r\sqrt{M}) \approx 0.04$ at $M = 196$, so every coordinate of $v$ moves the network by about the same amount along a nearly orthogonal direction, at every $u$.

Equal dimension is not equal performance: the subspace differs, and both optimizers depend on the coordinates, Adam through its per-coordinate steps and \llmbo{} through its search box and per-dimension lengthscales. Our TinyLoRA accuracies at a given parameter count therefore need not match those of the best $(u, n_\mathrm{tie})$ configuration and learning rate of \citet{morris2026learning}, and we do not claim they do. The deviation does not affect our analysis, however: GRPO and \llmbo{} run on the same $\Phi$, seed for seed, so they search the same subspace in the same coordinates and differ only in the optimizer. The choice of tying only fixes a simple, one-axis scaling protocol, in which the parameter count, the search dimension and the subspace dimension are all $u$.

\subsection{Results}
\label{app:scaling-results}

\begin{figure}[!htbp]
\centering
\includegraphics[width=0.9\linewidth]{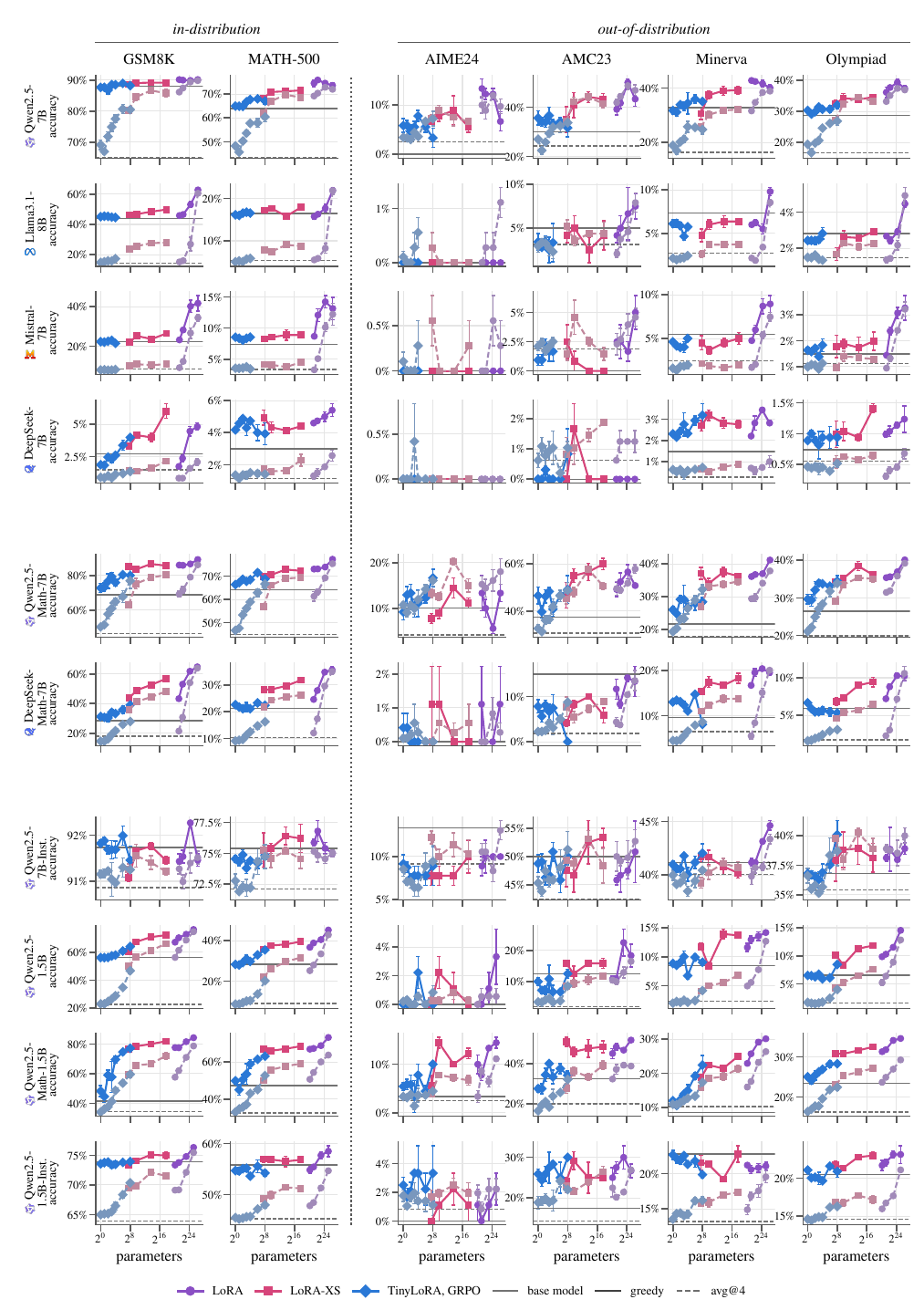}
\caption{Greedy (solid) and avg@4 (dashed) accuracy at the end of training against trainable parameters under GRPO, for every model (rows, in three blocks: the one 7-8B base model per family; the two math-specialized 7B models; Qwen2.5-7B-Instruct and the Qwen2.5 1.5B models, which train on the easy tier), on each benchmark (columns: in-distribution GSM8K and MATH-500, then the four out-of-distribution benchmarks right of the dotted line), each panel on its own accuracy axis: LoRA and LoRA-XS at $r \in \{1,2,8,32\}$, TinyLoRA at $u \in \{1,\dots,256\}$. Gray: the base model, in the same line style. Mean $\pm$ s.e.\ over seeds.}
\label{fig:scaling-base}
\end{figure}

Figure~\ref{fig:scaling-base} places TinyLoRA among the larger adapters, all trained with GRPO, on every model, in avg@4 and in greedy accuracy, on the two in-distribution tasks and the four out-of-distribution benchmarks. The OOD benchmarks follow the in-distribution tasks at smaller magnitude and with more noise, as expected from 30-675 problems each.

TinyLoRA's gains vary across model families. On Qwen2.5-7B, TinyLoRA retains a large part of LoRA's avg@4 gain even with only a few parameters, while greedy accuracy starts higher and has less room to improve. On Llama3.1-8B, Mistral-7B, and DeepSeek-7B, TinyLoRA with $u \le 4$ changes accuracy by at most \othersTinyBest{} points on either primary task. On Llama3.1-8B and Mistral-7B LoRA at rank 32 still gains \loraGainGreedyLlamaMathFive{} and \loraGainGreedyMistralMathFive{} points on greedy MATH-500 and \loraGainGreedyLlamaGsm{} and \loraGainGreedyMistralGsm{} on greedy GSM8K: for these families a useful direction is apparently not in TinyLoRA's random subspace. DeepSeek-7B moves little under LoRA as well (\loraGainGreedyDeepseekBaseMathFive{} and \loraGainGreedyDeepseekBaseGsm{} points).

As for the Math variants TinyLoRA retains a large part of LoRA's avg@4 gain down to a few parameters n Qwen2.5-Math-7B and on Qwen2.5-7B, with GRPO gaining up to \gainGrpoMeanMathUFourMathFive{} points on MATH-500 avg@4 at $u = 4$; on DeepSeek-Math-7B, as on the other non-Qwen models, it moves accuracy little.

For the instruct models like Qwen2.5-7B-Instruct and Qwen2.5-1.5B-Instruct leave little to gain in the ultra-low-parameter regime: GRPO moves their avg@4 on GSM8K and MATH-500 by at most \grpoBestInstruct{} points at $u \le 4$ (Appendix~\ref{app:instruct}).

\FloatBarrier
\section{Implementation details}
\label{app:impl}
This appendix collects the settings of both optimizers, the evaluation harness, the compute used, and the licenses of the assets, then details \llmbo{}'s search, its observation model, how we keep the frozen SVD bases identical across GPU models, and the one change we make to the standard GRPO loss. All are implemented in the anonymized code in the supplementary material.

\subsection{Hyperparameters}
\label{app:hparams}

\begin{table}[!htbp]
\centering
\caption{Settings of both optimizers. $N$: prompts in the training set; $S$: GRPO's optimizer steps; $s$ of $n$: a candidate's solved prompts.}
\label{tab:hparams}
\small
\begin{tabular}{@{}lp{0.66\linewidth}@{}}
\toprule
\multicolumn{2}{@{}l}{\emph{Shared}} \\
Adapter & TinyLoRA, $r = 2$, one $v \in \mbR^u$ tied across all $M$ modules (Appendix~\ref{app:param}) \\
Prompt / completion length & 512 / 1024 tokens \\
Completions per step or batch & 256 \\
Reward & 1 if math-verify finds the final answer equivalent to the reference, else 0 \\
\midrule
\multicolumn{2}{@{}l}{\emph{GRPO}} \\
Prompts $\times$ rollouts per step & $64 \times 4$, at $T = 1$ \\
Steps $S$ & $3 \lceil N / 64 \rceil$ (three epochs) \\
Loss & token-level clipped objective, $\epsilon = 0.2$, dual clip $\delta = 2$, no KL penalty \\
Optimizer & 8-bit AdamW, constant learning rate after 10 warm-up steps, gradient-norm clip 1 \\
Learning rate $\eta$ & $10^{-3}/(r\sqrt{u})$ TinyLoRA, $10^{-3}/r$ LoRA-XS, $5 \cdot 10^{-6}$ LoRA ($\alpha = 2r$) \\
Micro-batch & 4 sequences (one prompt's group), gradients accumulated over the step's 64 prompts \\
\midrule
\multicolumn{2}{@{}l}{\emph{\llmbo{}}} \\
Candidates per batch $b$ & 1 (ablation: 4) \\
Prompts per candidate $n$ & $256 / b$, fresh random prompts, one completion each \\
Objective & pass rate at $T = 1$ (for avg@4) or greedy (for greedy accuracy) \\
Batches & $S$ GP-guided, plus the initial and restart designs \\
Initial / restart design & 8 replicates of $\theta = 0$ and 8 Sobol points / 8 Sobol points \\
Search box & $[-\eta S, \eta S]^u$, with GRPO's TinyLoRA $\eta$ and $S$ \\
Observation & logit of the $\mathrm{Beta}(1 + s, 1 + n - s)$ posterior: digamma mean, trigamma variance \\
GP & Mat\'ern 5/2, one lengthscale per dimension, fixed per-trial noise plus a learned floor, MAP fit \\
Lengthscale prior & $\mathrm{LogNormal}(\sqrt{2} + \log(L\sqrt{u}), \sqrt{3})$ \\
Trust region & side $L$ from 0.8, at most 1.6; doubled after 3 successes, halved after $\lceil \max(4, u)/b \rceil$ failures; restart below $0.5^7$ \\
Acquisition & Thompson sampling over $\min(5000, \max(2000, 200u))$ Sobol candidates in the region \\
GP data & every trial since the trust region started, plus the $\theta = 0$ replicates \\
Incumbent and output & the evaluated trial of highest posterior mean, also the trust-region center \\
Scoring & the $b$ candidates of a batch in one vLLM call, each a rank-$r$ LoRA with $B = U\Sigma\sum_i v_i P_i$, $A = V^\top$ \\
\bottomrule
\end{tabular}
\end{table}

Table~\ref{tab:hparams} collects the settings of Sections~\ref{sec:method} and~\ref{sec:experiments} and Appendix~\ref{app:param}. On the SimpleRL-Zoo tiers $S \approx 390$-$400$.

\subsection{Evaluation harness}
\label{app:harness}

Every model is prompted with its SimpleRL-Zoo template: a plain question-answer template for Qwen2.5-1.5B, Llama3.1-8B and Mistral-7B, and for every other model the ChatML template that asks to reason step by step and put the final answer within \verb|\boxed{}|. vLLM \citep{kwon2023vllm} generates up to 1024 tokens, greedily for greedy accuracy and four samples at $T = 1$ for avg@4. math-verify (0.9.0) extracts the final answer, the last \verb|\boxed{}| of the completion or else its last mathematical expression, and checks it for symbolic or numeric equivalence with the reference answer; a completion with no answer, or whose grading times out, counts as wrong. Training rewards and every evaluation, of the base models included, use this one grader. The benchmarks are MATH-500 (500 problems), GSM8K (1319), AIME24 (30), AMC23 (40), Minerva Math (272) and OlympiadBench (675).

\subsection{Compute}
\label{app:infra}

Every run, of either optimizer, uses one GPU of a shared cluster, scheduled on whichever card was free: NVIDIA L40S (48\,GB), A100 (40 and 80\,GB), A40, A16, H100, L4 and RTX A4000. The \nModels{} models take over 14k GPU-hours of training (logged hours undercount runs resumed after preemption), plus evaluation. Since the two runs of a seed pair mostly ran on different cards (84 of the 336 avg@4 pairs at $b = 1$ share one), Section~\ref{sec:efficiency} compares FLOPs rather than wall-clock time.

\subsection{Licenses}
\label{app:licenses}

Models: the Qwen2.5 and Qwen2.5-Math models used here and Mistral-7B are released under Apache 2.0, Llama3.1-8B under the Llama 3.1 Community License, and DeepSeek-7B and DeepSeek-Math-7B under the DeepSeek Model License. Data: MATH, GSM8K and MATH-500 (from PRM800K) are MIT-licensed; the SimpleRL-Zoo tiers are drawn from MATH and GSM8K and released under Apache 2.0. OlympiadBench is MIT-licensed, and Minerva Math consists of MIT OpenCourseWare problems (CC BY-NC-SA 4.0). AIME24 and AMC23 are competition problems of the Mathematical Association of America; the AIME24 copy we use is released under Apache 2.0, the AMC23 copy carries no license, and both serve for evaluation only. Software: vLLM, TRL and Unsloth are Apache 2.0, BoTorch and GPyTorch MIT, and math-verify Apache 2.0.

\subsection{Observation model}
\label{app:obsmodel}

\begin{figure}[!htbp]
\centering
\includegraphics[width=\linewidth]{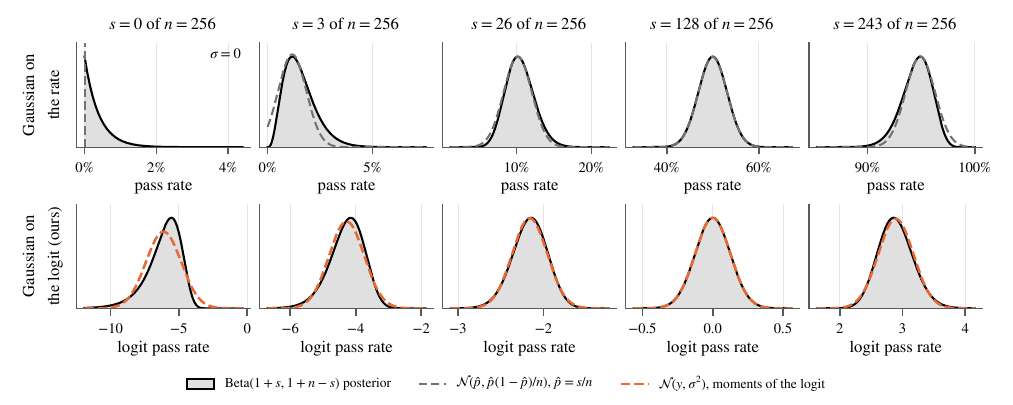}
\caption{The Gaussian observation a GP receives for a candidate that solves $s$ of $n = 256$ prompts, against the $\mathrm{Beta}(1 + s, 1 + n - s)$ posterior on its pass rate. \emph{Top}: a Gaussian on the rate with the plug-in binomial variance, each in pass-rate space. \emph{Bottom}: \llmbo{}'s Gaussian with the exact mean $y$ and variance $\sigma^2$ of the posterior's logit (Section~\ref{sec:method}), in logit space, the space the GP models.}
\label{fig:obsmodel}
\end{figure}

A candidate's score is a count, $s$ solved prompts out of $n$, whereas a GP with a closed-form posterior needs a Gaussian observation. The common choice, the rate $\hat p = s/n$ with the plug-in binomial variance $\hat p(1 - \hat p)/n$, fails where it matters most for us (Figure~\ref{fig:obsmodel}, top). Near a pass rate of 0 or 1, where the Instruct models train, the binomial posterior is skewed and the symmetric Gaussian misplaces its mass, and at $s = 0$ or $s = n$ the variance vanishes: the GP would treat one noisy evaluation as exact. A Gaussian on the rate also leaves the GP free to predict rates outside $[0, 1]$.

\llmbo{} instead matches a Gaussian to the logit of the Beta posterior (Figure~\ref{fig:obsmodel}, bottom). On the logit scale the posterior is close to Gaussian over the whole range of rates; at $s = 0$ it is skewed, but its variance stays finite and grows as the rate approaches the boundary, so an all-fail or all-pass evaluation reads as weak evidence rather than as certainty. The uniform prior keeps both shape parameters at least 1, which limits this skew; under the Jeffreys prior, with $\alpha = \frac12$ at $s = 0$, it is much stronger. The GP thus models an unbounded quantity with a known noise per trial, inside a stock fixed-noise GP whose Thompson samples stay exact. The construction is the one of Dirichlet-based GP classification \citep{milios2018dirichlet} and of Laplace Matching \citep{hobbhahn2021laplace}, which feed Gamma and Beta pseudo-observations to a GP as heteroscedastic Gaussians in a transformed space; we use the exact moments of the logit, $\psi(\alpha) - \psi(\beta)$ and $\psi_1(\alpha) + \psi_1(\beta)$, rather than those of a Laplace approximation. A latent GP with an exact binomial likelihood would avoid the approximation altogether but loses the closed-form posterior, and with it exact Thompson sampling, for a gain the figure shows to be small at $n = 256$.

\subsection{SVD bases across GPU models}
\label{app:svd-signs}

It is important to remember that the truncated SVD is defined only up to signs. TinyLoRA and LoRA-XS build every module's update on the frozen factors of its rank-$r$ truncated SVD, $\Delta W_m = U_m \Sigma_m R_m V_m^\top$ (Appendix~\ref{app:param}). For distinct singular values, these factors are unique only up to a joint sign flip of each singular pair $(u_i, v_i) \mapsto (-u_i, -v_i)$, and which sign a solver returns is an implementation detail. On our cluster, \texttt{torch.linalg.svd} in fp32 picks different signs on different GPU models. We computed the leading two singular pairs of the seven layer-0 projections of the six Qwen2.5 models on six GPU models: A100-PCIe-40GB and A100-SXM4-80GB agree exactly, and any two of the remaining five (A100, A40, H100, L4, L40S) disagree on the sign of 6 to 13 of the 84 singular vectors, 7-15\%, while the factors agree to within $4 \cdot 10^{-3}$ once the signs are aligned.

A sign flip changes TinyLoRA's subspace. Write the signs one solver picks relative to another as a diagonal matrix $D_m \in \{\pm 1\}^{r \times r}$, so that it returns $U_m D_m$ and $V_m D_m$. Since $D_m$ commutes with $\Sigma_m$,
\begin{equation}
    U_m D_m \Sigma_m R_m (V_m D_m)^\top = U_m \Sigma_m (D_m R_m D_m) V_m^\top,
\end{equation}
so the second solver's adapter with coefficients $R_m$ is the first solver's with $D_m R_m D_m$; at $r = 2$, a flip of one of the two pairs negates the off-diagonal entries of $R_m$. For LoRA-XS this is a change of coordinates, the set of reachable updates is unchanged, although a trained $R_m$ must still be read with the bases it was trained in. For TinyLoRA it is not: with $R_m = \sum_i v_i P^{(m)}_i$, the flipped solver yields the projections $D_m P^{(m)}_i D_m$, with $D_m$ differing between modules, and so a different $u$-dimensional subspace of the LoRA-XS space for every $u < M r^2$ (Appendix~\ref{app:tying}). Its distribution is the same, but seed $k$ on one card model is not seed $k$ on another. Left alone, this would silently break three things: the seed pairing of \llmbo{} and GRPO whenever the two runs land on different card models, a run preempted and requeued onto a different card model, which would continue in a different subspace from the same $v$, and the evaluation of an exported adapter.

We pin the basis by computing the SVD once, in the GRPO run of each seed, and store $V_m^\top$ as the adapter's LoRA $A$ matrix, which the exported adapter holds anyway. \llmbo{}'s search, resumed runs and the evaluation rebuild the remaining factors from the stored $V_m$ and the frozen weight alone: $\Sigma_m$ from the column norms of $W_m V_m$ and $U_m = W_m V_m \Sigma_m^{-1}$. Because $\sigma_i u_i = W_m v_i$, the sign of each $u_i$ follows from that of $v_i$, and the bases are identical, up to floating-point error, on any hardware.

\subsection{A bounded GRPO loss}
\label{app:dual-clip}

On the base models, a few TinyLoRA runs under the standard clipped GRPO loss diverged: the pre-clipping gradient norm spiked to $10^2$-$10^{11}$, after which the reward fell to zero and did not recover (Qwen2.5-7B at $u = 64$, training reward from 0.59 to 0.01; Qwen2.5-Math-7B at $u = 16$, GSM8K accuracy from 82\% to 40\%). Plain LoRA and the Instruct models were never affected.

The cause is a known gap in the clipped objective. Each GRPO step takes one optimizer step on its rollouts, so $\pi_{\theta_\mathrm{old}} = \pi_\theta$ and the ratio $\rho_{j,t}$ is 1 in exact arithmetic. The old log-probabilities, however, come from a separate bf16 forward pass, so $\rho_{j,t}$ carries numerical noise that can reach several nats on low-probability tokens. For a token with $A_j < 0$ and $\rho_{j,t} > 1 + \epsilon$, the minimum in the GRPO objective selects the unclipped term $\rho_{j,t} A_j$, and the loss $\rho_{j,t} |A_j|$ is unbounded: a single noisy token can dominate the gradient. Gradient-norm clipping bounds the step size but not its direction, and TinyLoRA's tied $v$ then moves every module at once. On a base model, a perturbed policy soon produces completions that run past the token limit, every reward in a group becomes zero, and no signal is left to recover from.

We therefore use the dual clip of \citet{ye2020mastering}, which bounds the objective of negative-advantage tokens from below by $\delta A_j$ with $\delta = 2$:
\begin{equation}
    \max\!\big( \min\!\big( \rho_{j,t} A_j,\ \mathrm{clip}(\rho_{j,t}, 1 - \epsilon, 1 + \epsilon) A_j \big),\ \delta A_j \big) \quad \text{for } A_j < 0.
\end{equation}
It changes the loss only of tokens with $\rho_{j,t} > \delta$, which in a healthy run, where the ratio departs from 1 only by numerical noise, are rare. With it, the largest gradient norm of these runs stays below 5 throughout training.

\end{document}